\documentclass[twocolumn]{article}

\usepackage{geometry}
\usepackage{amsmath}
\usepackage{amsthm}
\usepackage{amssymb}

\usepackage[utf8]{inputenc}
\usepackage{hyperref}

\usepackage[sort&compress,numbers,square]{natbib}
\usepackage{graphicx, color}

\usepackage{mathrsfs} 

\usepackage{adjustbox}
\usepackage{multirow}
\usepackage{subcaption}
\usepackage{array}
\usepackage[font=small,skip=1pt]{caption}
\newcolumntype{C}[1]{>{\centering\let\newline\\\arraybackslash\hspace{0pt}}m{#1}}

\title{Monocular Depth Estimators: Vulnerabilities and Attacks}
\author{Alwyn Mathew\thanks{Equal contribution.} \and Aditya Prakash Patra\footnotemark[1] \and Jimson Mathew}
\date{
	Indian Institute of Technology Patna \\ \texttt{\{alwyn.pcs16, aditya.cs16, jimson\}@iitp.ac.in}\\[2ex]%
}

\begin{document}
	
\setlength{\abovedisplayskip}{0pt}
\setlength{\belowdisplayskip}{0pt}

\maketitle
	
\begin{abstract}
	Recent advancements of neural networks lead to reliable monocular depth estimation. Monocular depth estimated techniques have the upper hand over traditional depth estimation techniques as it only needs one image during inference. Depth estimation is one of the essential tasks in robotics, and monocular depth estimation has a wide variety of safety-critical applications like in self-driving cars and surgical devices. Thus, the robustness of such techniques is very crucial. It has been shown in recent works that these deep neural networks are highly vulnerable to adversarial samples for tasks like classification, detection and segmentation. These adversarial samples can completely ruin the output of the system, making their credibility in real-time deployment questionable. In this paper, we investigate the robustness of the most state-of-the-art monocular depth estimation networks against adversarial attacks. Our experiments show that tiny perturbations on an image that are invisible to the naked eye (perturbation attack) and corruption less than about 1\% of an image (patch attack) can affect the depth estimation drastically. We introduce a novel deep feature annihilation loss that corrupts the hidden feature space representation forcing the decoder of the network to output poor depth maps. The white-box and black-box test compliments the effectiveness of the proposed attack. We also perform adversarial example transferability tests, mainly cross-data transferability.
	
\end{abstract}

\begin{figure}[ht]
\centering
\newcommand{\turnheightnew}{0.15\columnwidth}
\centering

\begin{center}

\newcommand{\turnwidth}{0.485\columnwidth}
\newcommand{\imlabel}[2]{\includegraphics[width=0.5\columnwidth]{#1}%
\raisebox{2pt}{\makebox[-2pt][r]{\footnotesize #2}}}

\begin{tabular}{@{\hskip 0mm}c@{\hskip 1.5mm}c}
\centering

\imlabel{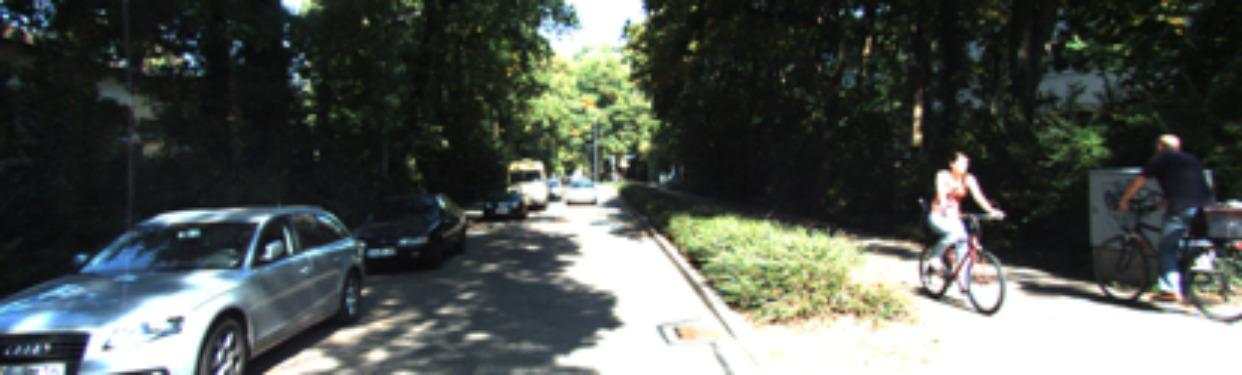}{\textcolor{red}{Clean input}} &
\imlabel{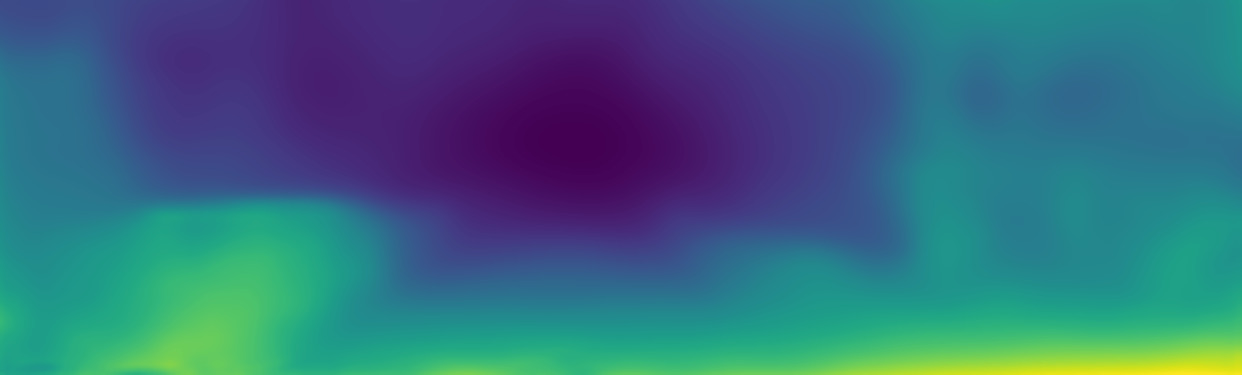}{\textcolor{red}{Clean depth}} \\

\imlabel{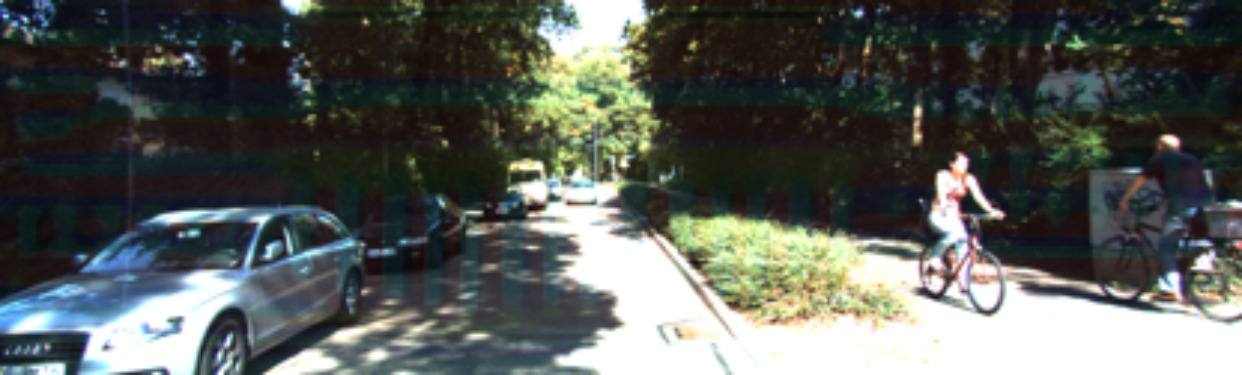}{\textcolor{red}{Perturbation input}} &
\imlabel{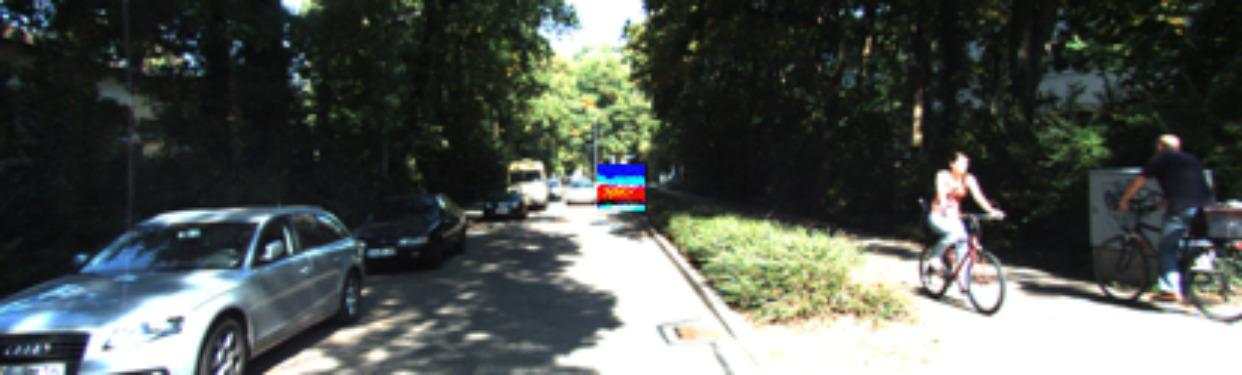}{\textcolor{red}{Patch input}} \\

\imlabel{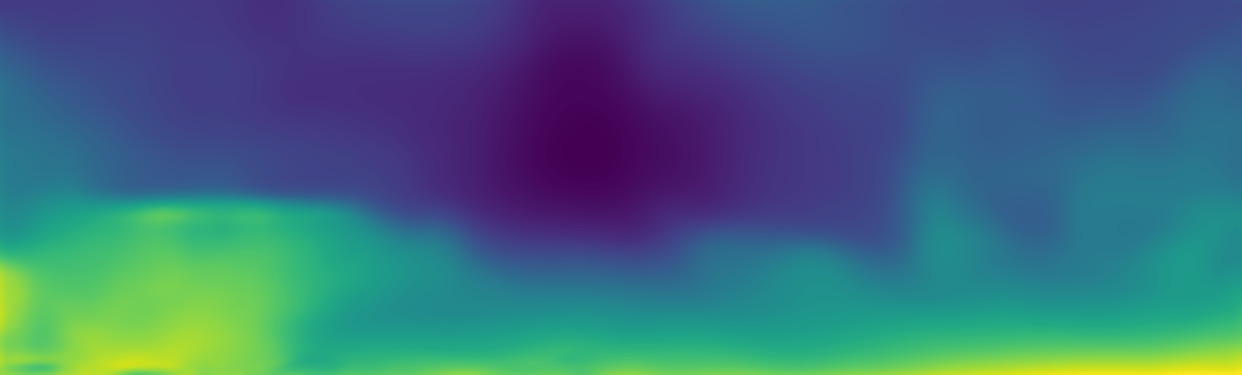}{\textcolor{red}{Perturbation depth}} &
\imlabel{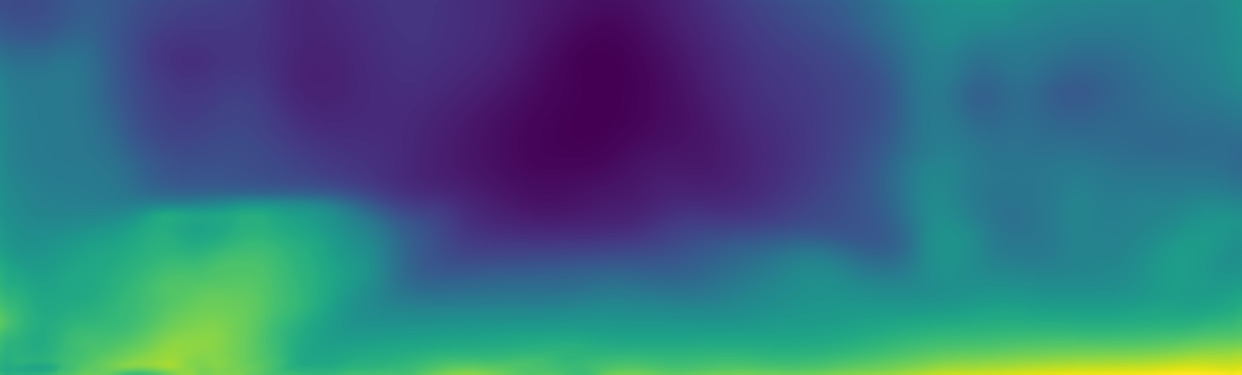}{\textcolor{red}{Patch depth}} \\

\imlabel{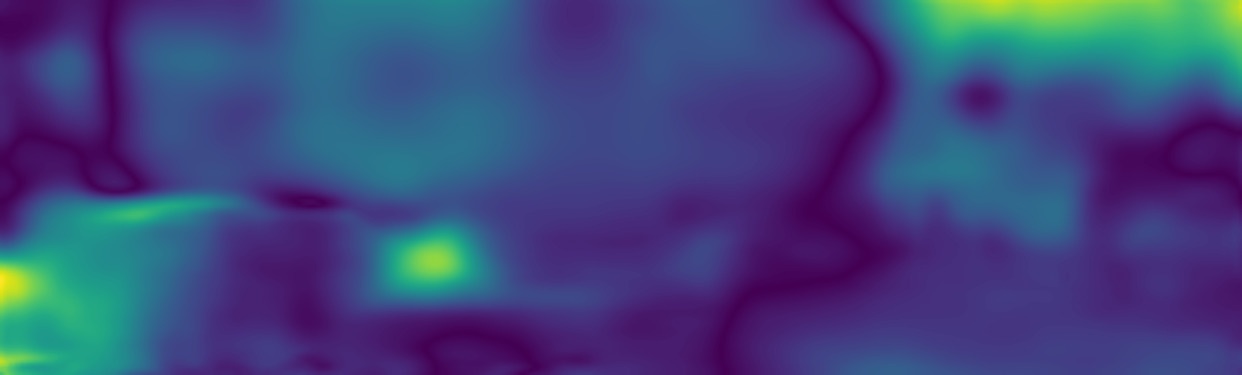}{\textcolor{red}{Pert depth gap}} &
\imlabel{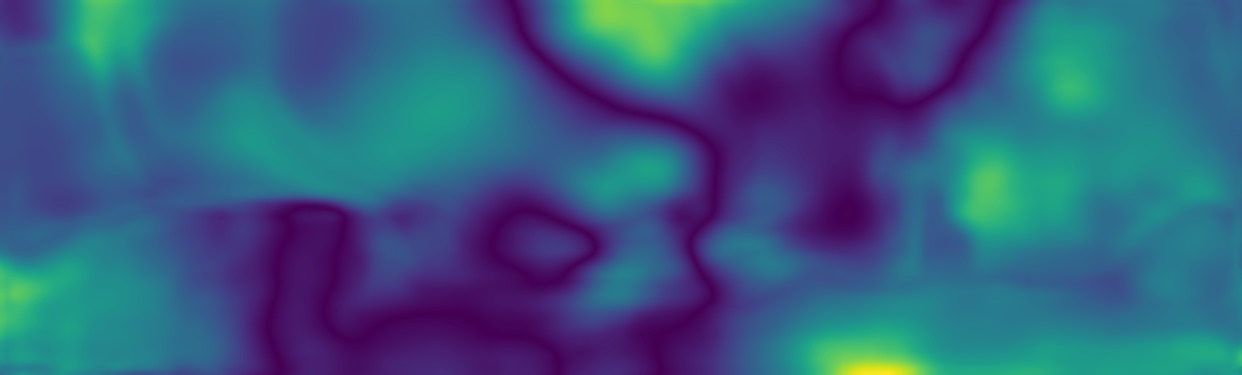}{\textcolor{red}{Patch depth gap}}

\end{tabular}
\end{center}

\caption{Adversarial attacks on monocular depth estimators. First row from left: Unattacked input image, Attacked input image with adversarial perturbation, Attacked input image with an adversarial patch. Second row from left: Depth estimated with SFM~\cite{zhou2017unsupervised} with Unattacked input image, Attacked input image with adversarial perturbation, Attacked input image with an adversarial patch. Third row from left: Ground truth depth from LiDAR, Gap between unattacked depth and attacked depth with perturbation and patch attack.}
\label{fig:teaser}

\end{figure}

\section{Introduction}

Per-pixel depth estimation from more than one two-dimensional images is a well-explored research field in computer vision. Recently the surge in estimating per-pixel depth from a single two-dimensional image has increased in the research community. Monocular depth estimation poses a variety of real-world applications like autonomous driving, robot navigation, and surgical devices. Monocular depth estimation has been studied in numerous settings like supervised, semi-supervised, and self-supervised. The introduction of convolution neural network (CNN) for depth estimation has helped to gain significant improvement over other traditional methods. The supervised method considered monocular depth estimation as a regression problem with ground truth depth which are collected using devices like LiDAR and RGBD camera. However ground truth depth collection from these devices is a tedious task. The devices used for this purpose has its own downfalls. LiDAR produces a sparse depth map, and RGBD cameras have limitations on the maximum depth it can capture. The self-supervised method studies monocular depth estimation as a view synthesis problem using stereo images and/or video sequences. This view synthesis acts as the supervision of the system. The semi-supervised setting takes the best of both worlds by using sparse depth maps and multi view images for depth estimation. 

Even though deep learning-based methods have shown astonishing results in most computer vision task, it has been shown that they are vulnerable to attacks. The vulnerability of deep learning models for tasks like image classification, detection, and segmentation is extensively studied in the literature. However, attacks on safety-critical models like depth estimation are not investigated broadly. In this paper, we investigate the robustness of monocular depth estimators against adversarial attacks. We mainly examine pixel perturbation and patch attack against a diverse set of depth estimators. There are few works on adversarial attacks on monocular depth estimation. But these attacks only focus on classical perturbation attacks like FGSM (Fast Gradient Sign Attack) and IFGSM (Iterative FGSM). FGSM is one of the most straightforward white-box adversarial attacks introduced in \cite{goodfellow2014explaining} which uses the gradient of the loss with respect to the input data, then adjusts the input data to maximize the loss. FGSM is a one-shot method, whereas IFGSM takes more than one gradient steps to find the perturbation. Some of the drawbacks of these attacks are:

\begin{enumerate}
    \item Perturbation attacks like FGSM are image dependent and are not ideal for real-world attacks as it needs access to the image captured by the system to attack the model.
    \item Defense against FGSM is more natural than adaptive perturbations generated by a network.
\end{enumerate}

Existing adversarial attacks on classification, object detection, and segmentation aim to maximize non-essential softmax and minimize other softmax predictions of the network, thereby flipping the final prediction of the deep neural network.  However, these attacks don't have a significant effect on the scene understanding models like depth estimation. Despite its success in basic vision tasks like classification ,regression models can be attacked by these attacks to a very limited extent. In this work, we introduce a deep feature annihilation loss for both perturbation and patch attacks as shown in Figure~\ref{fig:teaser}. Deep feature annihilation (DFA) loss corrupts the internal representation of a deep neural network, rather than just the final layers. Attacks with DFA have demonstrated a more substantial effect on resultants. The main contributions of this paper are:

\begin{enumerate}
    \item Show the vulnerability of monocular depth estimators against precisely calculated perturbation added to the input image.
    \item Study the effect of patch attack on monocular depth estimators even when less than one percent of the image pixels are corrupted.
    \item Introdunce a novel Deep Feature Annihilation loss for stronger attacks. 
    \item Conducted extensive white and black-box testing for both perturbation and patch attack for monocular depth estimation.
    \item Study the transferability of adversarial samples, mainly cross-model transferability and cross data transferability.
\end{enumerate}

\section{Related Works}

\subsection{Monocular Depth Estimation}

Zhou \textit{et al}., \cite{zhou2017unsupervised} proposed an unsupervised depth estimation end-to-end framework from video sequence where view synthesis act as the supervisory signal. Godard \textit{et al}., \cite{godard2017unsupervised} estimate disparity from stereo images with a novel LR consistency loss.
Mahjourian \textit{et al}., \cite{mahjourian2018unsupervised} introduced a 3D loss to enforce consistency of the estimated 3D point cloud. Yin \textit{et al}., \cite{yin2018geonet} jointly learned monocular depth, optical flow and ego motion from videos in unsupervised manner. Luo \textit{et al}., \cite{luo2018every} also jointly learned monocular depth, optical flow and ego motion from video but used these result to produce motion mask, occlusion mask, 3D motion map for rigid background and dynamic objects. Wang \textit{et al}., \cite{wang2018learning} proposed novel depth normalization strategy that substantially improves performance of the estimator. Godard \textit{et al}., \cite{godard2019digging} achieved state of the art results in KITTI rectified dataset with minimum reprojection loss, full-resolution mutli-scale sampling and a mask that ignores training pixels that violate camera motion assumptions. Bian \textit{et al}., \cite{bian2019unsupervised} proposes a scale consistent loss and a mask to handle moving and occluded objects. Lee \textit{et al}., \cite{lee2019big} introduce a network architecture that utilizes novel local planer guidance layers to guide densely encoded features to estimate depth.

\subsection{Adversarial Attack}

In \cite{szegedy2013intriguing} Szegedy \textit{et al}. shows that numerous state of the art neural networks are vulnerable to adversarial samples. Attacking a neural network with adversarial samples is called an adversarial attack. Adversarial attacks strive to make small perturbation in the network input so that it causes erroneous results. Attacks are mainly two types: perturbation attack and patch attack. The perturbation attack adds minute changes to pixels values so that its indistinguishable from the clean image but affects the prediction accuracy of the model. Patch attack adds a small patch that usually occupies less than one percent of the image and makes the result inaccurate. The drawback of the perturbation attack is that the attack needs to have access to the system to modify the image with the adversarial sample. Patch attacks succeed in this issue and can be placed at an apparent location in the scene to attack the model.

Goodfellow \textit{et al.} \cite{goodfellow2014explaining} introduced a fast and straightforward perturbation attack called Fast Gradient Sign Attack (FGSM) that uses gradient with respect to input and adjust the input data to maximize the loss. Nguyen \textit{et al.} \cite{nguyen2015deep} shows how easy it is to generate adversarial samples that are undetectable to the human eye. They have also shown completely unrecognizable attacked images classified with a high degree of confidence. Su \textit{et al.} \cite{su2019one} reveals that even attacking a single pixel can change the classification prediction. \cite{kurakin2016adversarial} demonstrates a real physical adversarial attack in which a printed adversarial example can also affect the prediction correctness. \cite{athalye2017synthesizing} reflects the existence of 3D adversarial samples, which can fool classification models in different viewpoints, transformations, and cameras. \cite{sharif2016accessorize} fooled facial recognition system with adversarial glasses and \cite{evtimov2017robust} ticked stop sign classifier by placing adversarial stickers over it.

\subsection{Adversarial Attack on Scene Understanding}

Adversarial attacks in tasks like detection, semantic segmentation, and reinforcement learning are studied in \cite{hendrik2017universal, xie2017adversarial, behzadan2017vulnerability}. We mainly focus on attacks toward scene understanding tasks like depth estimation. Ranjan \textit{et al.} \cite{ranjan2019attacking}, shows the effect of adversarial patch attack on optical flow estimation. They have demonstrated that corrupting less than one percent of the pixels with a patch can deteriorate the prediction quality. The effect of adversarial patch extends beyond the region of the attack in some of the optical flow estimation models. In essence, the encoder-decoder architecture based models found to be more sensitive than spatial pyramid architecture. They also visualized feature maps of successful and failed attacks to understand the essence under the hood. Hu \textit{et al.} \cite{hu2019analysis} studies basic perturbation attacks like IFGSM (Iterative FGSM) on monocular depth estimation and also proposes a defense mechanism based on \cite{hu2019visualization}. The perturbations non-salient parts are marked out to defend against the attack. Zhang \textit{et al.} \cite{zhang2020adversarial} explores basic perturbation attacks with FGSM, IFGSM, and MI-FGSM (Momentum IFGSM) for non-targetted, targeted, and universal attack in monocular depth estimation. In contrast, we investigate global and image dependent perturbation with a novel DFA loss that attacks deep feature maps rather than final activation like \cite{hu2019analysis} and \cite{zhang2020adversarial}. We also conducted an extensive study adversarial patch attacks on monocular depth estimators, which are more significant in a real-world attack. The transferability of both global perturbation and patch are studied in this work.

\section{Approach}

The robustness of a model that predicts depth from one or more images is considered to be a safety-critical task. We aim to optimize pixel intensity change in an image so that it affects the final prediction of the depth estimation models. These pixel intensity changes can occur in mainly two ways: 

\begin{enumerate}
    \item Adding a small perturbation to the clean image in a way that it's indistinguishable from the clean image as briefed in Section~\ref{pertattack}.
    \item Corrupting few pixels, usually less than 1\% in the image as briefed in Section~\ref{patchattack}.
\end{enumerate}

Say $D \in \mathbb{R}^{h \times w}$ where $D(x,y)$ is the ground truth depth value at pixel location $(x,y)$ where $x \in \mathbb{Z}_+^{h}, y \in \mathbb{Z}_+^{w}$ of a clean image $I \in \mathbb{R}^{h \times w \times 3}$, the attacked adversarial image $\tilde{I} \in \mathbb{R}^{h \times w \times 3}$ will force the model to predict a wrong estimate $\tilde{D} \in \mathbb{R}^{h \times w}$ with a constraint that $I(x,y) \approx \tilde{I}(x,y)$ and $h, w$ is the height and width of the image.  Though an adversarial sample can be created against the objective by maximizing the loss between predicted and ground-truth value, but ground truth depth maps are scarce. Widely popular autonomous dataset like KITTI \cite{geiger2013vision} provides sparse LiDAR ground truth but are very limited in number. The indoor NYU dataset \cite{silberman2012indoor} consists of dense depth ground truth, but the depth range are limited to the RGBD camera.  

Ground truth data collected from devices like Sonar, Radar, and Lidar can be used to train a supervised depth estimation model. But supervised depth learning requires a vast amount of ground truth depth data, and these depth sensors have their error and noise characteristics, which will affect the learning process of the deep neural network. 3D lasers depth sensors measurements are typically much sparser than the image and thus lack detailed depth inference. These sensors also require accurate calibration and synchronization with the cameras. Rather than relying on ground truth depth information, an alternative approach to train a model to estimate depth is by self-supervised fashion using stereo image pairs and/or video streams from a single camera. The self-supervision signal comes from the loss between the synthesized view and the target view. View synthesis is another extensively explored problem in computer vision, where the objective is to reconstruct a specific view from one or more given views. 

As all state of the art approaches are self-supervised, in this investigation we study five self-supervised monocular depth estimators, namely, SFM (Structure from motion)~\cite{zhou2017unsupervised}, SCSFM (Scale-Consistent SFM)~\cite{bian2019unsupervised}, Monodepth1~\cite{godard2017unsupervised}, DDVO (Direct Depth Visual Odometry)~\cite{wang2018learning}, Monodepth1~\cite{godard2017unsupervised}, Monodepth2~\cite{godard2019digging}, and B2F~\cite{janai2018unsupervised}.  SFM, SCSFM, DDVO, B2F and Monodepth2 learns depth from monocular video but Monodepth1 uses stereo images. Monodepth2 also extends there work to stereo and stereo + videos training. Due to the limited availability of ground truth data, the depth network prediction of the clean image is taken as pseudo ground truth. Using these pseudo ground truths for attack makes it easy to attack a model in the absence of actual ground truth.

\begin{figure}[!ht]
\newcommand{\turnheightnew}{0.15\columnwidth}
\centering

\begin{tabular}{@{\hskip 0.5mm}c@{\hskip 0.5mm}c@{\hskip 0.5mm}c@{\hskip 0.5mm}c@{}}

Models & \multicolumn{3}{c}{Perturbation constrain} \\
\hline \\

 &
{\hspace{0mm} $\eta=0.01$} &
{\hspace{0mm} $\eta=0.05$} &
{\hspace{0mm} $\eta=0.1$} \\

\cline{2-4} \\

{SFM~\cite{zhou2017unsupervised}} &
\includegraphics[width=0.12\textwidth]{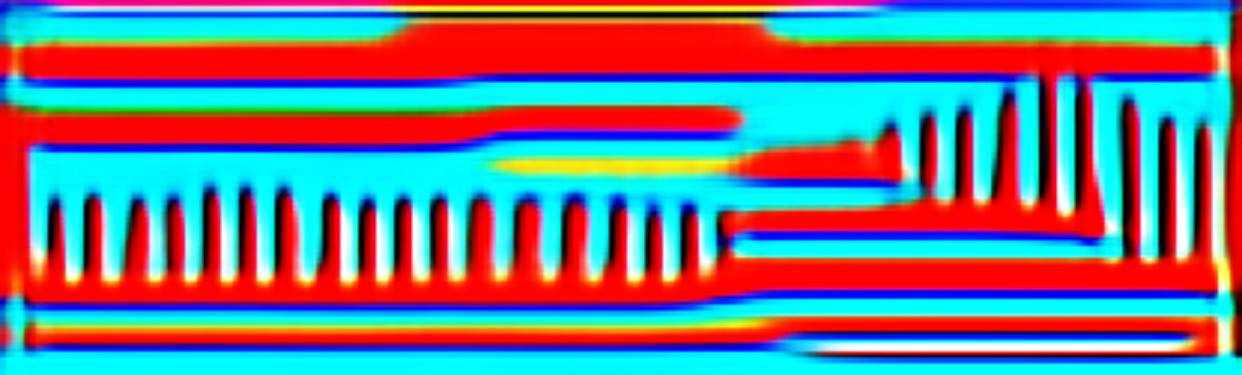} &
\includegraphics[width=0.12\textwidth]{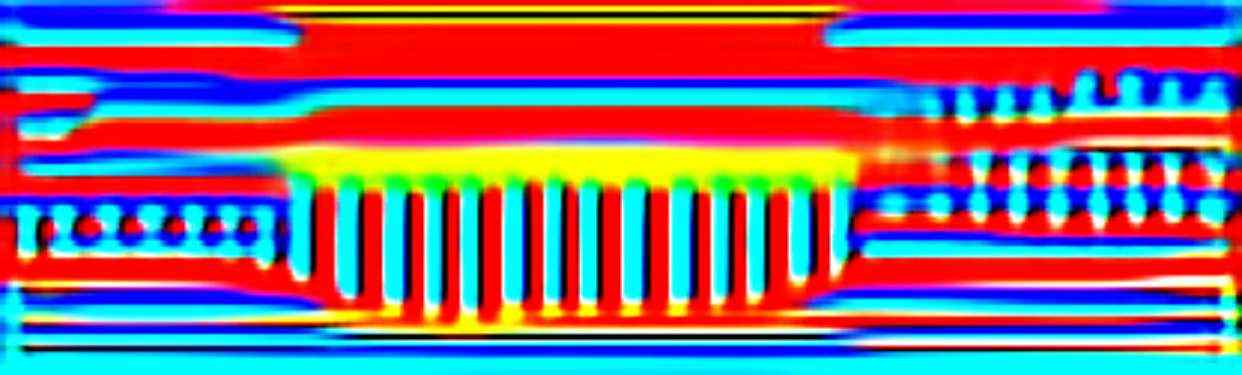} &
\includegraphics[width=0.12\textwidth]{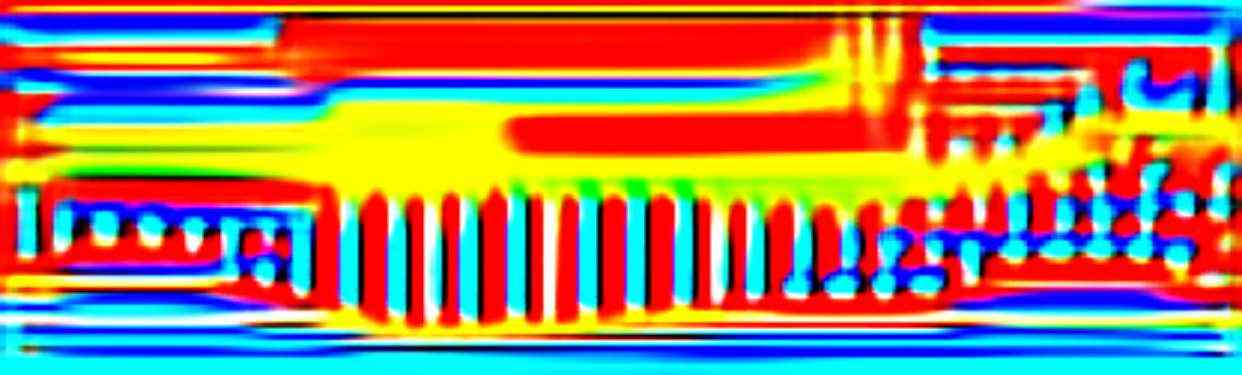} \\

{DDVO~\cite{wang2018learning}} &
\includegraphics[width=0.12\textwidth]{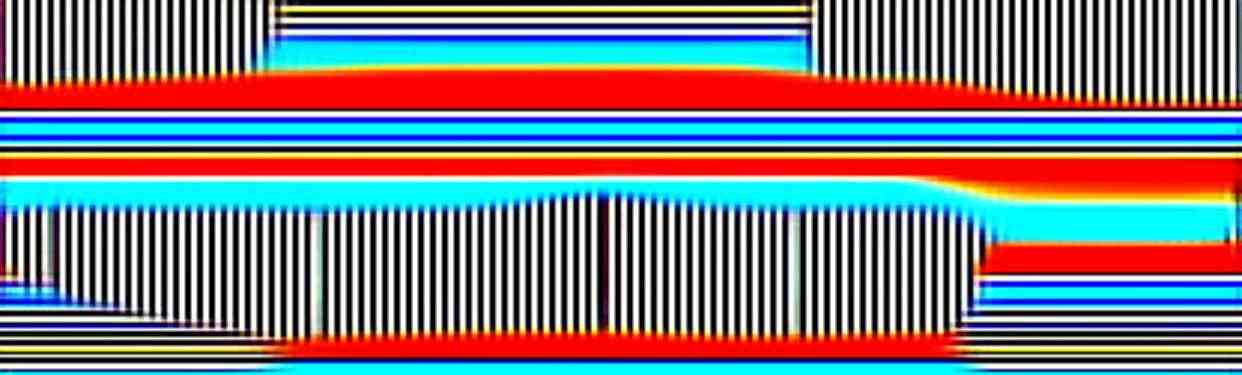} &
\includegraphics[width=0.12\textwidth]{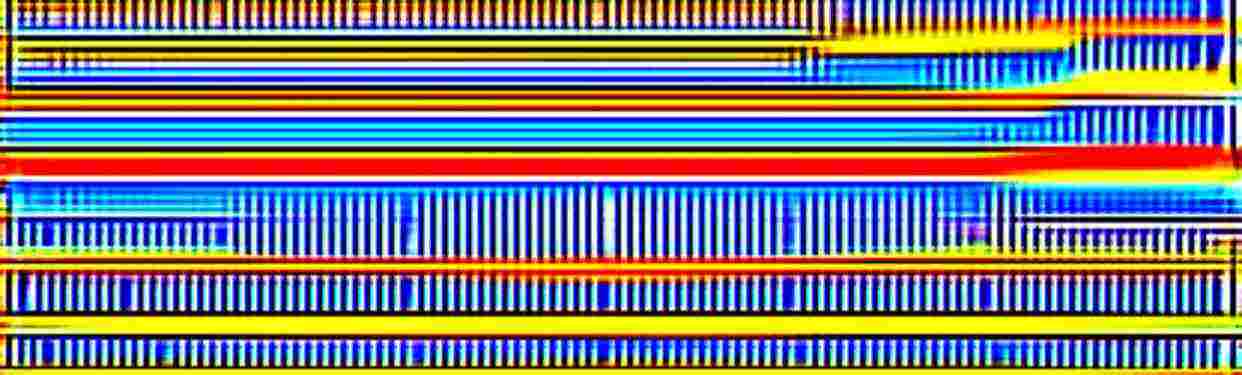} &
\includegraphics[width=0.12\textwidth]{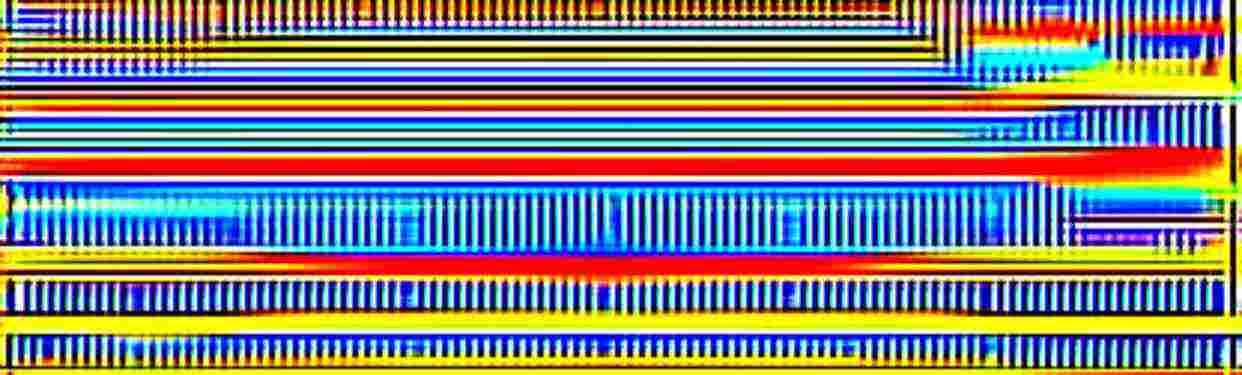} \\

{B2F~\cite{janai2018unsupervised} } &
\includegraphics[width=0.12\textwidth]{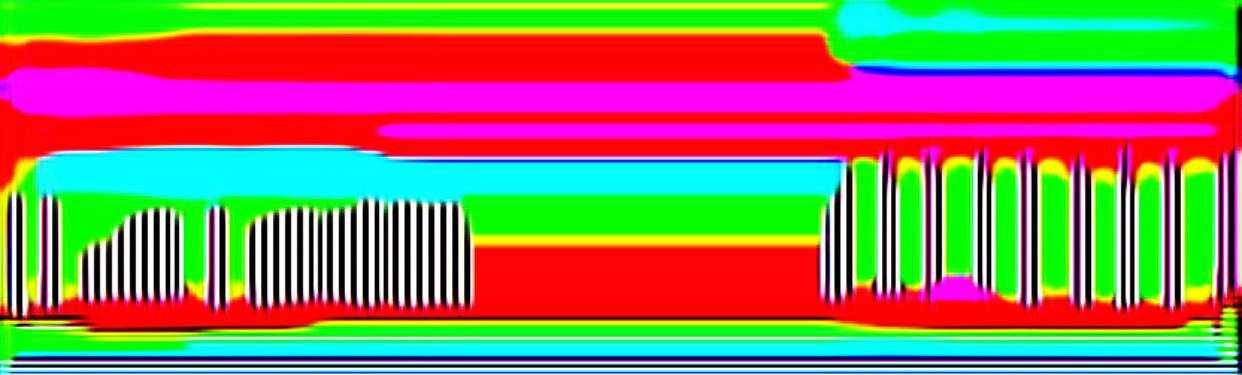} &
\includegraphics[width=0.12\textwidth]{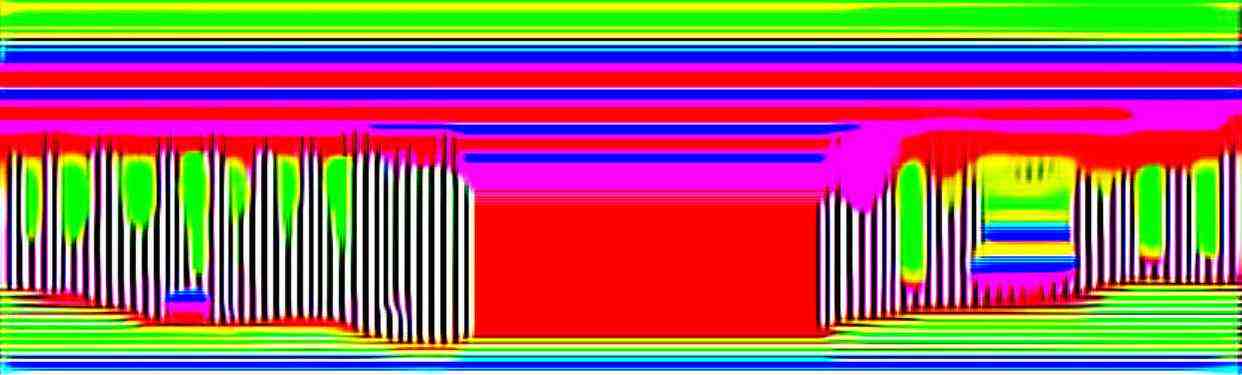} &
\includegraphics[width=0.12\textwidth]{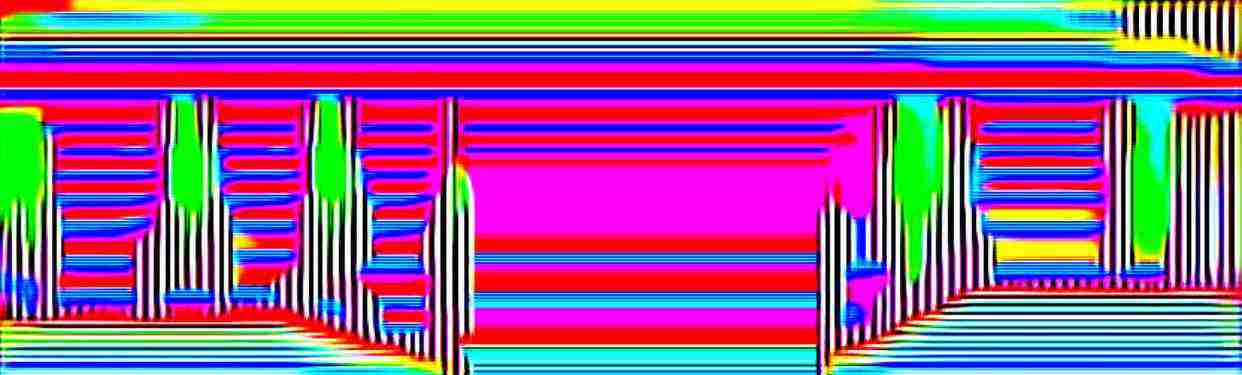} \\

{SCSFM~\cite{bian2019unsupervised}} &
\includegraphics[width=0.12\textwidth]{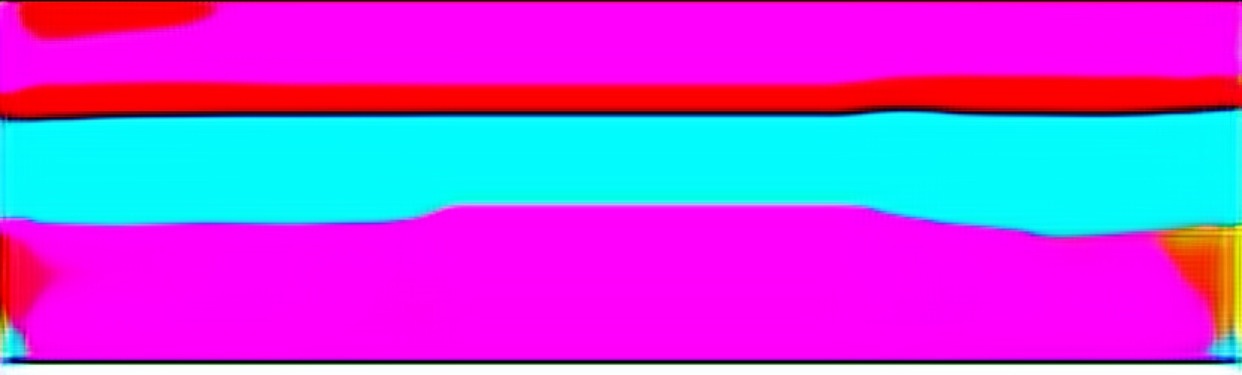} &
\includegraphics[width=0.12\textwidth]{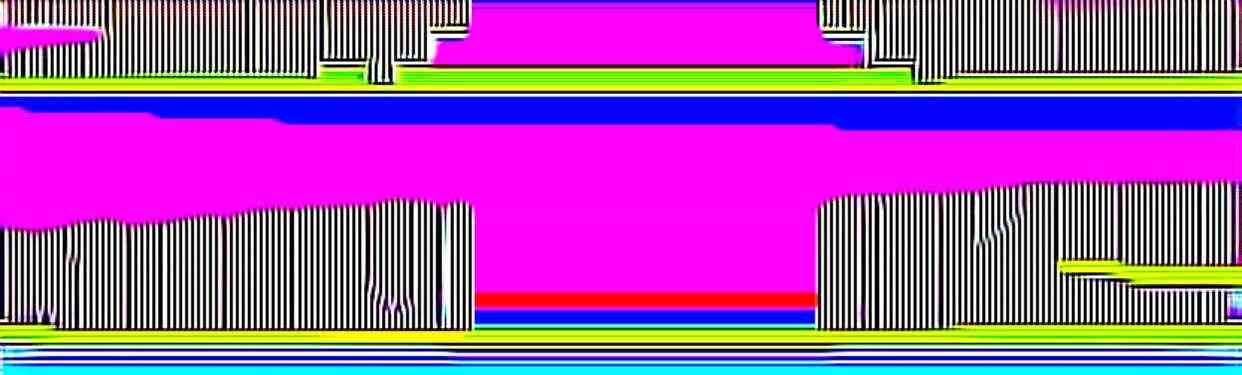} &
\includegraphics[width=0.12\textwidth]{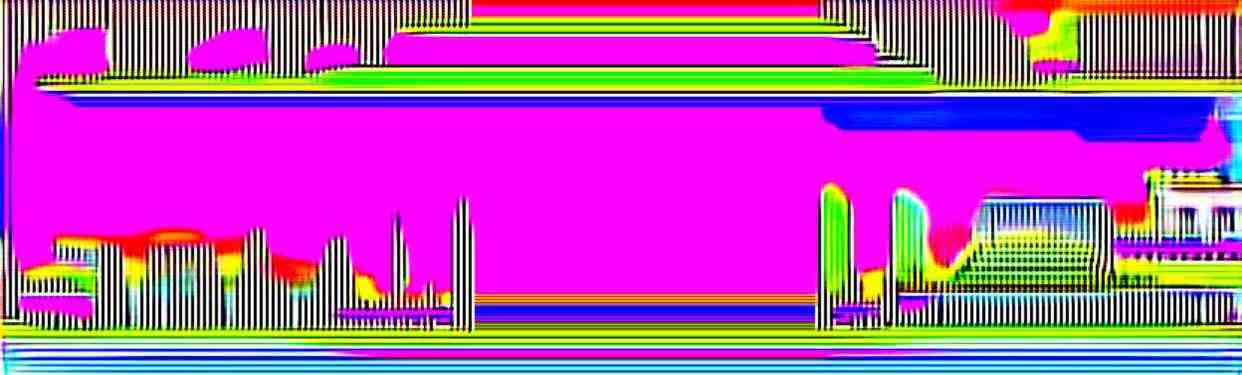} \\

{Mono1~\cite{godard2017unsupervised}} &
\includegraphics[width=0.12\textwidth]{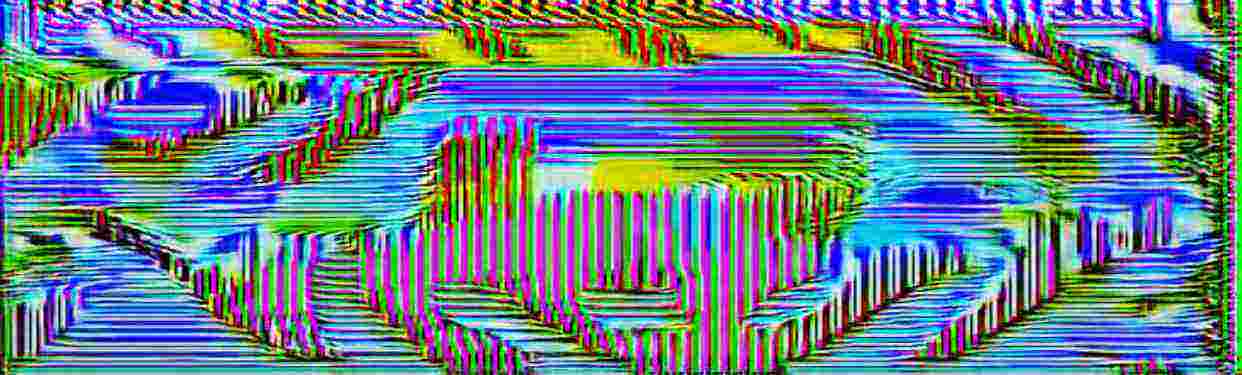} & 
\includegraphics[width=0.12\textwidth]{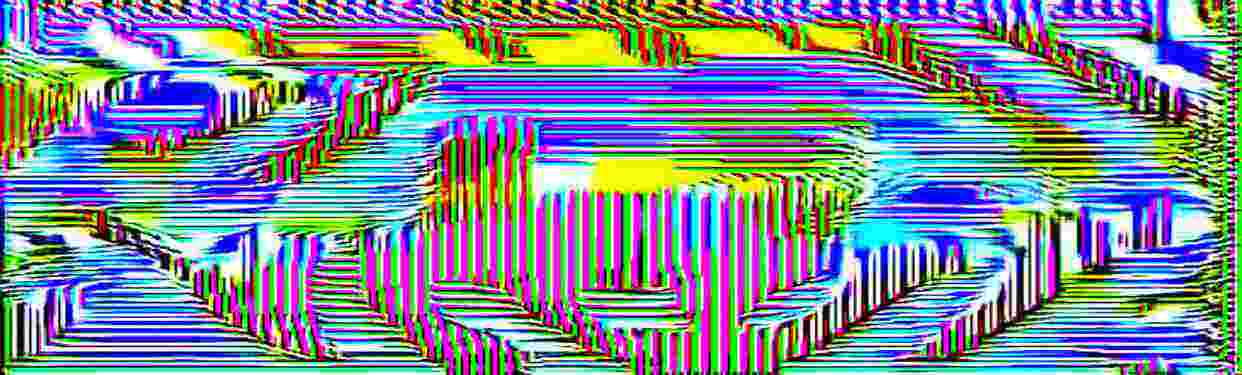} &
\includegraphics[width=0.12\textwidth]{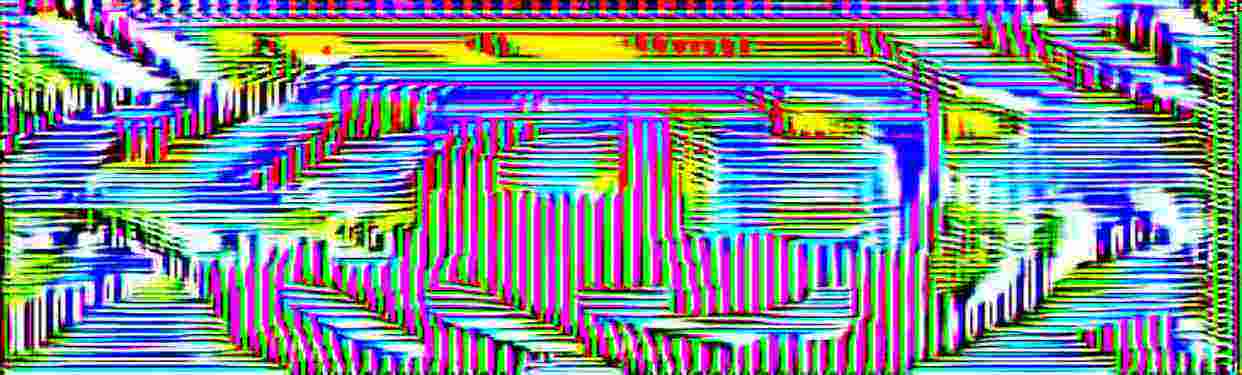} \\

{Mono2~\cite{godard2019digging}} &
\includegraphics[width=0.12\textwidth]{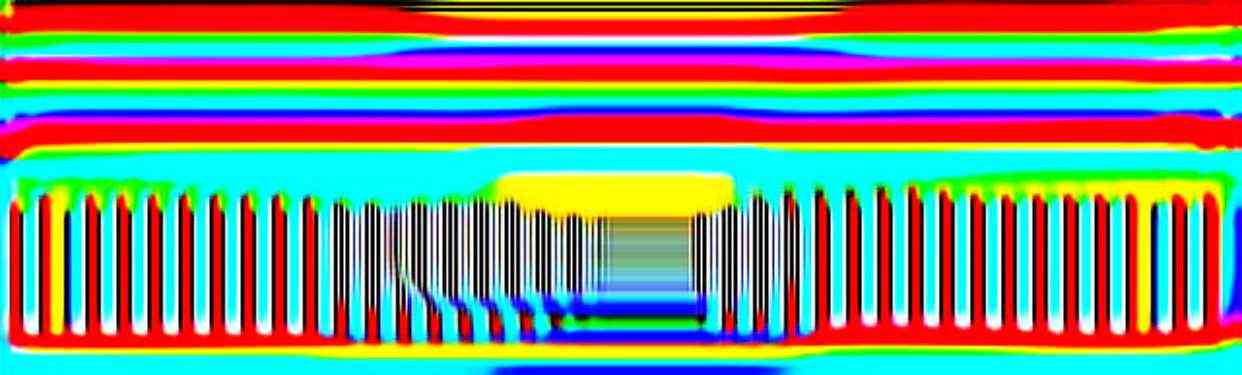} &
\includegraphics[width=0.12\textwidth]{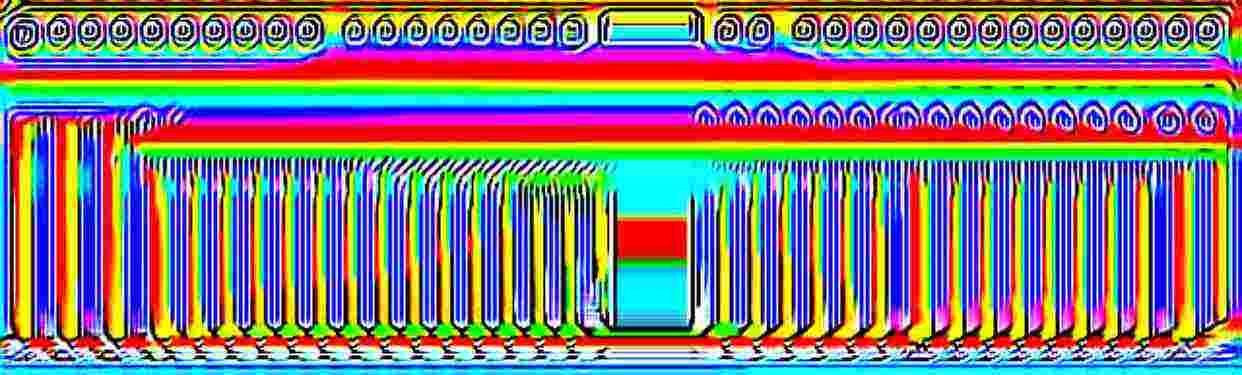} &
\includegraphics[width=0.12\textwidth]{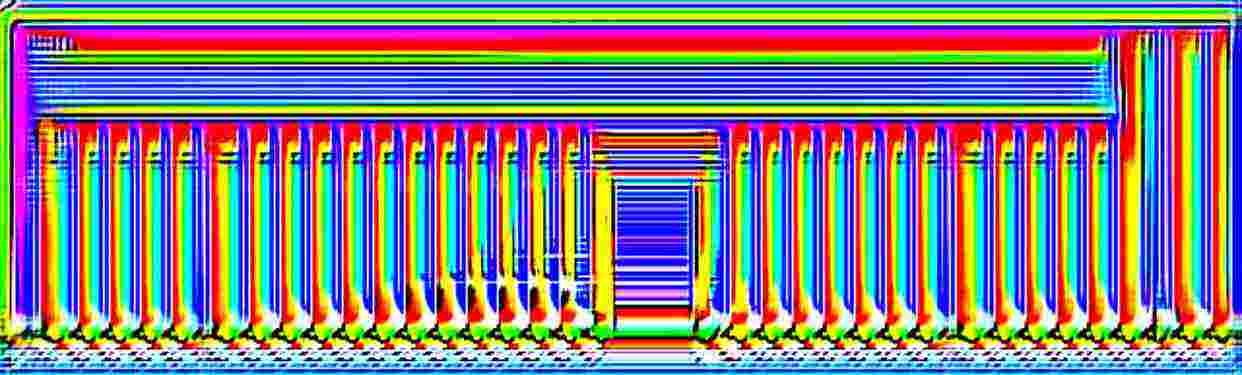} \\

\end{tabular}
\caption{Adversarial perturbation generated to attack each models under different $\eta$.}
\label{fig:pert_adv}
\end{figure}

\subsection{Perturbation Attack} \label{pertattack}

A depth network $M$ that estimates depth $D$ from a single image $I \in \mathbb{R}^{h \times w \times 3}$ as shown in Equ.~\ref{eq1}, is attacked by adding a perturbation $\alpha \in \mathbb{R}^{h \times w}$ to $I$ as shown in Equ.~\ref{eq2}.  The perturbation $\alpha$ is constrained with $\eta$ when added to an image to make the added perturbation unapparent to the naked eye as shown in Equ.~\ref{eq3}. The perturbation is designed to be undetected in the attacked image $\tilde{I}$, and the attacked image looks as close as possible to the clean image $I$. The response of the network $M$ at attacked image $\tilde{I}$ is $\tilde{D}$.

\begin{gather}
	D = M(I) \label{eq1} \\
	\tilde{I} = Adv(I, \alpha, \eta) = I + \eta * \alpha \label{eq3} \\
	\tilde{D} = M(Adv(I, \alpha, \eta)) \label{eq2}
\end{gather}

We basically experiment with two forms of pertubation namely global $\alpha_G$ and image-specific $\alpha_I$  pertubation. A perturbation network $P$ is used to generate image-specific or global perturbation. $P$ is a trivial encoder-decoder architecture based deep neural network. The image-specific perturbation network is trained with an unattacked image as input like $\alpha_I = P(I)$. This perturbation needs access to the input image to generate a perturbation noise which makes it less suitable for real-world attack (qualitative analysis in Supplementary material). However the global perturbation network is trained with random noise as input like $\alpha_G = P(\mathcal{N})$ where $\mathcal{N} \sim \mathcal{U}(0,1) \in \mathbb{R}^{h \times w}$. This perturbation has a higher chance of attacking a system in real world as compared to image-specific perturbation, thus in this investigation we focus on global pertubation. Both image-specific and global perturbation are optimized by minimizing DFA loss (see Section \ref{loss}). The optimized global pertubations for monocular depth estimators with different $\eta$ is shown in Figure~\ref{fig:pert_adv}.

\subsection{Patch Attack} \label{patchattack}

A depth network $M$ which estimates depth $D$ from a single image $I \in \mathbb{R}^{h \times w \times 3}$ is attacked by placing an adversarial patch $\beta \in \mathbb{R}^{H \times W}$ on the image $I$ at location $\xi$ as shown in Equ.~\ref{eq4} where $H,W$ is less than 1\% of image size $h,w$. Transformations $\omega$ includes rotation and scaling are applied on the patch $\beta$ randomly to make the patch invariant to these transformations. The response of the network $M$ at attacked image $Adv(I, \beta, \omega, \xi)$ is $\tilde{D}$.

\begin{gather}
    \tilde{D} = M(Adv(I, \beta, \omega, \xi)) \label{eq4}
\end{gather}

\begin{figure}[!ht]
  \centering
\newcommand{\turnheightnew}{0.15\columnwidth}
\centering

\begin{center}

{\def\arraystretch{1}\tabcolsep=0pt
\begin{tabular}{@{}p{0.2\linewidth}@{}C{0.2\linewidth}@{}C{0.2\linewidth}@{}C{0.2\linewidth}@{}C{0.2\linewidth}@{}}

Models & \multicolumn{4}{c}{Patch size} \\
\hline

 &
{\hspace{0mm} $50 \times 50$} &
{\hspace{0mm} $60 \times 60$} &
{\hspace{0mm} $72 \times 72$} &
{\hspace{0mm} $100 \times 100$} \\

\cline{2-5} \\

{SFM~\cite{zhou2017unsupervised}} & 
\includegraphics[width=0.4\linewidth]{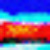} &
\includegraphics[width=0.48\linewidth]{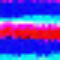} &
\includegraphics[width=0.576\linewidth]{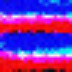} &
\includegraphics[width=0.8\linewidth]{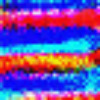} \\

{DDVO~\cite{wang2018learning}} &
\includegraphics[width=0.4\linewidth]{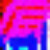} &
\includegraphics[width=0.48\linewidth]{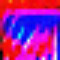} &
\includegraphics[width=0.576\linewidth]{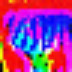} &
\includegraphics[width=0.8\linewidth]{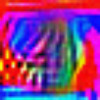} \\

{B2F~\cite{janai2018unsupervised}} &
\includegraphics[width=0.4\linewidth]{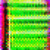} &
\includegraphics[width=0.48\linewidth]{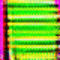} &
\includegraphics[width=0.576\linewidth]{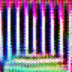} &
\includegraphics[width=0.8\linewidth]{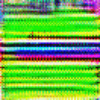} \\

{SCSFM~\cite{bian2019unsupervised}} &
\includegraphics[width=0.4\linewidth]{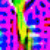} &
\includegraphics[width=0.48\linewidth]{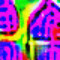} &
\includegraphics[width=0.576\linewidth]{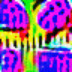} &
\includegraphics[width=0.8\linewidth]{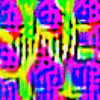} \\

{Mono1~\cite{godard2017unsupervised}} &
\includegraphics[width=0.4\linewidth]{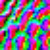} &
\includegraphics[width=0.48\linewidth]{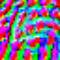} &
\includegraphics[width=0.576\linewidth]{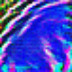} &
\includegraphics[width=0.8\linewidth]{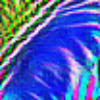} \\

{Mono2~\cite{godard2019digging}} &
\includegraphics[width=0.4\linewidth]{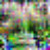} &
\includegraphics[width=0.48\linewidth]{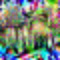} &
\includegraphics[width=0.576\linewidth]{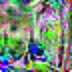} &
\includegraphics[width=0.8\linewidth]{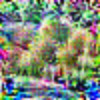} \\

\end{tabular}

}
\end{center}
  \caption{Adversarial patches of different sizes optimized to attack each models.}
  \label{fig:patch_adv}
\end{figure}

Adversarial patches are generated by optimizing random noise initialized with $\mathcal{N} \sim \mathcal{U}(0,1) \in \mathbb{R}^{H \times W}$ by a neural network. The optimization minimizes DFA (see Section~\ref{loss}) to create global patches that are apt for real-world attacks. The  square patches we optimized for monocular depth estimators are of different sizes i.e., $50 \times 50$, $60 \times 60$, $72 \times 72$ and $100 \times 100$ which are approximately 0.5\%, 0.75\%, 1\% and 2\% of the original KITTI image size $1242 \times 375$ as shown in Figure~\ref{fig:patch_adv}.

\subsection{Deep Feature Annihilation loss} \label{loss}

As compared to networks for tasks like classification and detection, regression networks are much more difficult to attack, especially depth estimation networks. To deal with this issue, we introduce a novel DFA loss $L_{dfa}$, which attacks the internal feature representation of the input image. Note that we are optimizing the following objective after every convolution layer sparing few initial encoder layers.

\begin{gather}
F_l = F_l^{att} \times F_l^{org} \\
L_{dfa} = \sum_{l}^{} \log [1+ \mathbb{E}(F_l^2) - \mathbb{E}(F_l)^2) ]  W_l
\end{gather}

The aim is to minimize the variance of correlation between the features maps $F_l^{org}$ and $F_l^{att}$ where $F_l^{org}, F_l^{att}$ are the activations obtained from layer $l$ in the depth network $M$ after passing the unattacked image and attacked image through the network, respectively weighted with an empirically found $W_l$ . This objective forces non-dominant activation values to increase while forcing the dominant activation values to decrease, resulting in the reduction of variance as shown in Figure~\ref{fig:feat_lvis_sfm}. Variance reduction ends up costing network much semantic information, which makes the decoder incompetent to propagate useful information to output proper depth map. We have experimented by limiting the loss to just a variance reduction of $F_l^{att}$; however, our objective of minimizing the variance of correlations works much better in practice.

\begin{figure*}[ht]
  \centering
\newcommand{\turnheightnew}{0.15\columnwidth}
\centering

\begin{center}

\begin{tabular}{@{\hskip 0.5mm}c@{\hskip 0.5mm}c@{\hskip 0.5mm}c@{\hskip 0.5mm}c@{\hskip 0.5mm}c@{}}

& Attacked image & Clean depth & Attacked depth & Depth gap \\

\rotatebox[origin=l]{45}{SFM} & 
\includegraphics[width=0.22\linewidth]{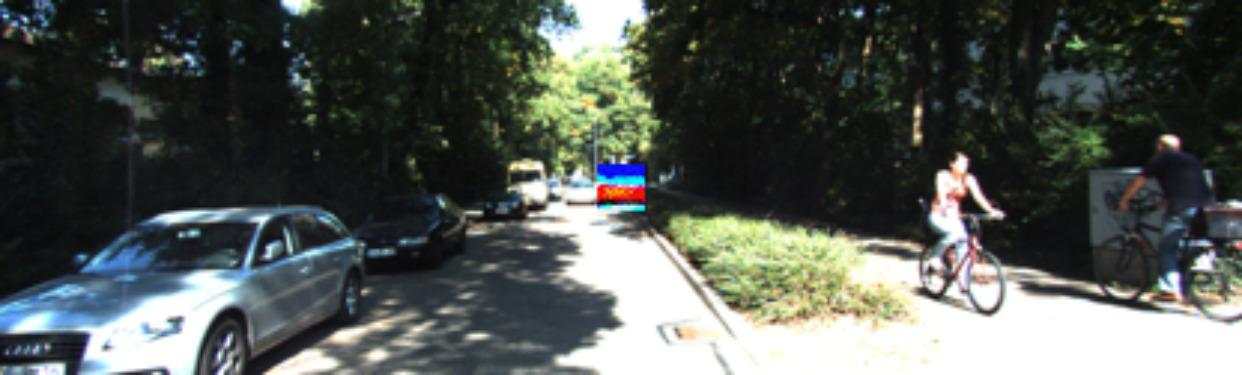} &
\includegraphics[width=0.22\linewidth]{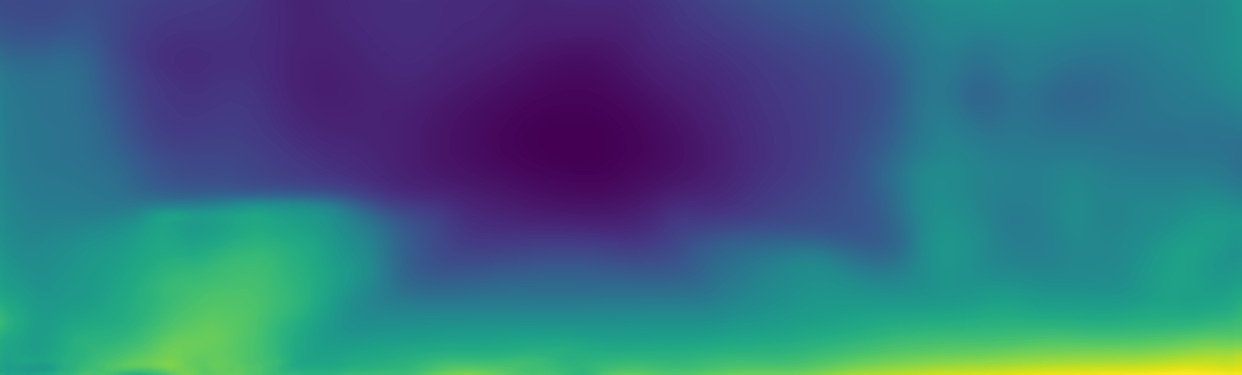} &
\includegraphics[width=0.22\linewidth]{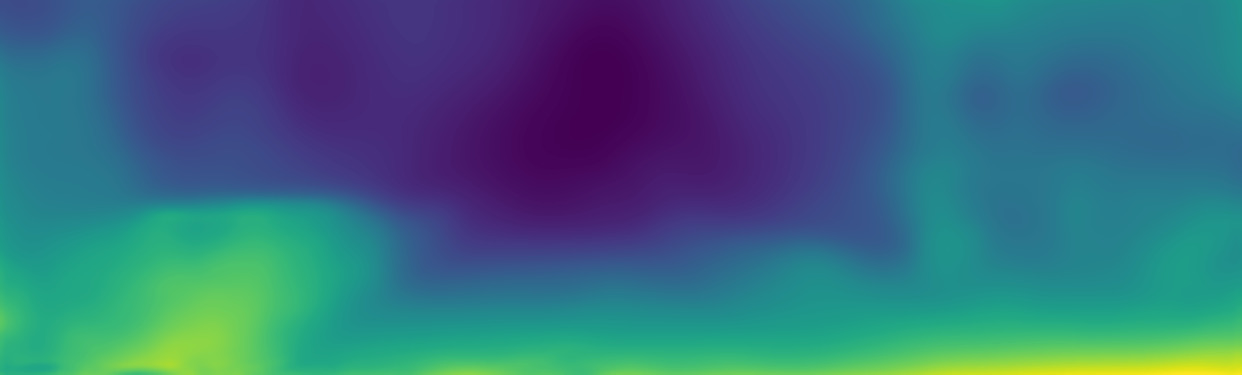} &
\includegraphics[width=0.22\linewidth]{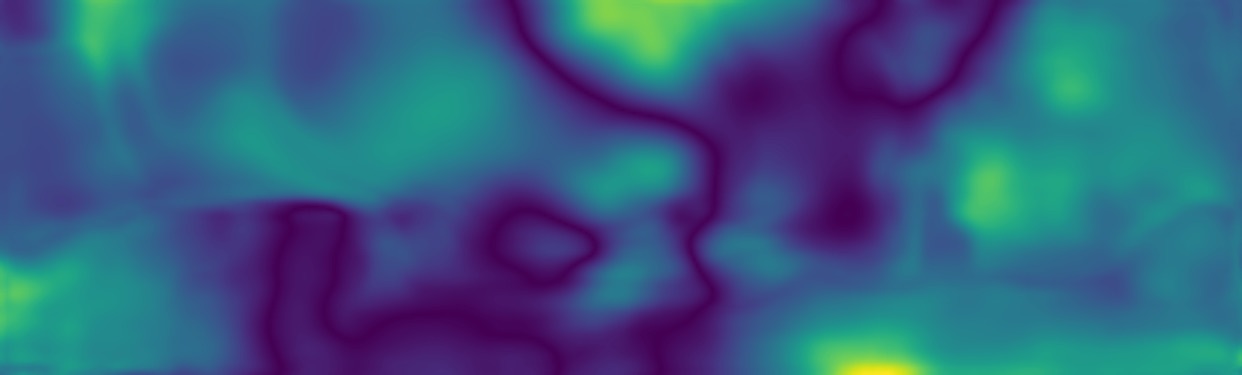} \\

\rotatebox[origin=l]{45}{DDVO} & 
\includegraphics[width=0.22\linewidth]{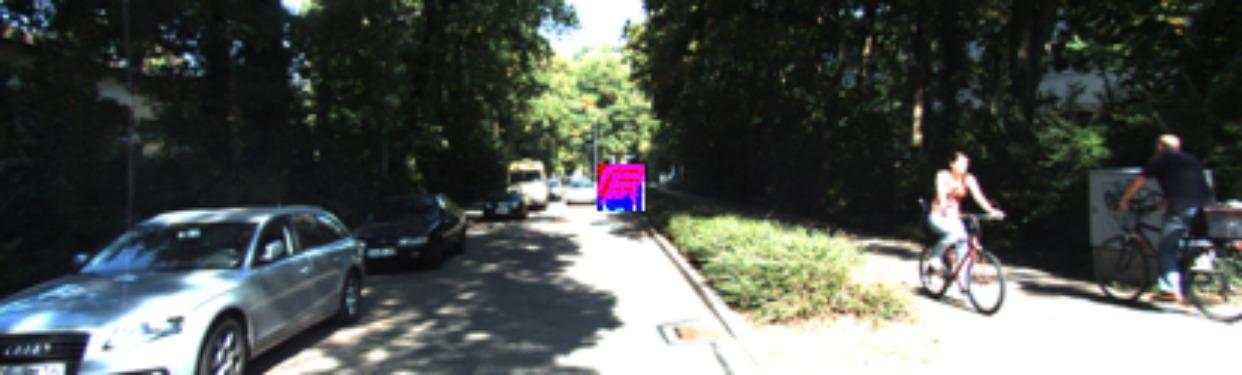} &
\includegraphics[width=0.22\linewidth]{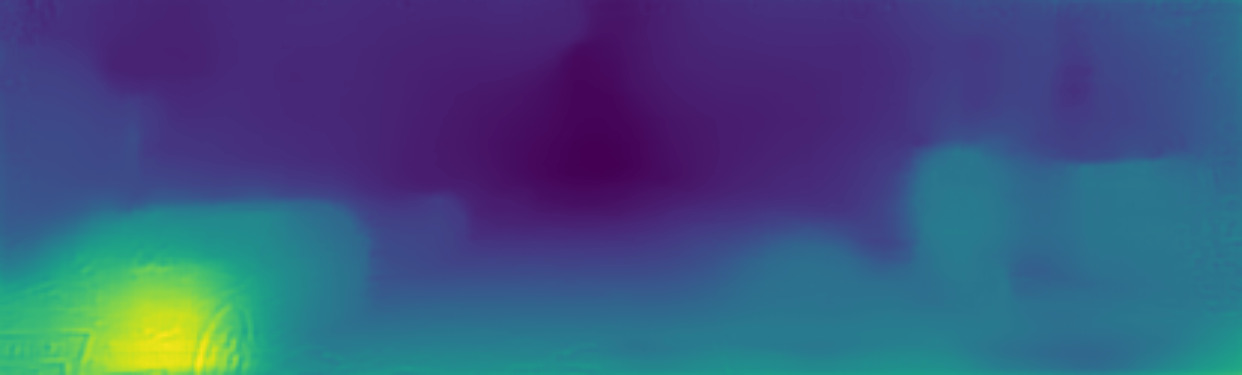} &
\includegraphics[width=0.22\linewidth]{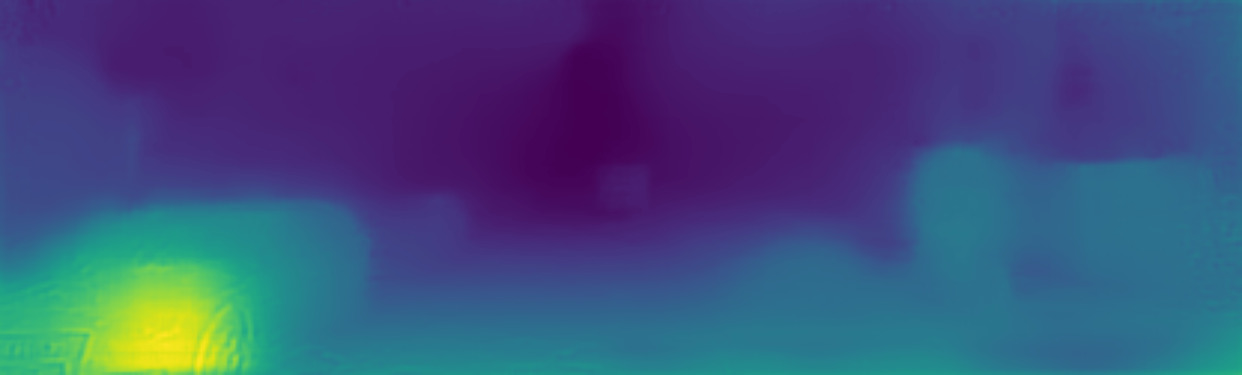} &
\includegraphics[width=0.22\linewidth]{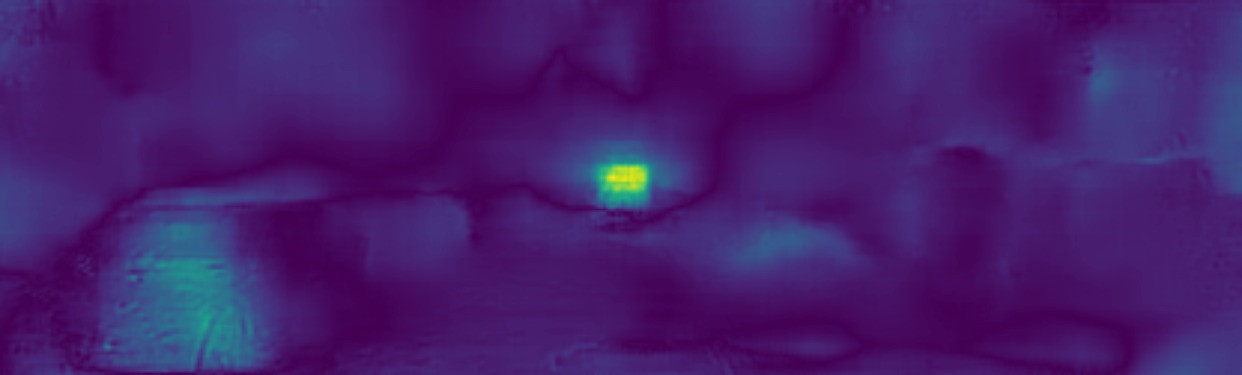} \\

\rotatebox[origin=l]{45}{B2F} & 
\includegraphics[width=0.22\linewidth]{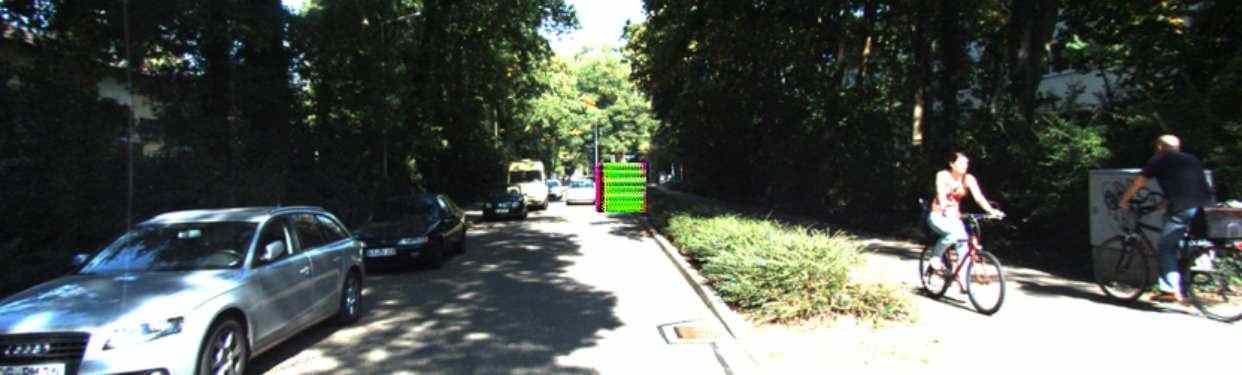} &
\includegraphics[width=0.22\linewidth]{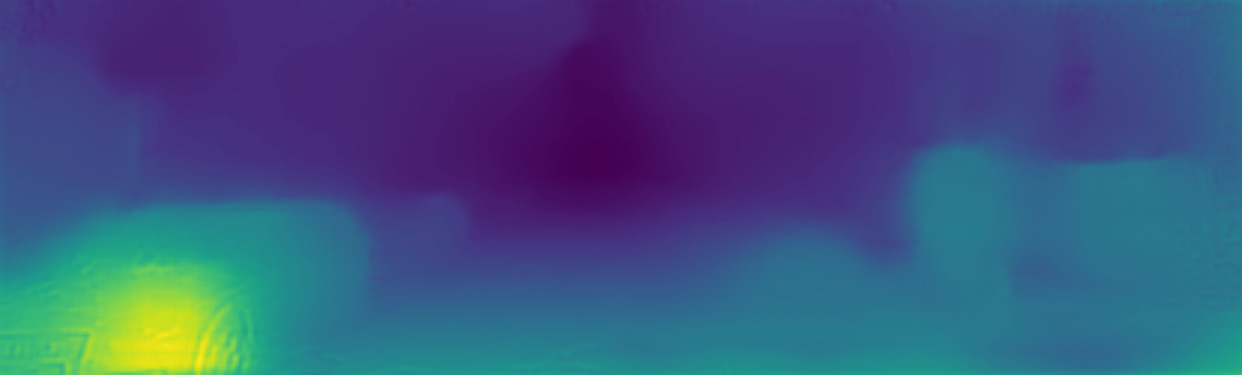} &
\includegraphics[width=0.22\linewidth]{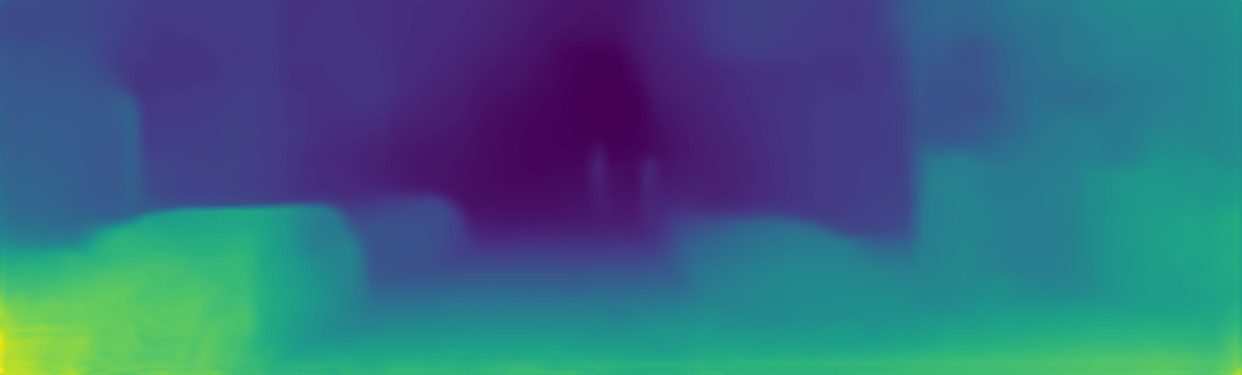} &
\includegraphics[width=0.22\linewidth]{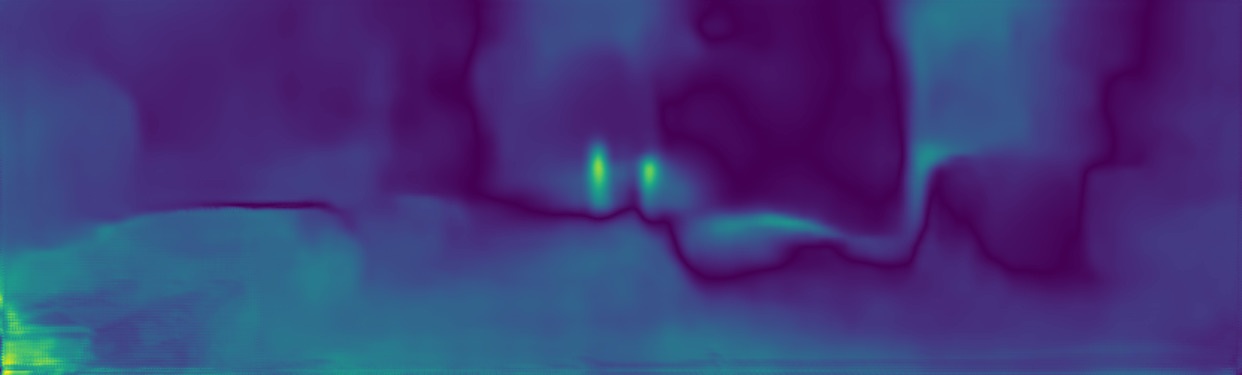} \\

\rotatebox[origin=l]{45}{SCSFM} & 
\includegraphics[width=0.22\linewidth]{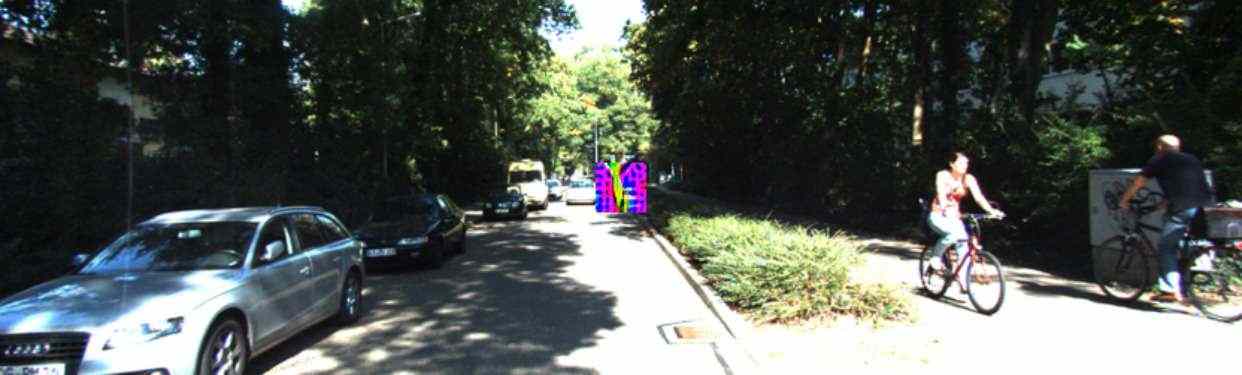} &
\includegraphics[width=0.22\linewidth]{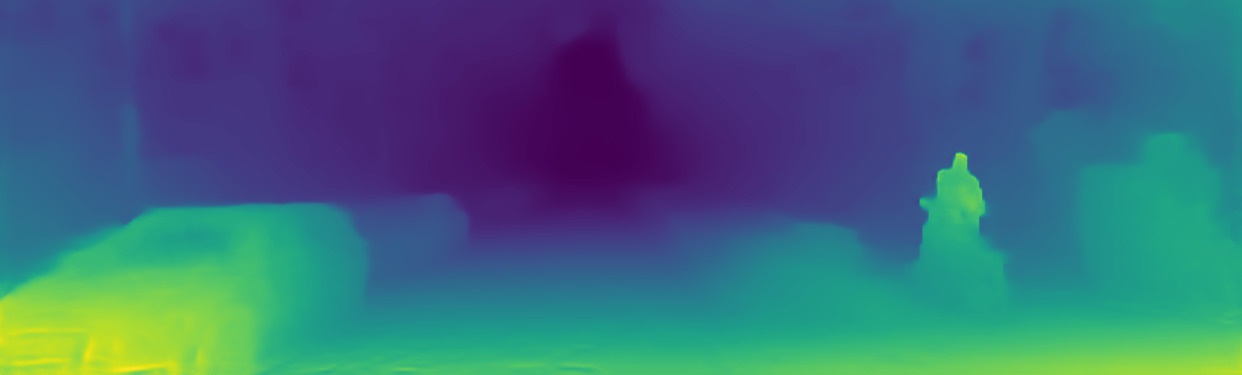} &
\includegraphics[width=0.22\linewidth]{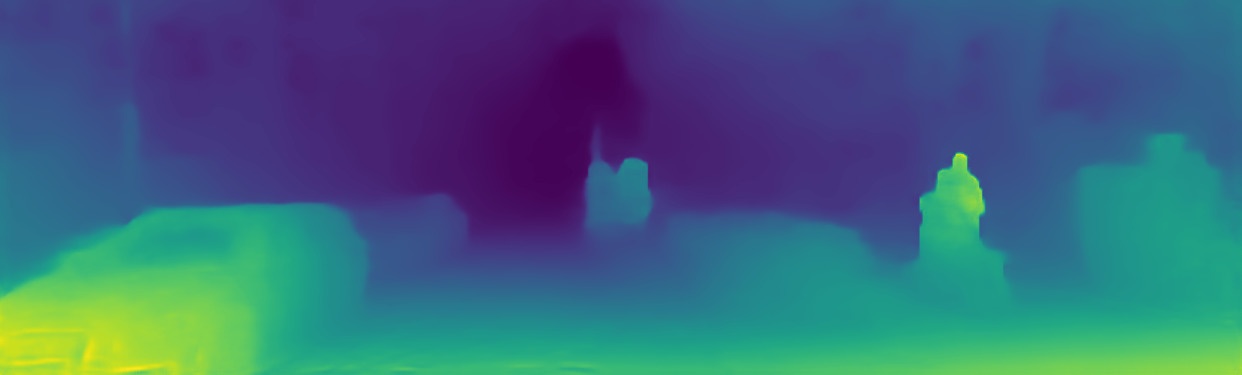} &
\includegraphics[width=0.22\linewidth]{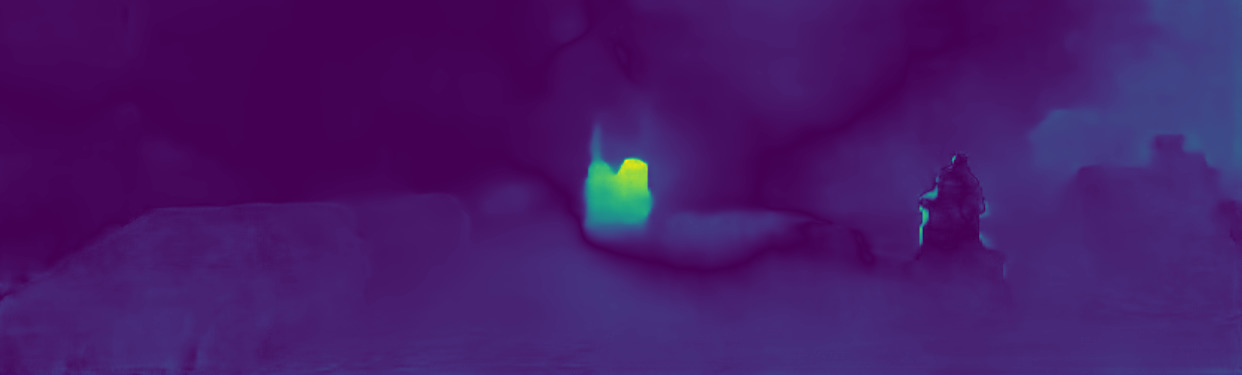} \\

\rotatebox[origin=l]{45}{Mono1} & 
\includegraphics[width=0.22\linewidth]{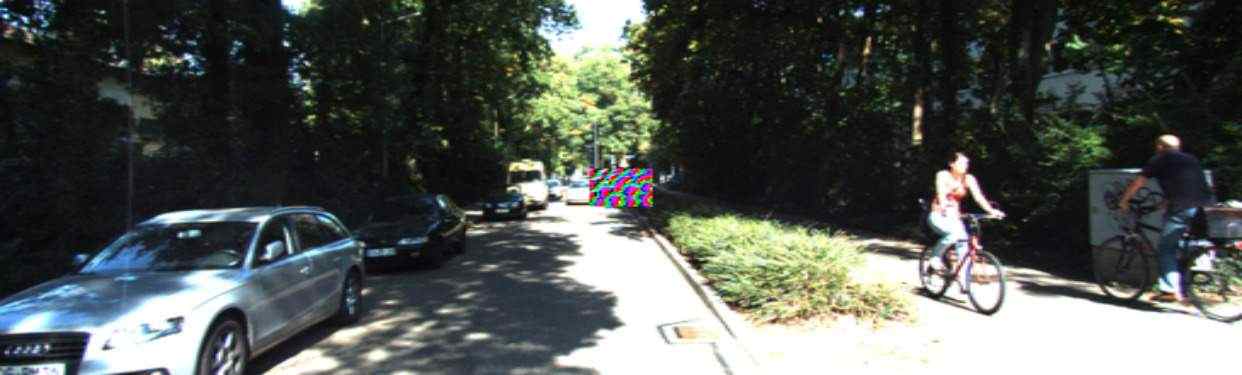} &
\includegraphics[width=0.22\linewidth]{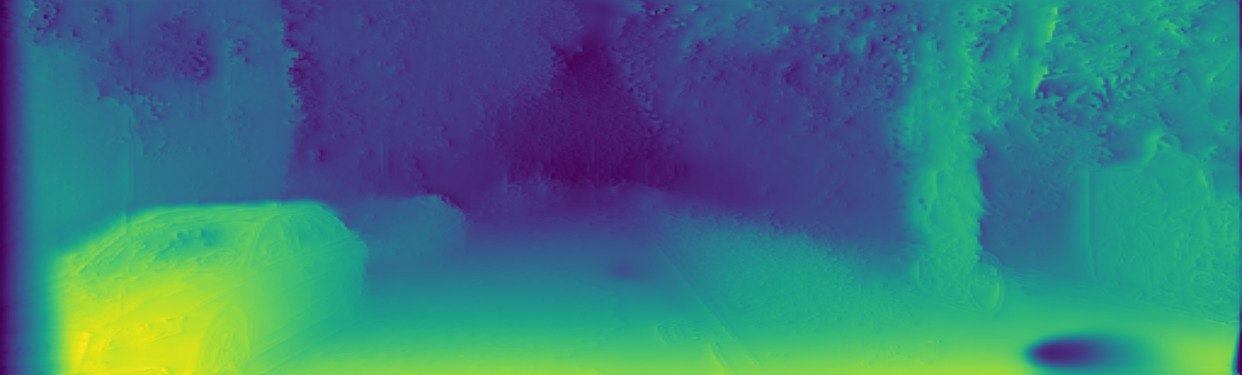} &
\includegraphics[width=0.22\linewidth]{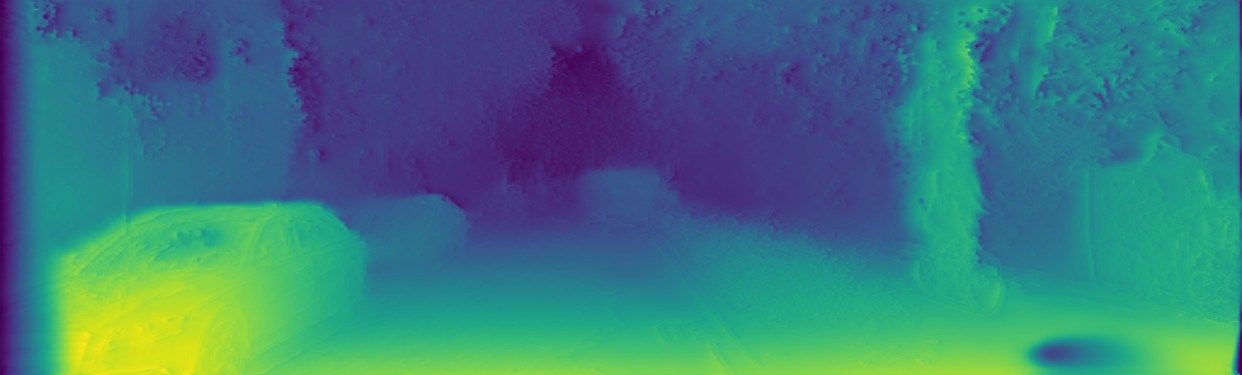} &
\includegraphics[width=0.22\linewidth]{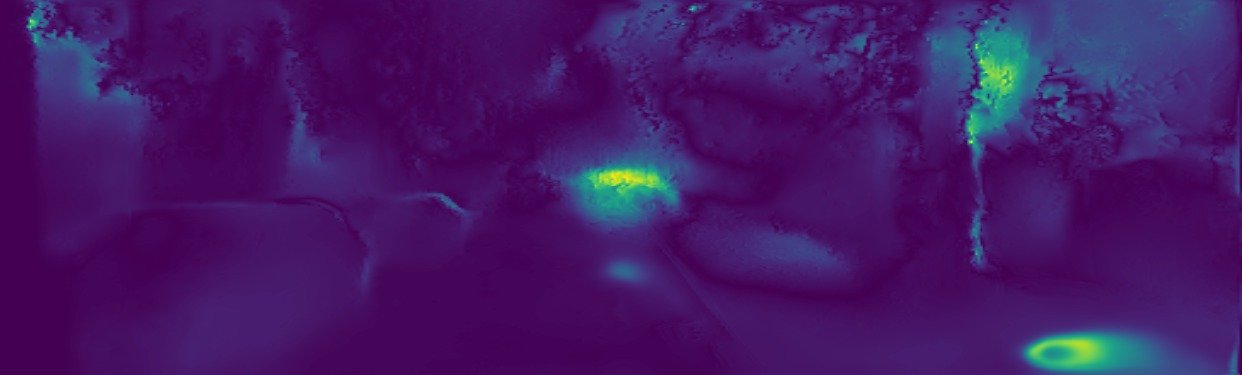} \\

\rotatebox[origin=l]{45}{Mono2} & 
\includegraphics[width=0.22\linewidth]{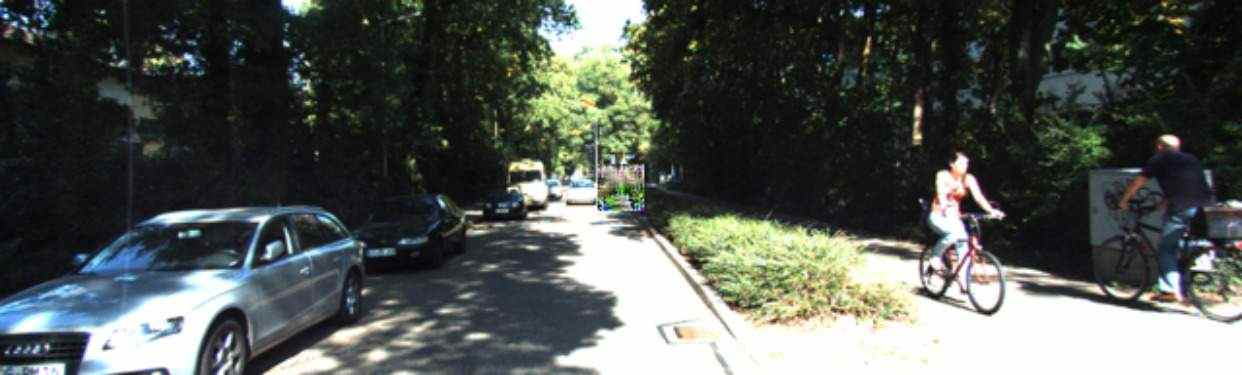} &
\includegraphics[width=0.22\linewidth]{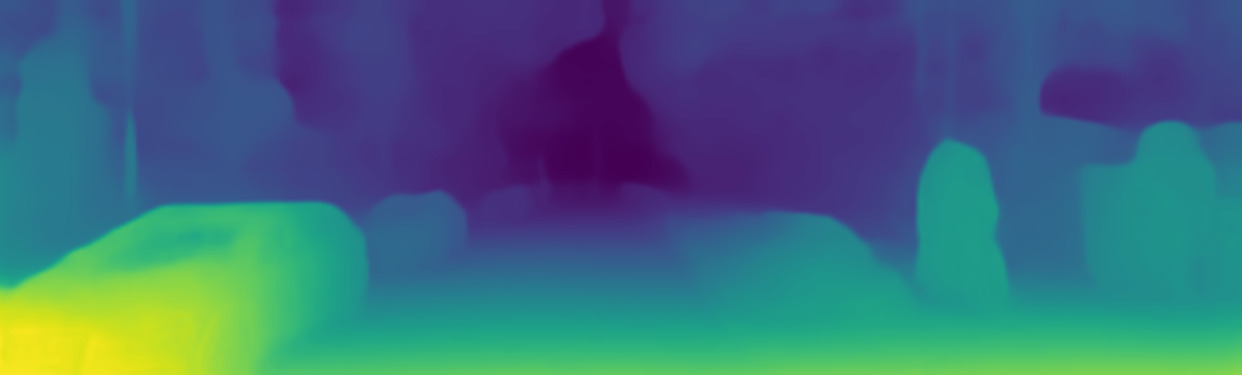} &
\includegraphics[width=0.22\linewidth]{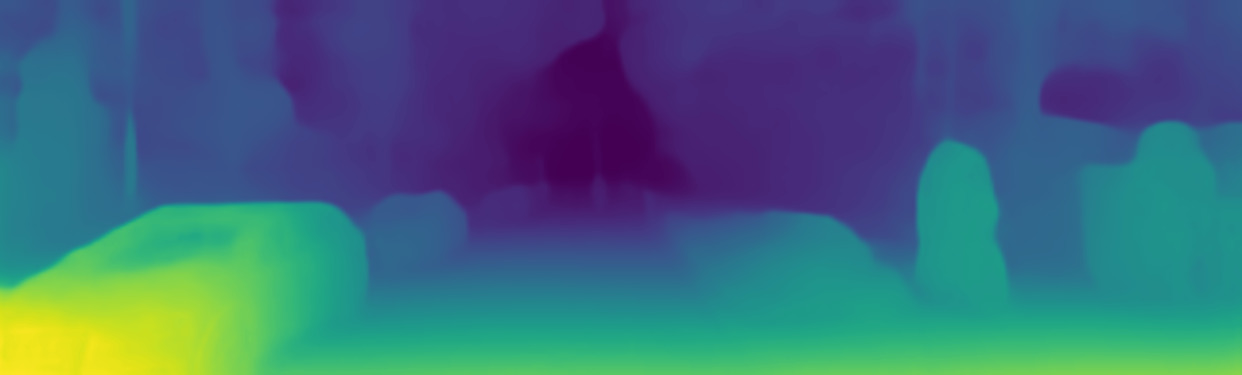} &
\includegraphics[width=0.22\linewidth]{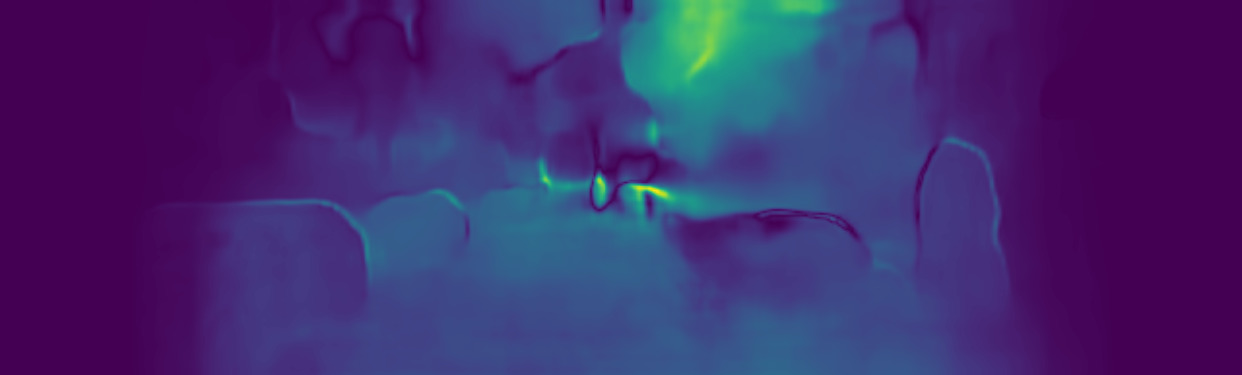} \\

\end{tabular}
\end{center}
\caption{White-box patch test with patch size $50 \times 50$.} 
\label{fig:white_patch_adv50}
\end{figure*}
\begin{table*}[!ht]
	\centering
	\begin{tabular}{l|c|c|c|c|c|c|c|c|c}
		\hline
		\multirow{2}{*}{Methods} & \multirow{2}{*}{Clean} & \multicolumn{8}{c}{Attacked} \\
		\cline{3-10}
		&    & \multicolumn{2}{c|}{$50 \times 50$} & \multicolumn{2}{c|}{$60 \times 60$} & \multicolumn{2}{c|}{$72 \times 72$} & \multicolumn{2}{c}{$100 \times 100$}  \\
		\cline{3-10}
		& RMSE   & RMSE & Rel (\%)             & RMSE & Rel (\%)             & RMSE & Rel (\%)             & RMSE & Rel (\%)  \\
		\hline
		SFM~\cite{zhou2017unsupervised}                       & 6.1711                 & 6.6232 & 8               & 7.844 & 28                & 8.8061 & 43               & 9.2719 & 51    \\
		DDVO~\cite{wang2018learning}                     & 5.5072                 & 5.963 & 9                & 6.3477 & 16               & 6.9496 & 27               & 7.5986 & 45    \\
		B2F~\cite{janai2018unsupervised}                      & 5.1615                 & 5.5533  & 8              & 5.7199 & 11               & 6.1145 & 18                & 6.6022 & 28   \\
		SCSFM~\cite{bian2019unsupervised}                    & 5.2271                 & 6.4036 & 23              & 6.7562 & 30               & 7.1606 & 37               & 8.0283 & 54    \\
		Mono1~\cite{godard2017unsupervised}                    & 5.1973                 & 5.7658 & 11              & 6.0623 & 17               & 6.2326 & 20               & 7.1151 & 37    \\
		Mono2~\cite{godard2019digging}                    & 4.9099                 & 5.3133 & 9               & 5.4127 & 11               & 5.5343 & 13               & 6.1714 & 26    \\
		\hline
	\end{tabular}
	\caption{White-box patch attack at different patch sizes. Root mean square error (RMSE) of each model with and without adversarial attack and relative degradation (Rel) of RMSE on KITTI 2015.}
	\label{tab:whitebox_patch_rmse}
\end{table*}

\begin{figure*}[!ht]
\centering
\newcommand{\turnheightnew}{0.15\columnwidth}
\centering

\begin{center}

\begin{tabular}{@{\hskip 0.5mm}c@{\hskip 0.5mm}c@{\hskip 0.5mm}c@{\hskip 0.5mm}c@{\hskip 0.5mm}c@{}}

& Attacked image & Clean depth & Attacked depth & Depth gap \\

\rotatebox[origin=l]{45}{SFM} & 
\includegraphics[width=0.22\linewidth]{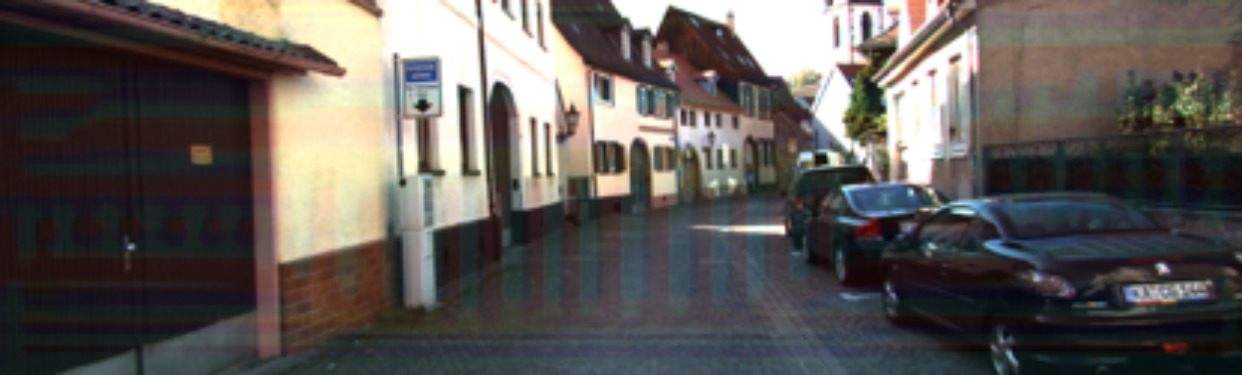} &
\includegraphics[width=0.22\linewidth]{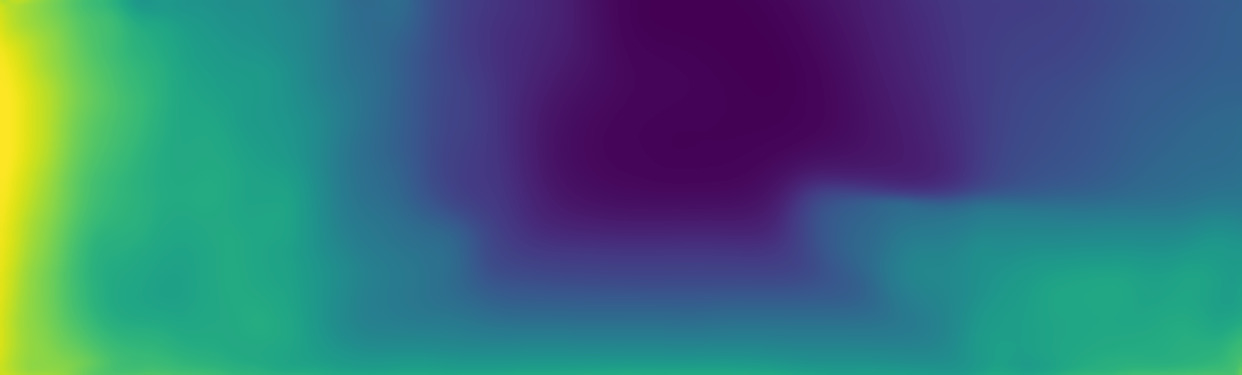} &
\includegraphics[width=0.22\linewidth]{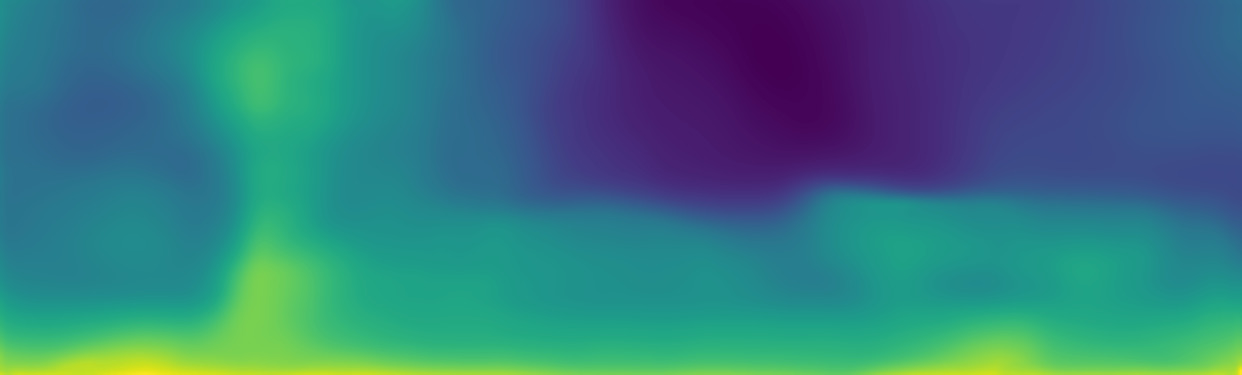} &
\includegraphics[width=0.22\linewidth]{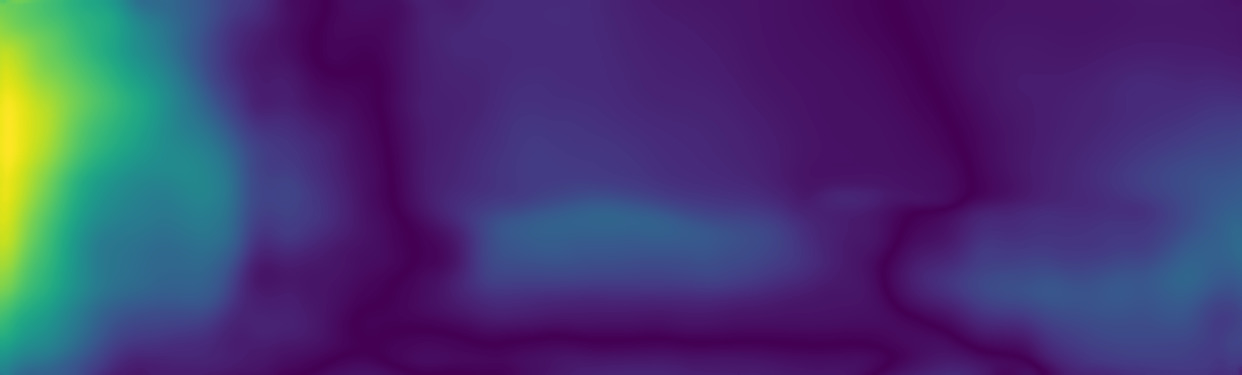} \\

\rotatebox[origin=l]{45}{DDVO} & 
\includegraphics[width=0.22\linewidth]{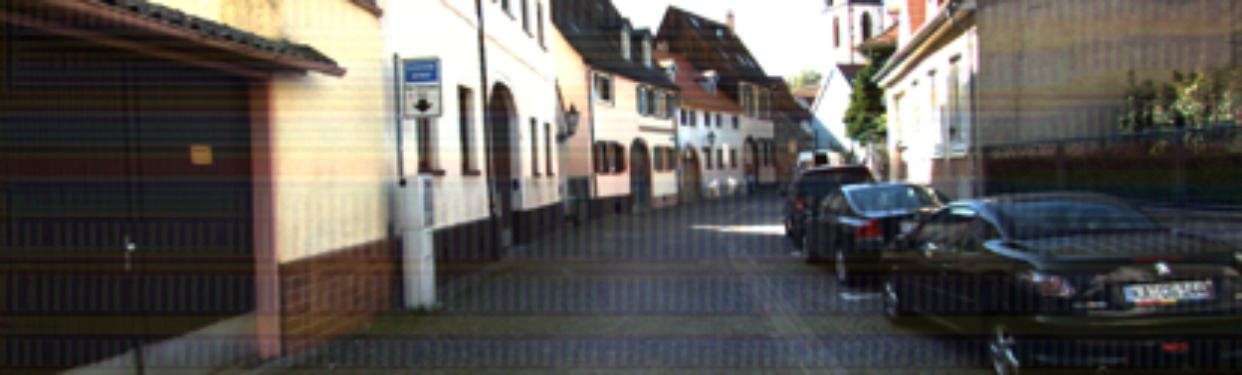} &
\includegraphics[width=0.22\linewidth]{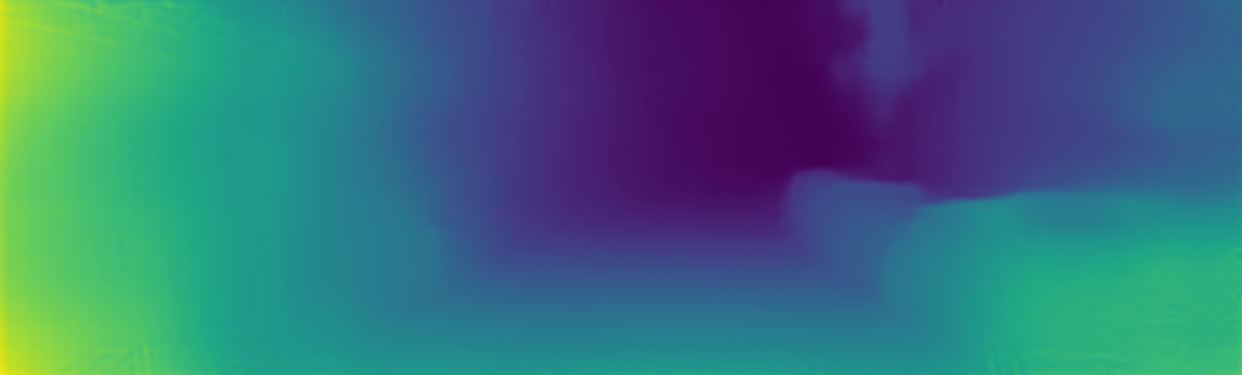} &
\includegraphics[width=0.22\linewidth]{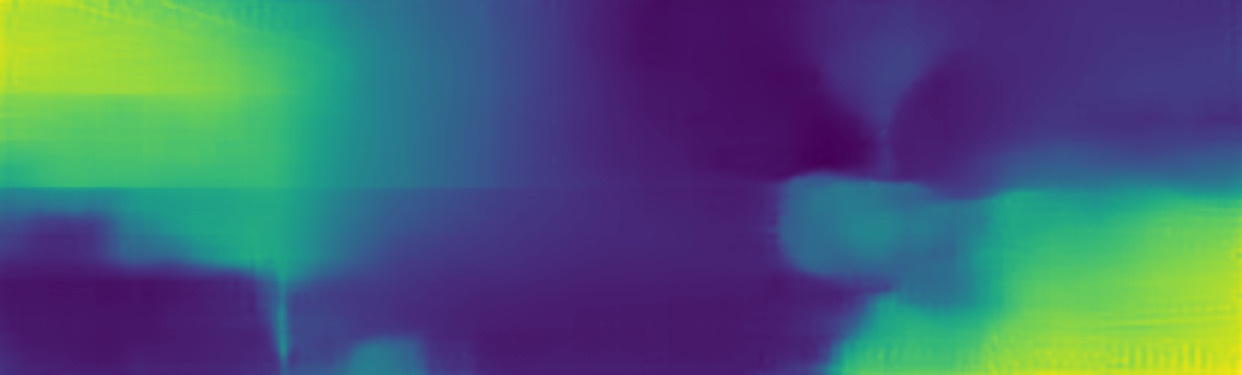} &
\includegraphics[width=0.22\linewidth]{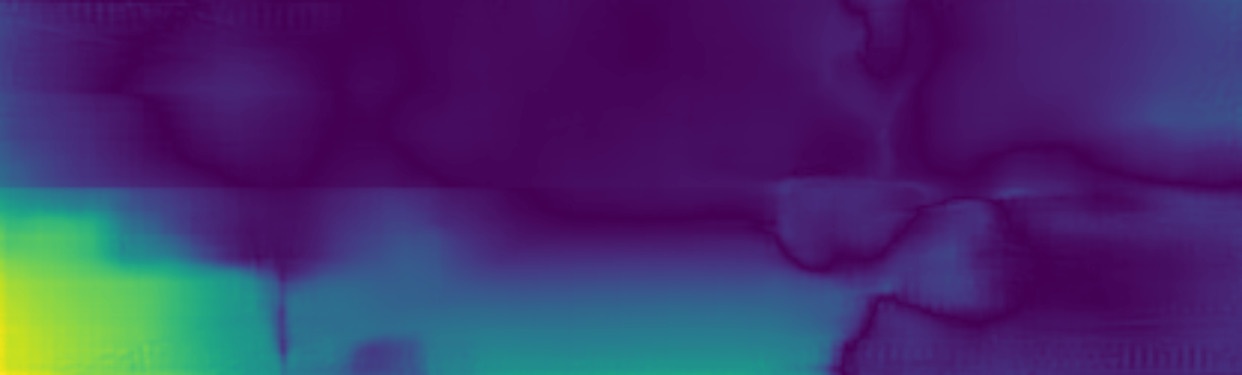} \\

\rotatebox[origin=l]{45}{B2F} & 
\includegraphics[width=0.22\linewidth]{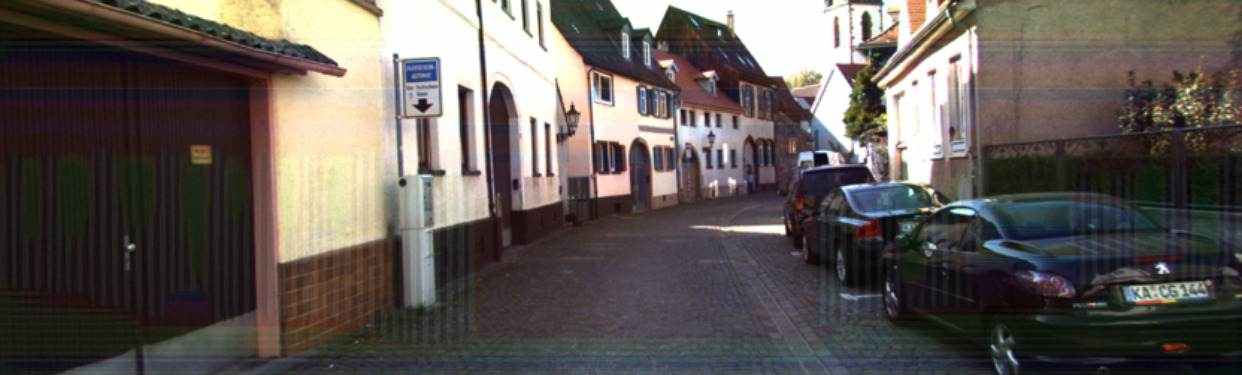} &
\includegraphics[width=0.22\linewidth]{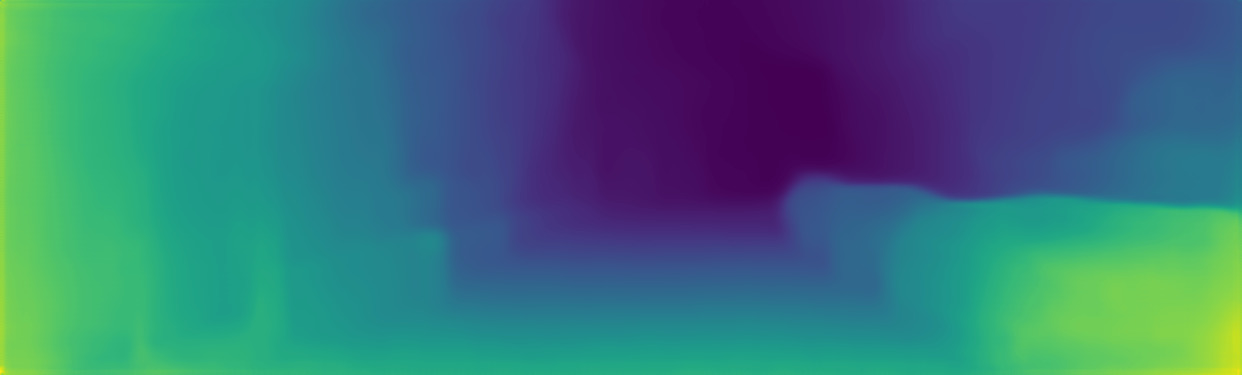} &
\includegraphics[width=0.22\linewidth]{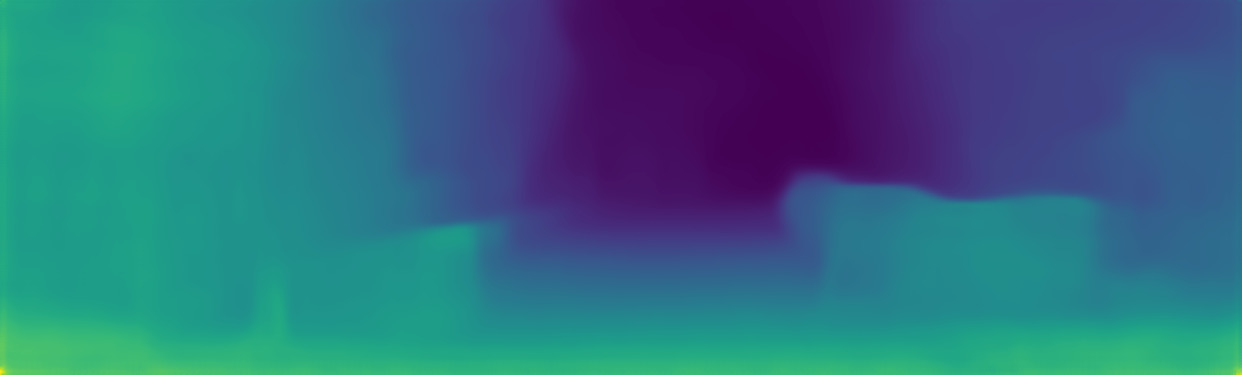} &
\includegraphics[width=0.22\linewidth]{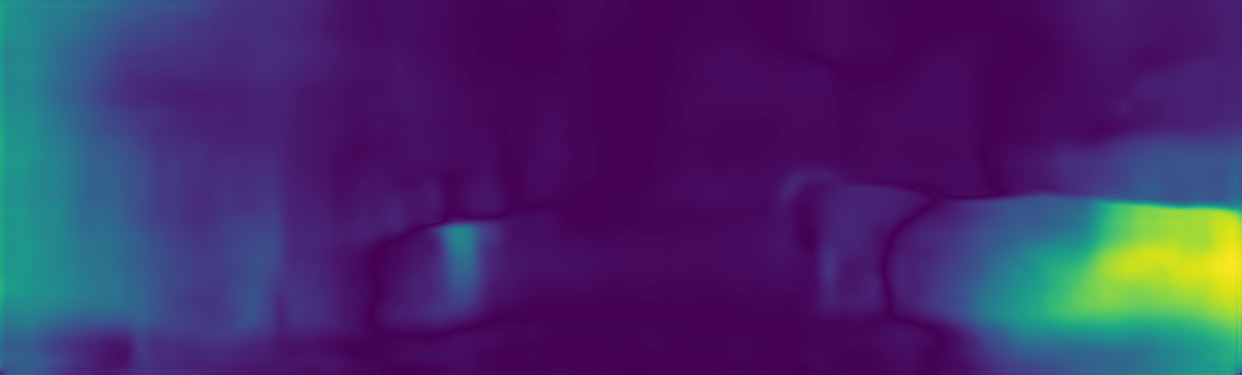} \\

\rotatebox[origin=l]{45}{SCSFM} & 
\includegraphics[width=0.22\linewidth]{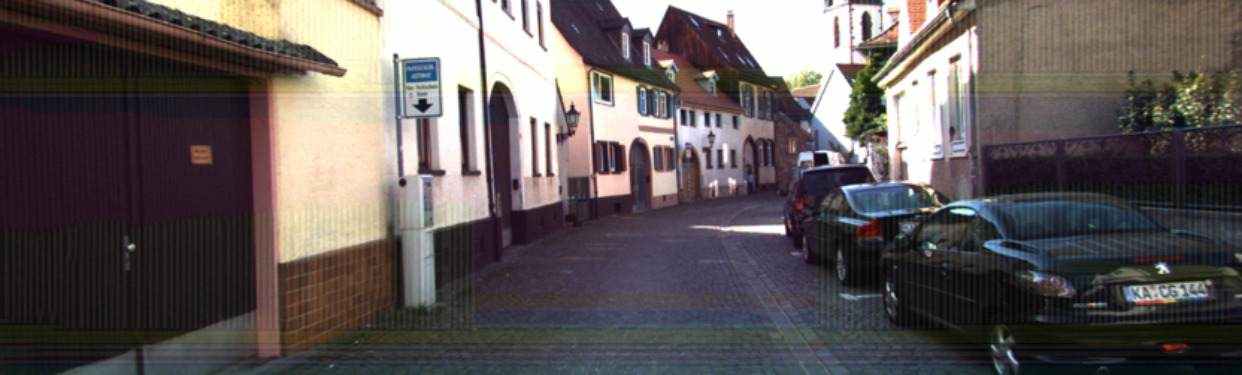} &
\includegraphics[width=0.22\linewidth]{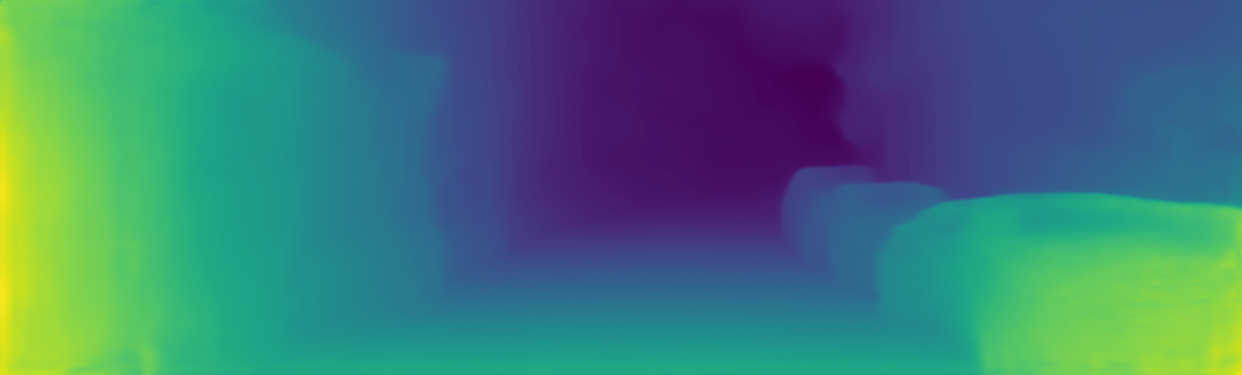} &
\includegraphics[width=0.22\linewidth]{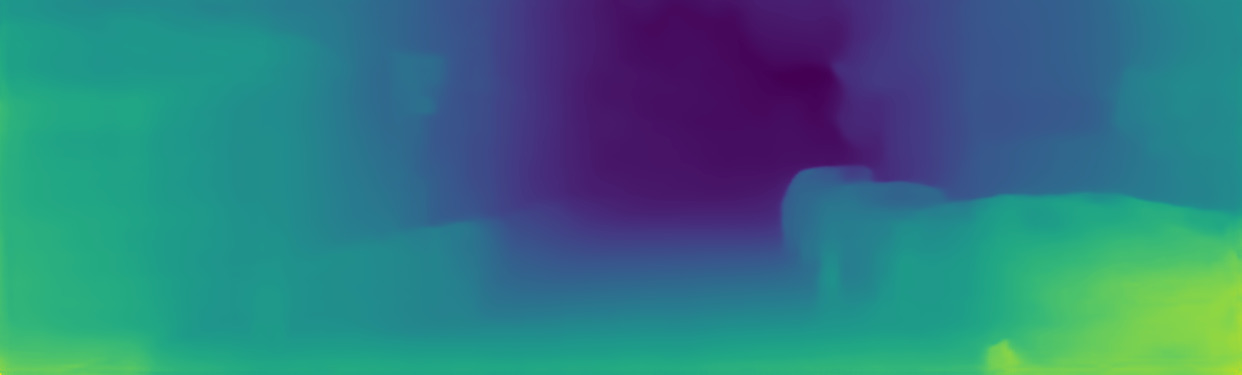} &
\includegraphics[width=0.22\linewidth]{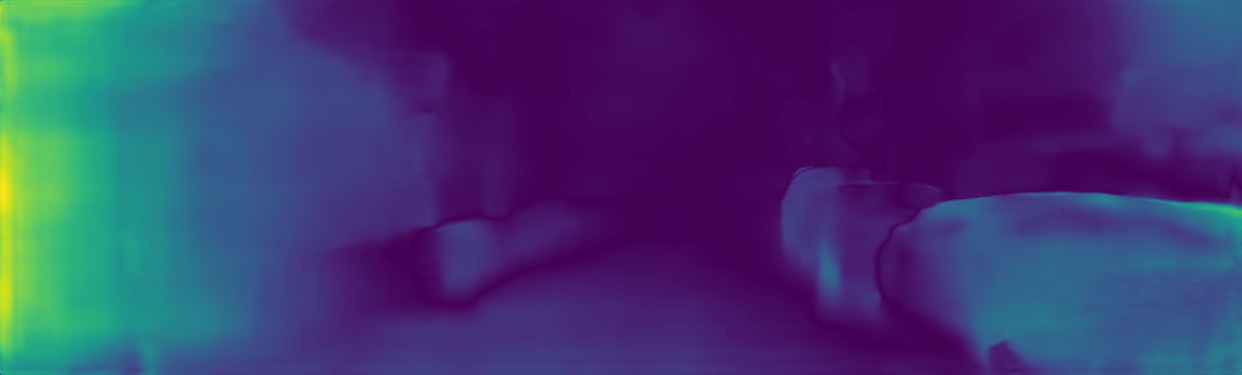} \\

\rotatebox[origin=l]{45}{Mono1} & 
\includegraphics[width=0.22\linewidth]{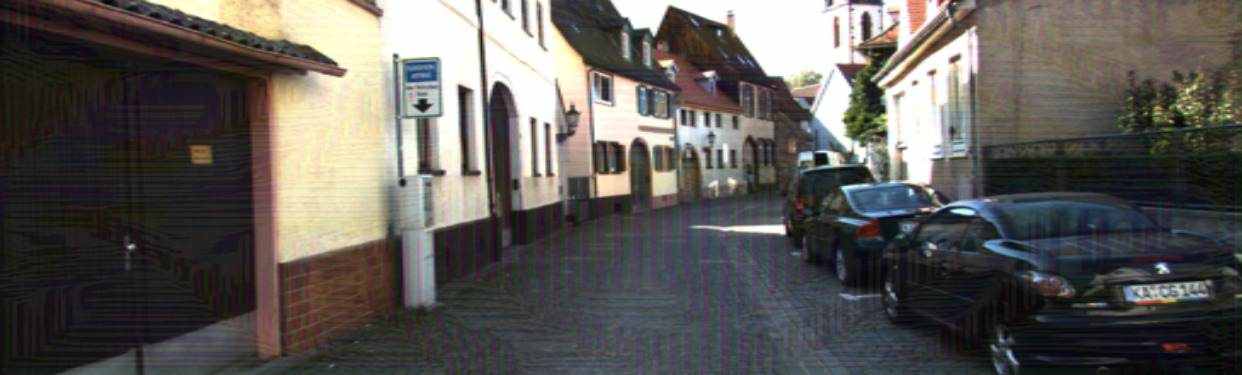} &
\includegraphics[width=0.22\linewidth]{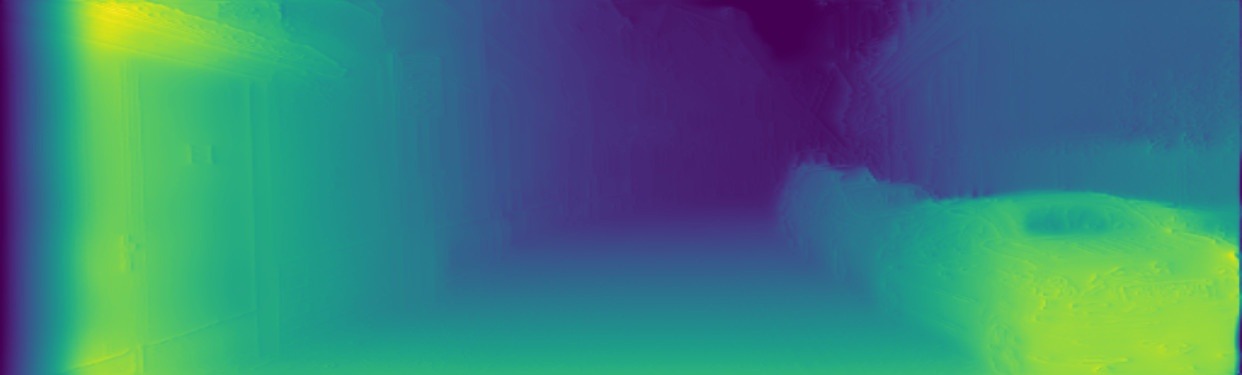} &
\includegraphics[width=0.22\linewidth]{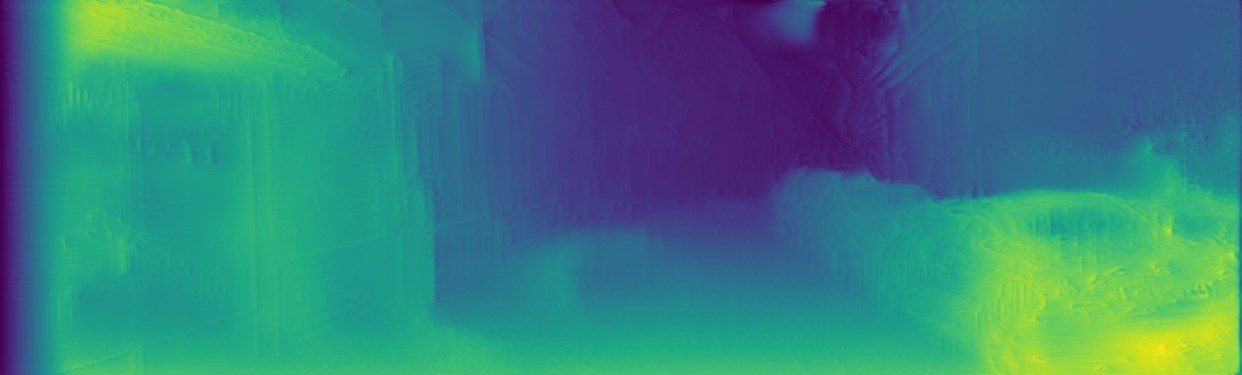} &
\includegraphics[width=0.22\linewidth]{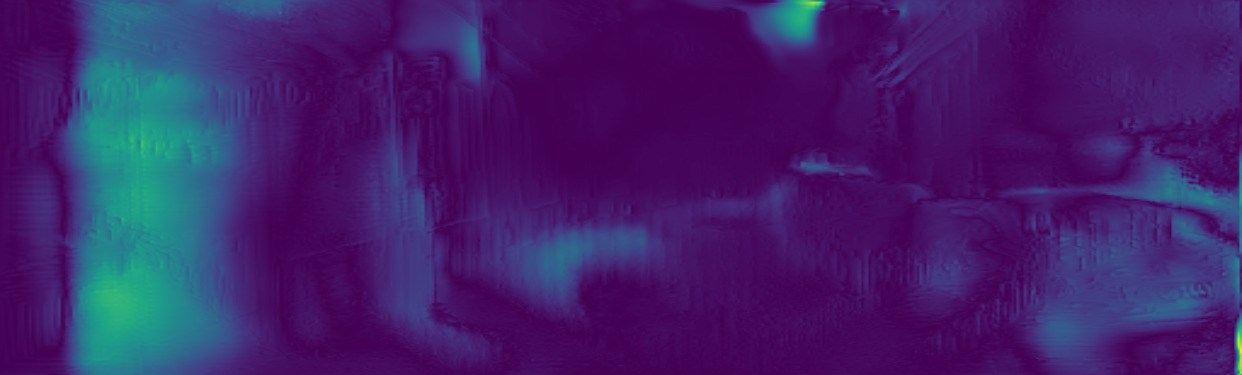} \\

\rotatebox[origin=l]{45}{Mono2} & 
\includegraphics[width=0.22\linewidth]{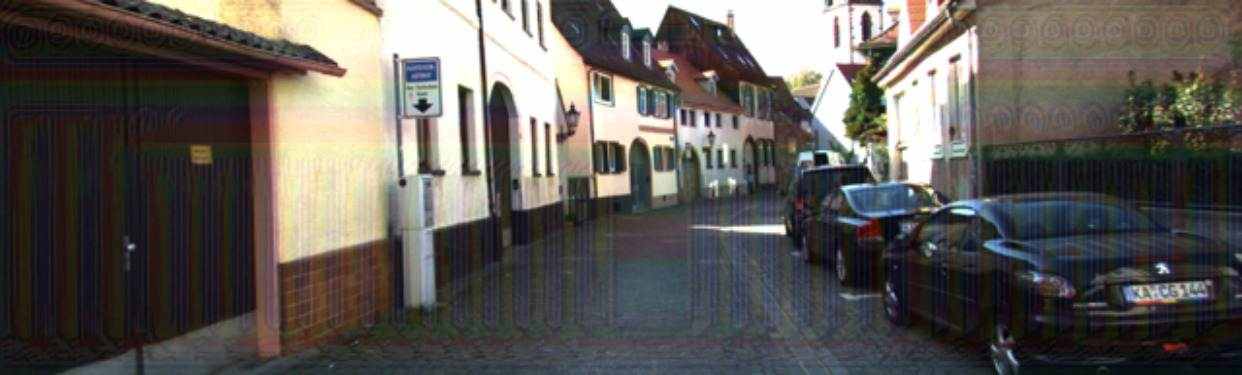} &
\includegraphics[width=0.22\linewidth]{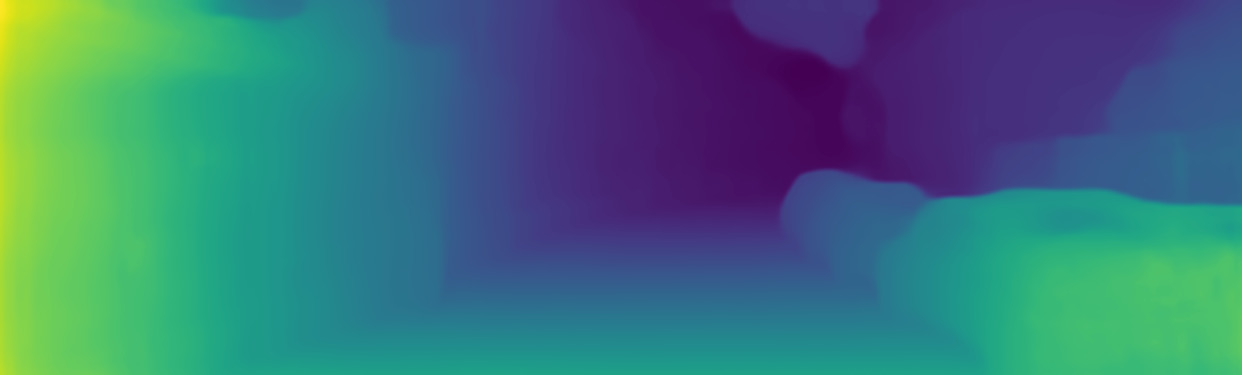} &
\includegraphics[width=0.22\linewidth]{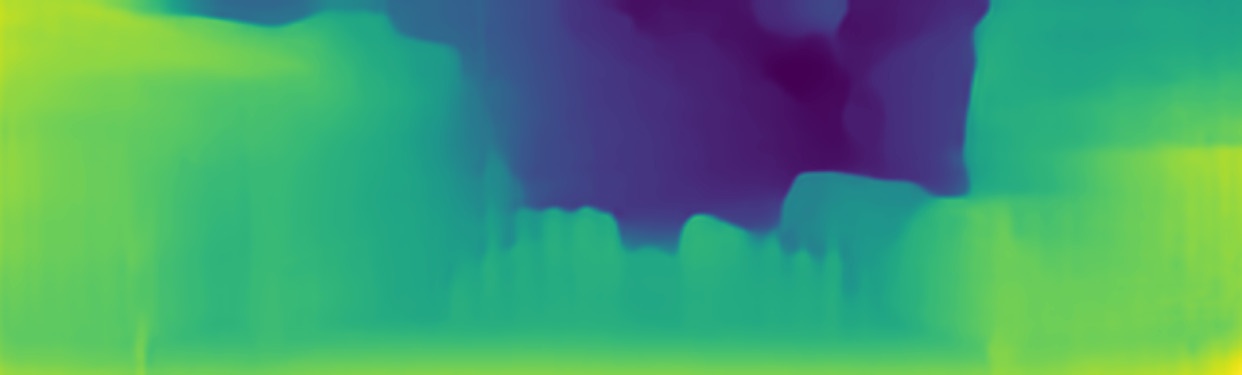} &
\includegraphics[width=0.22\linewidth]{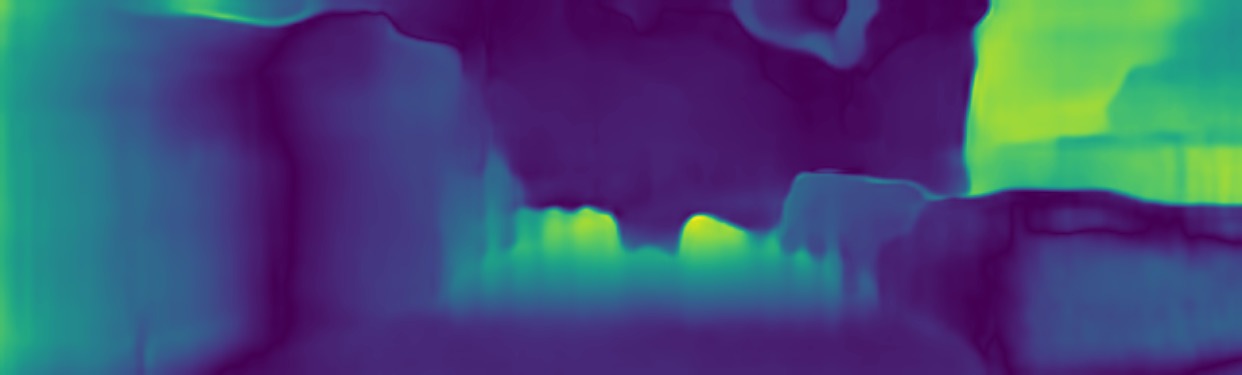} \\

\end{tabular}
\end{center}

\caption{White-box perturbation test when $\eta = 0.05$.} 
\label{fig:white_pert_adv5}
\end{figure*}
\begin{table*}
	\centering
	\begin{tabular}{l|c|c|c|c|c|c|c}
		\hline
		\multirow{2}{*}{Methods} & \multirow{2}{*}{Clean} & \multicolumn{6}{c}{Attacked} \\
		\cline{3-8}
		&    & \multicolumn{2}{c|}{$\eta=0.01$} & \multicolumn{2}{c|}{$\eta=0.05$} & \multicolumn{2}{c}{$\eta=0.1$} \\
		\cline{3-8}
		& RMSE   & RMSE & Rel (\%)             & RMSE & Rel  (\%)            & RMSE & Rel (\%) \\
		\hline
		SFM~\cite{zhou2017unsupervised}                      & 6.1711                 & 6.2619 & 2              & 8.2988 & 35        & 10.1711 & 75        \\
		DDVO~\cite{wang2018learning}                     & 5.5072                 & 5.6649 & 3              & 10.9875 & 100      & 18.3904 & 234       \\
		B2F~\cite{janai2018unsupervised}                      & 5.1615                 & 5.2876 & 3              & 6.7341 & 31        & 10.2199 & 99        \\
		SCSFM~\cite{bian2019unsupervised}                    & 5.2271                 & 5.277  & 1              & 6.5294 & 25        & 9.8324 & 89         \\
		Mono1~\cite{godard2017unsupervised}                    & 5.1973                 & 5.2217 & 1              & 6.8002 & 31        & 12.0708 & 133       \\
		Mono2~\cite{godard2019digging}                    & 4.9099                 & 4.9984 & 2              & 9.8473 & 101       & 13.5497 & 176       \\
		\hline
	\end{tabular}
	\caption{White-box perturbation attack at different pertubation contrains $\eta$.}
	\label{tab:whitebox_gpert_rmse}
\end{table*}

\begin{figure*}[!ht]
\centering
\newcommand{\turnheightnew}{0.15\columnwidth}
\centering

\begin{center}

\begin{tabular}{@{\hskip 0.5mm}c@{\hskip 0.5mm}c@{\hskip 0.5mm}c@{\hskip 0.5mm}c@{}}

 & Clean image & Attacked image &  \\

 & 
\includegraphics[width=0.22\linewidth]{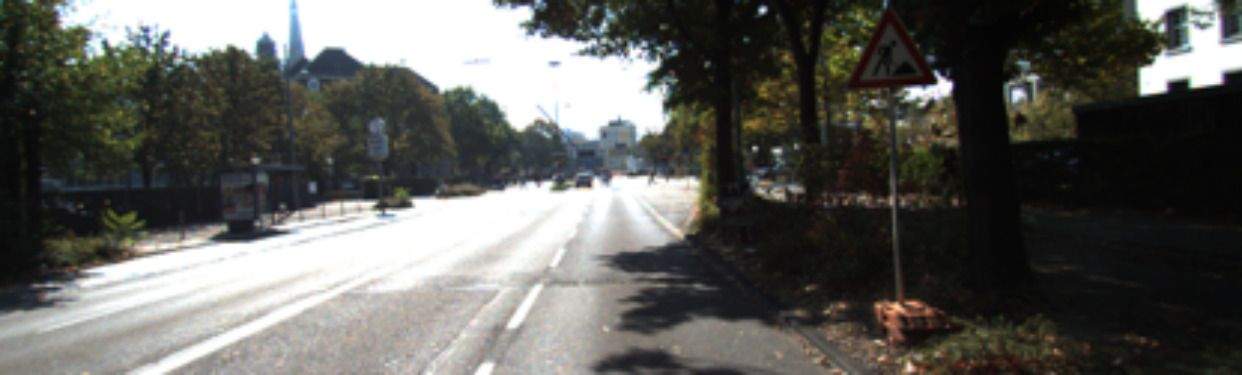} &
\includegraphics[width=0.22\linewidth]{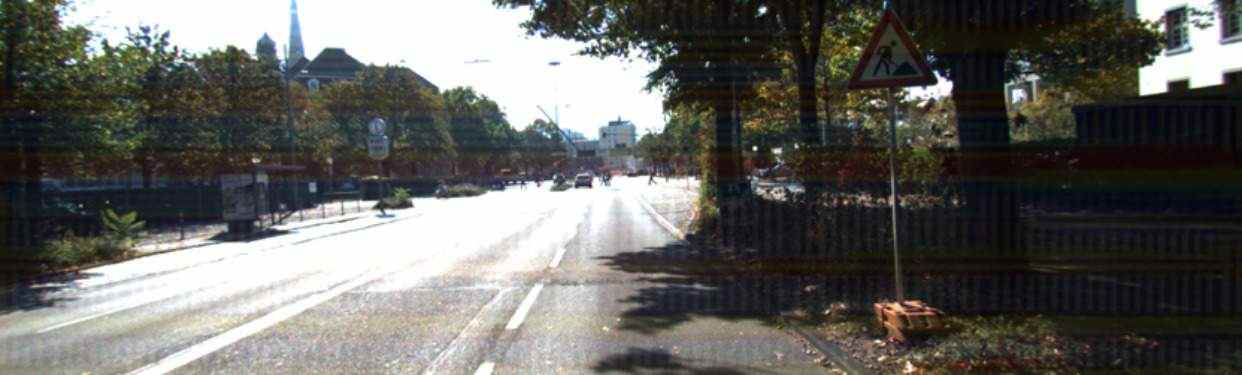} &  \\

 & Clean depth & Attacked depth & Depth gap \\

\rotatebox[origin=l]{45}{SFM} & 
\includegraphics[width=0.22\linewidth]{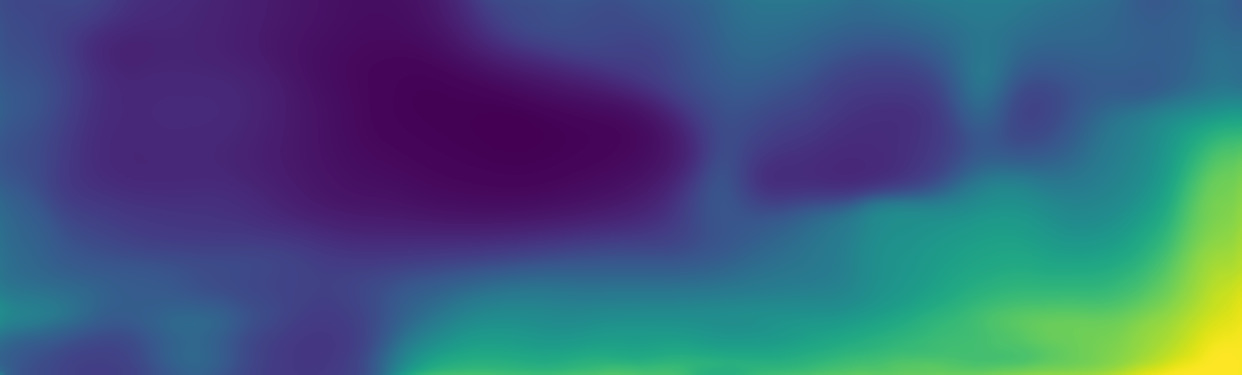} &
\includegraphics[width=0.22\linewidth]{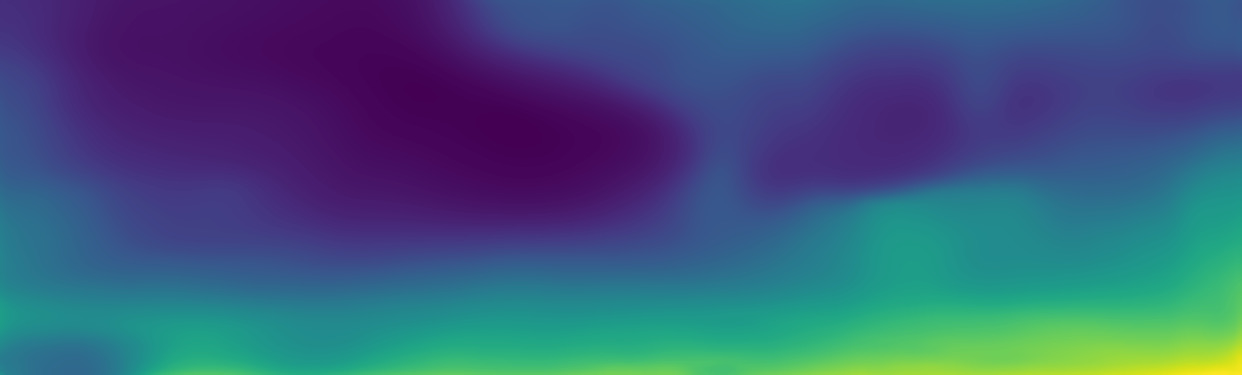} &
\includegraphics[width=0.22\linewidth]{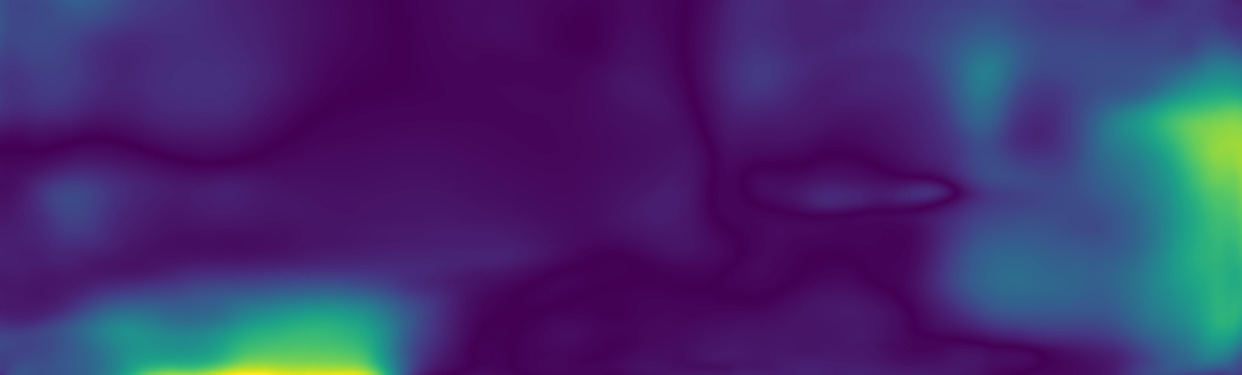} \\

\rotatebox[origin=l]{45}{B2F} & 
\includegraphics[width=0.22\linewidth]{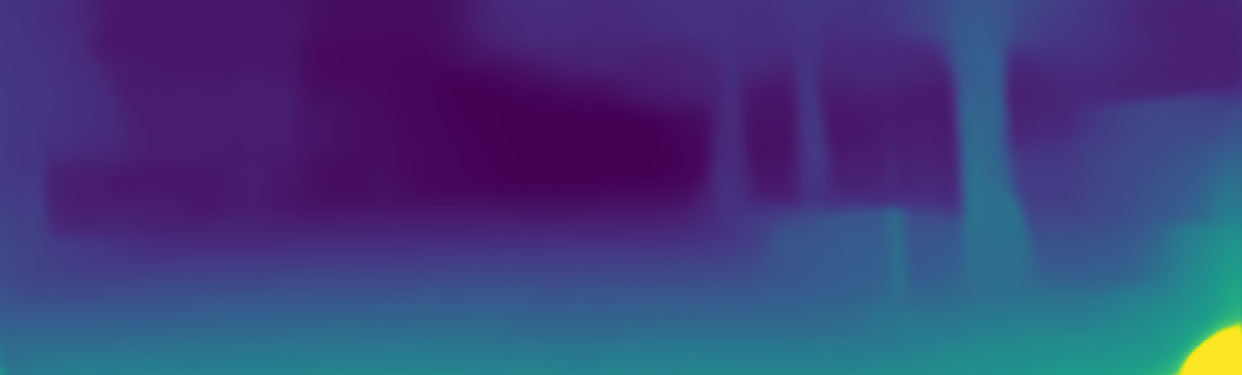} &
\includegraphics[width=0.22\linewidth]{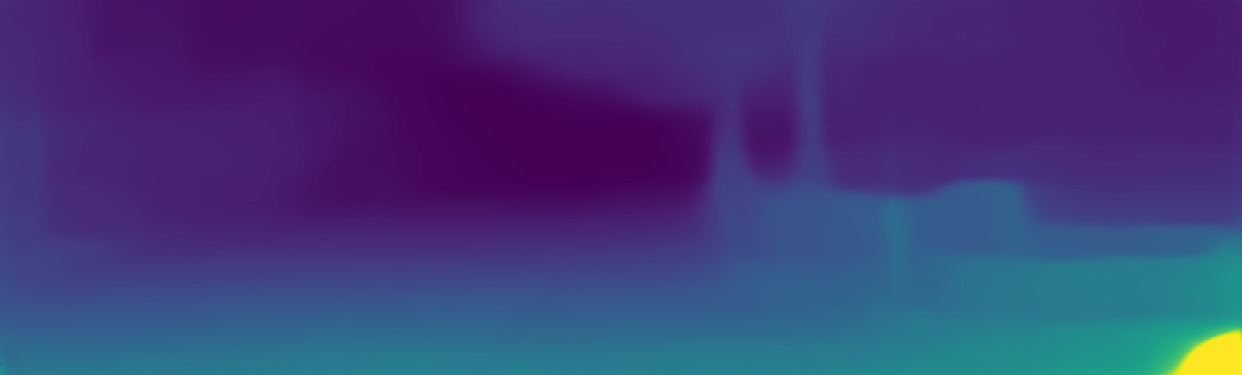} &
\includegraphics[width=0.22\linewidth]{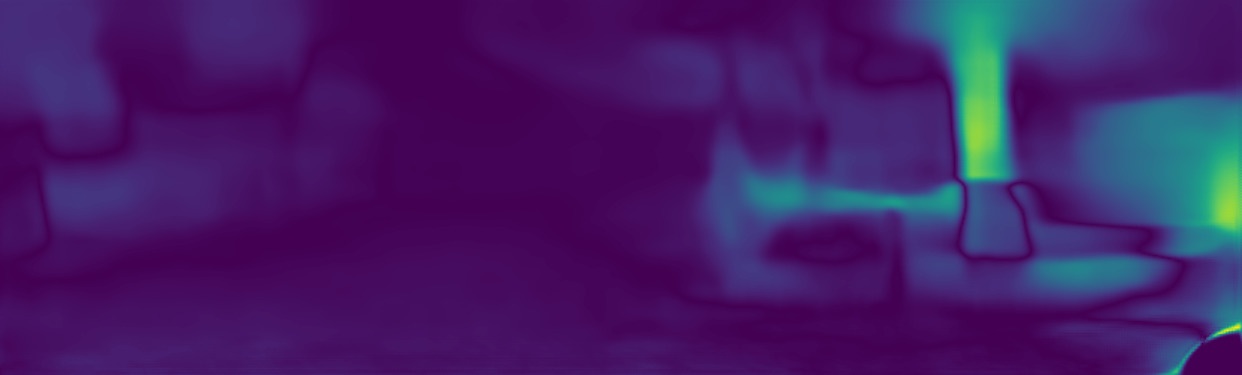} \\

\rotatebox[origin=l]{45}{SCSFM} & 
\includegraphics[width=0.22\linewidth]{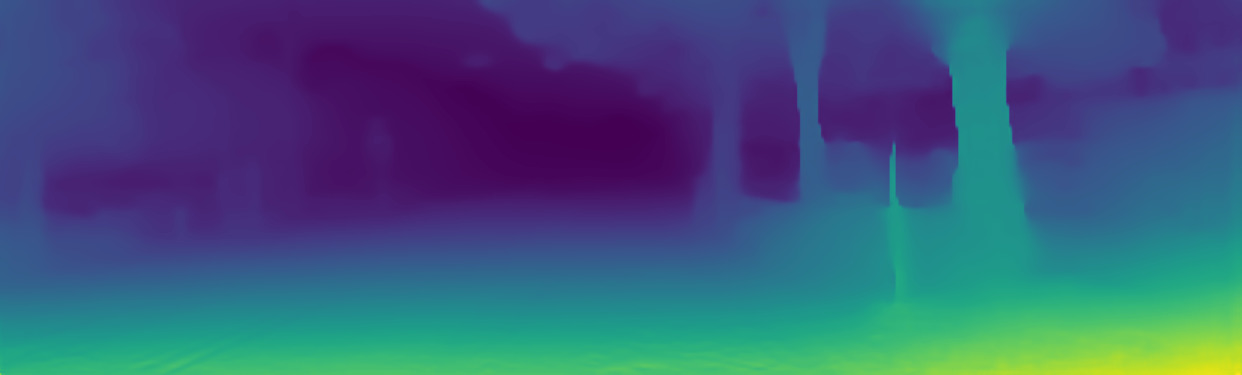} &
\includegraphics[width=0.22\linewidth]{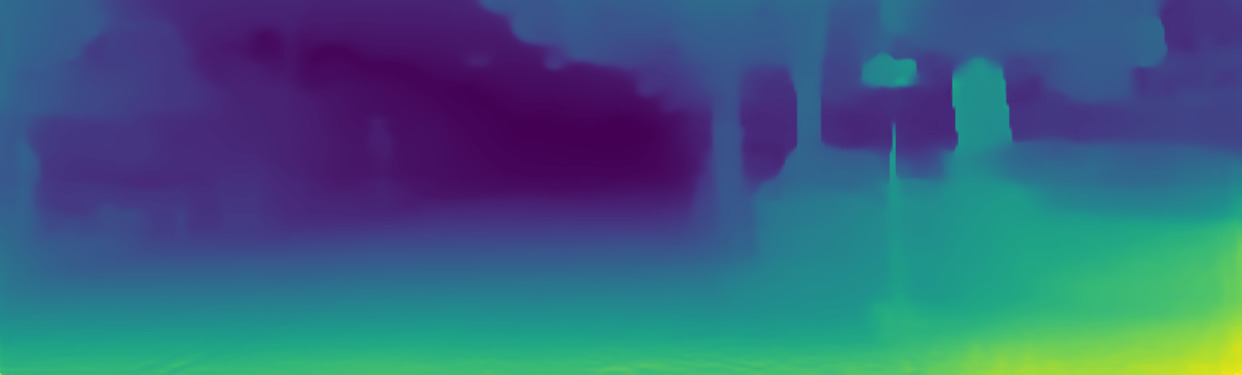} &
\includegraphics[width=0.22\linewidth]{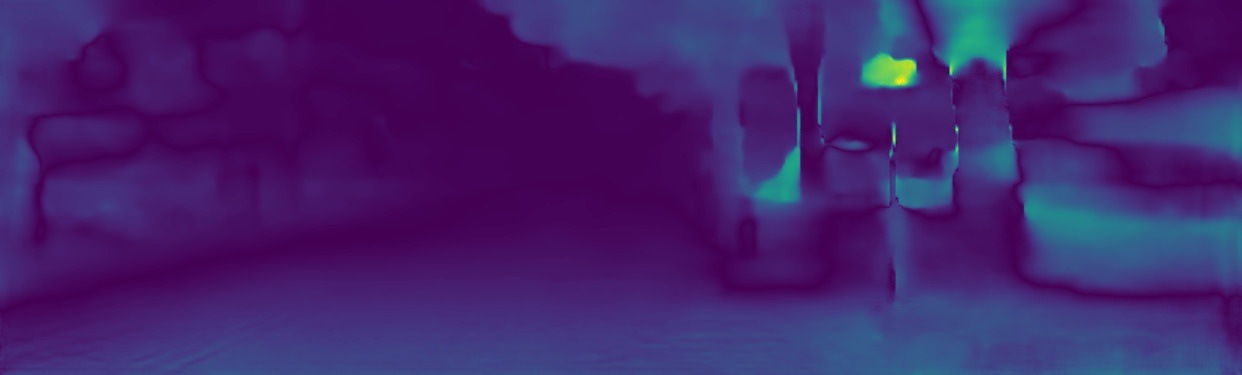} \\

\rotatebox[origin=l]{45}{Mono1} & 
\includegraphics[width=0.22\linewidth]{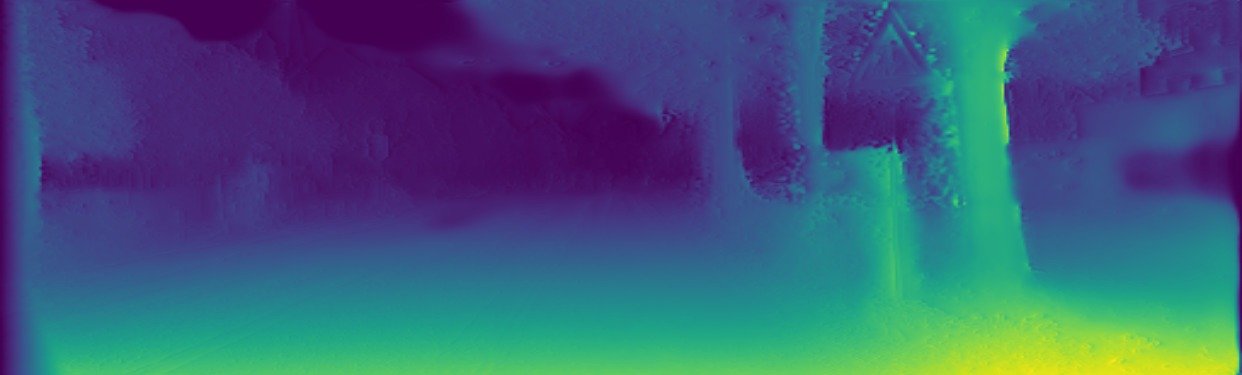} &
\includegraphics[width=0.22\linewidth]{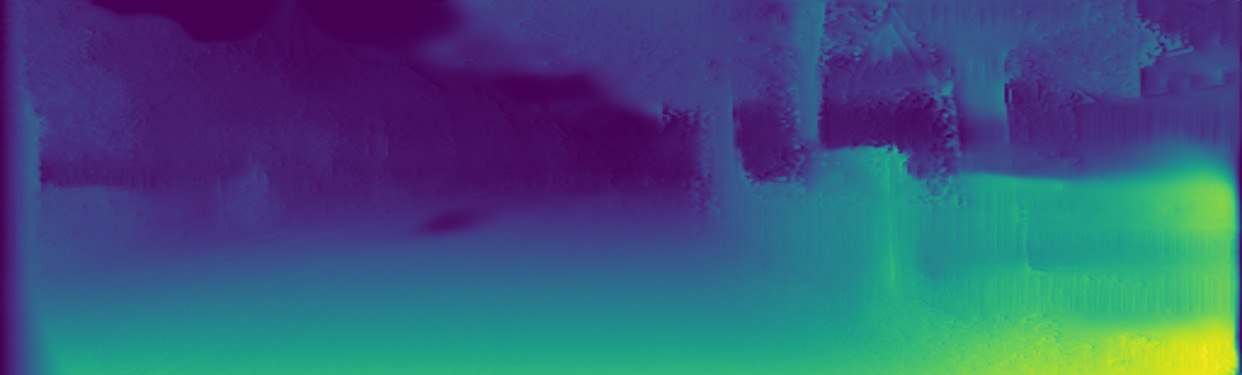} &
\includegraphics[width=0.22\linewidth]{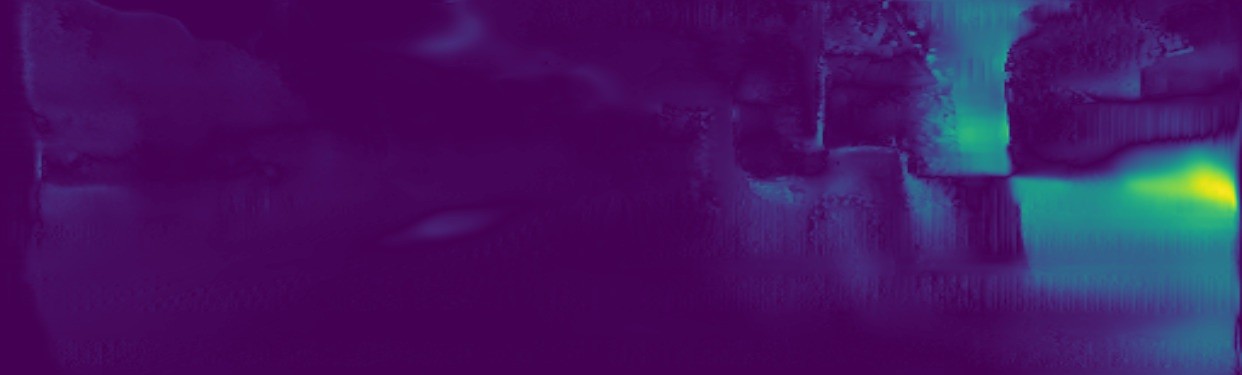} \\

\rotatebox[origin=l]{45}{Mono2} & 
\includegraphics[width=0.22\linewidth]{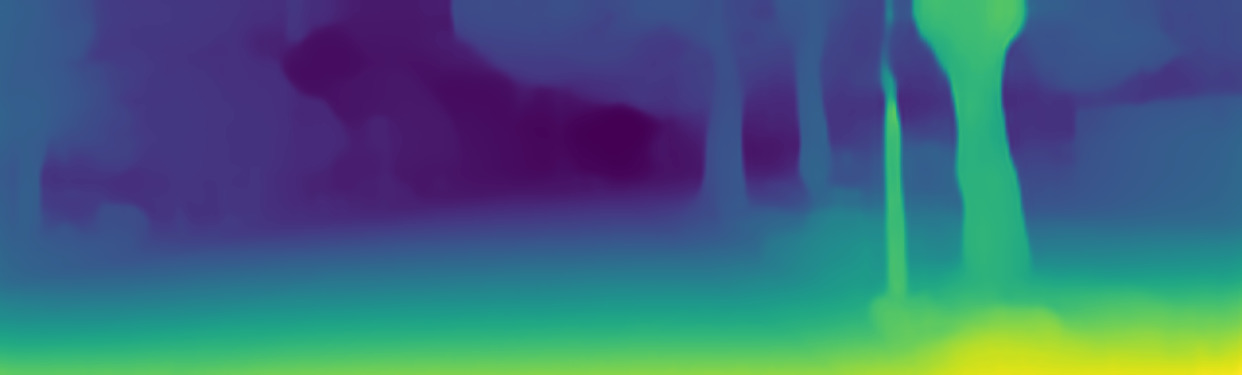} &
\includegraphics[width=0.22\linewidth]{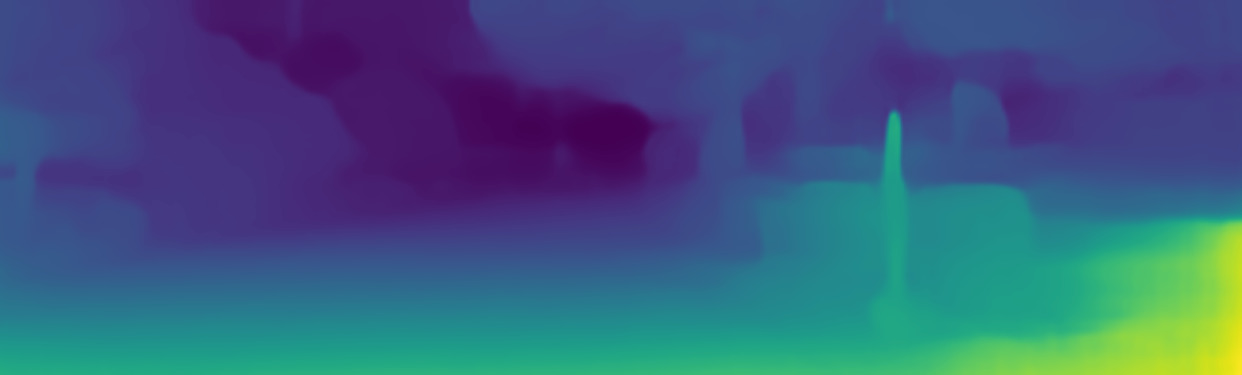} &
\includegraphics[width=0.22\linewidth]{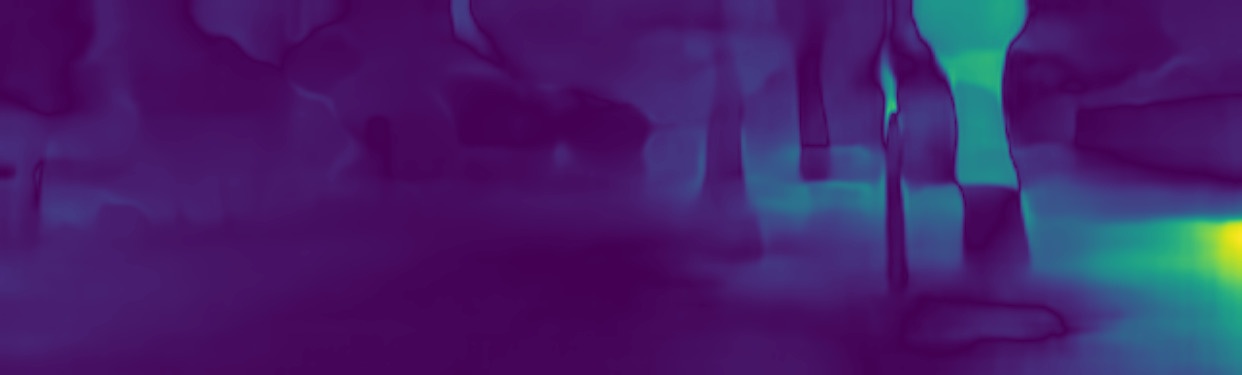} \\

\end{tabular}
\end{center}
\caption{Black box perturbation test with $\eta = 0.05$.}
\label{fig:black_pert_adv5}
\end{figure*}
\begin{table*}
	\begin{adjustbox}{max width=\textwidth}
	\begin{tabular}{l|c|c|c|c|c|c|c|c|c|c|c|c}
		\hline
		\multicolumn{1}{l}{} & \multicolumn{10}{|c}{Perturbation}  \\
		\cline{2-13}
		\multicolumn{1}{l|}{} & \multicolumn{2}{c|}{SFM~\cite{zhou2017unsupervised}} & \multicolumn{2}{c|}{DDVO~\cite{wang2018learning}} & \multicolumn{2}{c|}{B2F~\cite{janai2018unsupervised}} & \multicolumn{2}{c|}{SCSFM~\cite{bian2019unsupervised}} & \multicolumn{2}{c|}{Mono1~\cite{godard2017unsupervised}} & \multicolumn{2}{c}{Mono2~\cite{godard2019digging} }\\
		\cline{2-13}
		Methods    & RMSE   & Rel (\%)   & RMSE     & Rel (\%)   & RMSE      & Rel (\%)  & RMSE     & Rel (\%)  & RMSE    & Rel (\%)     & RMSE     & Rel (\%) \\
		\hline
		SFM~\cite{zhou2017unsupervised}      & -            & -       & 6.818         & 11      & 6.3894         & 4         & 6.2981        & 3        & 6.2668      & 2     & 6.5082  & 6   \\
		DDVO~\cite{wang2018learning}     & 6.3922       & 17      & -             & -       & 6.1185         & 12        & 5.7431        & 5        & 5.7929      & 6     & 6.0563  & 10  \\
		B2F~\cite{janai2018unsupervised}      & 5.8541       & 14      & 5.9961        & 17      & -              & -         & 6.0593        & 18       & 5.5322      & 8     & 5.772   & 12   \\
		SCSFM~\cite{bian2019unsupervised}    & 5.6802       & 9       & 5.7216        & 10      & 5.9632         & 15        & -             & -        & 5.4092      & 4     & 5.6468  & 9    \\
		Mono1~\cite{godard2017unsupervised}    & 5.2791       & 2       & 5.5388        & 7       & 5.4201         & 5         & 5.4602        & 6        & -           & -     & 5.4485  & 5   \\
		Mono2~\cite{godard2019digging}     & 5.7929       & 18      & 5.7408        & 17      & 5.526          & 13        & 5.6306        & 15       & 5.2432      & 7     & -       & -   \\
		\hline
	\end{tabular}
	\end{adjustbox}
	\caption{Black-box perturbation attack when $\eta=0.05$.}
	\label{tab:trans_pert_rmse}
\end{table*}

\begin{figure*}[!ht]
\centering
\newcommand{\turnheightnew}{0.15\columnwidth}
\centering

\begin{center}

\begin{tabular}{@{\hskip 0.5mm}c@{\hskip 0.5mm}c@{\hskip 0.5mm}c@{\hskip 0.5mm}c@{}}

 & Clean image & Attacked image &  \\

 & 
\includegraphics[width=0.22\linewidth]{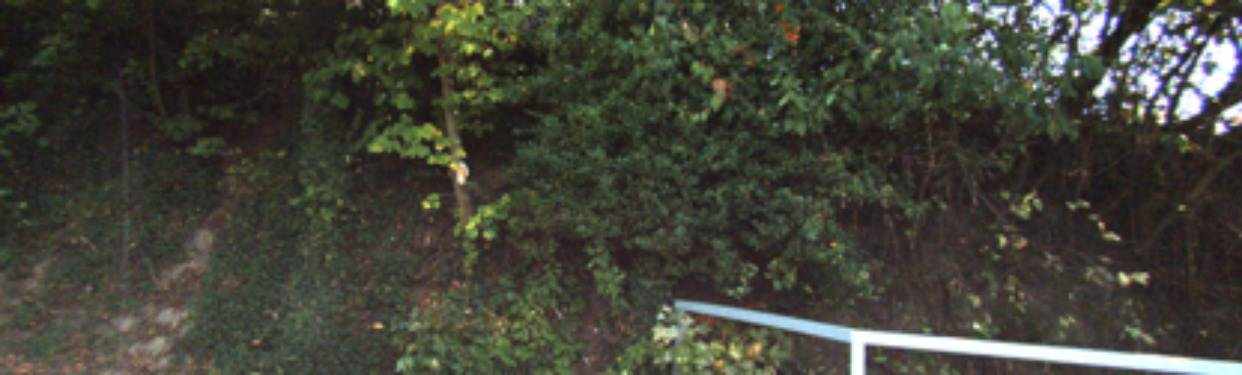} &
\includegraphics[width=0.22\linewidth]{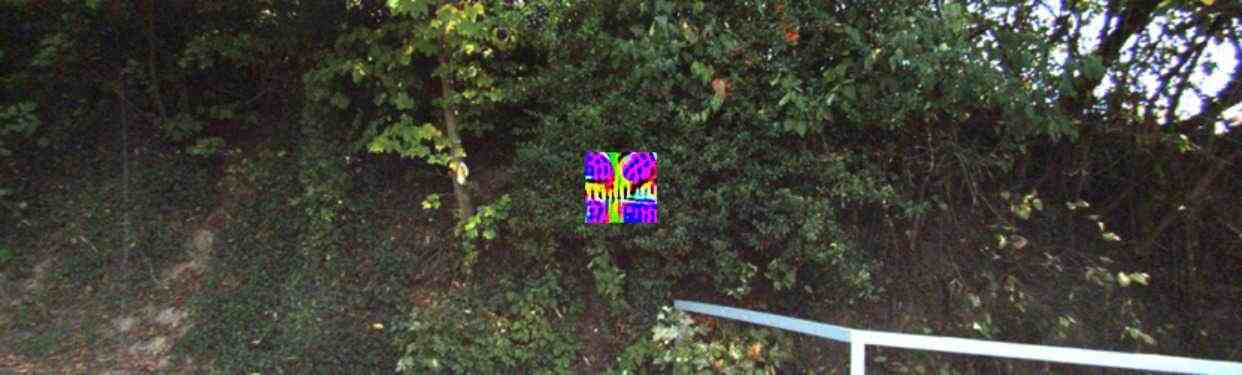} &  \\

 & Clean depth & Attacked depth & Depth gap \\

\rotatebox[origin=l]{45}{SFM} & 
\includegraphics[width=0.22\linewidth]{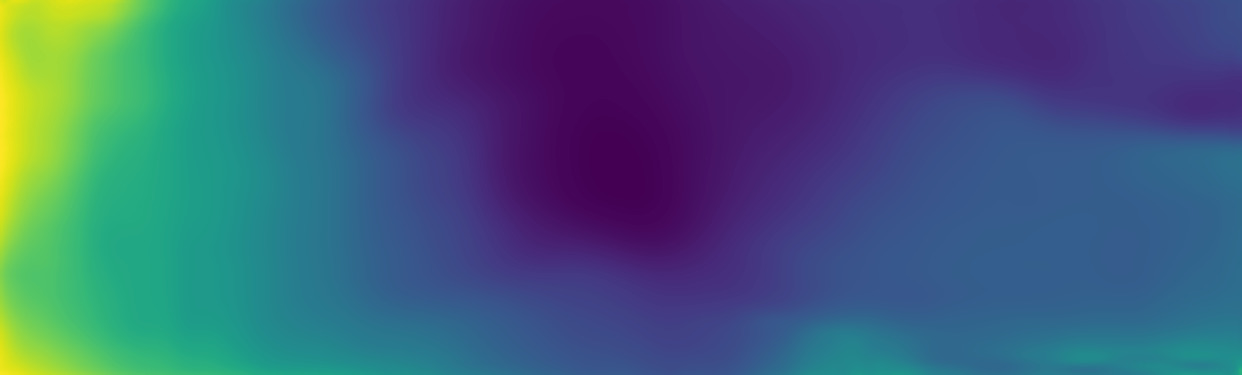} &
\includegraphics[width=0.22\linewidth]{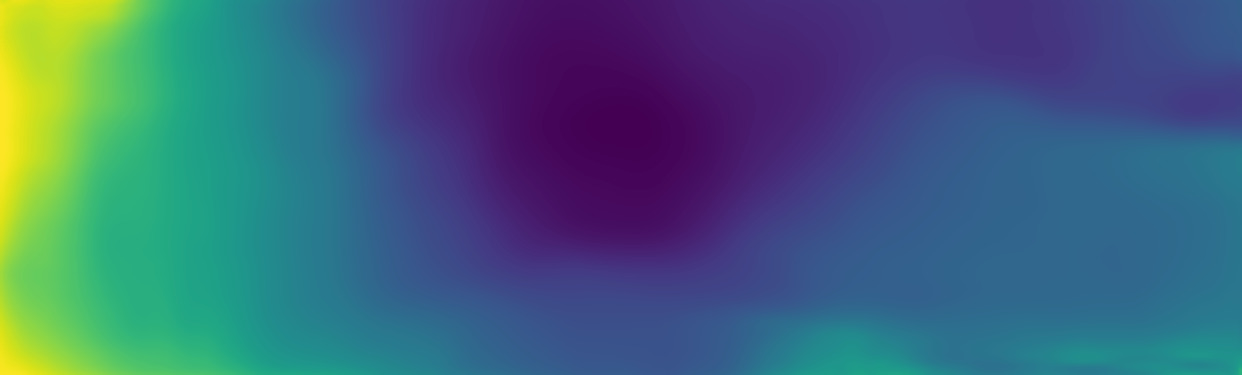} &
\includegraphics[width=0.22\linewidth]{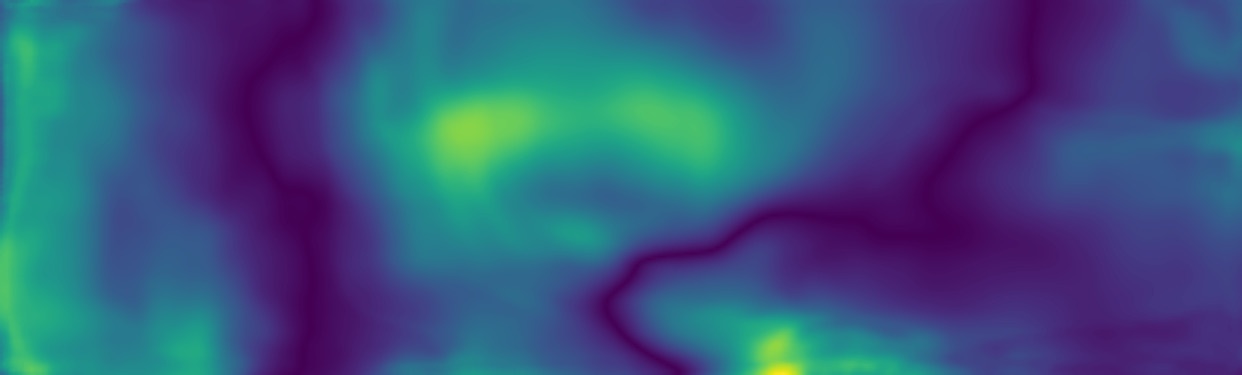} \\

\rotatebox[origin=l]{45}{DDVO} & 
\includegraphics[width=0.22\linewidth]{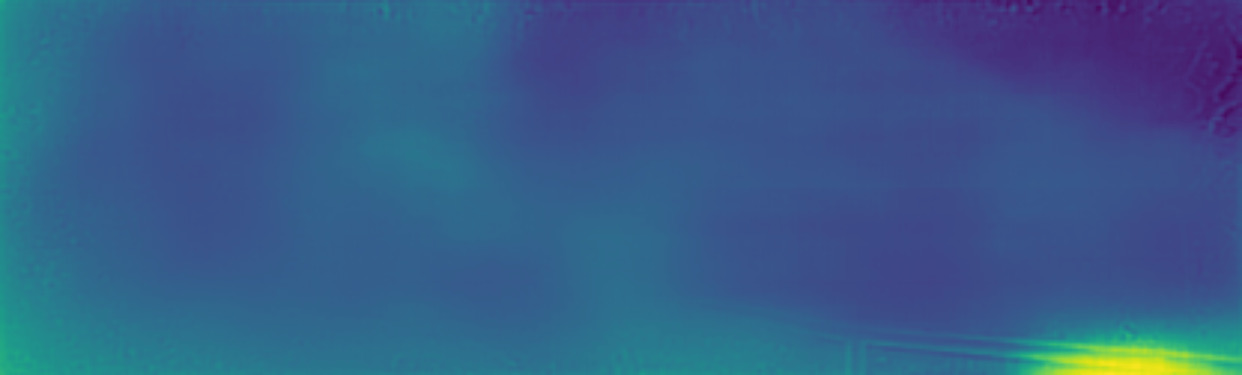} &
\includegraphics[width=0.22\linewidth]{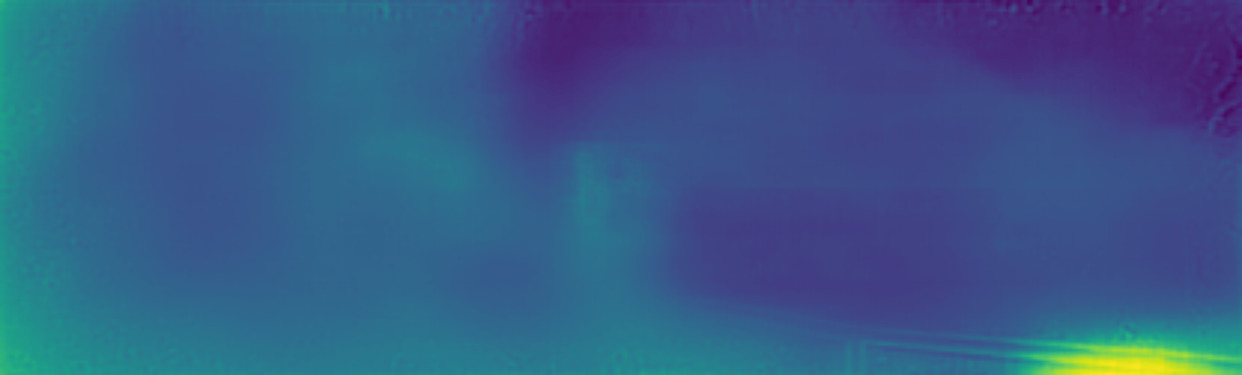} &
\includegraphics[width=0.22\linewidth]{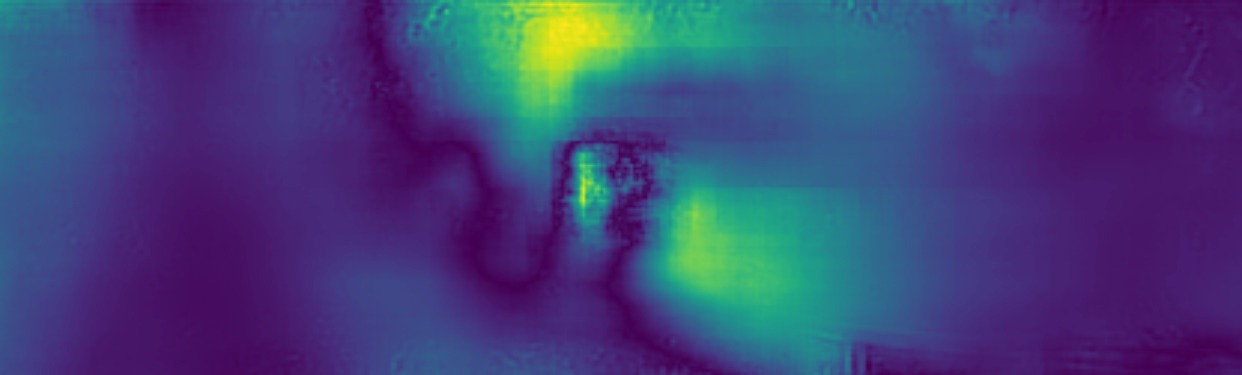} \\

\rotatebox[origin=l]{45}{B2F} & 
\includegraphics[width=0.22\linewidth]{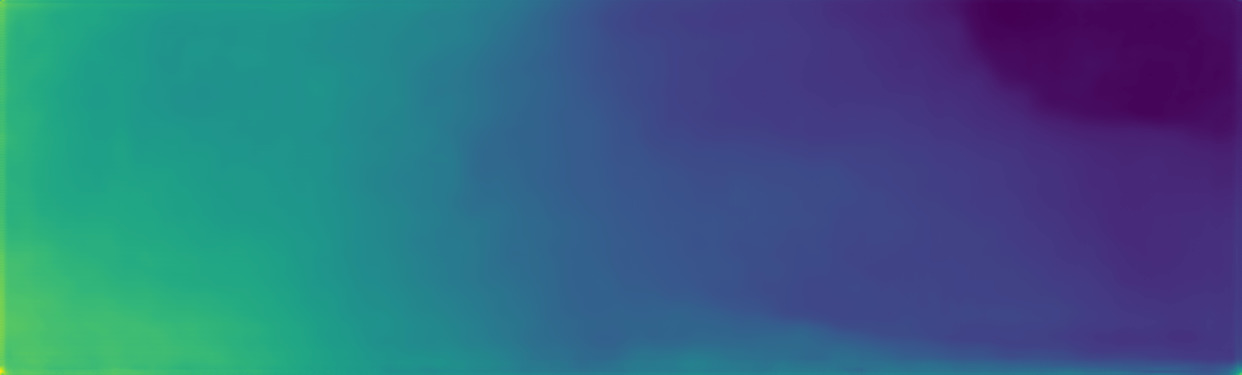} &
\includegraphics[width=0.22\linewidth]{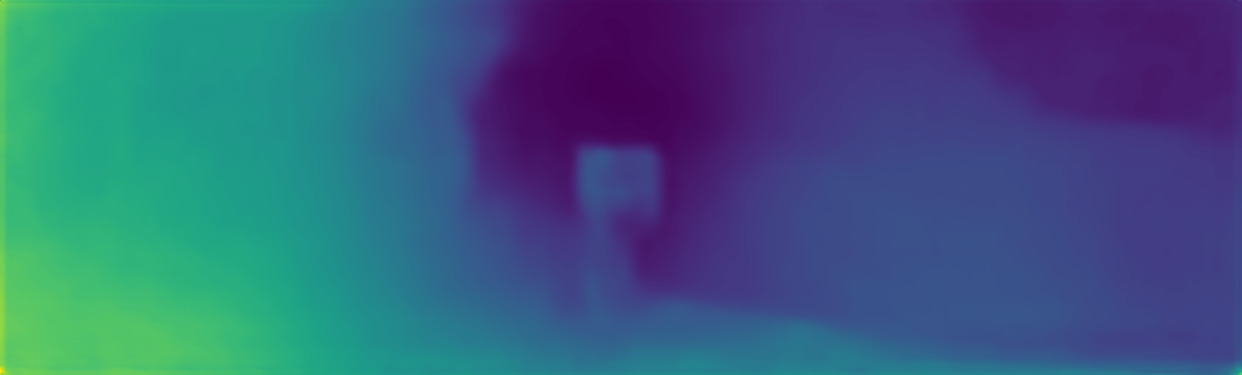} &
\includegraphics[width=0.22\linewidth]{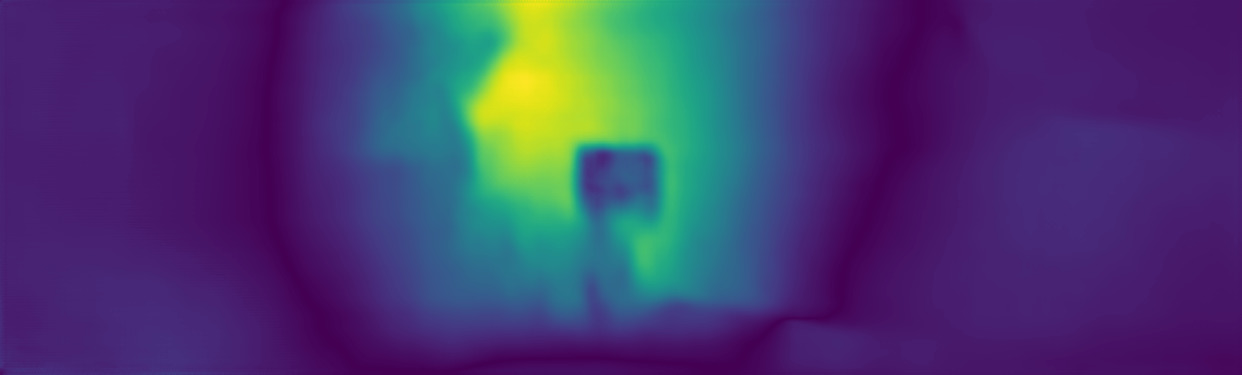} \\

\rotatebox[origin=l]{45}{Mono1} & 
\includegraphics[width=0.22\linewidth]{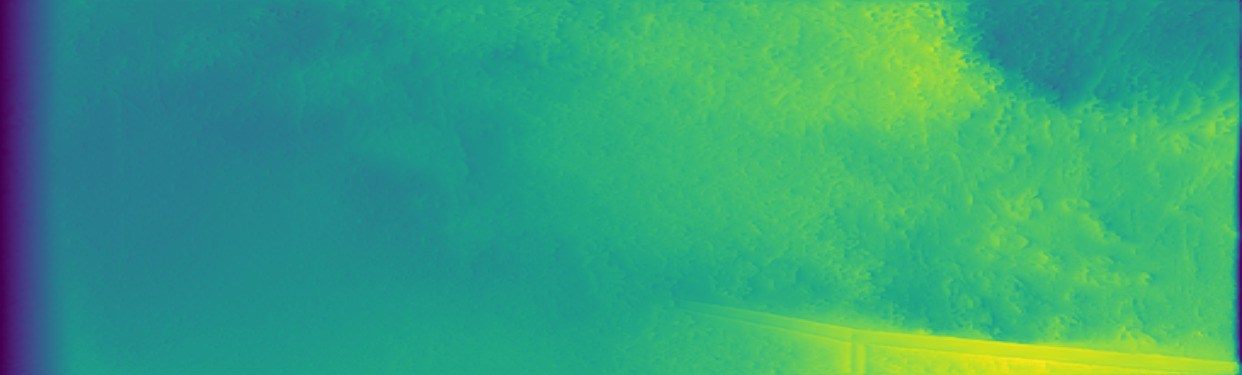} &
\includegraphics[width=0.22\linewidth]{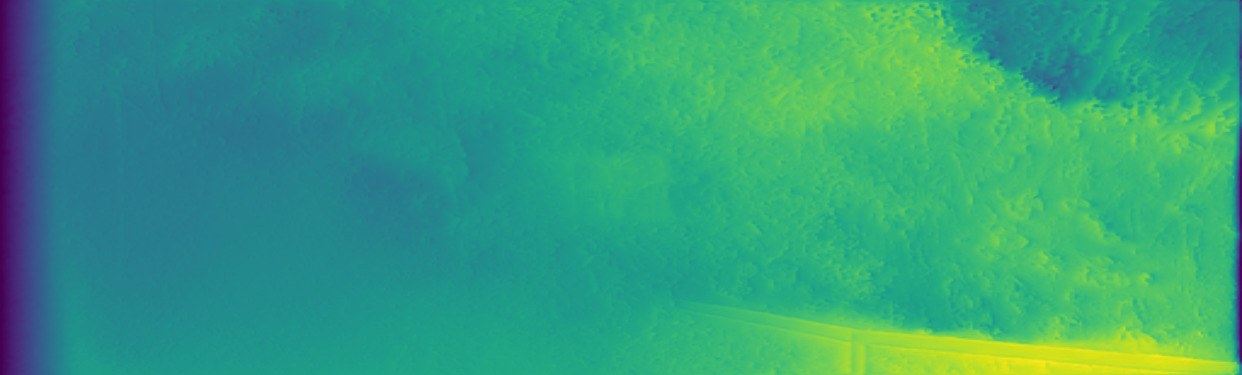} &
\includegraphics[width=0.22\linewidth]{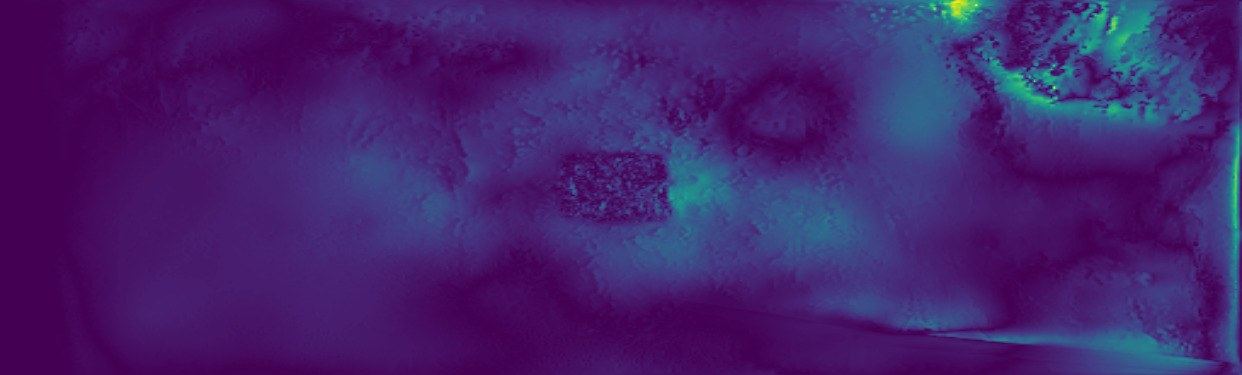} \\

\rotatebox[origin=l]{45}{Mono2} & 
\includegraphics[width=0.22\linewidth]{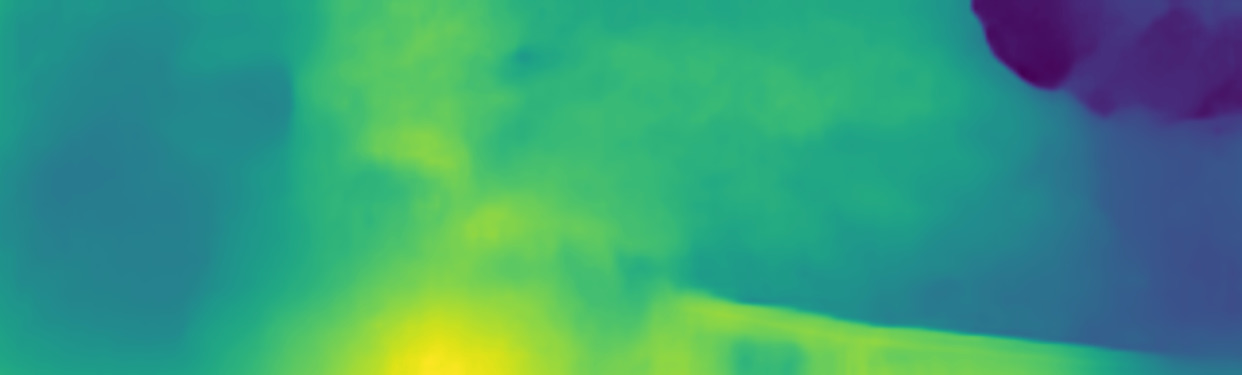} &
\includegraphics[width=0.22\linewidth]{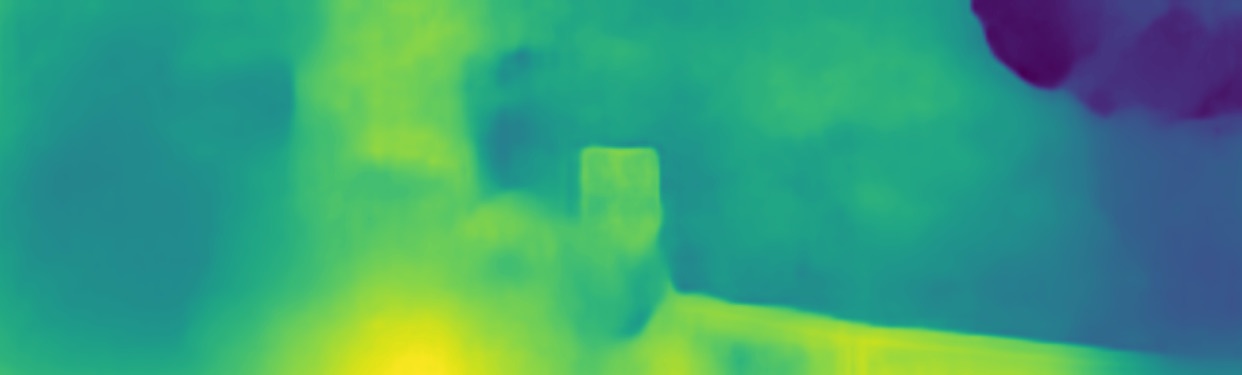} &
\includegraphics[width=0.22\linewidth]{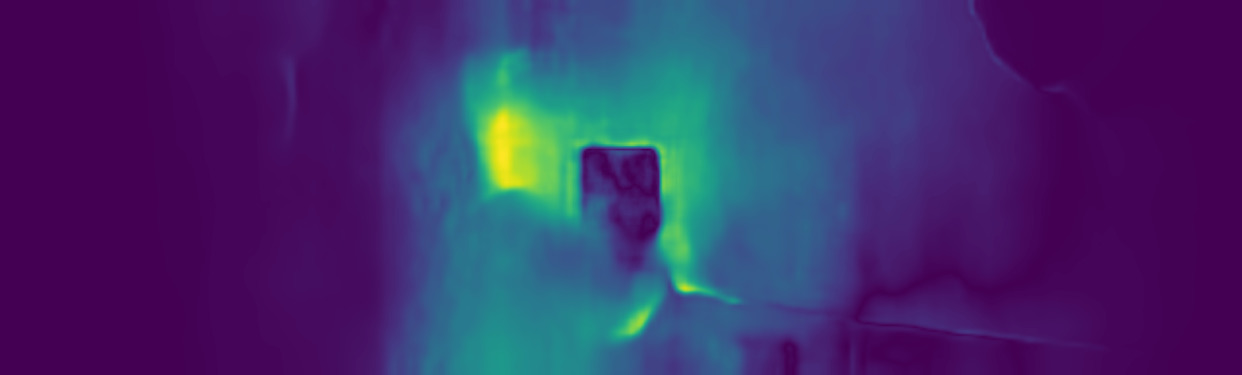} \\

\end{tabular}
\end{center}
\caption{Black box for adversarial patch of size $72 \times 72$.}
\label{fig:black_patch_adv72}
\end{figure*}
\begin{table*}
	\begin{adjustbox}{max width=\textwidth}
	\begin{tabular}{l|c|c|c|c|c|c|c|c|c|c|c}
		\hline
		\multicolumn{1}{l}{} & \multicolumn{1}{|l}{} & \multicolumn{10}{|c}{Patch}  \\
		\cline{3-12}
		\multicolumn{1}{l|}{} & \multicolumn{1}{c|}{Clean} & \multicolumn{2}{c|}{SFM~\cite{zhou2017unsupervised}} & \multicolumn{2}{c|}{DDVO~\cite{wang2018learning}} & \multicolumn{2}{c|}{SCSFM~\cite{bian2019unsupervised}} & \multicolumn{2}{c|}{Mono1~\cite{godard2017unsupervised}} & \multicolumn{2}{c}{Mono2~\cite{godard2019digging}}\\
		\cline{2-12}
		Methods  & RMSE        & RMSE       & Rel (\%)     & RMSE        & Rel (\%)     & RMSE    & Rel (\%)      & RMSE       & Rel (\%)       & RMSE      & Rel (\%)   \\
		\hline
		SFM~\cite{zhou2017unsupervised}      &6.1711 		& -            & -        & 7.5362        & 23       & 6.3677         & 4        & 6.7395        & 10        & 6.2181         & 1        \\
		DDVO~\cite{wang2018learning}     &5.5072 		& 5.9095       & 8        & -             & -        & 5.6998         & 4        & 5.7942        & 6         & 5.5974         & 2        \\
		B2F~\cite{janai2018unsupervised}      &5.1615 		& 5.7002       & 11       & 5.6992        & 11       & 5.6442         & 10       & 5.5763        & 9         & 5.374          & 5        \\
		SCSFM~\cite{bian2019unsupervised}    &5.2271 		& 6.599        & 27       & 6.5846        & 26       & -              & -        & 6.5564        & 26        & 5.2272         & 0        \\
		Mono1~\cite{godard2017unsupervised}    &5.1973 		& 6.1638       & 19       & 5.9163        & 14       & 5.8969         & 14       & -             & -         & 5.5173         & 3        \\
		Mono2~\cite{godard2019digging}    &4.9099 		& 5.4502       & 12       & 5.4289        & 11       & 5.4155         & 11       & 5.5037        & 13        & -              & -        \\
		\hline
	\end{tabular}
	\end{adjustbox}
	\caption{Black-box attack for adversarial patch of size $72 \times 72$.}
	\label{tab:trans_patch_rmse}
\end{table*}

\begin{figure*}[!t]
	\centering
	
	\begin{tabular}{ccccc}
		
		Clean image & Attacked image & Clean depth & Attacked depth & Depth gap\\
		
		\includegraphics[width=0.17\linewidth]{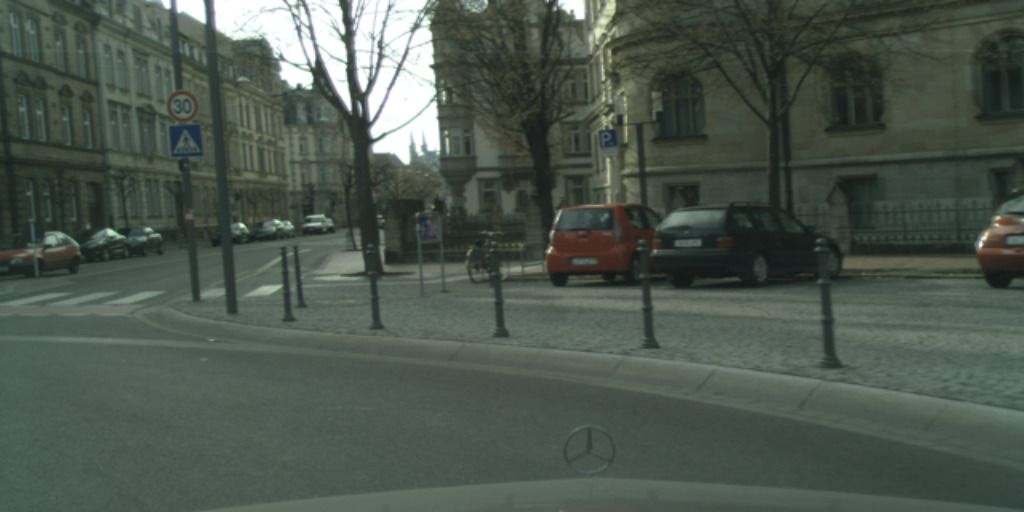} &
		\includegraphics[width=0.17\linewidth]{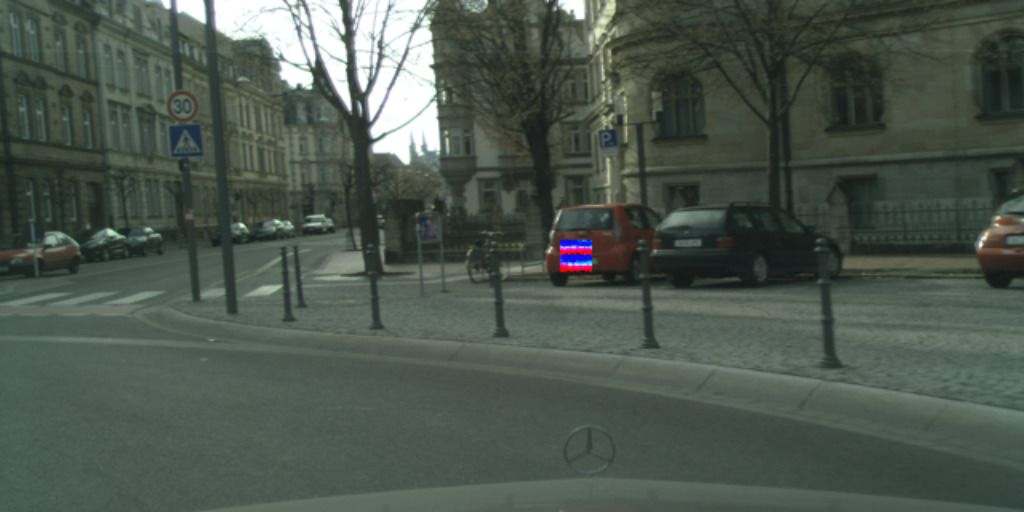} &
		\includegraphics[width=0.17\linewidth]{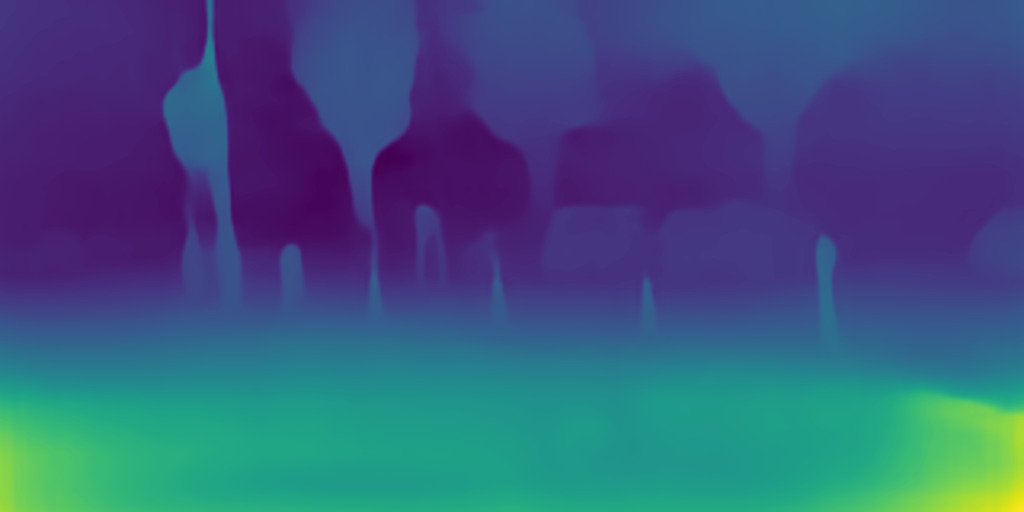} &
		\includegraphics[width=0.17\linewidth]{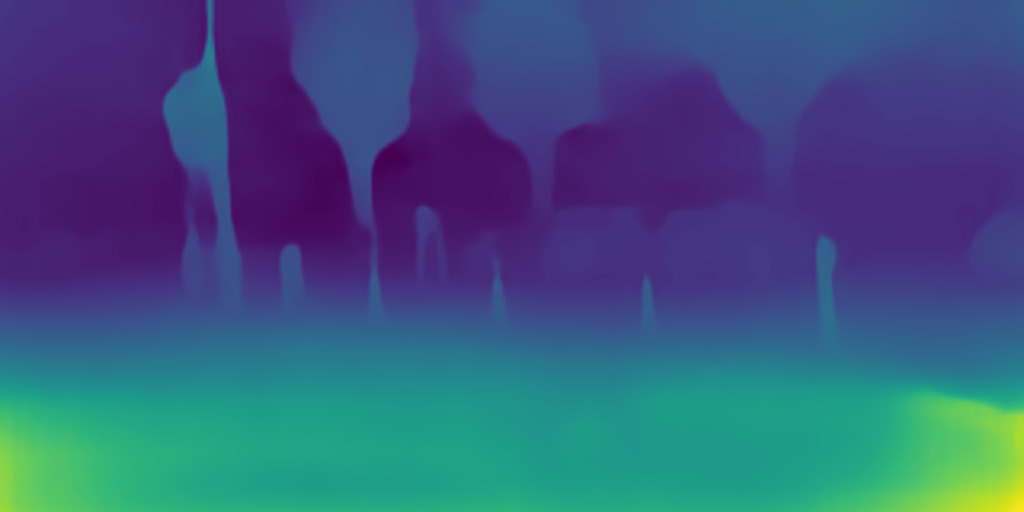} &
		\includegraphics[width=0.17\linewidth]{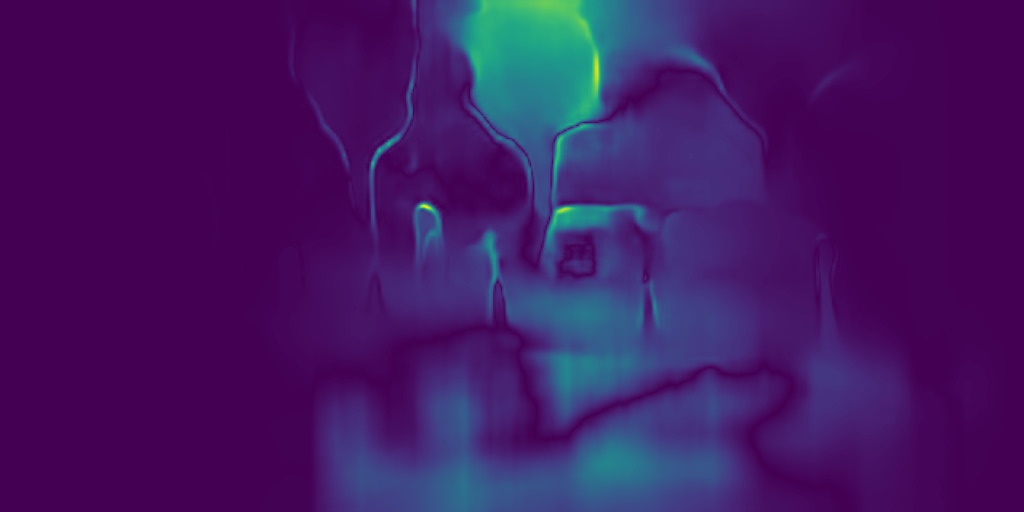} \\
		
		\includegraphics[width=0.17\linewidth]{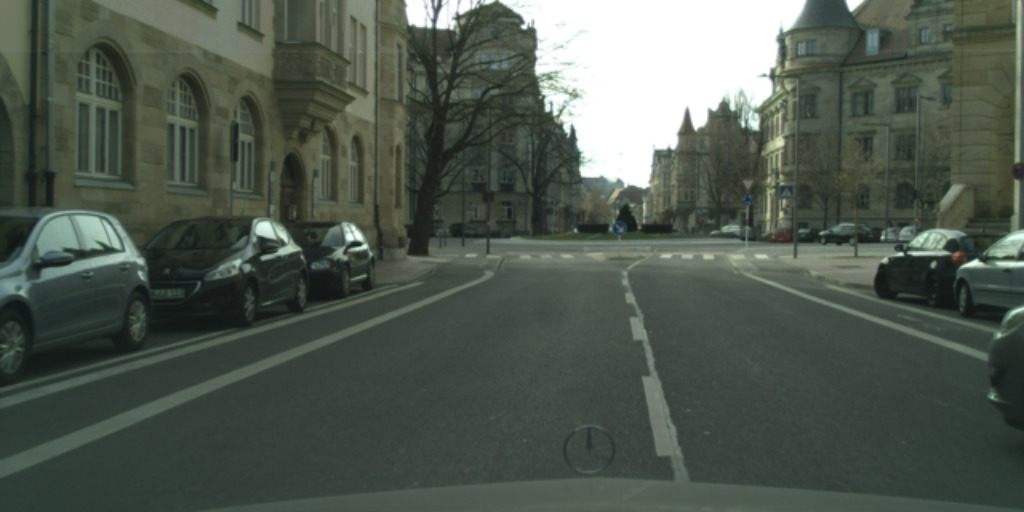} &
		\includegraphics[width=0.17\linewidth]{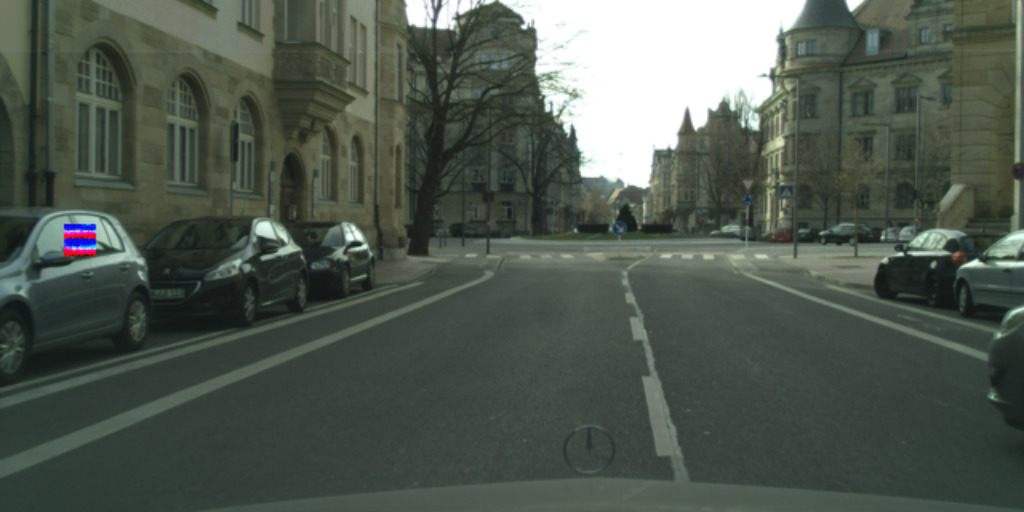} &
		\includegraphics[width=0.17\linewidth]{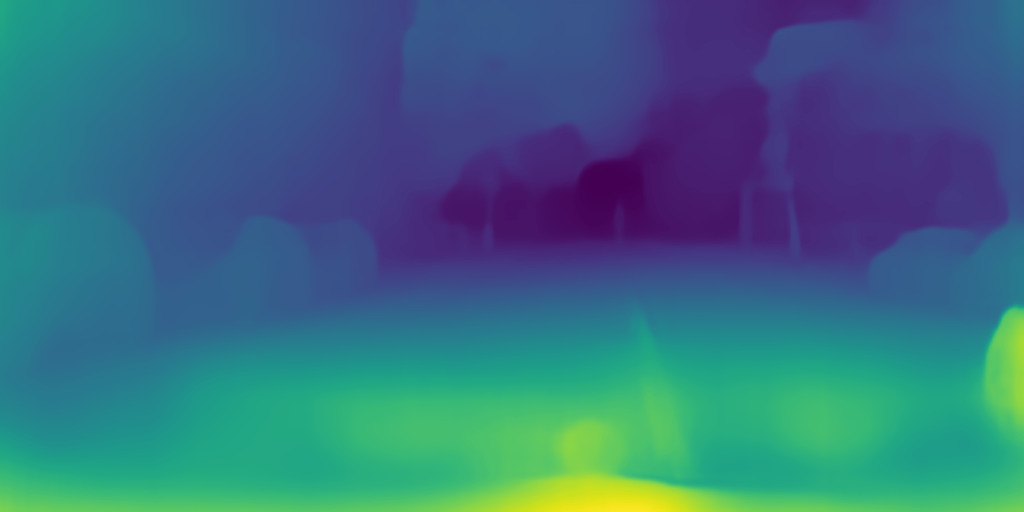} &
		\includegraphics[width=0.17\linewidth]{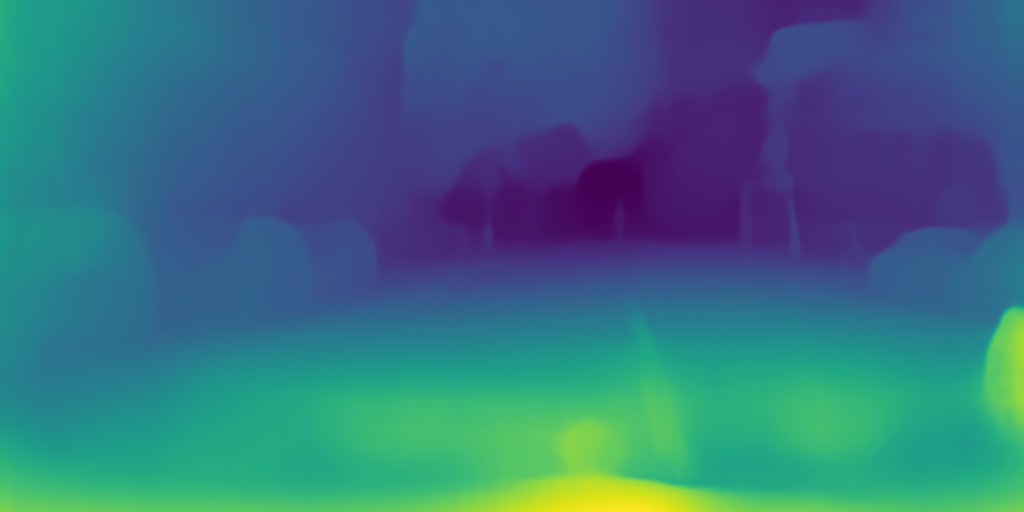} &
		\includegraphics[width=0.17\linewidth]{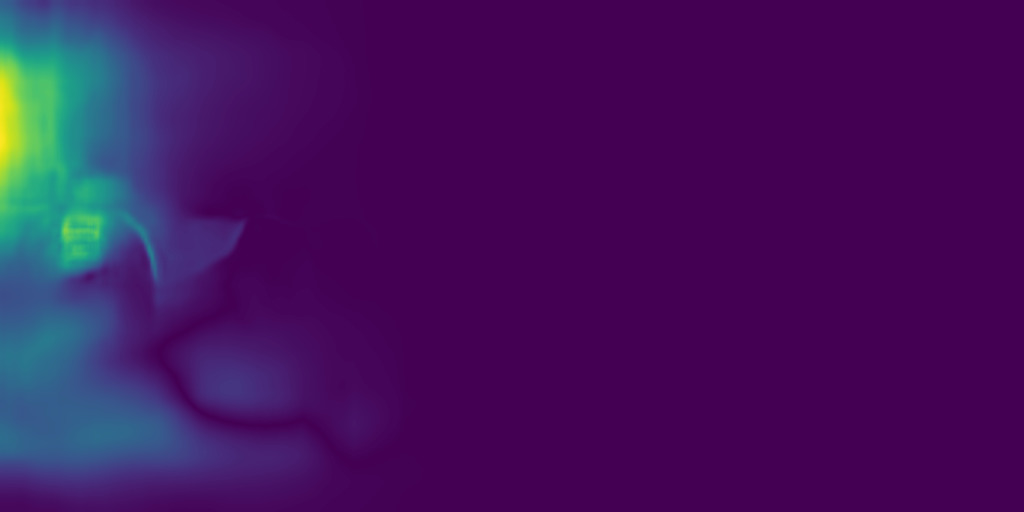} \\
		
		\includegraphics[width=0.17\linewidth]{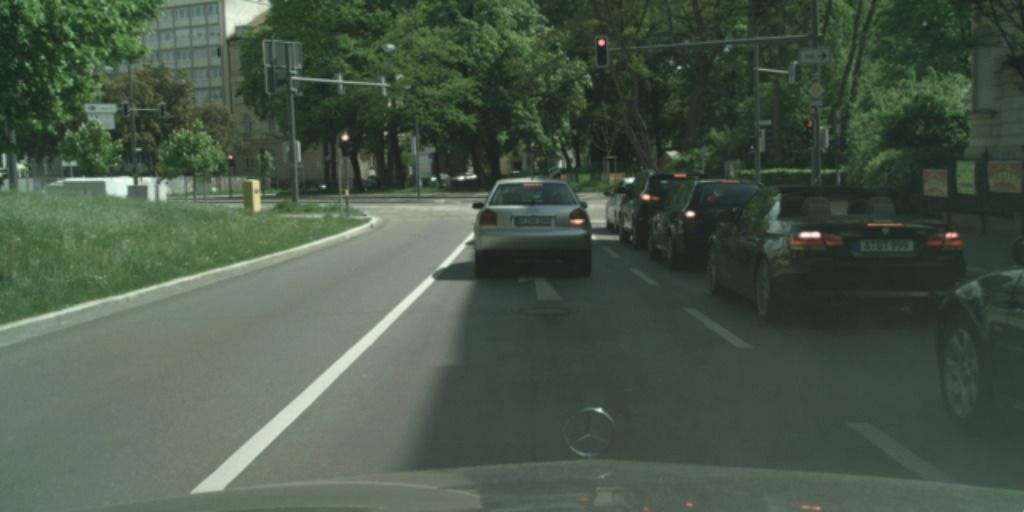} &
		\includegraphics[width=0.17\linewidth]{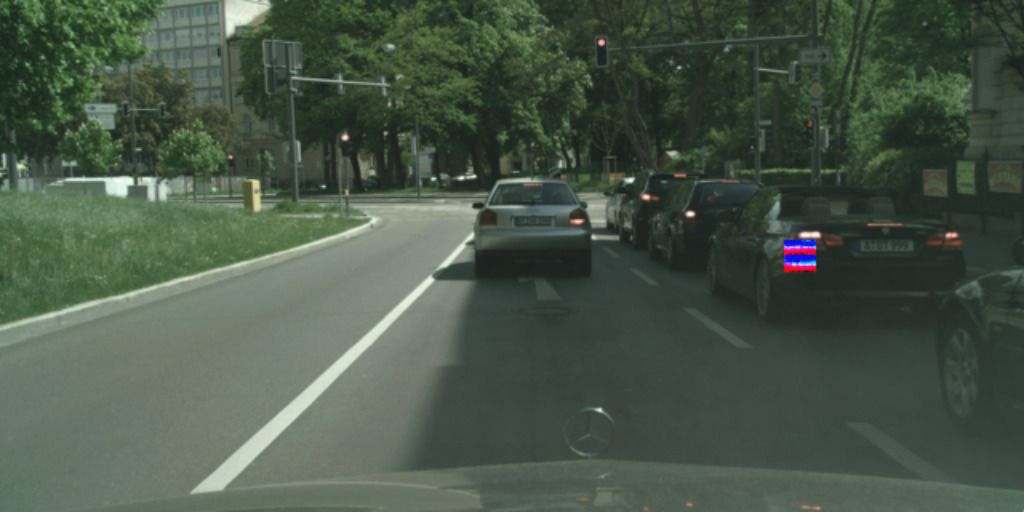} &
		\includegraphics[width=0.17\linewidth]{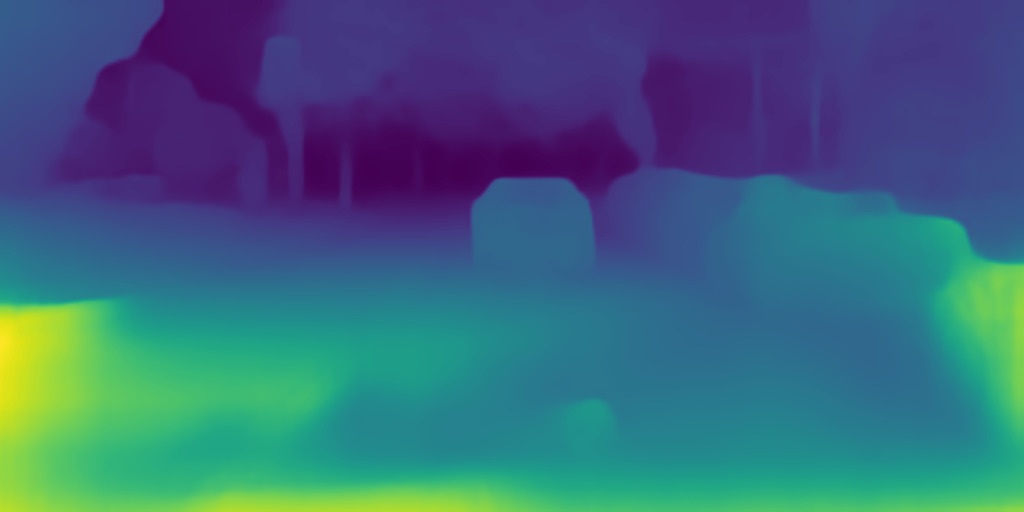} &
		\includegraphics[width=0.17\linewidth]{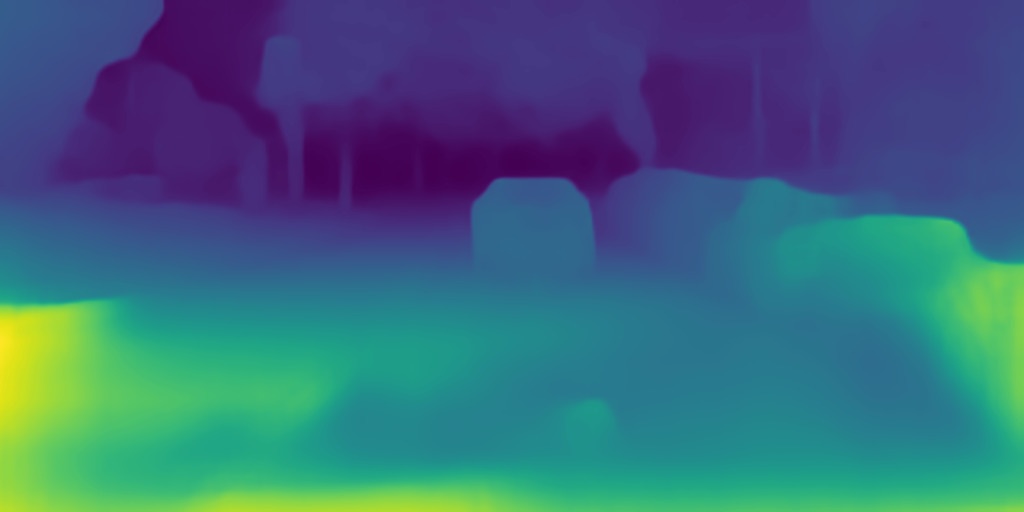} &
		\includegraphics[width=0.17\linewidth]{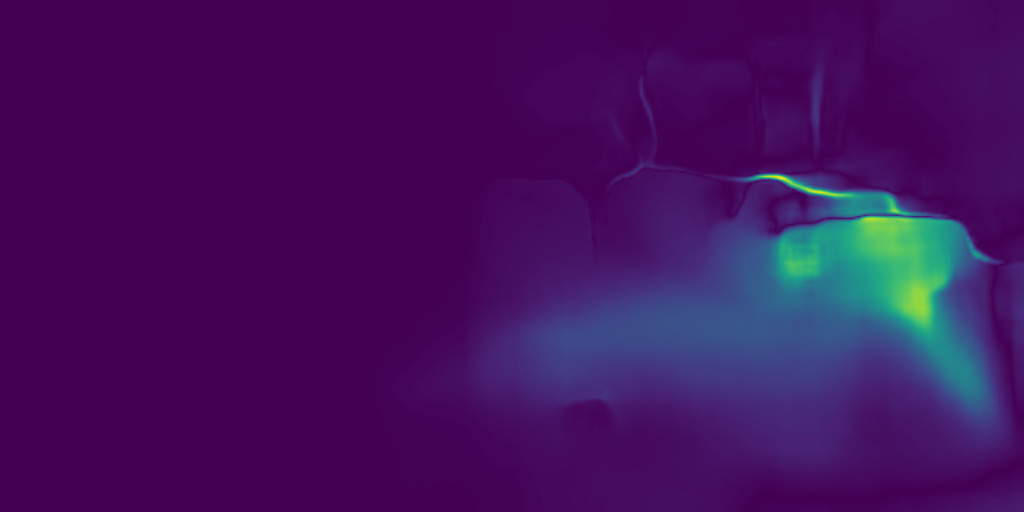} \\
		
		\includegraphics[width=0.17\linewidth]{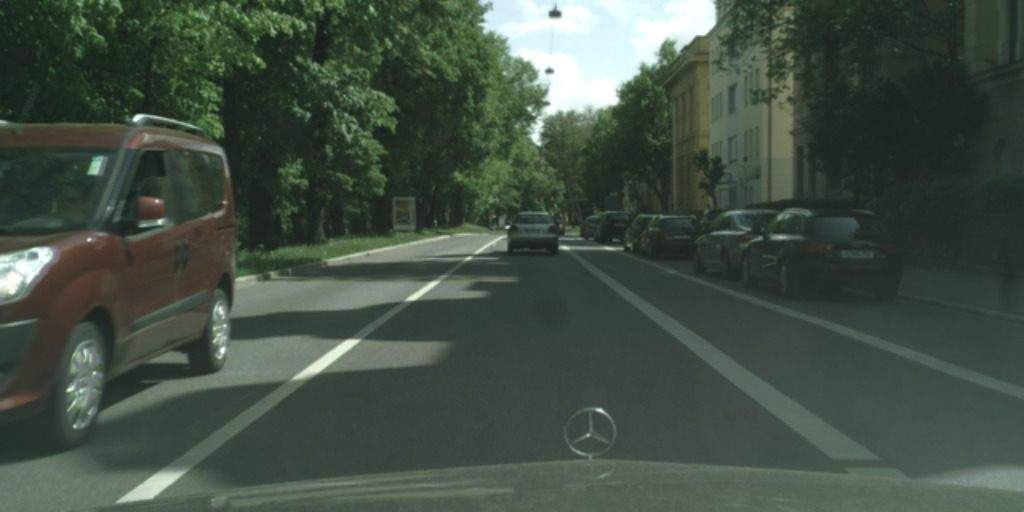} &
		\includegraphics[width=0.17\linewidth]{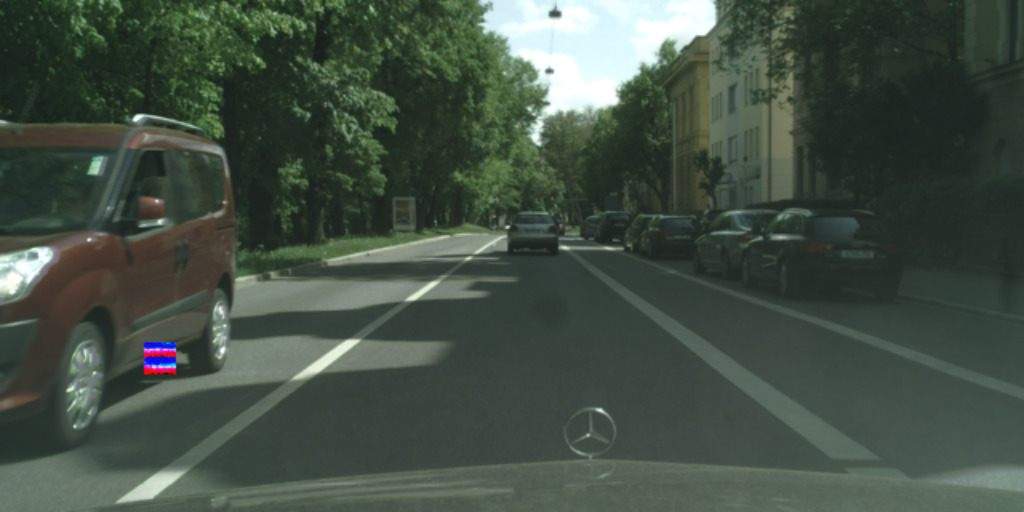} &
		\includegraphics[width=0.17\linewidth]{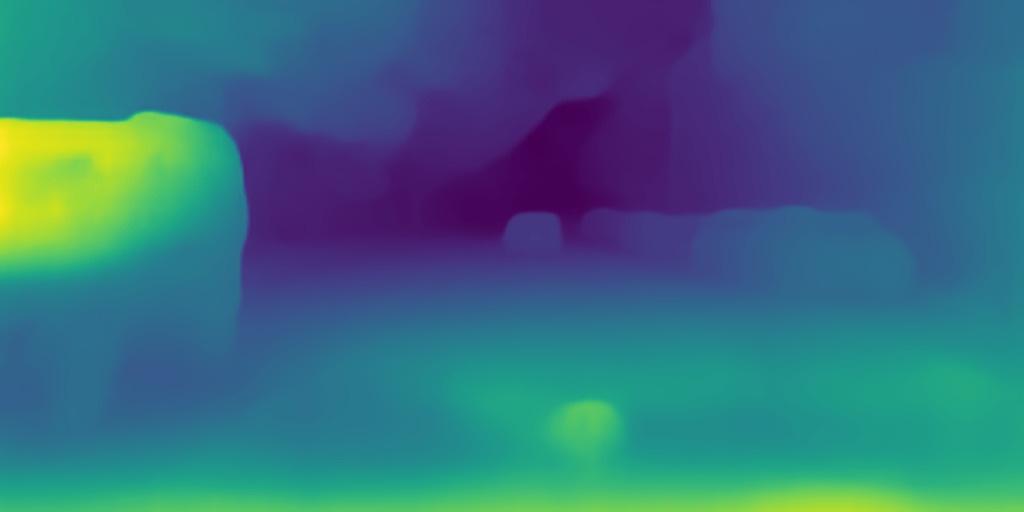} &
		\includegraphics[width=0.17\linewidth]{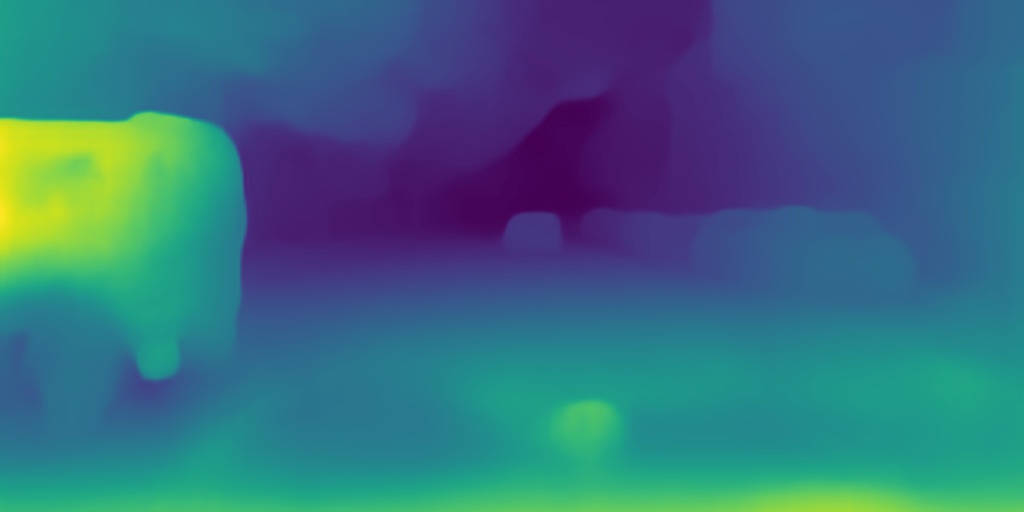} &
		\includegraphics[width=0.17\linewidth]{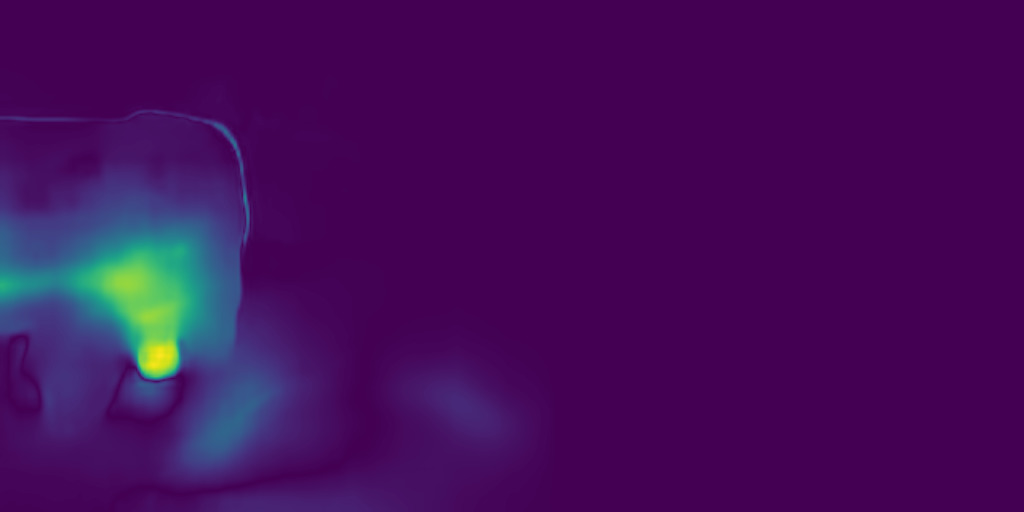} \\
		
		\includegraphics[width=0.17\linewidth]{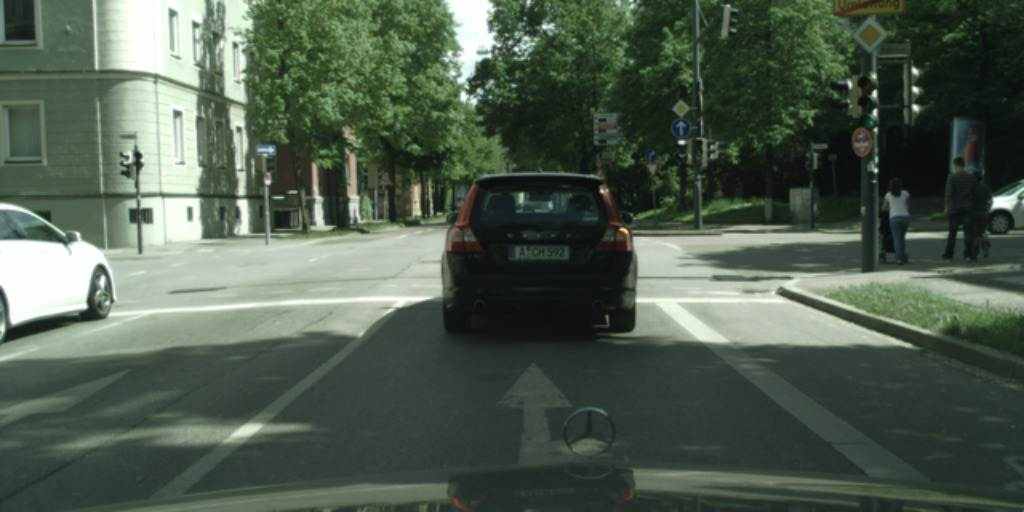} &
		\includegraphics[width=0.17\linewidth]{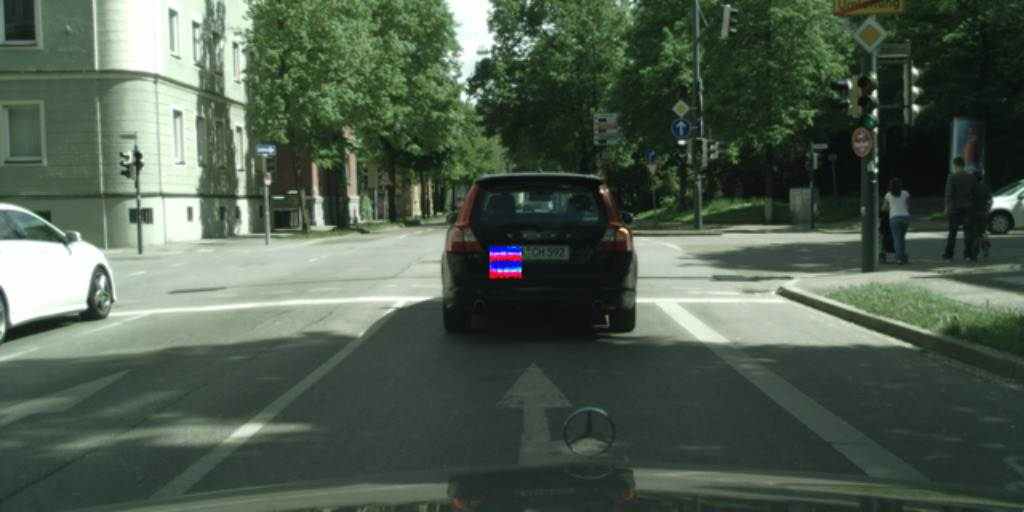} &
		\includegraphics[width=0.17\linewidth]{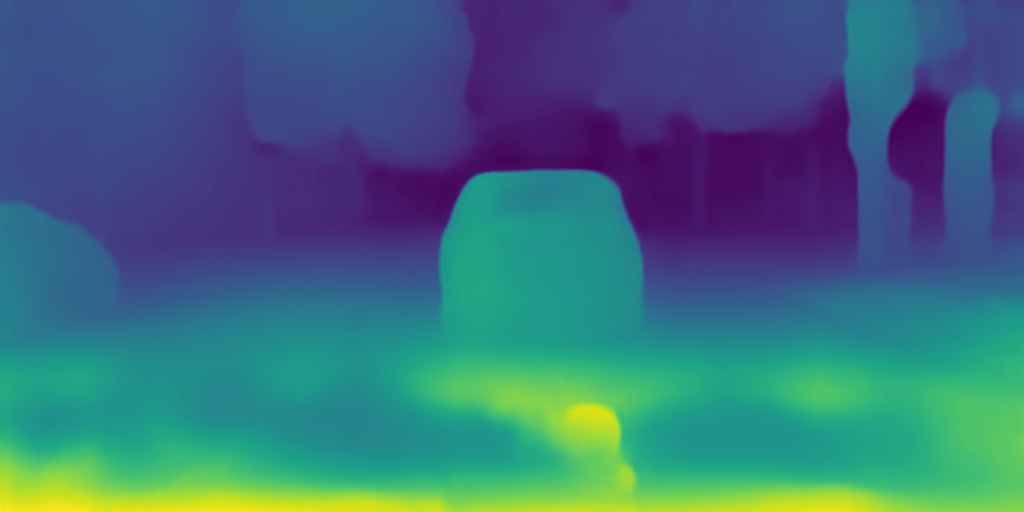} &
		\includegraphics[width=0.17\linewidth]{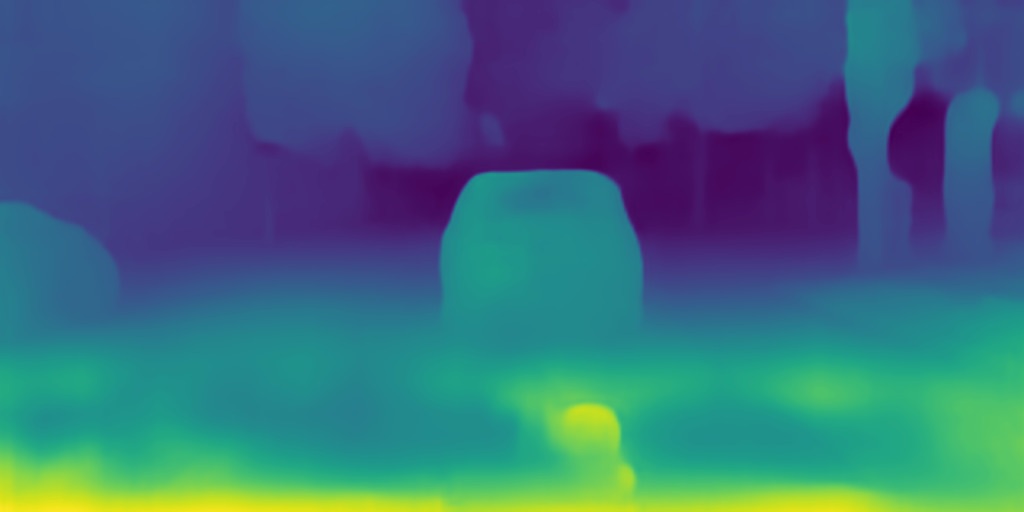} &
		\includegraphics[width=0.17\linewidth]{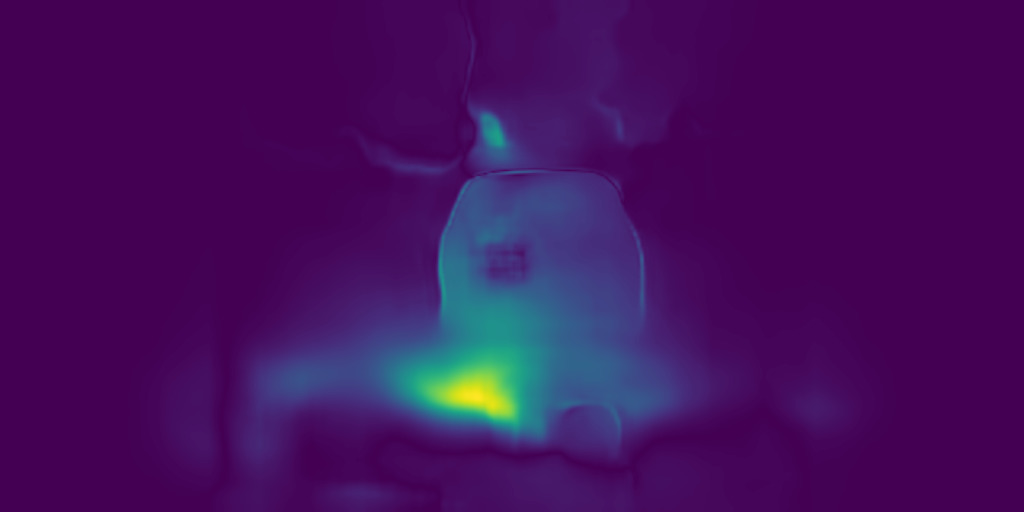} \\
		
	\end{tabular}
	\caption{Cross-data transferability. KITTI~\cite{geiger2013vision} dataset adversarial patch used to attack images from Cityscape~\cite{cordts2016cityscapes} dataset.}
	\label{fig:cross_data}
\end{figure*}

\section{Experiments}

Adversarial attacks can be categorized mainly in two ways: the desired perturbation type and assumption of the attacker's knowledge. Two perturbation types used in this investigation are briefed in Section~\ref{pertattack} and Section~\ref{patchattack}. In this section, the kinds of assumptions of the attacker's knowledge will be discussed, particularly, white-box and black-box attack.

\subsection{White-box Attack}

In a white-box attack, the attacker has full knowledge and access to the target model, including network architecture, input, output, and weights. We optimized adversarial samples in both perturbation and patch type. Generated perturbation is the same size as the input image of the target model. The perturbation was optimized in such a way that it is imperceivable to our naked eye. The perturbation added to the input image is constrained in our experiments in the range of $[0.01, 0.1]$ as shown in Figure~\ref{fig:white_pert_adv5} and Table~\ref{tab:whitebox_gpert_rmse}. For patch type attack we generate square adversarial patches on each network independently. All patches estimated for these networks are global, i.e., doesn't depend on the image where it is pasted. Each patch is augmented to make it invariant to rotation and location. Rotation is sampled randomly from $n \times \frac{\pi}{2}$ where $n \in \mathbb{Z}_+: 0 \leq n \leq 4$ and location from $\xi(x, y)$ where $(x,y) \in \mathbb{Z}_+: 0 \leq x \leq w, 0 \leq y \leq h$. The patch and perturbation are optimized with 45000+ image frames from the KITTI dataset. The output from the respective pretrained model in response to the clean image is treated as pseudo ground truth to optimize the adversarial sample. Then the DFA loss between network's responses from a clean image and an attacked image is minimized.
Stochastic Gradient Descent (SDG) \cite{sutskever2013importance} with learning rate $0.01$ and Adam \cite{kingma2014adam} with learning rate $0.001$, $\beta_1 = 0.9$, $\beta_2 = 0.999$ are used to optimize perturbation and patch, respectively, in PyTorch \cite{paszke2019pytorch}.

\begin{figure*}[!ht]
	\centering
	\begin{adjustbox}{max width=\textwidth}
	\begin{tabular}{cccccccccccccc}
		
		\multicolumn{14}{c}{Clean features} \\
		\hline
		
		conv5 & conv6 & conv7 & iconv1 & iconv2 & iconv3 & iconv4 & iconv5 & iconv6 & iconv7 & disp1 & disp2 & disp3 & disp4 \\

		\includegraphics[width=0.05\linewidth]{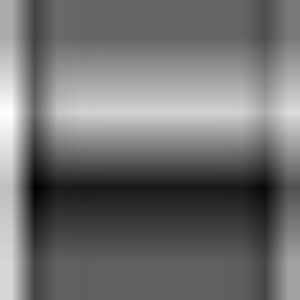} &
		\includegraphics[width=0.05\linewidth]{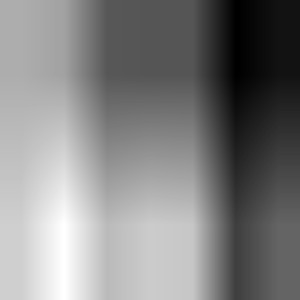} &
		\includegraphics[width=0.05\linewidth]{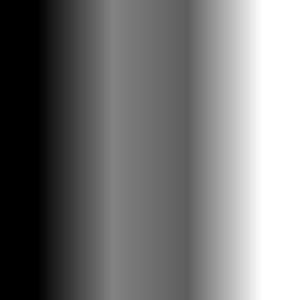} &
		\includegraphics[width=0.05\linewidth]{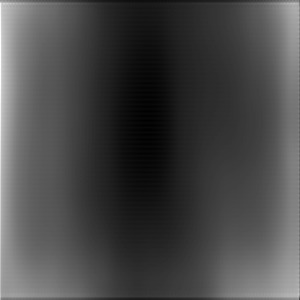} &
		\includegraphics[width=0.05\linewidth]{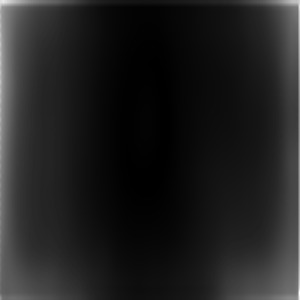} &
		
		\includegraphics[width=0.05\linewidth]{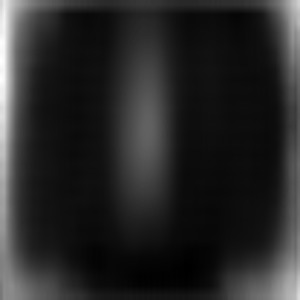} &
		\includegraphics[width=0.05\linewidth]{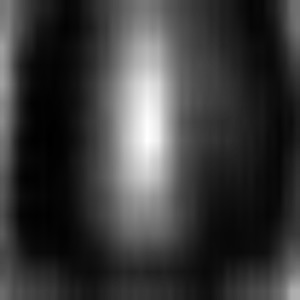} &
		\includegraphics[width=0.05\linewidth]{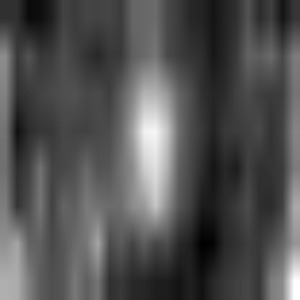} &
		\includegraphics[width=0.05\linewidth]{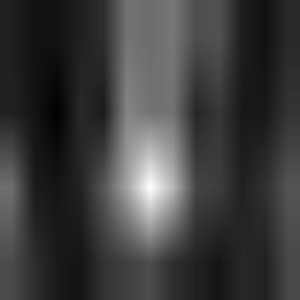} &
		\includegraphics[width=0.05\linewidth]{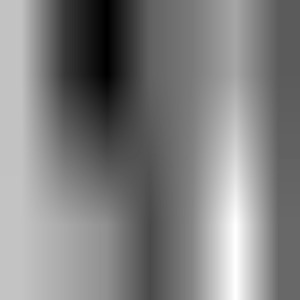} &
		\includegraphics[width=0.05\linewidth]{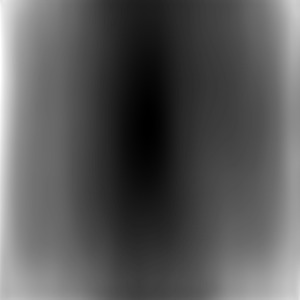} &
		\includegraphics[width=0.05\linewidth]{figs/feature/sfm_clean_disp1.jpg} &
		\includegraphics[width=0.05\linewidth]{figs/feature/sfm_clean_disp1.jpg} &
		\includegraphics[width=0.05\linewidth]{figs/feature/sfm_clean_disp1.jpg} \\
		
		\multicolumn{14}{c}{Attacked features} \\
		\hline
		
		conv5 & conv6 & conv7 & iconv1 & iconv2 & iconv3 & iconv4 & iconv5 & iconv6 & iconv7 & disp1 & disp2 & disp3 & disp4 \\

		\includegraphics[width=0.05\linewidth]{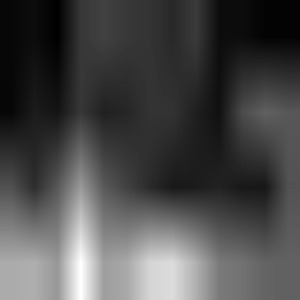} &
		\includegraphics[width=0.05\linewidth]{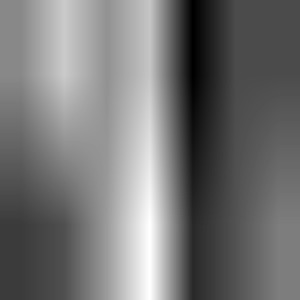} &
		\includegraphics[width=0.05\linewidth]{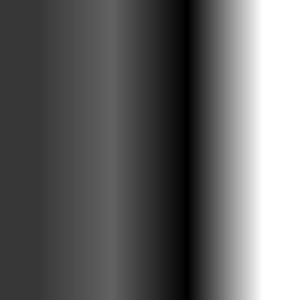} &
		\includegraphics[width=0.05\linewidth]{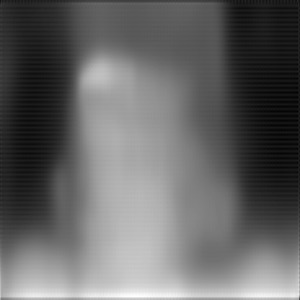} &
		\includegraphics[width=0.05\linewidth]{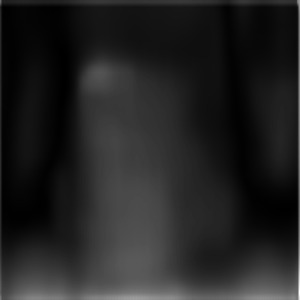} &
		\includegraphics[width=0.05\linewidth]{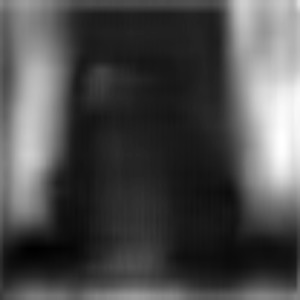} &
		\includegraphics[width=0.05\linewidth]{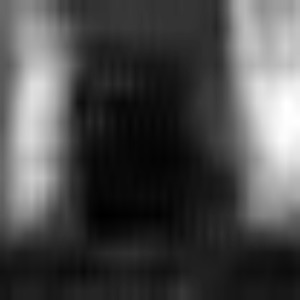} &
		\includegraphics[width=0.05\linewidth]{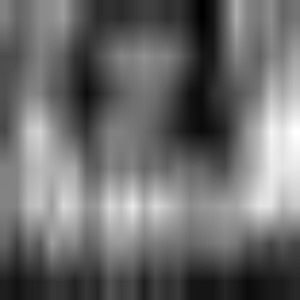} &
		\includegraphics[width=0.05\linewidth]{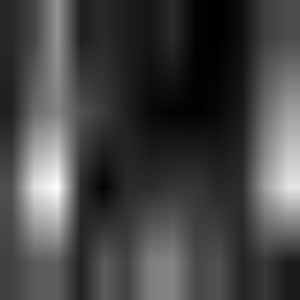} &
		\includegraphics[width=0.05\linewidth]{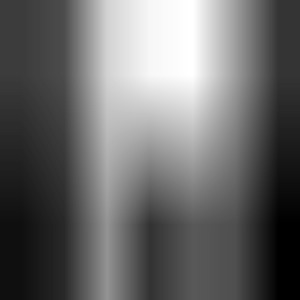} &
		\includegraphics[width=0.05\linewidth]{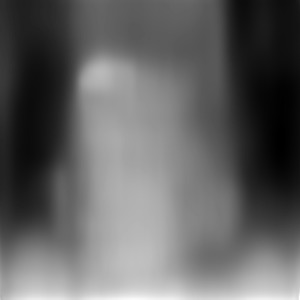} &
		\includegraphics[width=0.05\linewidth]{figs/feature/sfm_adv_disp1.jpg} &
		\includegraphics[width=0.05\linewidth]{figs/feature/sfm_adv_disp1.jpg} &
		\includegraphics[width=0.05\linewidth]{figs/feature/sfm_adv_disp1.jpg} \\
		
	\end{tabular}
	\end{adjustbox}
	\caption{Layer-wise SFM~\cite{zhou2017unsupervised} feature visualization. `convx' are encoder layers and `iconvx' are decoder layers.}
	\label{fig:feat_lvis_sfm}
\end{figure*}

\subsection{Black-box Attack}

In a black-box attack, the attacker has access to input and output but not to target network architecture and weights. The black-box attack is the most realistic attack, especially in safety-critical applications. These applications are designed in such a way that the attacker doesn't have direct access to the system. A global perturbation and patch learned from different models in the investigation are used for this attack. The global adversarial sample is optimized like a white-box attack. 
Even though adversarial samples are successful in attacking deep learning models with small perturbation or by corrupting few pixels, the transferability property of the adversarial samples makes it a severe threat. It has been shown that adversarial samples are cross-model transferable i.e., adversarial pertubation optimized to attack Model A can be used to attack Model B which deals with the same task. Let Model A be an open-sourced model trained to estimate per-pixel depth from an image, with the transferability property; an attacker can attack a closed Model B effectively using Model A's adversarial samples. In this study, we analyzed the transferability of the adversarial sample across all models under investigation. We also study cross-data transferability of the learned patch as shown in Figure~\ref{fig:cross_data}. Cross-data transferability examines how well an adversarial sampled can be optimized with a publicly available dataset A to attack a model on proprietary data. The adverisal patch learned from KITTI dataset was used to attack images from Cityscape~\cite{cordts2016cityscapes}, considering KITTI as open source data and cityscape as proprietary data. This transferability assures that publicly available data can be used to attack almost any related models.

\subsection{Evaluation}

The qualitative evalution of the attacks are conducted using metrics in \cite{eigen2014depth} with KITTI 2015 dataset \cite{geiger2013vision}. Root mean square error (RMSE) along with relative degradation (Rel) of RMSE are used to measure the vulnerability of the depth networks (Absolute relative error metric in supplementary material).
As shown in Table~\ref{tab:whitebox_patch_rmse} and Table~\ref{tab:whitebox_gpert_rmse}, its evident that all state-of-the-art models are vulnerable to adversarial samples in whitebox attacks. Our adversarial perturbation could attack all models very well and in few networks like DDVO and Mono2 just 5\% of corruption could cause more than 100\% damage to the depth maps. We can also see that corrupting less than even 1\% of the image using the adversarial patch can cause more than 20\% damage in the estimated depth map and also the damage extends significantly beyond the region of the patch thus making it a very dangerous real world attack.
The qualitative result in Table~\ref{tab:trans_pert_rmse} and Table~\ref{tab:trans_patch_rmse} assure successful black-box attack on all state-of-the-art monocular depth estimation networks.
\subsection{Analyze Deep Features}

Unlike most adversarial attacks, in this investigation, we corrupt the network's internal representation of an input image with our perturbation. In our experiment, striking the final layer alone didn't affect the prediction drastically for monocular depth estimation. The adversarial samples generated from existing methods still preserve a high level of input information in its deep hidden layers. These extracted features obstruct the desired attack in the model.  To counter this issue, we targeted deep features representation of the input image to generate adversarial noise. We visualized deep features of attacked and clean image in Figure~\ref{fig:feat_lvis_sfm}. For better visualization, feature outputs of zero image are taken as a clean image, and perturbation added to zero image is considered as an attacked image. It shows how each network layer reacts to the clean image and how the decoder layers react to an attacked image and exhibit an activation inversion property. From the visualization we can see that suppressed and non-dominant activations are exaggerated while dominant activations are weakened thus justifying our earlier claim.
DFA makes the internal feature representation hollow and thereby creating a more robust attack on the final prediction of the network.

\section{Conclusion}

In this paper, we proposed an effective adversarial attack on monocular depth estimators. We explored both perturbation and patch type attacks in this study. The adversarial samples are designed with DFA loss to destroy the internal representation of the network, thereby resulting in a more vigorous attack. The attacks are evaluated extensively with white-box and black-box testing on KITTI dataset. The deep features of the network are visualized for a better understanding of the proposed attack. Also, cross-data transferability of the attack is examined.
	
\bibliography{refs}


\clearpage
\onecolumn
\begin{center}
	{\Large Supplementary Material:\\ 
		Monocular Depth Estimators: Vulnerabilities and Attacks}
\end{center}

\setcounter{section}{0}

\section{White-box attack}

\subsection{Patch attack}

\begin{figure*}[!ht]
  \centering
\newcommand{\turnheightnew}{0.15\columnwidth}
\centering

\begin{center}

\begin{tabular}{@{\hskip 0.5mm}c@{\hskip 0.5mm}c@{\hskip 0.5mm}c@{\hskip 0.5mm}c@{\hskip 0.5mm}c@{}}

& Attacked image & Clean depth & Attacked depth & Depth gap \\

\rotatebox[origin=l]{45}{SFM} & 
\includegraphics[width=0.22\linewidth]{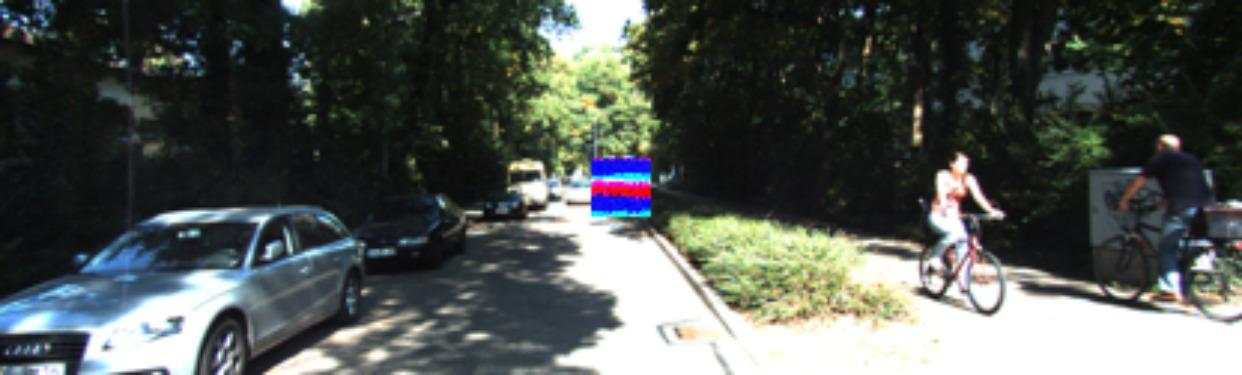} &
\includegraphics[width=0.22\linewidth]{figs/patch/qual/sfm/82_d.jpg} &
\includegraphics[width=0.22\linewidth]{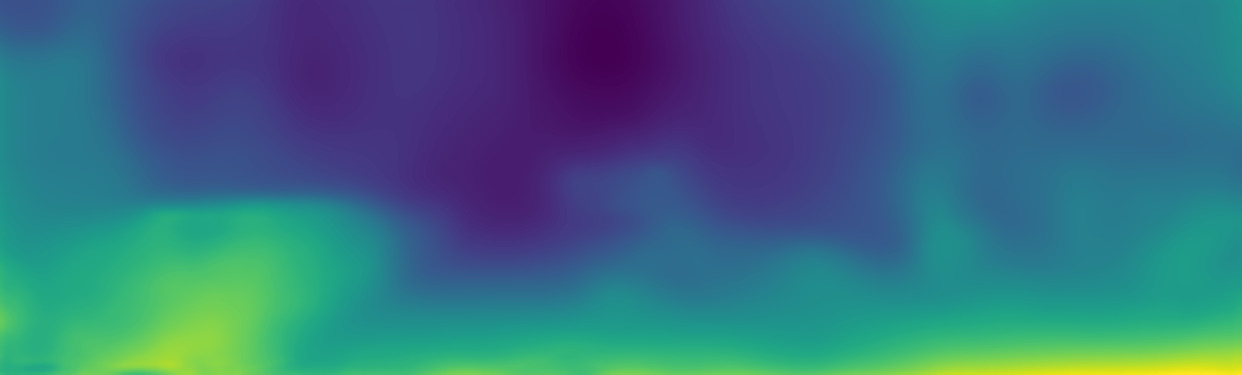} &
\includegraphics[width=0.22\linewidth]{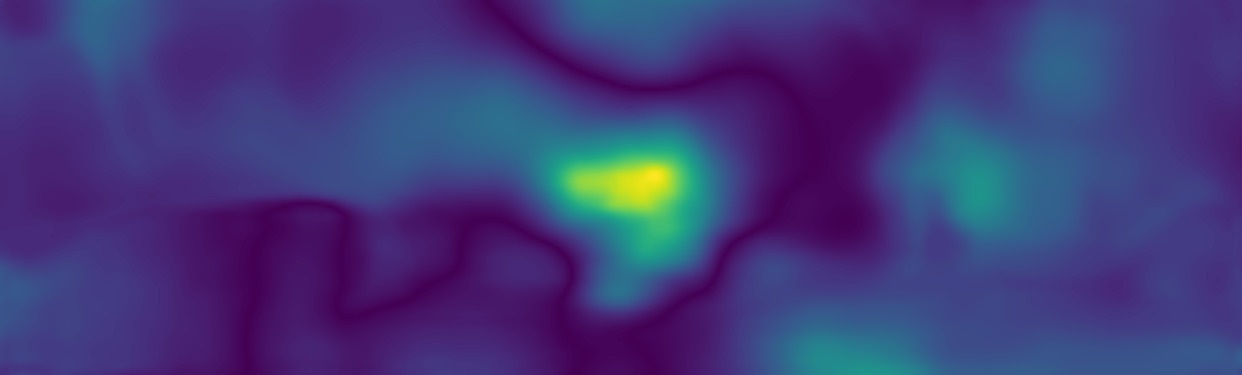} \\

\rotatebox[origin=l]{45}{DDVO} & 
\includegraphics[width=0.22\linewidth]{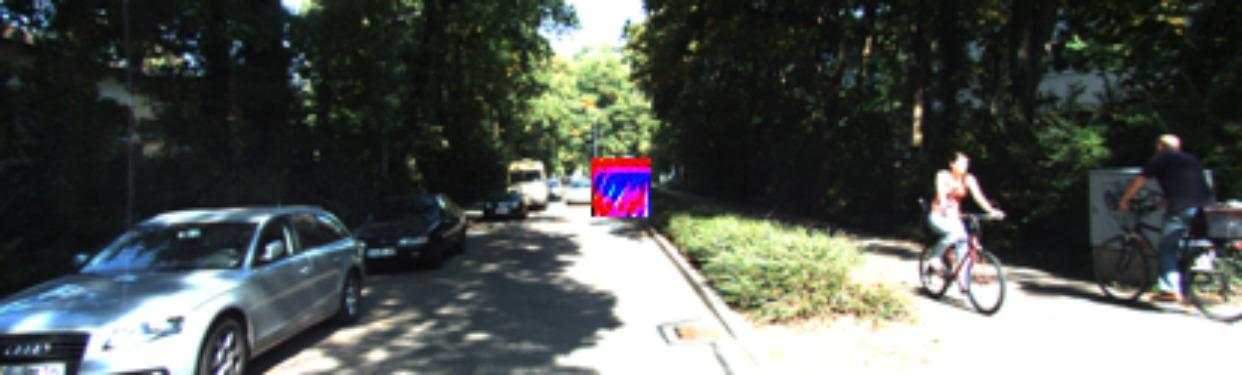} &
\includegraphics[width=0.22\linewidth]{figs/patch/qual/ddvo/82_d.jpg} &
\includegraphics[width=0.22\linewidth]{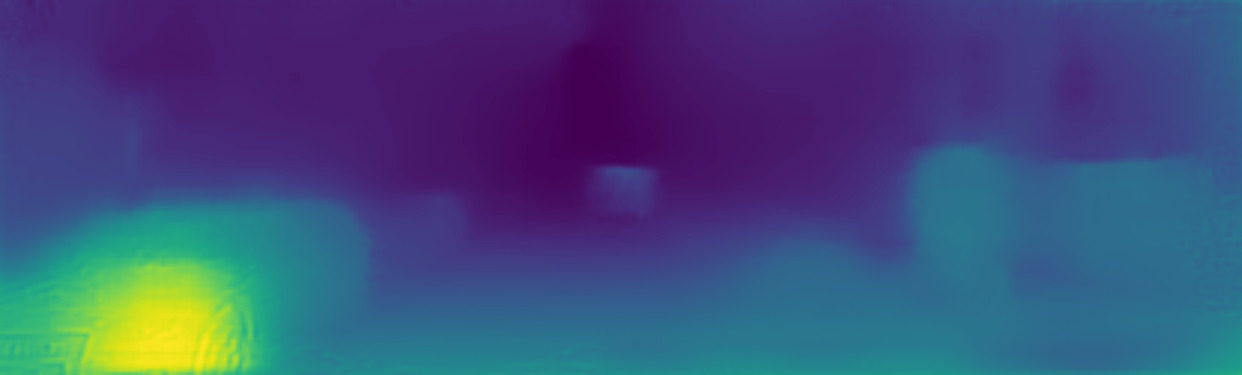} &
\includegraphics[width=0.22\linewidth]{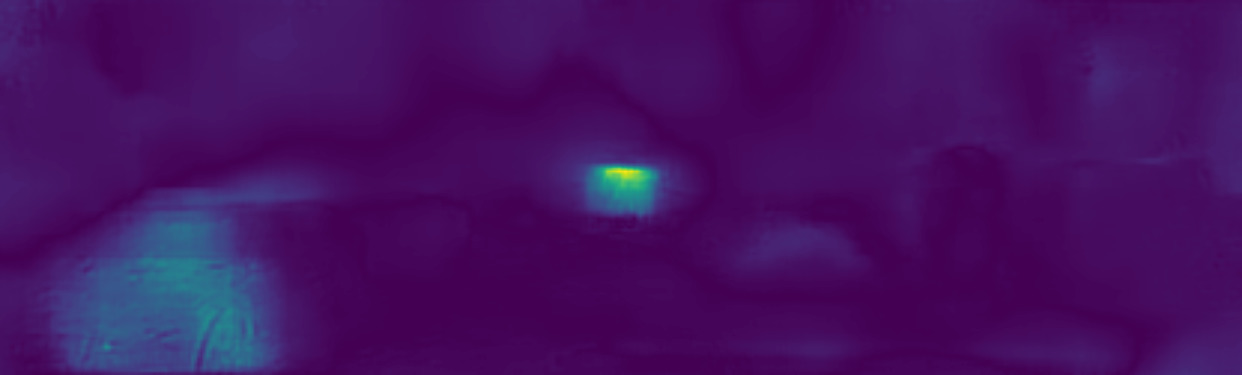} \\

\rotatebox[origin=l]{45}{B2F} & 
\includegraphics[width=0.22\linewidth]{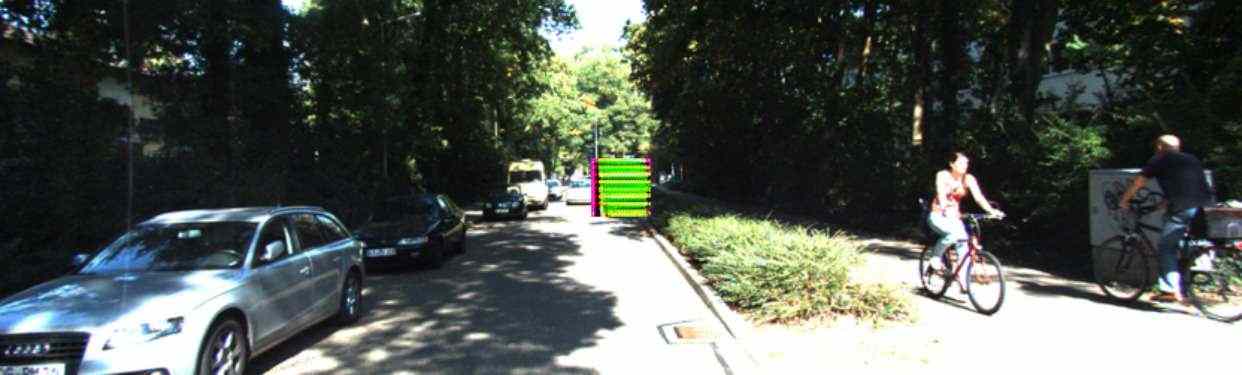} &
\includegraphics[width=0.22\linewidth]{figs/patch/qual/b2f/82_d.jpg} &
\includegraphics[width=0.22\linewidth]{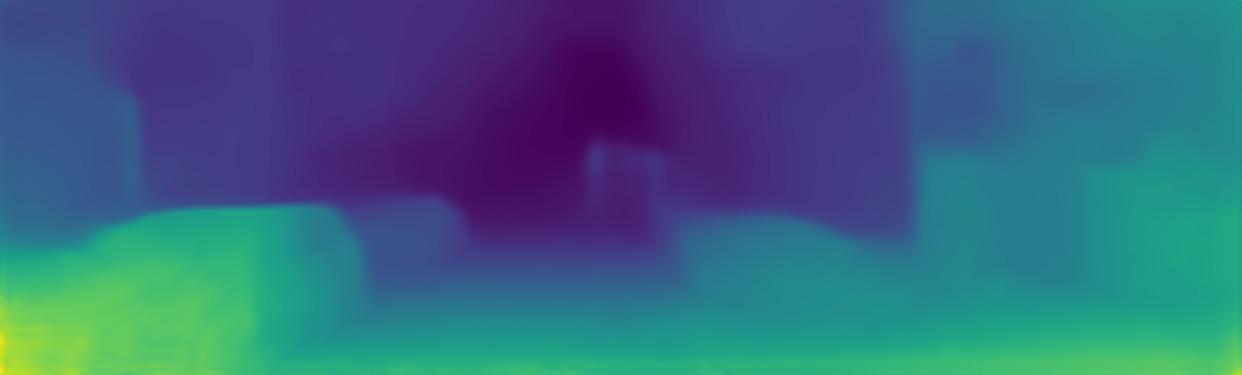} &
\includegraphics[width=0.22\linewidth]{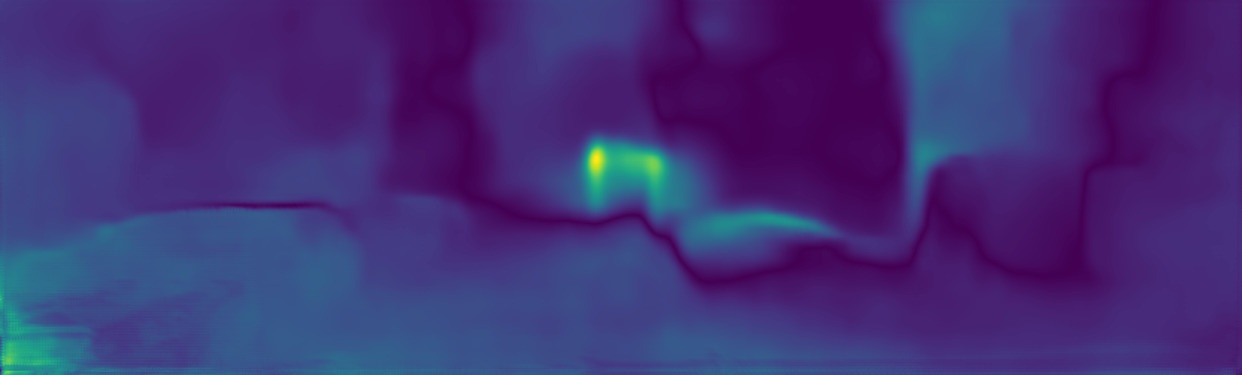} \\

\rotatebox[origin=l]{45}{SCSFM} & 
\includegraphics[width=0.22\linewidth]{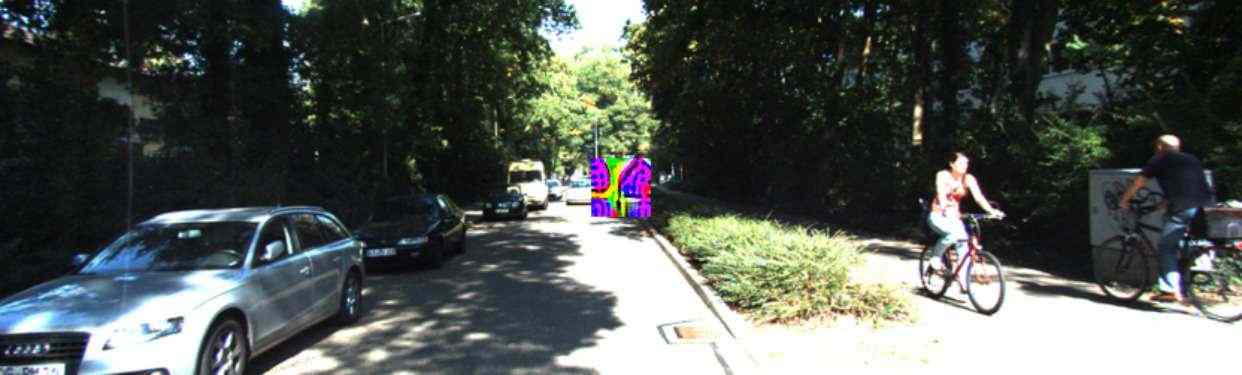} &
\includegraphics[width=0.22\linewidth]{figs/patch/qual/scsfm/82_d.jpg} &
\includegraphics[width=0.22\linewidth]{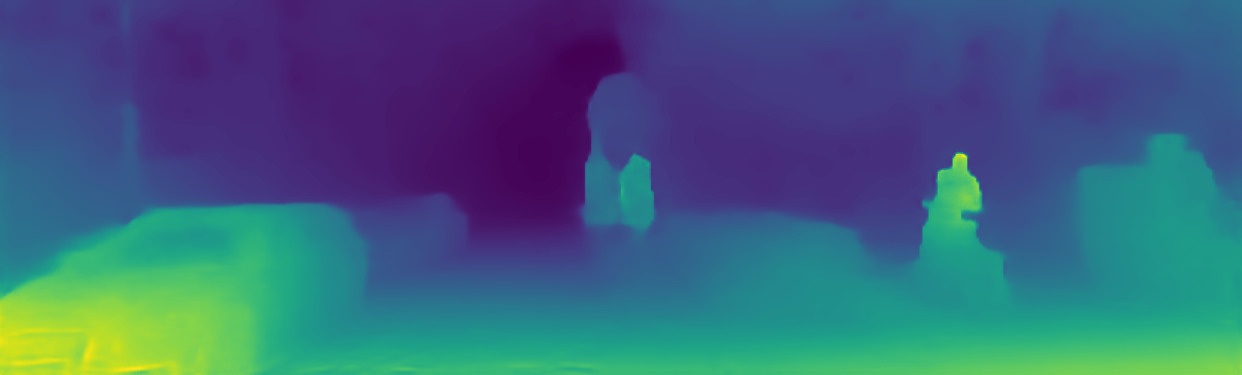} &
\includegraphics[width=0.22\linewidth]{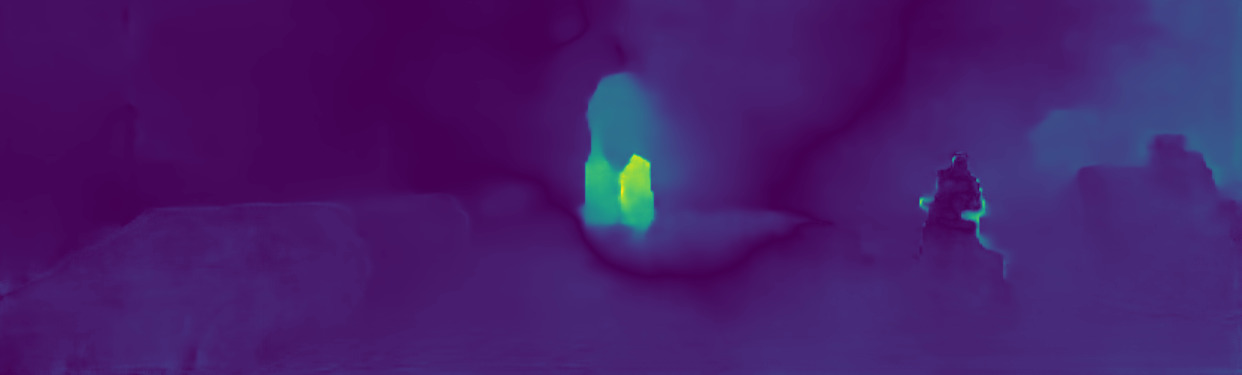} \\

\rotatebox[origin=l]{45}{Mono1} & 
\includegraphics[width=0.22\linewidth]{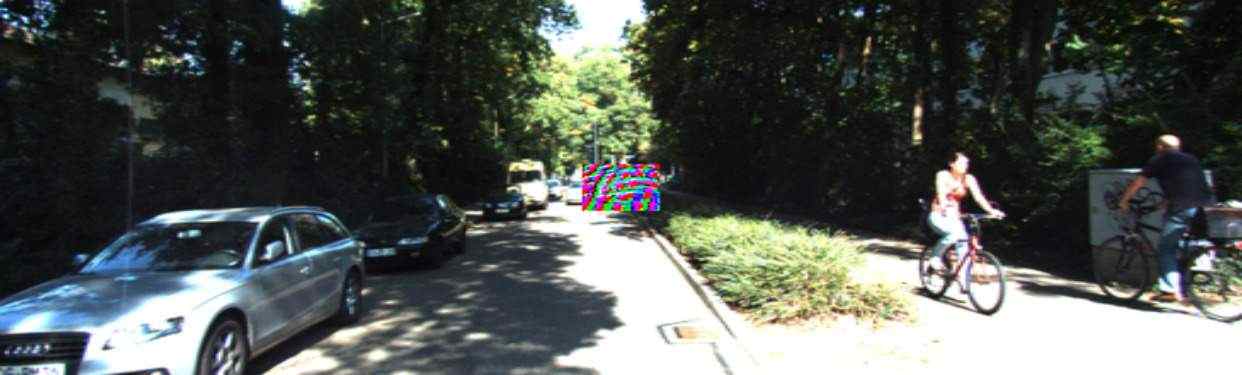} &
\includegraphics[width=0.22\linewidth]{figs/patch/qual/mono1/82_d.jpg} &
\includegraphics[width=0.22\linewidth]{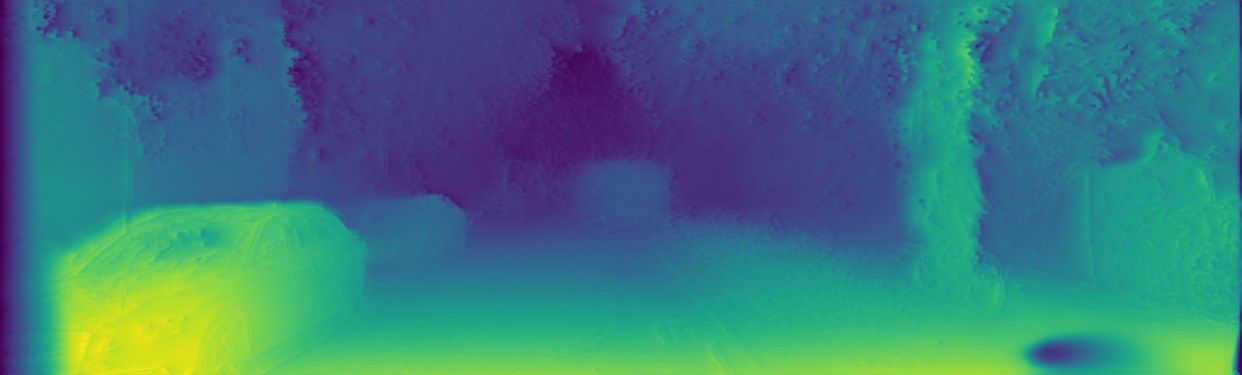} &
\includegraphics[width=0.22\linewidth]{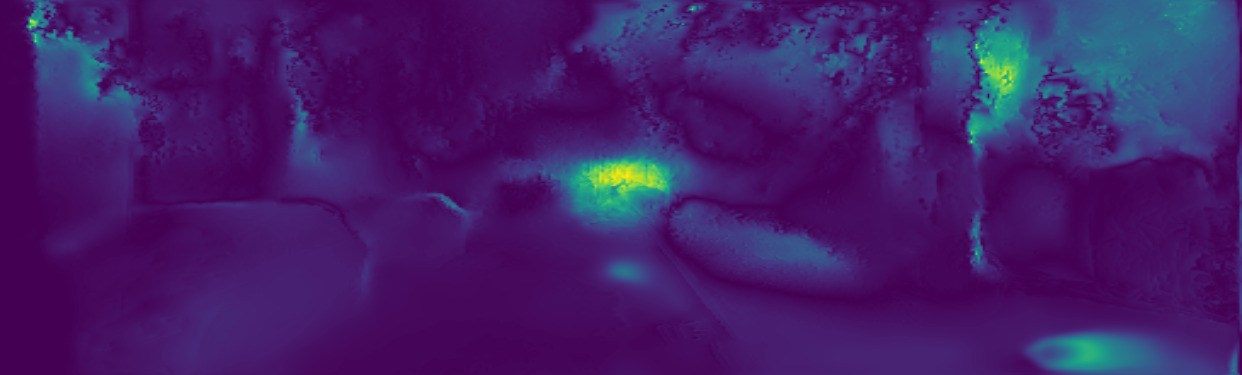} \\

\rotatebox[origin=l]{45}{Mono2} & 
\includegraphics[width=0.22\linewidth]{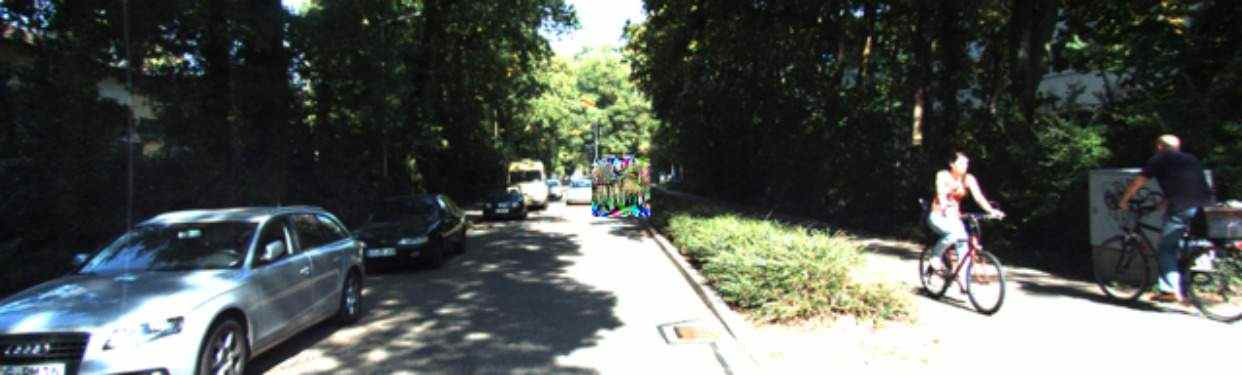} &
\includegraphics[width=0.22\linewidth]{figs/patch/qual/mono2/82_d.jpg} &
\includegraphics[width=0.22\linewidth]{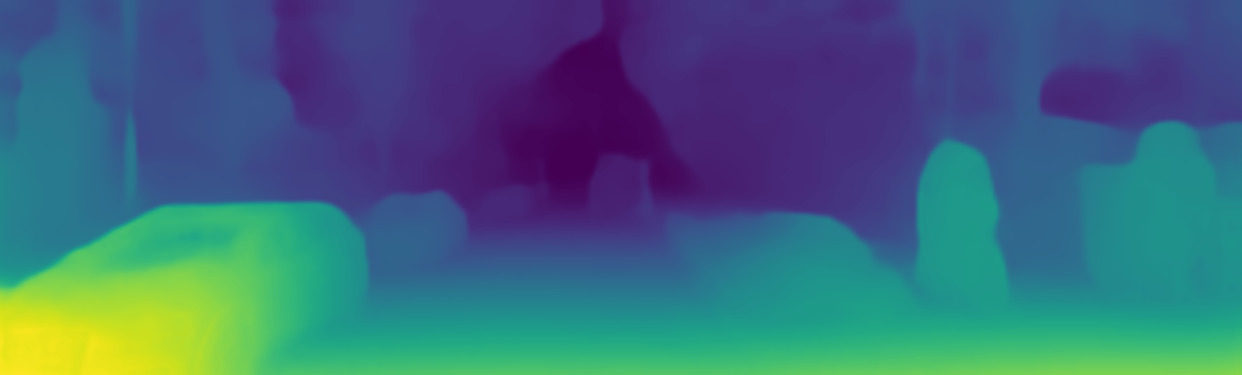} &
\includegraphics[width=0.22\linewidth]{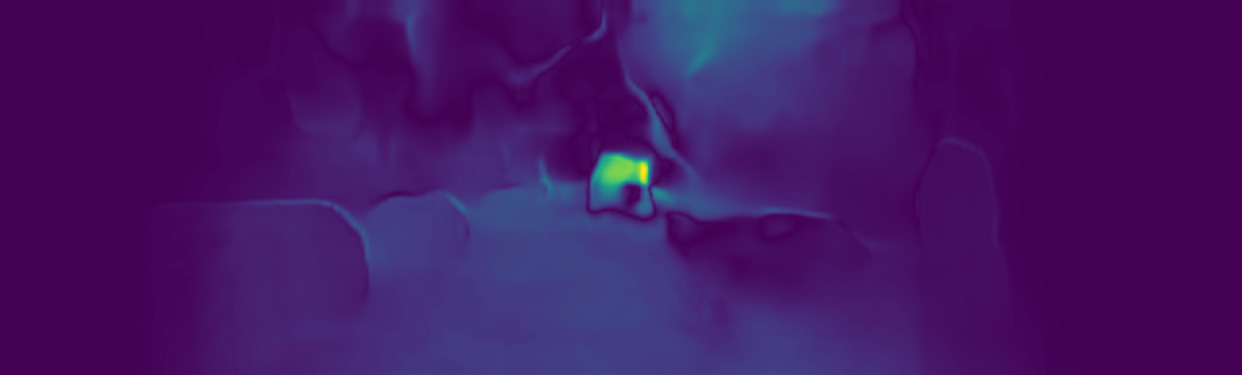} \\

\end{tabular}
\end{center}
\caption{White-box patch test with sample of size $60 \times 60$.}
\label{fig:white_patch_adv60}
\end{figure*}

\begin{figure*}[!ht]
  \centering
\newcommand{\turnheightnew}{0.15\columnwidth}
\centering

\begin{center}

\begin{tabular}{@{\hskip 0.5mm}c@{\hskip 0.5mm}c@{\hskip 0.5mm}c@{\hskip 0.5mm}c@{\hskip 0.5mm}c@{}}

& Attacked image & Clean depth & Attacked depth & Depth gap \\

\rotatebox[origin=l]{45}{SFM} & 
\includegraphics[width=0.22\linewidth]{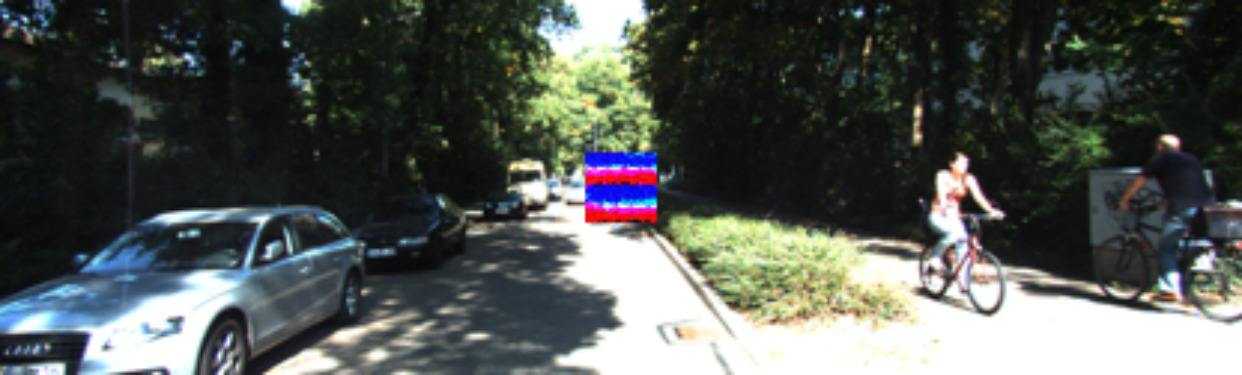} &
\includegraphics[width=0.22\linewidth]{figs/patch/qual/sfm/82_d.jpg} &
\includegraphics[width=0.22\linewidth]{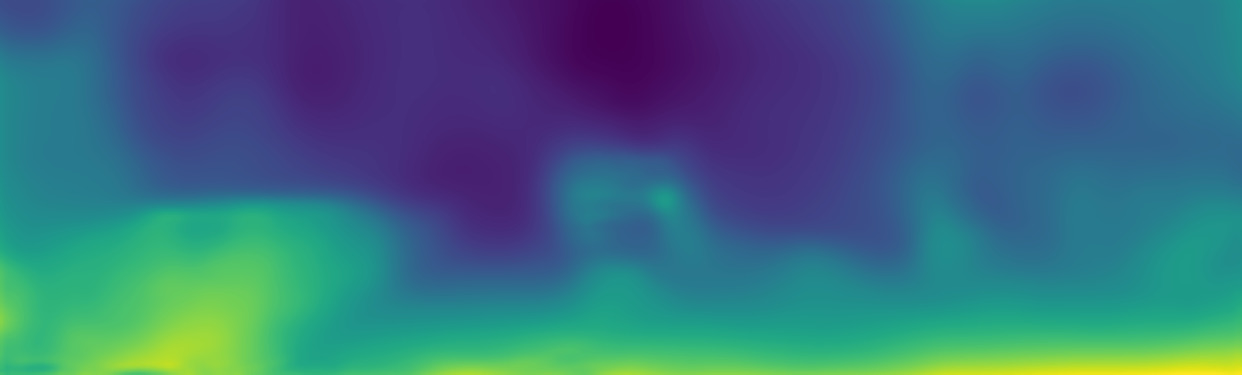} &
\includegraphics[width=0.22\linewidth]{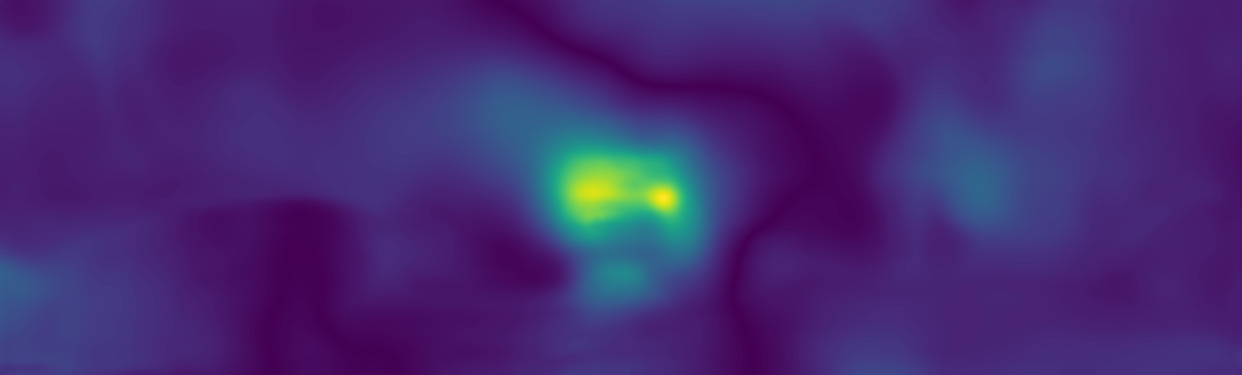} \\

\rotatebox[origin=l]{45}{DDVO} & 
\includegraphics[width=0.22\linewidth]{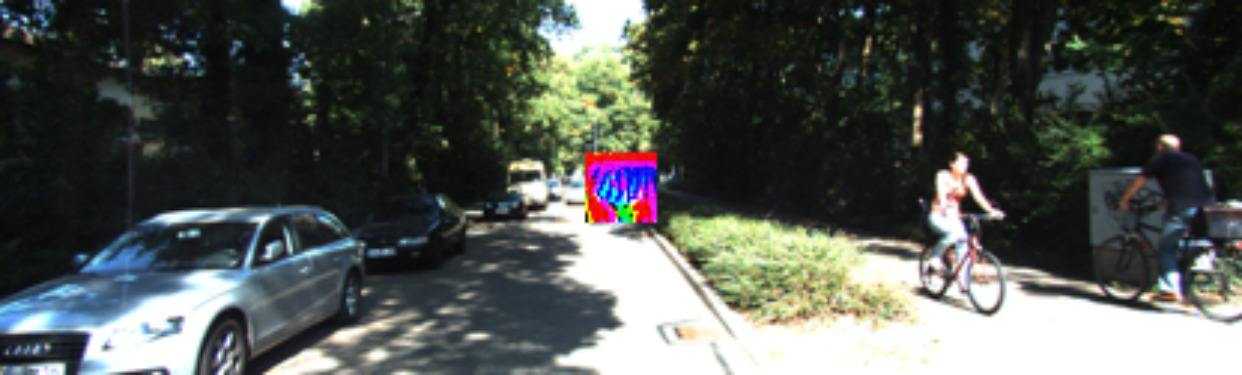} &
\includegraphics[width=0.22\linewidth]{figs/patch/qual/ddvo/82_d.jpg} &
\includegraphics[width=0.22\linewidth]{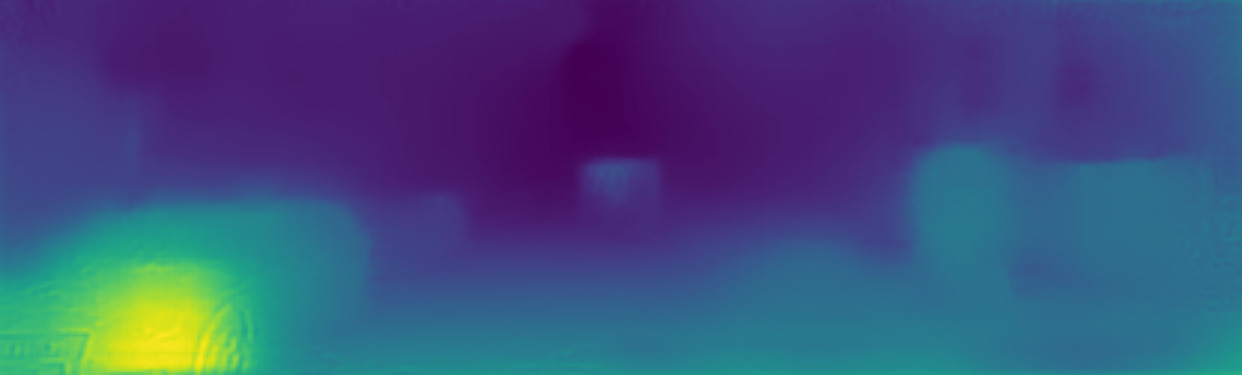} &
\includegraphics[width=0.22\linewidth]{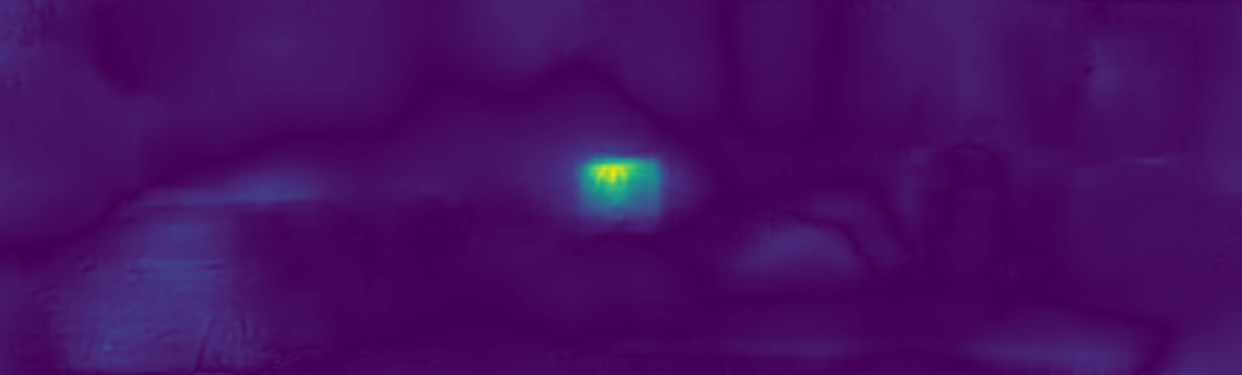} \\

\rotatebox[origin=l]{45}{B2F} & 
\includegraphics[width=0.22\linewidth]{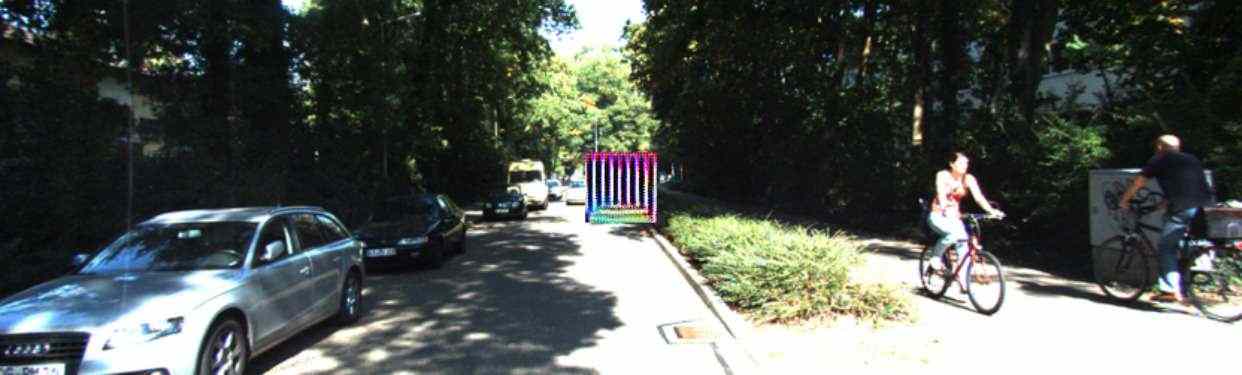} &
\includegraphics[width=0.22\linewidth]{figs/patch/qual/b2f/82_d.jpg} &
\includegraphics[width=0.22\linewidth]{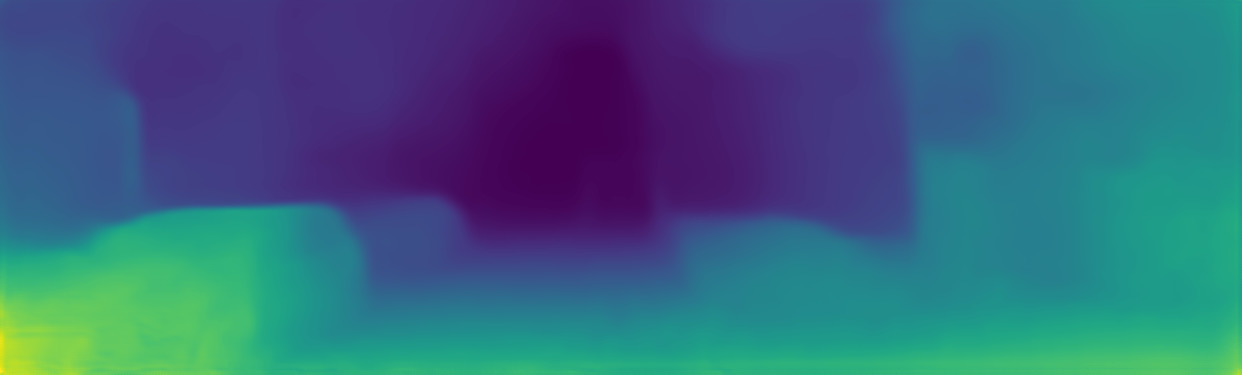} &
\includegraphics[width=0.22\linewidth]{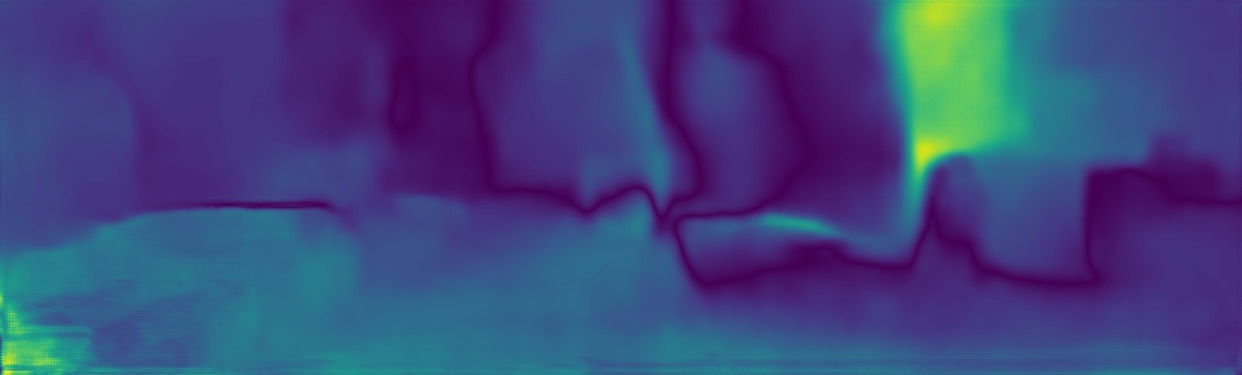} \\

\rotatebox[origin=l]{45}{SCSFM} & 
\includegraphics[width=0.22\linewidth]{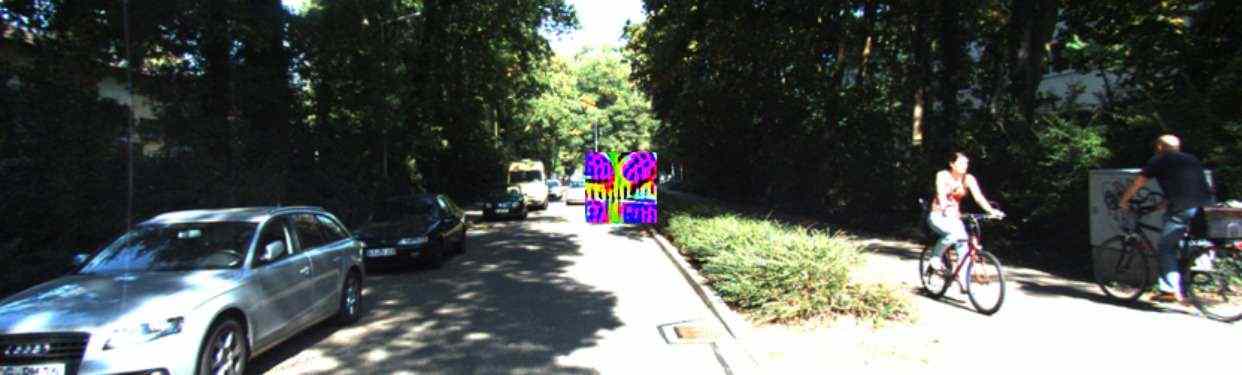} &
\includegraphics[width=0.22\linewidth]{figs/patch/qual/scsfm/82_d.jpg} &
\includegraphics[width=0.22\linewidth]{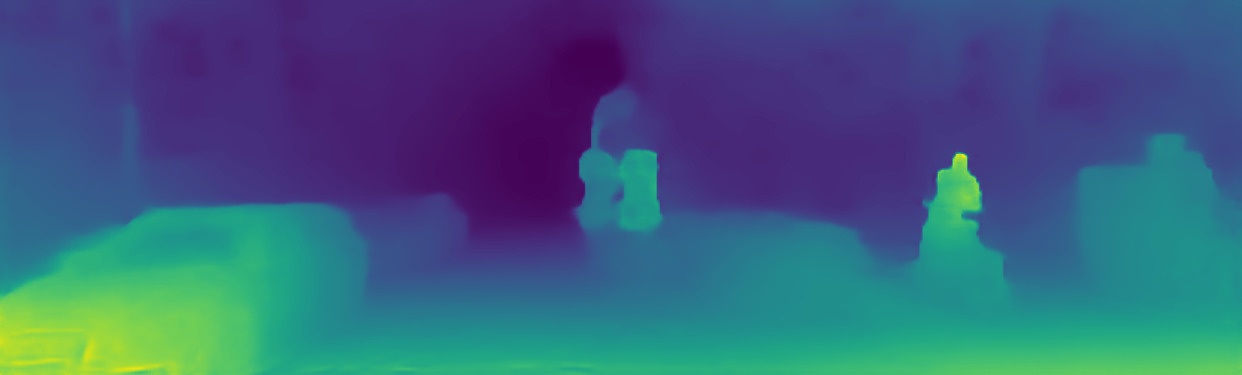} &
\includegraphics[width=0.22\linewidth]{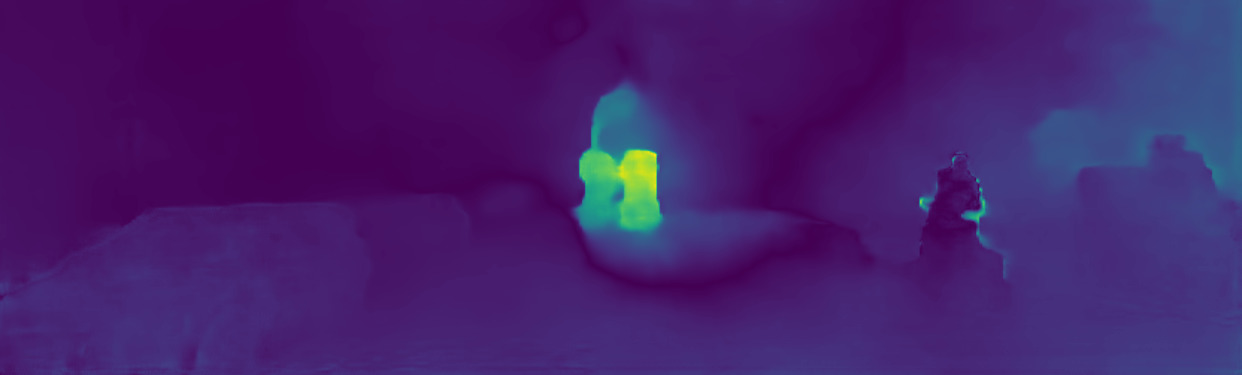} \\

\rotatebox[origin=l]{45}{Mono1} & 
\includegraphics[width=0.22\linewidth]{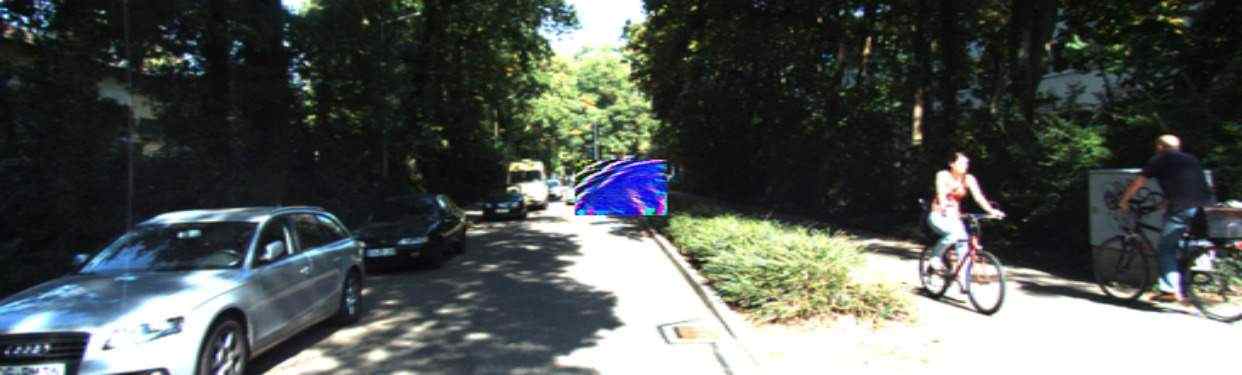} &
\includegraphics[width=0.22\linewidth]{figs/patch/qual/mono1/82_d.jpg} &
\includegraphics[width=0.22\linewidth]{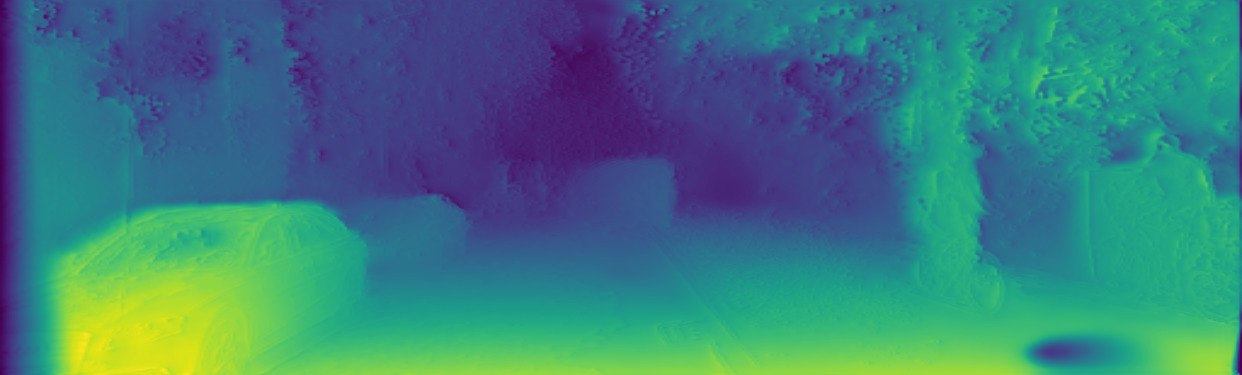} &
\includegraphics[width=0.22\linewidth]{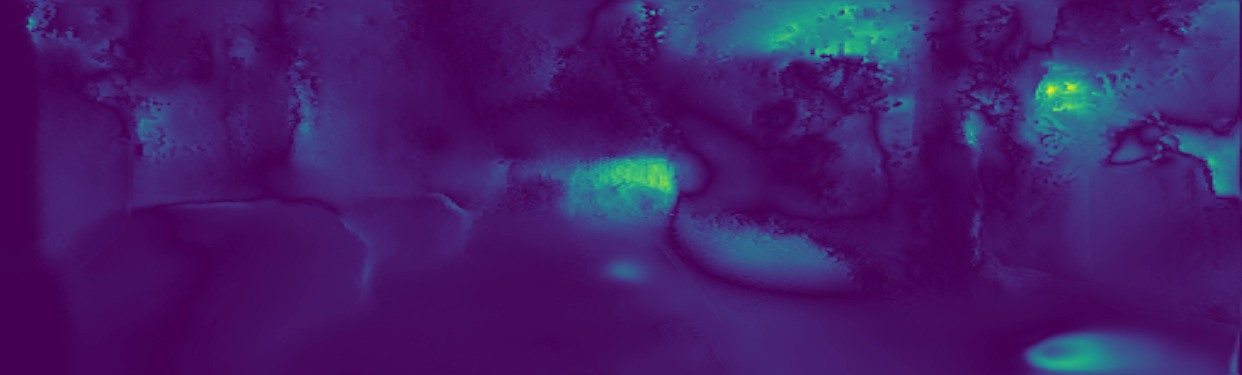} \\

\rotatebox[origin=l]{45}{Mono2} & 
\includegraphics[width=0.22\linewidth]{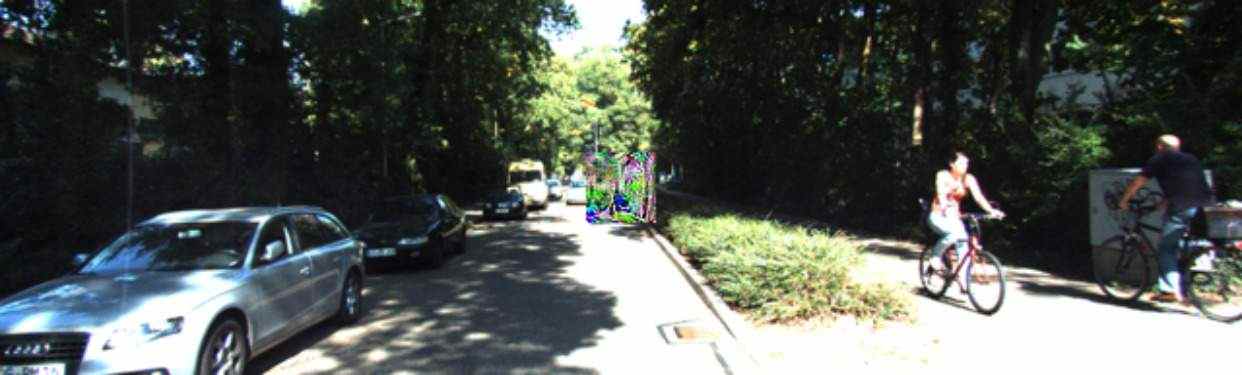} &
\includegraphics[width=0.22\linewidth]{figs/patch/qual/mono2/82_d.jpg} &
\includegraphics[width=0.22\linewidth]{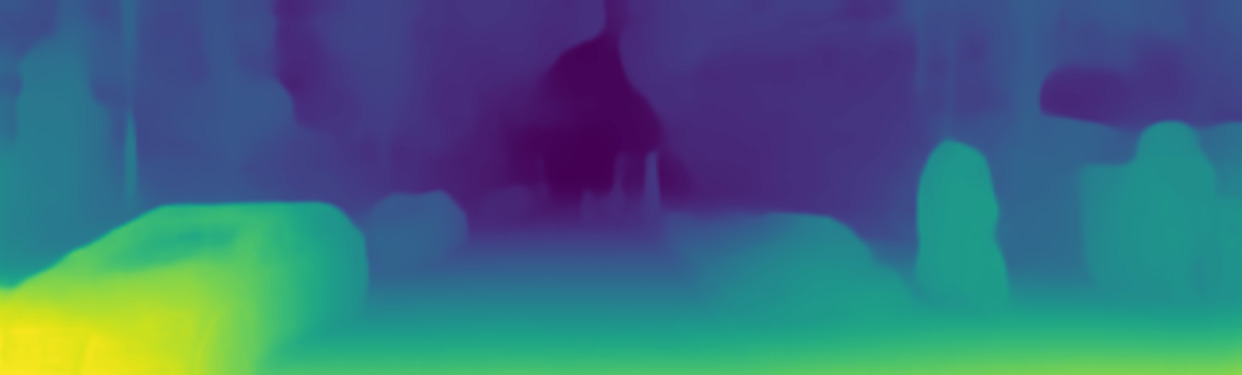} &
\includegraphics[width=0.22\linewidth]{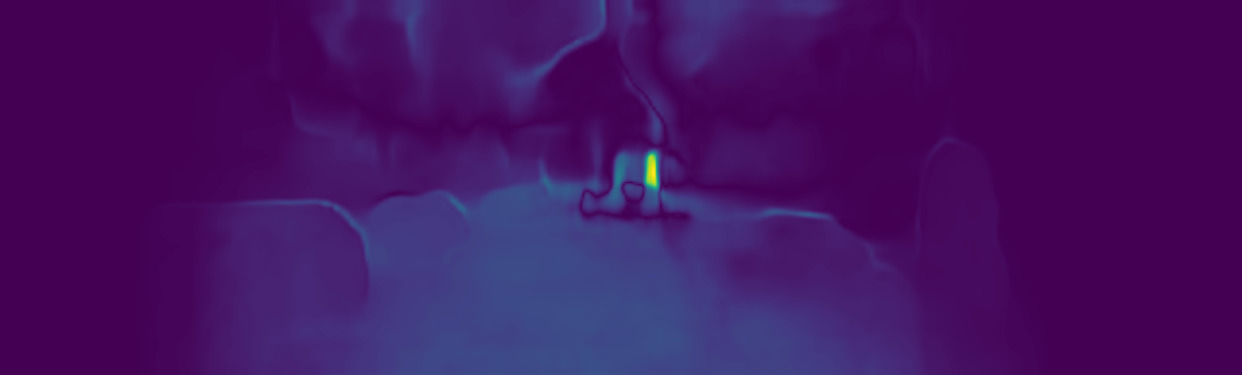} \\

\end{tabular}
\end{center}
\caption{White-box patch test with patch size $72 \times 72$.}
\label{fig:white_patch_adv72}
\end{figure*}

\begin{figure*}[!ht]
  \centering
\newcommand{\turnheightnew}{0.15\columnwidth}
\centering

\begin{center}

\begin{tabular}{@{\hskip 0.5mm}c@{\hskip 0.5mm}c@{\hskip 0.5mm}c@{\hskip 0.5mm}c@{\hskip 0.5mm}c@{}}

& Attacked image & Clean depth & Attacked depth & Depth gap \\

\rotatebox[origin=l]{45}{SFM} & 
\includegraphics[width=0.22\linewidth]{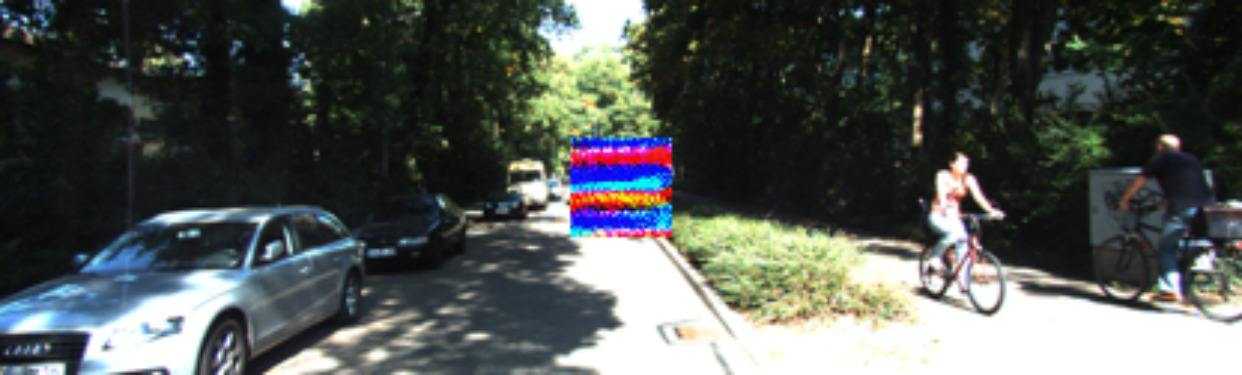} &
\includegraphics[width=0.22\linewidth]{figs/patch/qual/sfm/82_d.jpg} &
\includegraphics[width=0.22\linewidth]{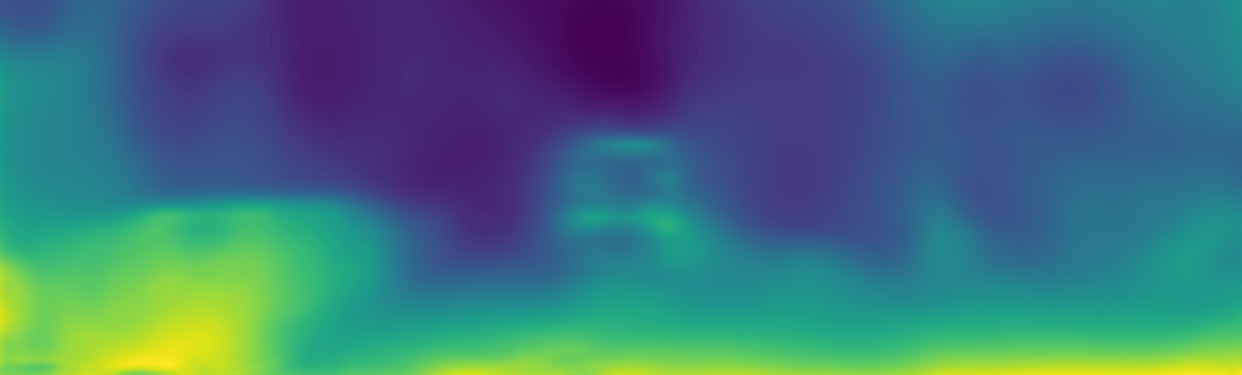} &
\includegraphics[width=0.22\linewidth]{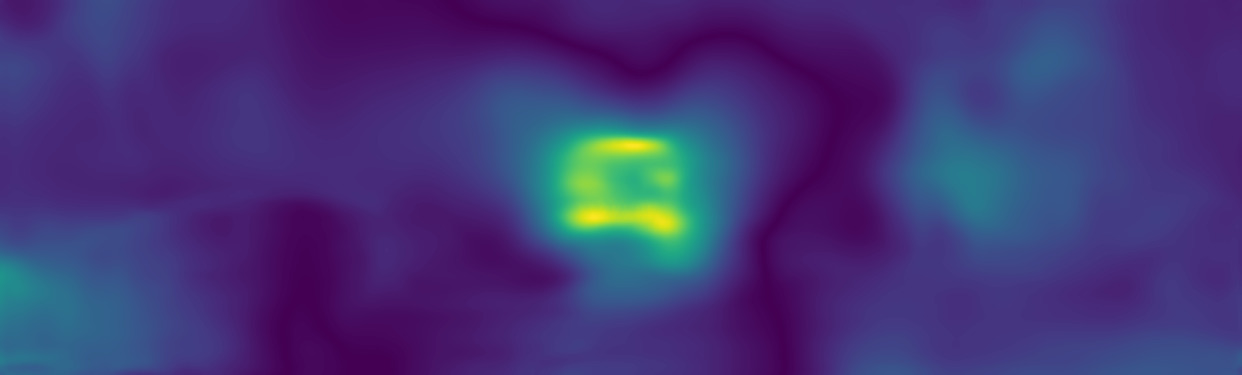} \\

\rotatebox[origin=l]{45}{DDVO} & 
\includegraphics[width=0.22\linewidth]{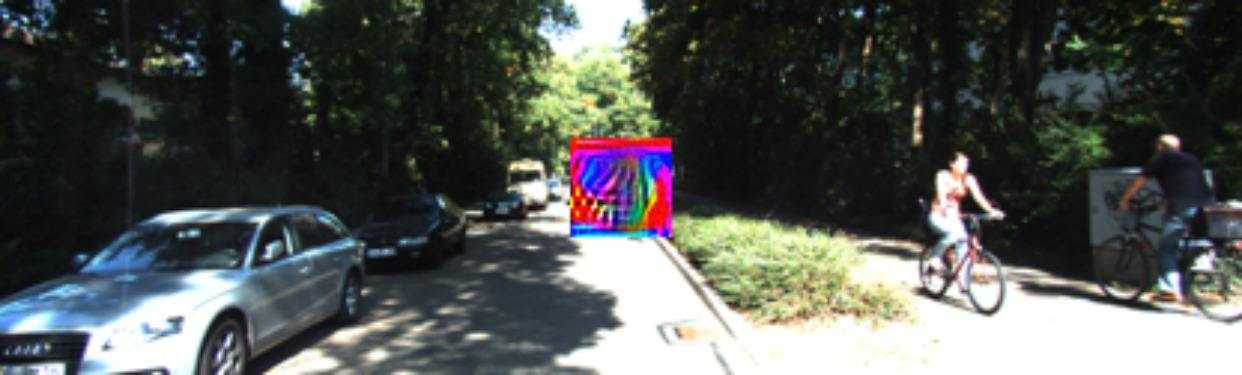} &
\includegraphics[width=0.22\linewidth]{figs/patch/qual/ddvo/82_d.jpg} &
\includegraphics[width=0.22\linewidth]{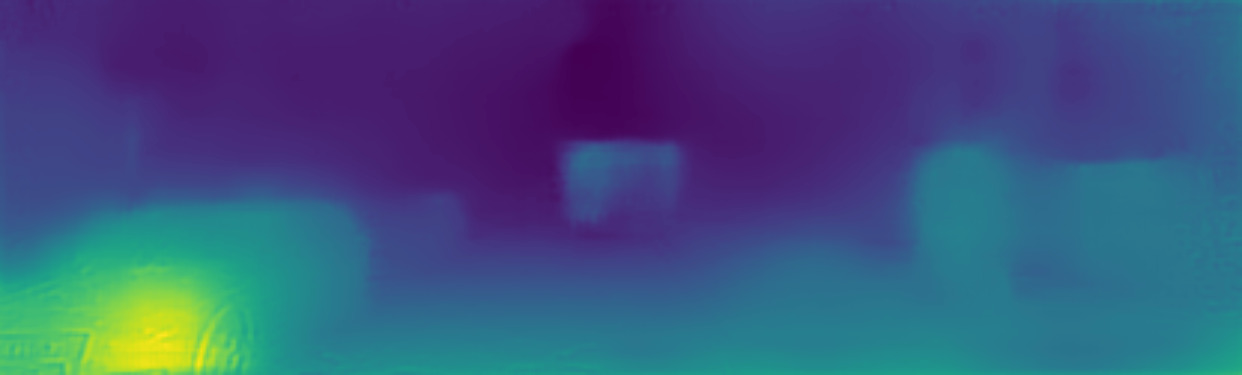} &
\includegraphics[width=0.22\linewidth]{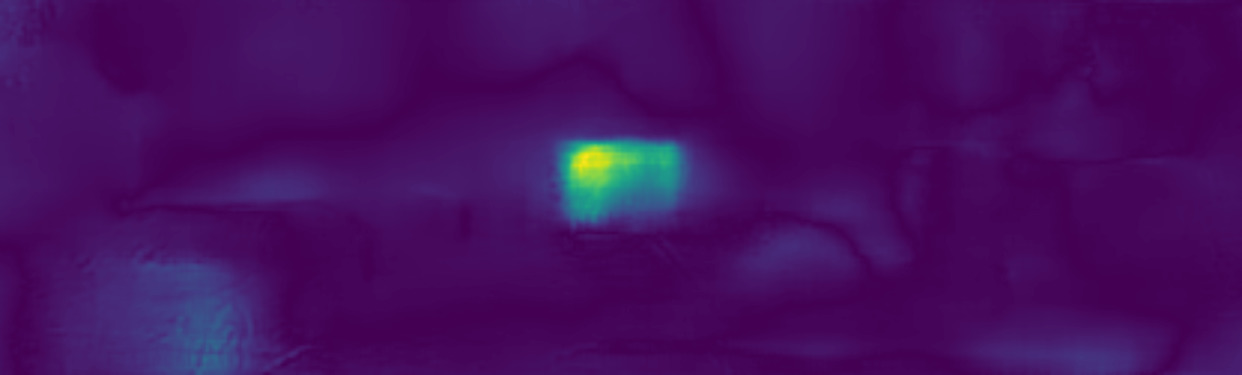} \\

\rotatebox[origin=l]{45}{B2F} & 
\includegraphics[width=0.22\linewidth]{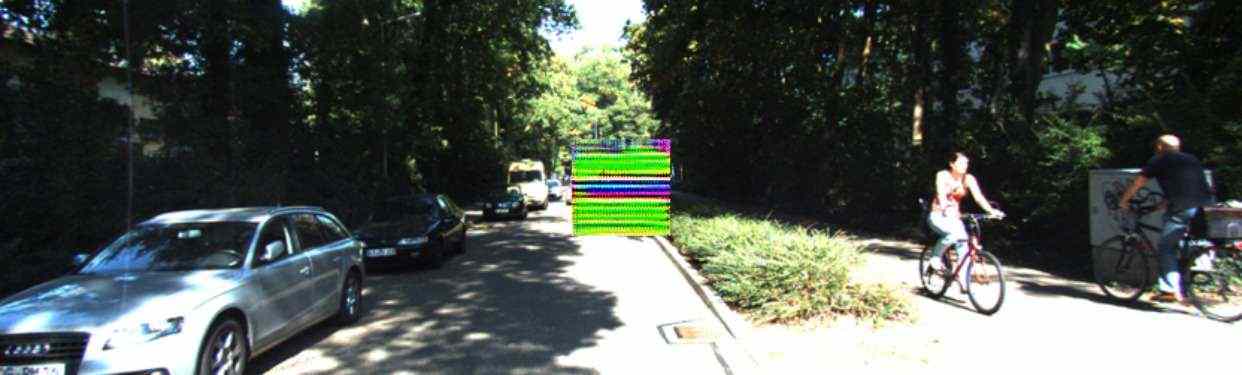} &
\includegraphics[width=0.22\linewidth]{figs/patch/qual/b2f/82_d.jpg} &
\includegraphics[width=0.22\linewidth]{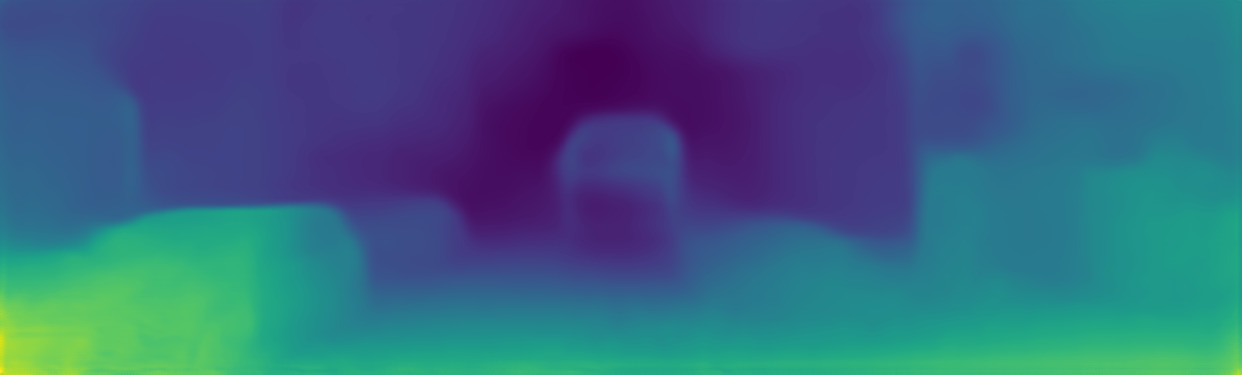} &
\includegraphics[width=0.22\linewidth]{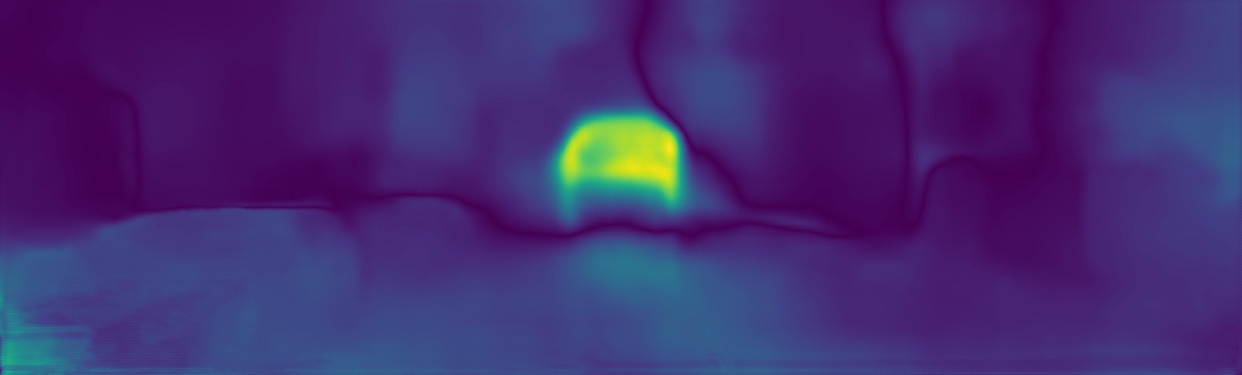} \\

\rotatebox[origin=l]{45}{SCSFM} & 
\includegraphics[width=0.22\linewidth]{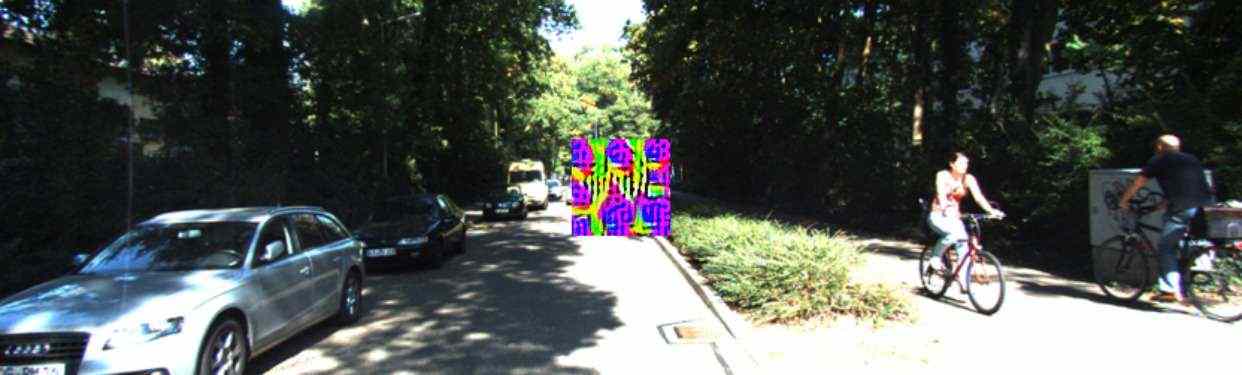} &
\includegraphics[width=0.22\linewidth]{figs/patch/qual/scsfm/82_d.jpg} &
\includegraphics[width=0.22\linewidth]{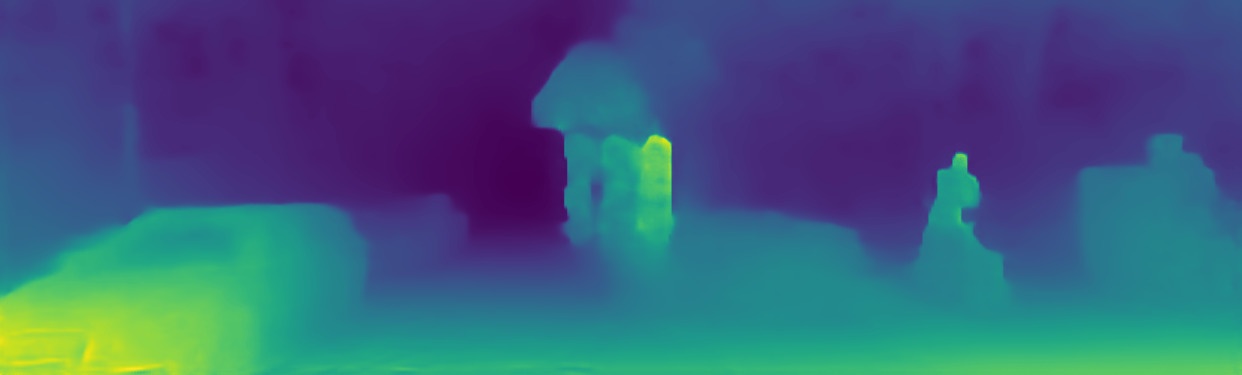} &
\includegraphics[width=0.22\linewidth]{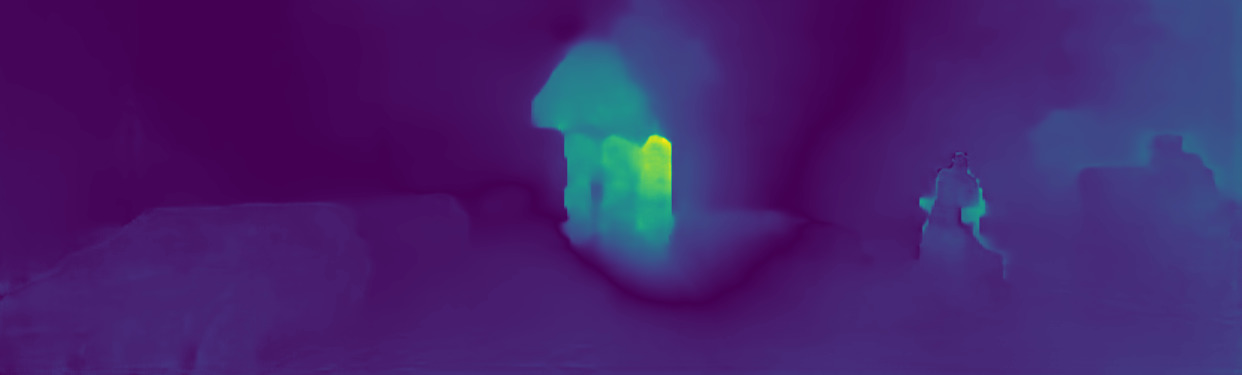} \\

\rotatebox[origin=l]{45}{Mono1} & 
\includegraphics[width=0.22\linewidth]{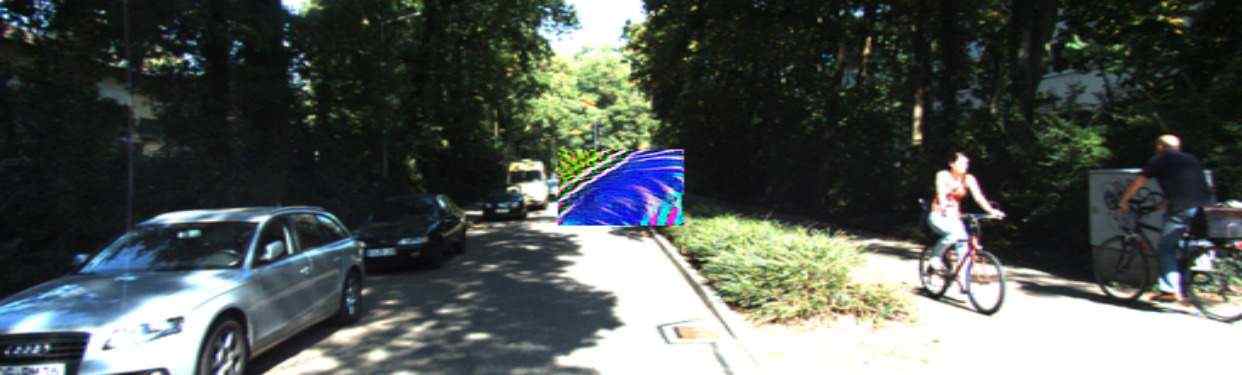} &
\includegraphics[width=0.22\linewidth]{figs/patch/qual/mono1/82_d.jpg} &
\includegraphics[width=0.22\linewidth]{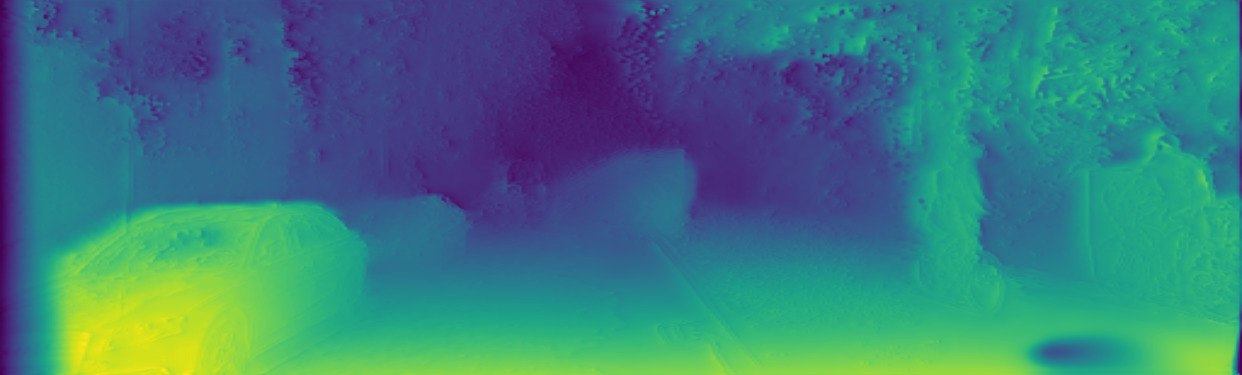} &
\includegraphics[width=0.22\linewidth]{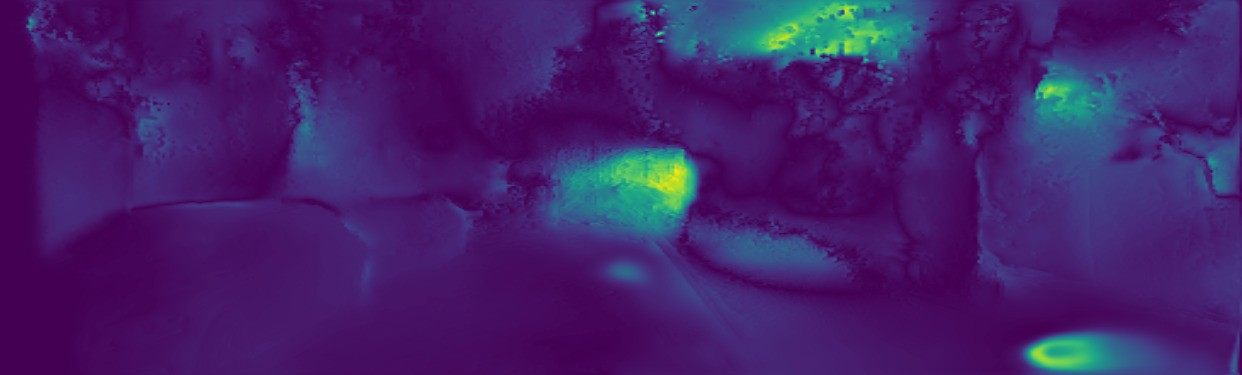} \\

\rotatebox[origin=l]{45}{Mono2} & 
\includegraphics[width=0.22\linewidth]{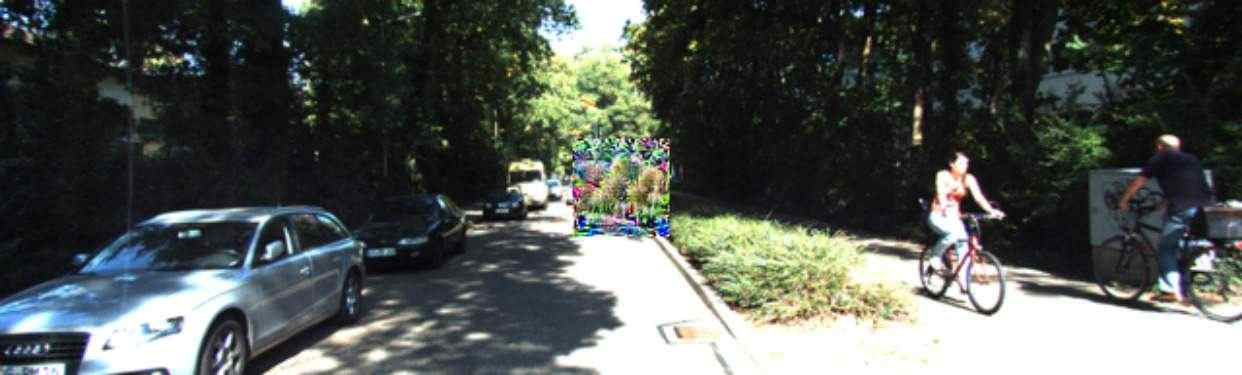} &
\includegraphics[width=0.22\linewidth]{figs/patch/qual/mono2/82_d.jpg} &
\includegraphics[width=0.22\linewidth]{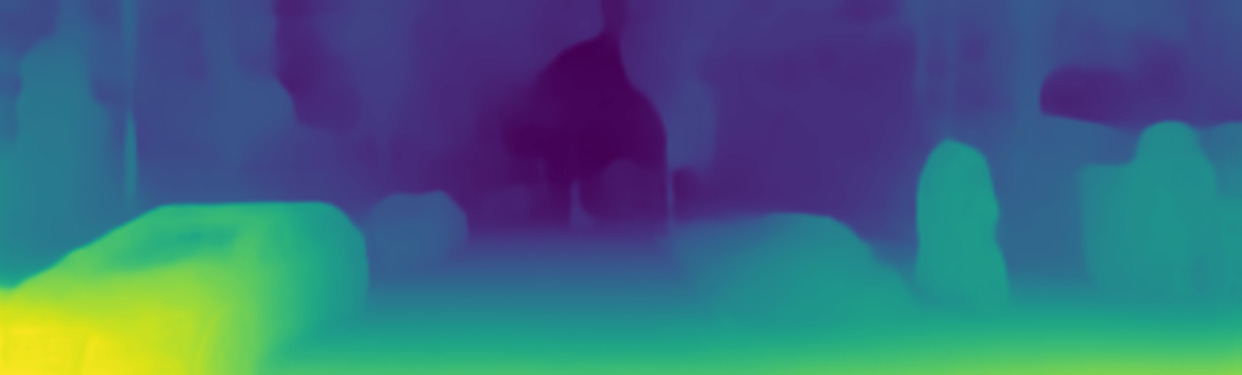} &
\includegraphics[width=0.22\linewidth]{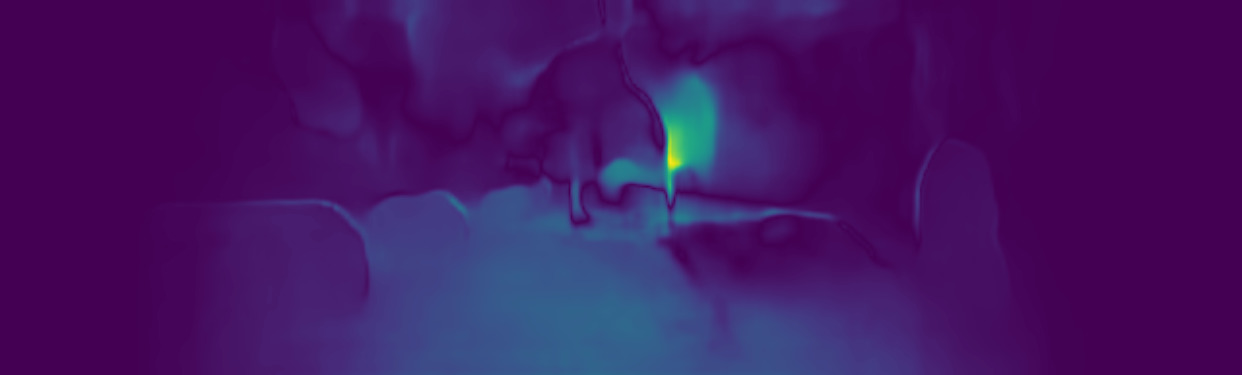} \\

\end{tabular}
\end{center}
\caption{White-box patch test with patch size $100 \times 100$.}
\label{fig:white_patch_adv100}
\end{figure*}

\begin{table*}[!ht]
	\centering
	\begin{adjustbox}{max width=\textwidth}
	\begin{tabular}{l|c|c|c|c|c|c|c|c|c}
		\hline
		\multirow{2}{*}{Methods} & \multirow{2}{*}{Clean} & \multicolumn{8}{c}{Attacked} \\
		\cline{3-10}
		&    & \multicolumn{2}{c|}{$50 \times 50$} & \multicolumn{2}{c|}{$60 \times 60$} & \multicolumn{2}{c|}{$72 \times 72$} & \multicolumn{2}{c}{$100 \times 100$}  \\
		\cline{3-10}
		& Absrel   & Absrel & Rel (\%)              & Absrel & Rel (\%)             & Absrel & Rel (\%)             & Absrel & Rel (\%)  \\
		\hline
		SFM~\cite{zhou2017unsupervised}                      & 0.1755                 & 0.1887 & 8                & 0.2088 & 19               & 0.2193 & 25               & 0.2345 & 34    \\
		DDVO~\cite{wang2018learning}                     & 0.1488                 & 0.1517 & 2                & 0.1519 & 3                & 0.1575 & 6                & 0.1712 & 16    \\
		B2F~\cite{janai2018unsupervised}                      & 0.1358                 & 0.153  & 13               & 0.1573 & 16               & 0.1573 & 16               & 0.1697 & 25    \\
		SCSFM~\cite{bian2019unsupervised}                    & 0.1283                 & 0.1533 & 20               & 0.1652 & 29               & 0.1759 & 38               & 0.2046 & 60    \\
		Mono1~\cite{godard2017unsupervised}                    & 0.1095                 & 0.1166 & 7                & 0.1205 & 11               & 0.1271 & 17               & 0.1423 & 30    \\
		Mono2~\cite{godard2019digging}                    & 0.1123                 & 0.1239 & 11               & 0.1263 & 13               & 0.1317 & 18               & 0.1513 & 35    \\
		\hline
	\end{tabular}
	\end{adjustbox}
	\caption{White-box patch attack at different patch sizes.}
	\label{whitebox_patch_absrel}
\end{table*}

\newpage

\subsection{Perturbation attack}

\begin{figure*}[!ht]
\centering
\newcommand{\turnheightnew}{0.15\columnwidth}
\centering

\begin{center}

\begin{tabular}{@{\hskip 0.5mm}c@{\hskip 0.5mm}c@{\hskip 0.5mm}c@{\hskip 0.5mm}c@{\hskip 0.5mm}c@{}}

& Attacked image & Clean depth & Attacked depth & Depth gap \\

\rotatebox[origin=l]{45}{SFM} & 
\includegraphics[width=0.22\linewidth]{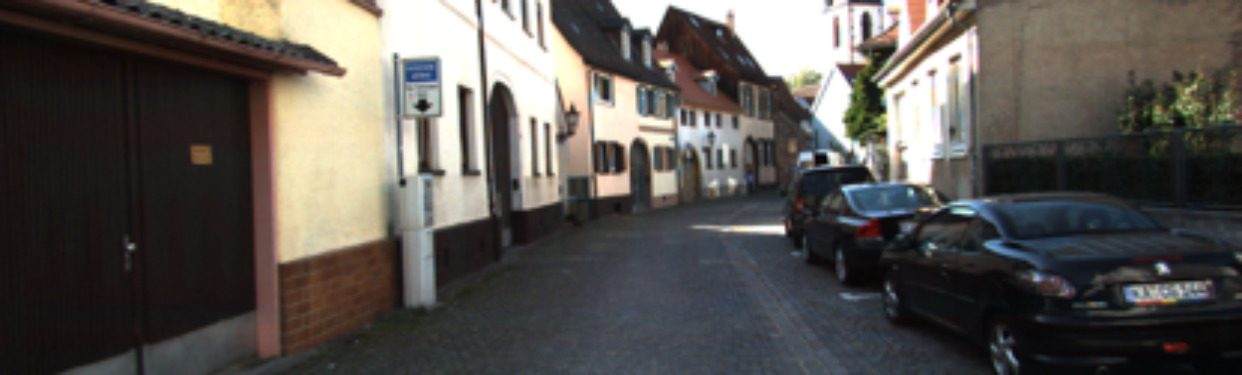} &
\includegraphics[width=0.22\linewidth]{figs/pert/qual/sfm/403_d.jpg} &
\includegraphics[width=0.22\linewidth]{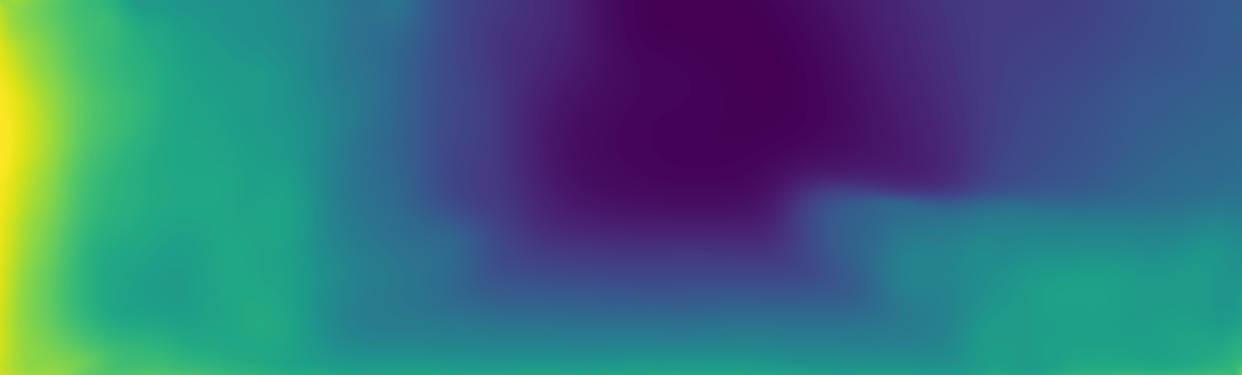} &
\includegraphics[width=0.22\linewidth]{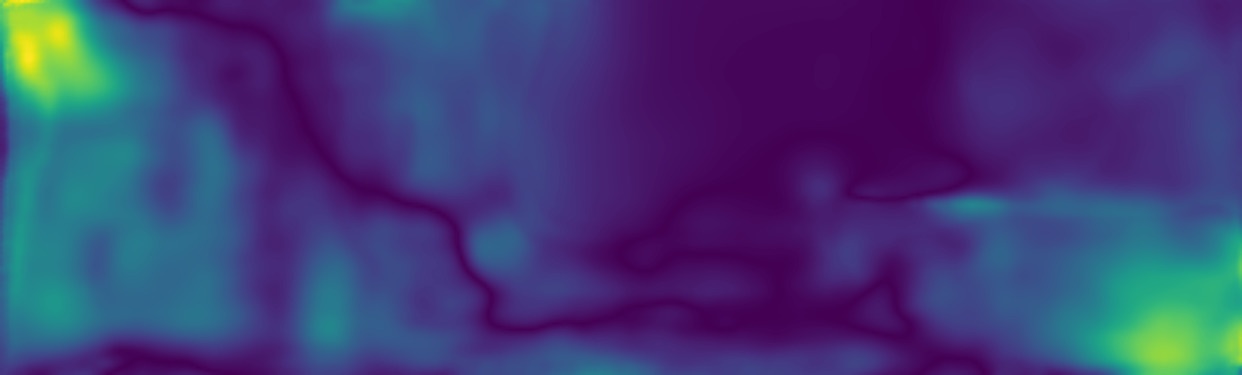} \\

\rotatebox[origin=l]{45}{DDVO} & 
\includegraphics[width=0.22\linewidth]{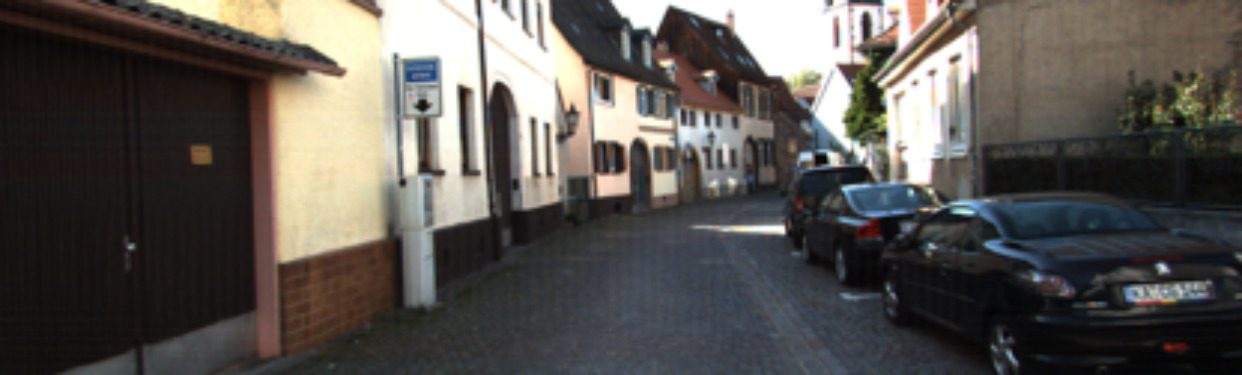} &
\includegraphics[width=0.22\linewidth]{figs/pert/qual/ddvo/403_d.jpg} &
\includegraphics[width=0.22\linewidth]{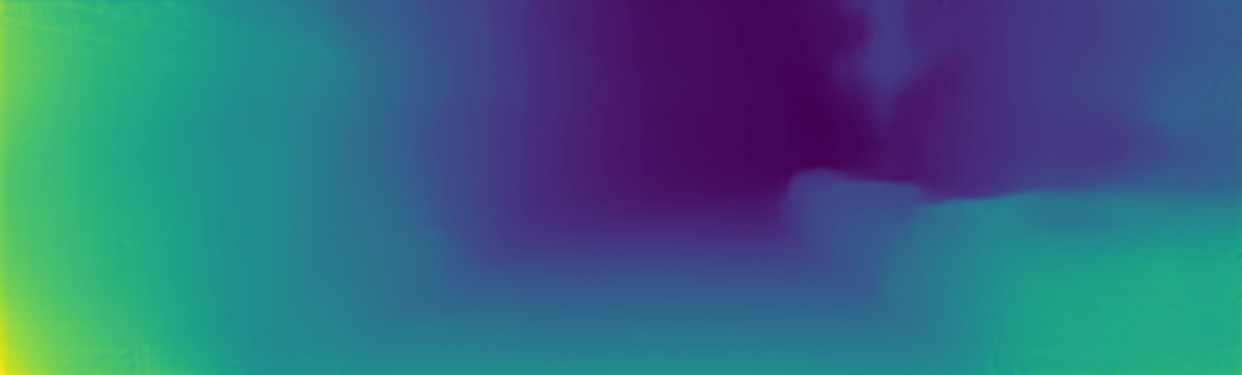} &
\includegraphics[width=0.22\linewidth]{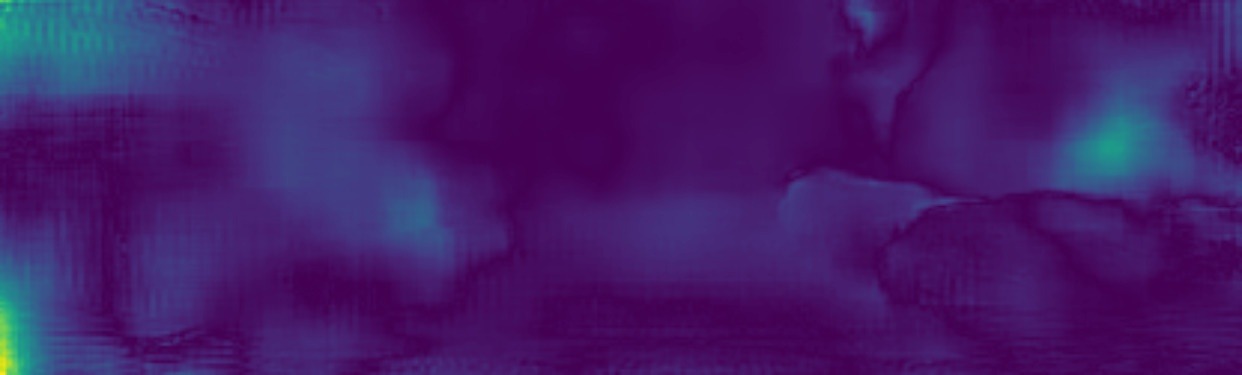} \\

\rotatebox[origin=l]{45}{B2F} & 
\includegraphics[width=0.22\linewidth]{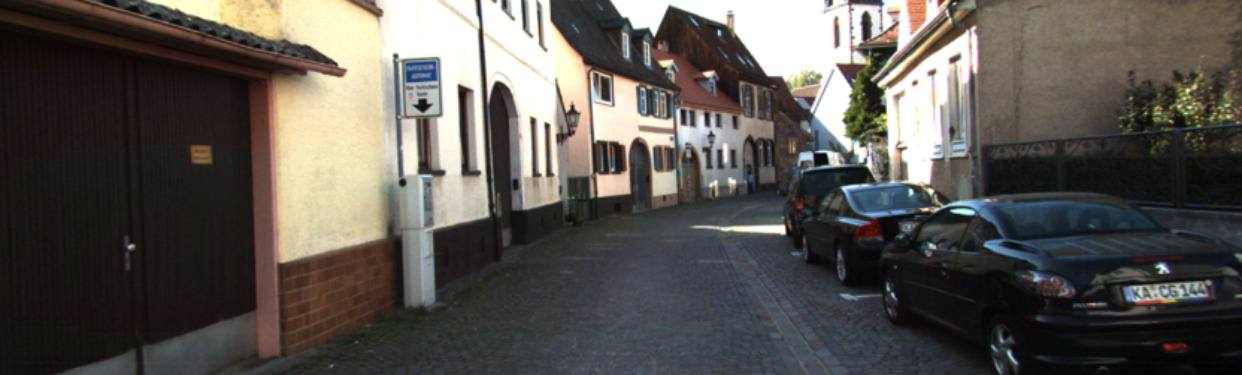} &
\includegraphics[width=0.22\linewidth]{figs/pert/qual/b2f/403_d.jpg} &
\includegraphics[width=0.22\linewidth]{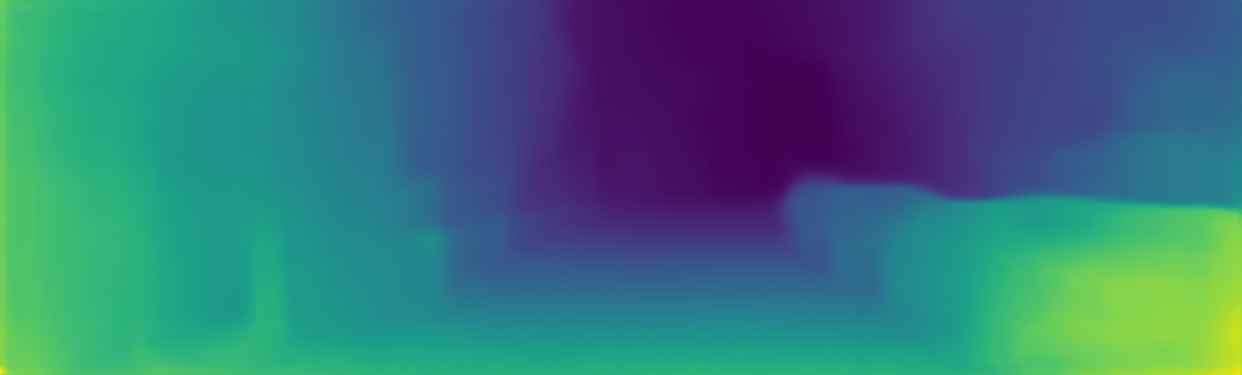} &
\includegraphics[width=0.22\linewidth]{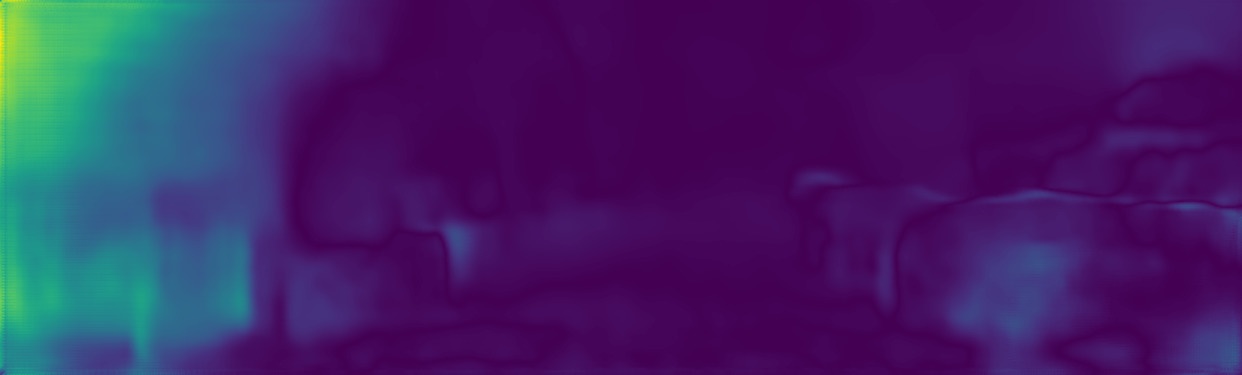} \\

\rotatebox[origin=l]{45}{SCSFM} & 
\includegraphics[width=0.22\linewidth]{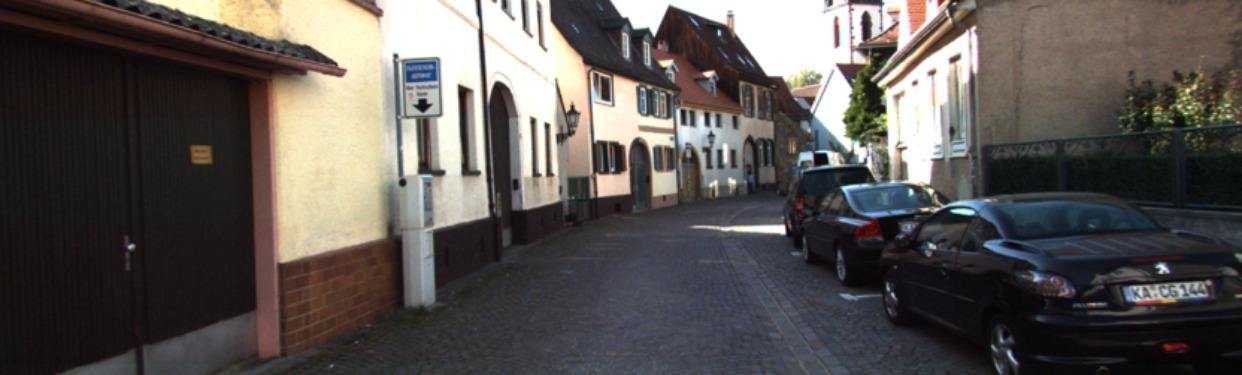} &
\includegraphics[width=0.22\linewidth]{figs/pert/qual/scsfm/403_d.jpg} &
\includegraphics[width=0.22\linewidth]{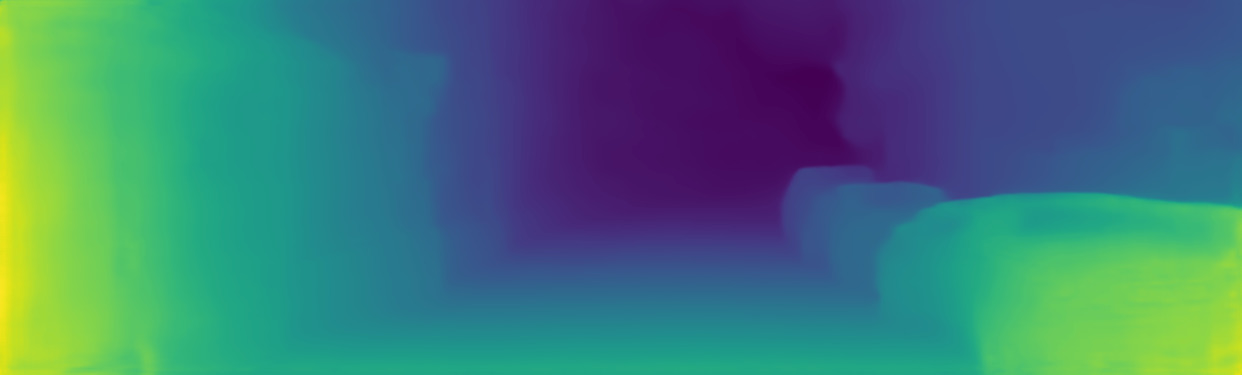} &
\includegraphics[width=0.22\linewidth]{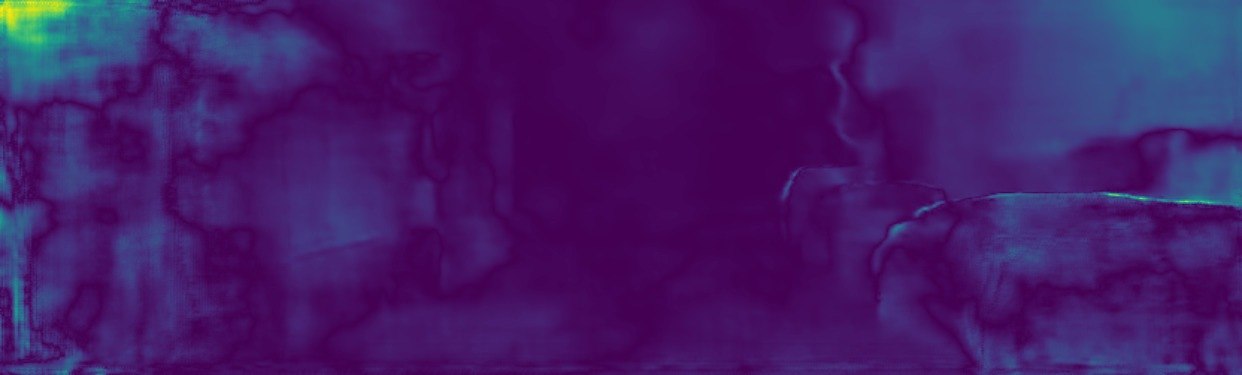} \\

\rotatebox[origin=l]{45}{Mono1} & 
\includegraphics[width=0.22\linewidth]{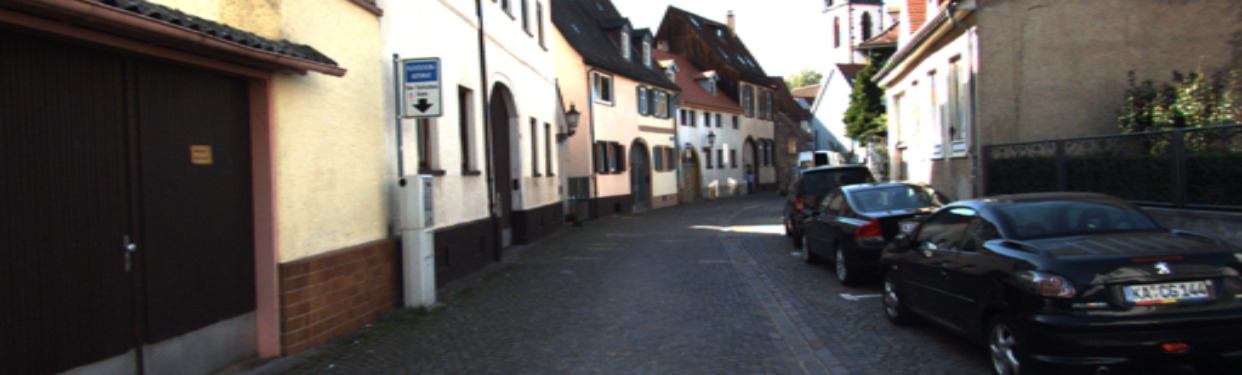} &
\includegraphics[width=0.22\linewidth]{figs/pert/qual/mono1/403_d.jpg} &
\includegraphics[width=0.22\linewidth]{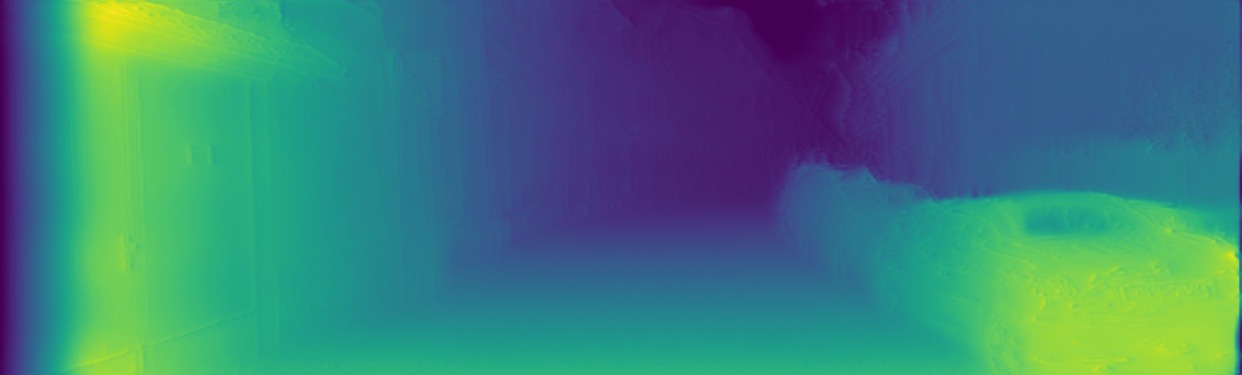} &
\includegraphics[width=0.22\linewidth]{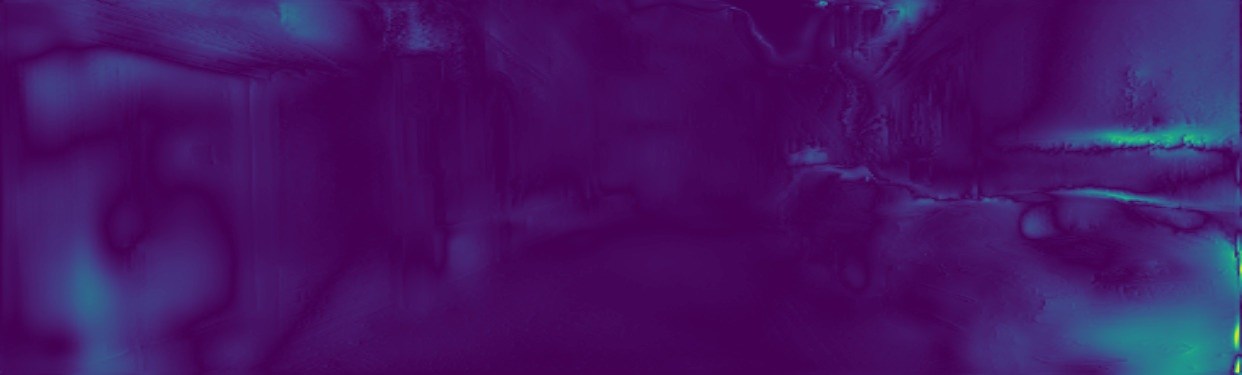} \\

\rotatebox[origin=l]{45}{Mono2} & 
\includegraphics[width=0.22\linewidth]{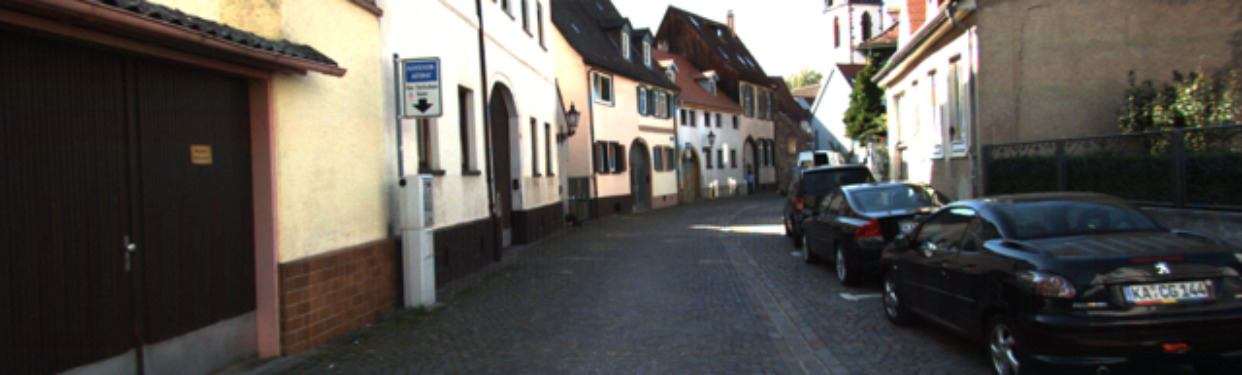} &
\includegraphics[width=0.22\linewidth]{figs/pert/qual/mono2/403_d.jpg} &
\includegraphics[width=0.22\linewidth]{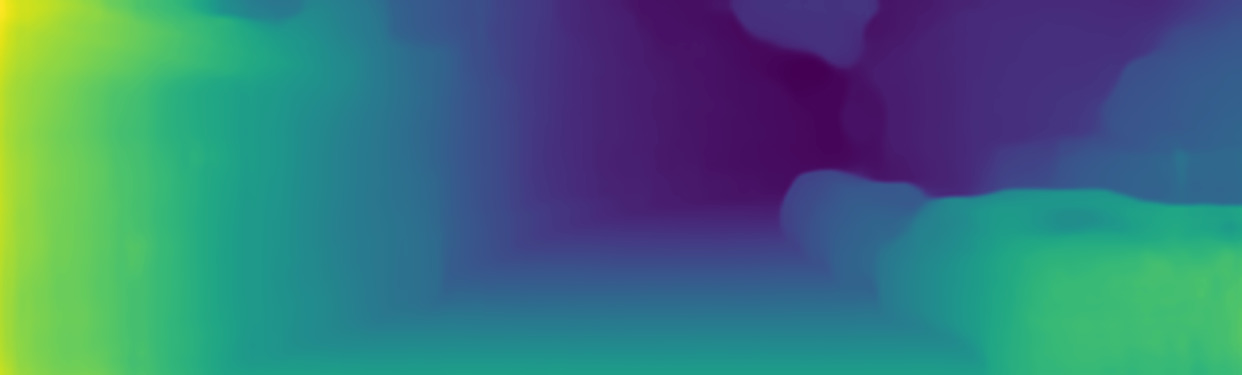} &
\includegraphics[width=0.22\linewidth]{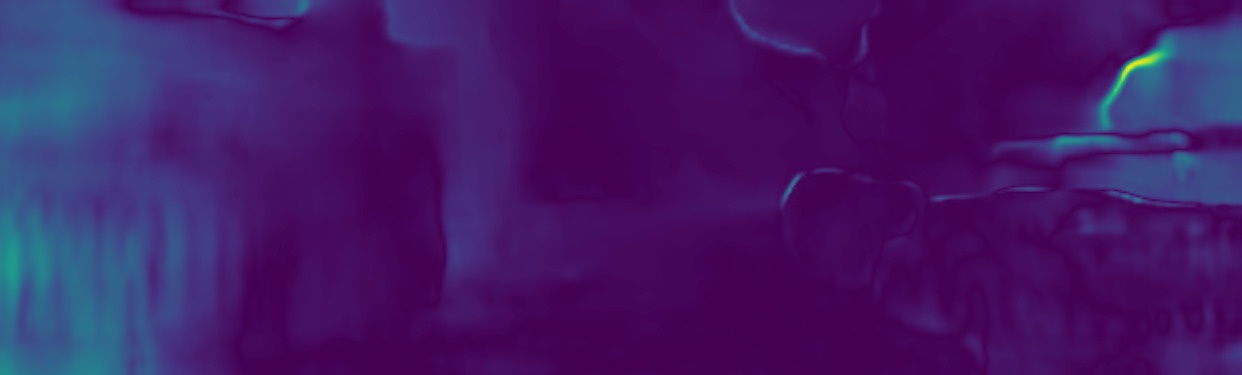} \\

\end{tabular}
\end{center}

\caption{White-box perturbation test when $\eta = 0.01$}
\label{fig:white_pert_adv1}
\end{figure*}


\begin{figure*}[!ht]
\centering
\newcommand{\turnheightnew}{0.15\columnwidth}
\centering

\begin{center}

\begin{tabular}{@{\hskip 0.5mm}c@{\hskip 0.5mm}c@{\hskip 0.5mm}c@{\hskip 0.5mm}c@{\hskip 0.5mm}c@{}}

& Attacked image & Clean depth & Attacked depth & Depth gap \\

\rotatebox[origin=l]{45}{SFM} & 
\includegraphics[width=0.22\linewidth]{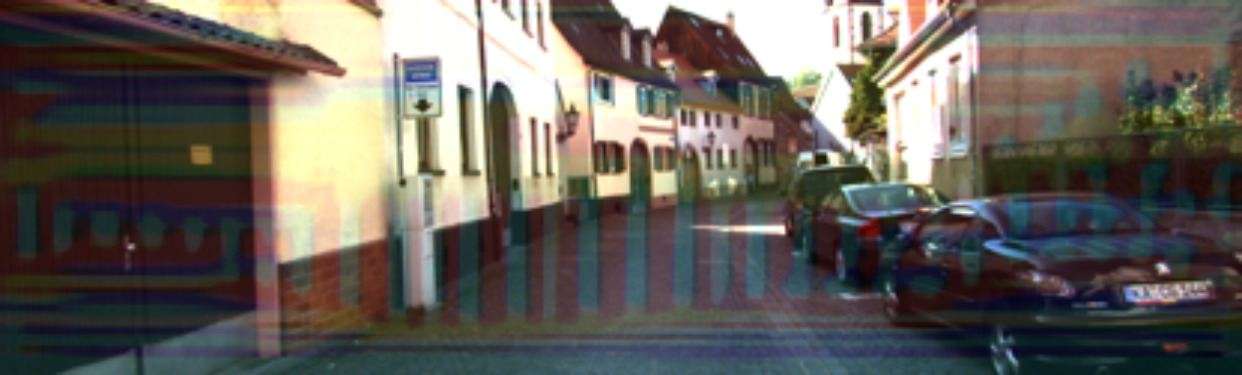} &
\includegraphics[width=0.22\linewidth]{figs/pert/qual/sfm/403_d.jpg} &
\includegraphics[width=0.22\linewidth]{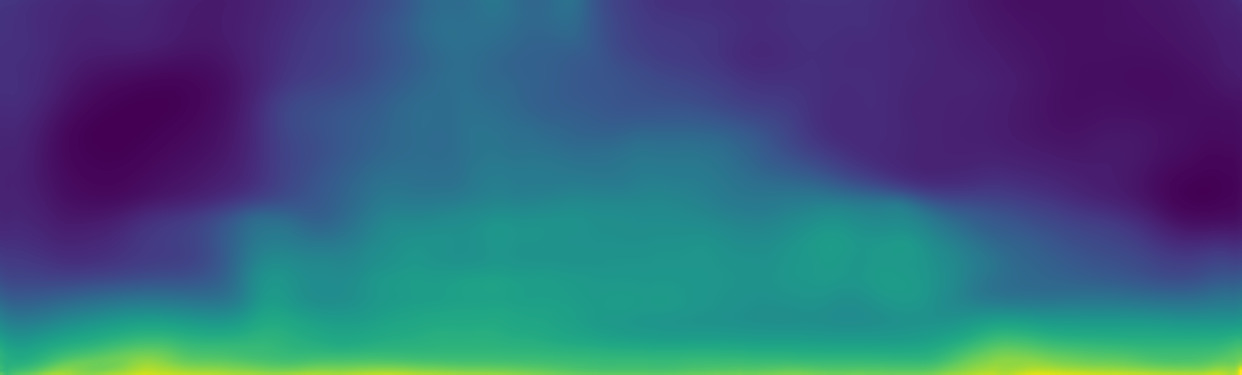} &
\includegraphics[width=0.22\linewidth]{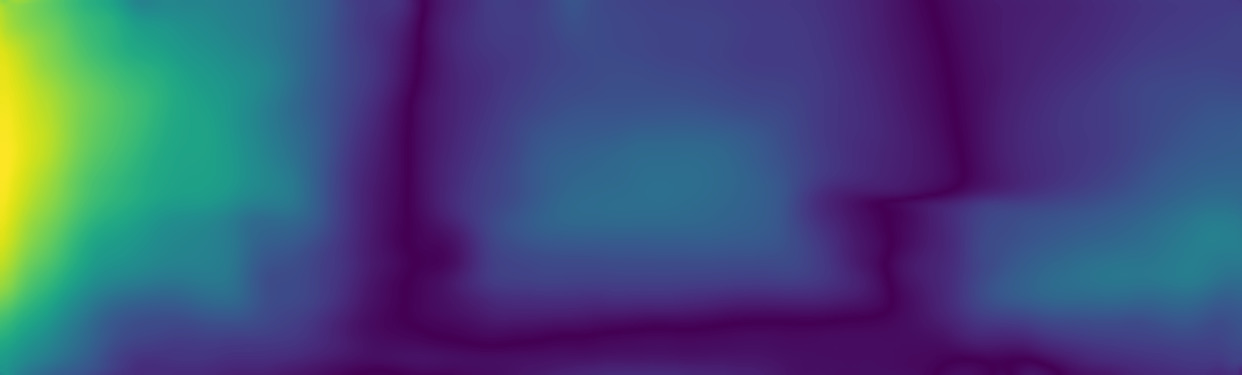} \\

\rotatebox[origin=l]{45}{DDVO} & 
\includegraphics[width=0.22\linewidth]{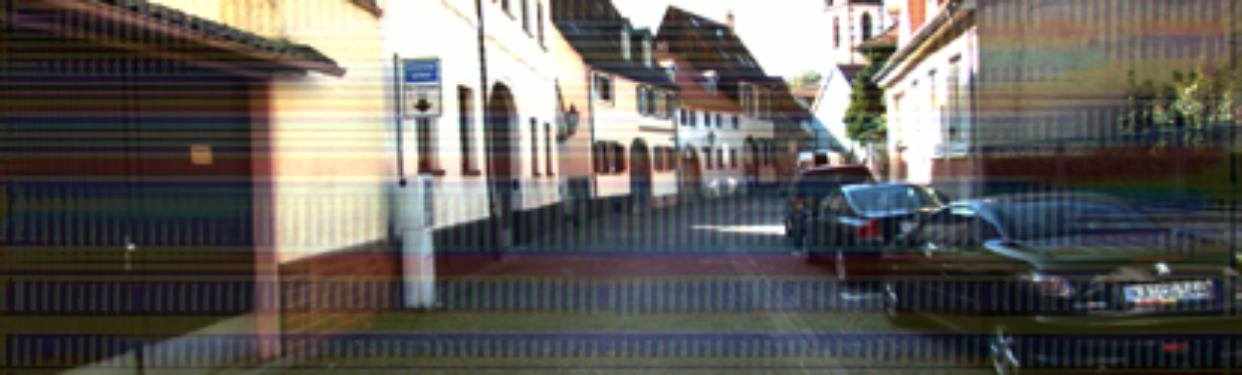} &
\includegraphics[width=0.22\linewidth]{figs/pert/qual/ddvo/403_d.jpg} &
\includegraphics[width=0.22\linewidth]{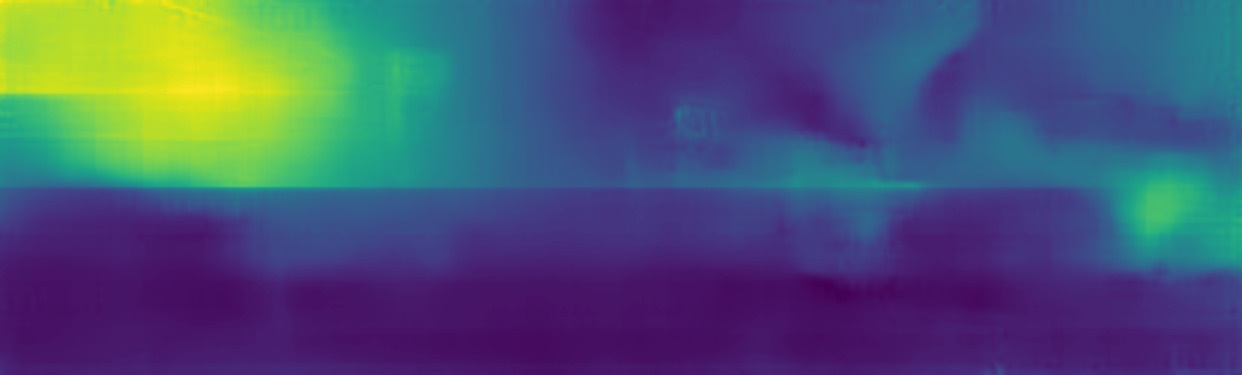} &
\includegraphics[width=0.22\linewidth]{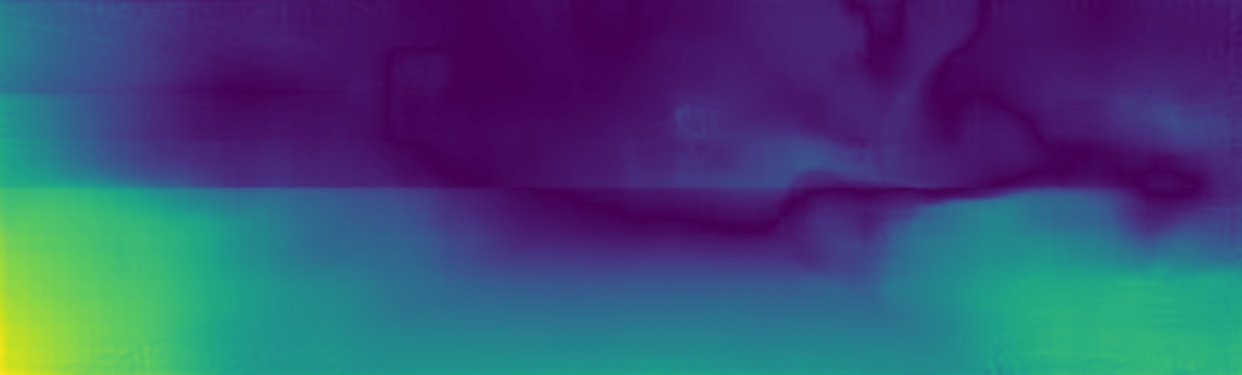} \\

\rotatebox[origin=l]{45}{B2F} & 
\includegraphics[width=0.22\linewidth]{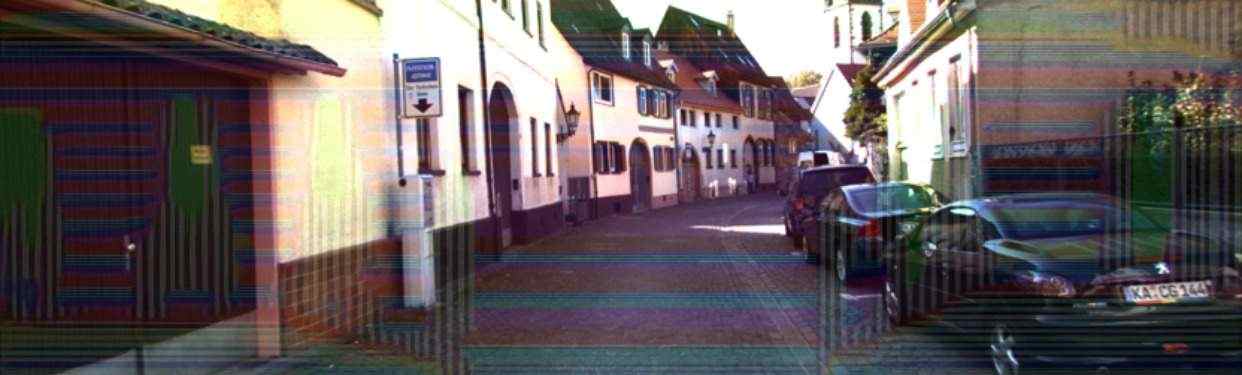} &
\includegraphics[width=0.22\linewidth]{figs/pert/qual/b2f/403_d.jpg} &
\includegraphics[width=0.22\linewidth]{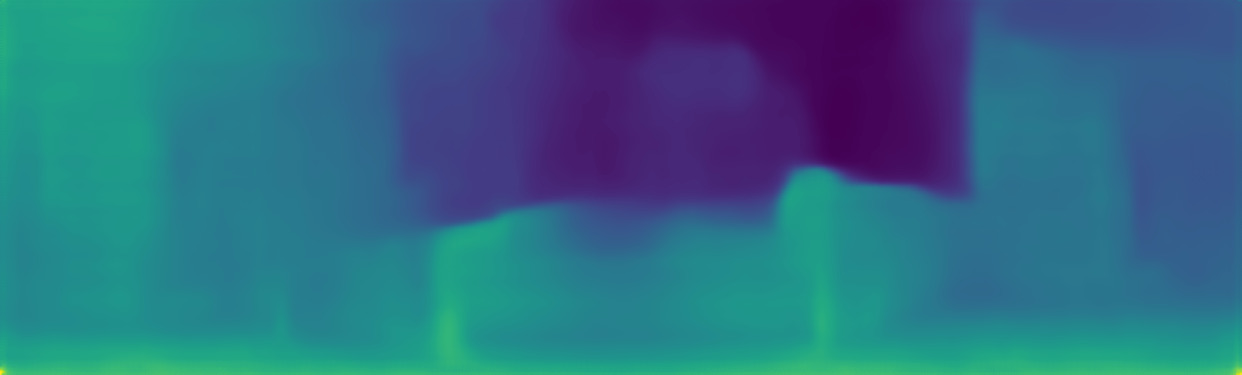} &
\includegraphics[width=0.22\linewidth]{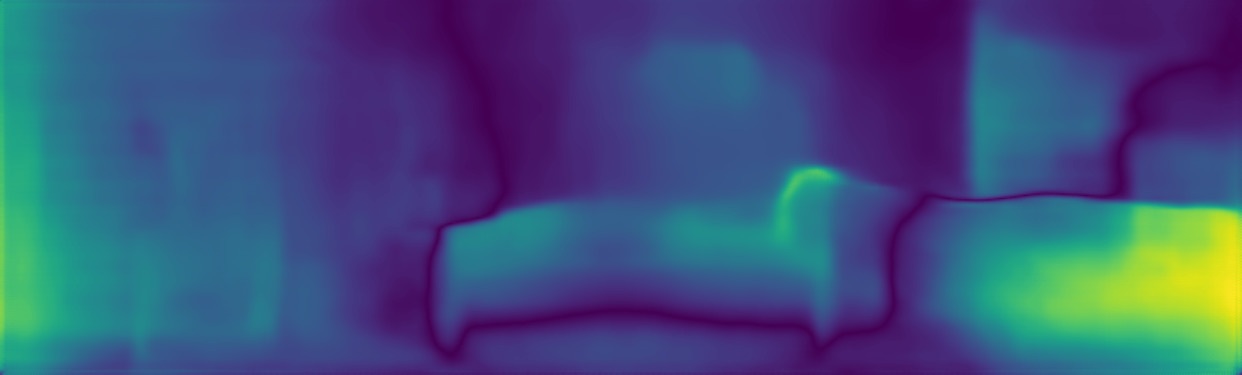} \\

\rotatebox[origin=l]{45}{SCSFM} & 
\includegraphics[width=0.22\linewidth]{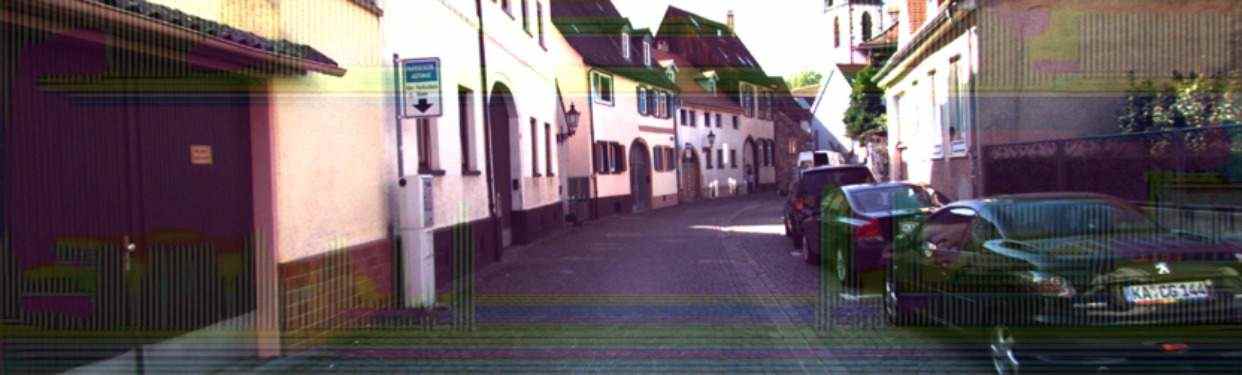} &
\includegraphics[width=0.22\linewidth]{figs/pert/qual/scsfm/403_d.jpg} &
\includegraphics[width=0.22\linewidth]{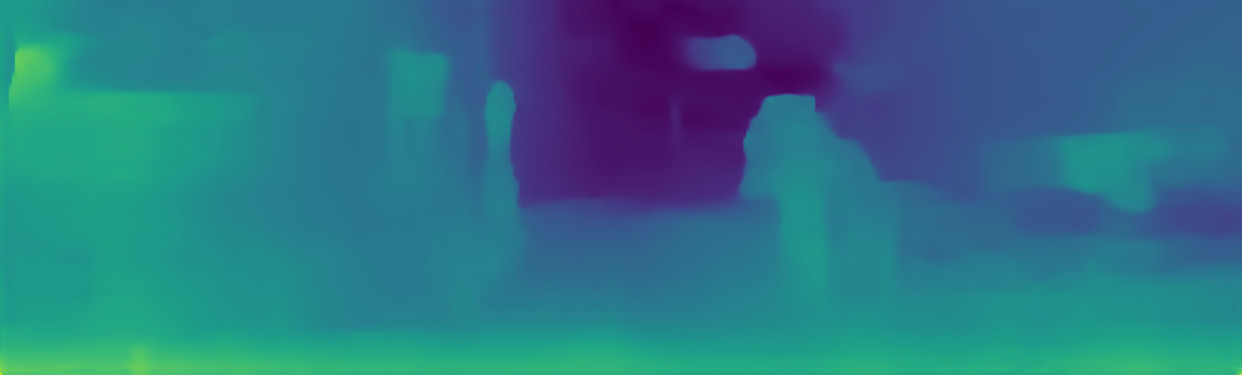} &
\includegraphics[width=0.22\linewidth]{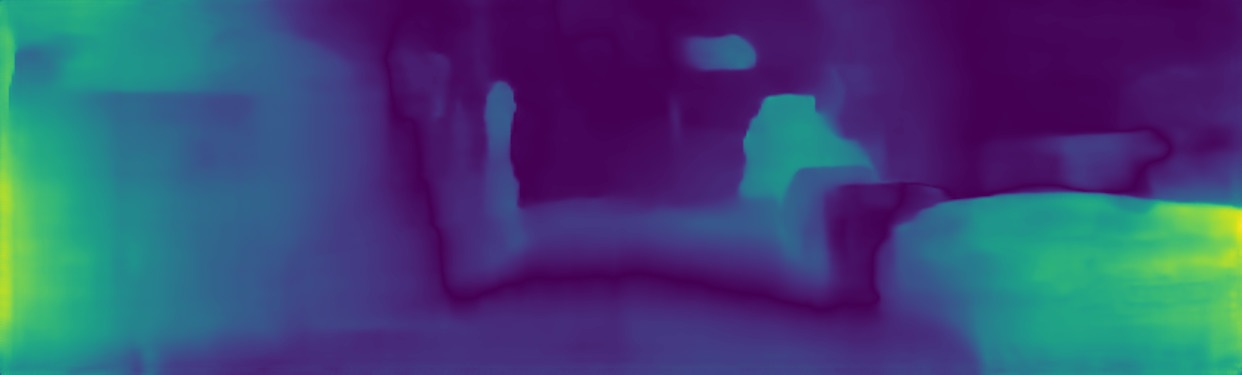} \\

\rotatebox[origin=l]{45}{Mono1} & 
\includegraphics[width=0.22\linewidth]{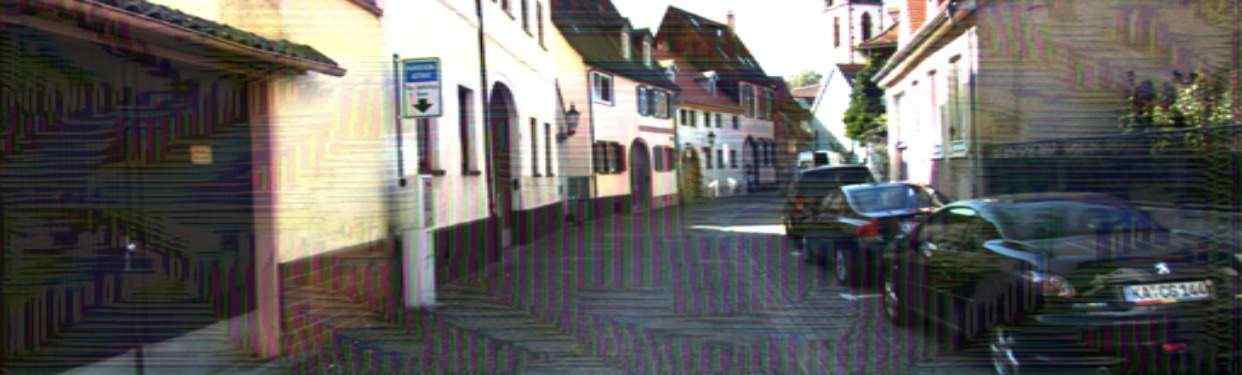} &
\includegraphics[width=0.22\linewidth]{figs/pert/qual/mono1/403_d.jpg} &
\includegraphics[width=0.22\linewidth]{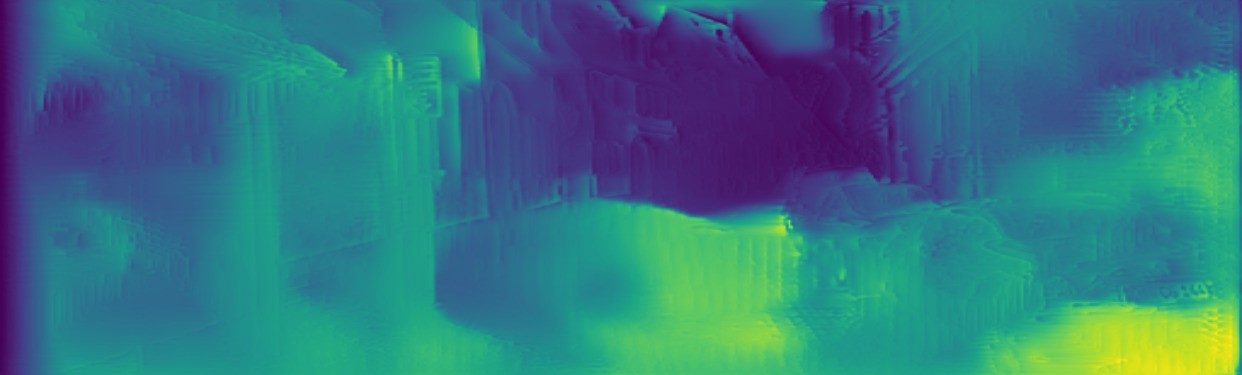} &
\includegraphics[width=0.22\linewidth]{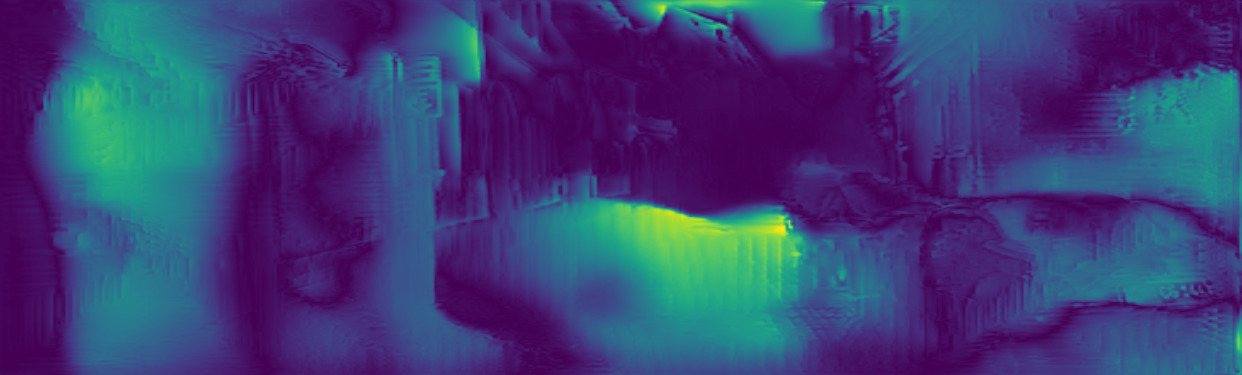} \\

\rotatebox[origin=l]{45}{Mono2} & 
\includegraphics[width=0.22\linewidth]{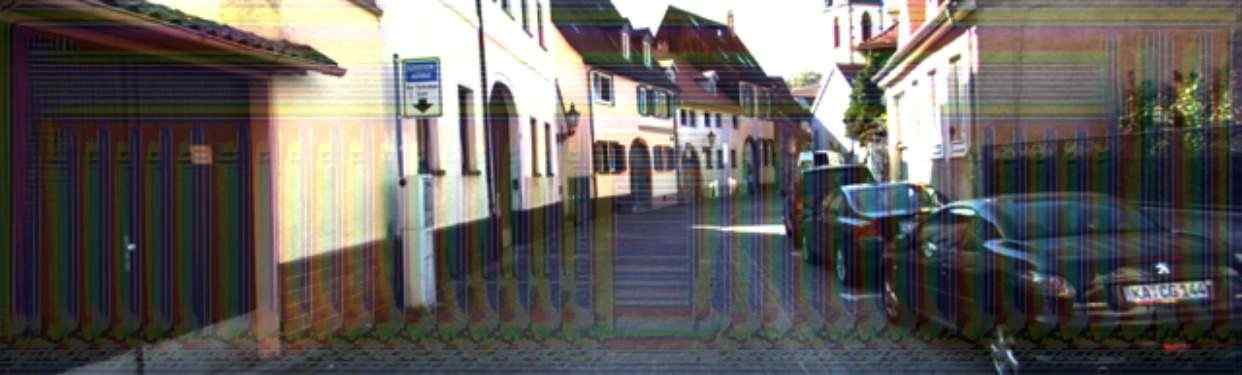} &
\includegraphics[width=0.22\linewidth]{figs/pert/qual/mono2/403_d.jpg} &
\includegraphics[width=0.22\linewidth]{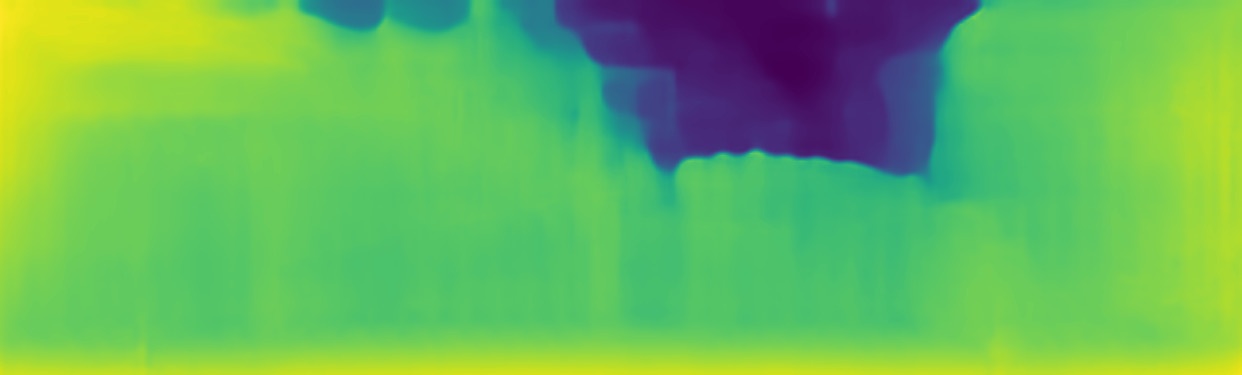} &
\includegraphics[width=0.22\linewidth]{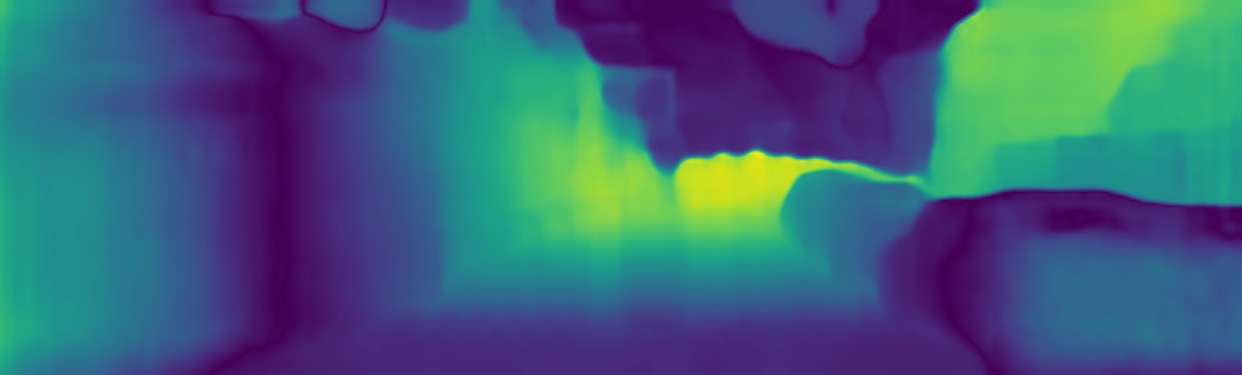} \\

\end{tabular}
\end{center}

\caption{White-box perturbation test when $\eta = 0.1$}
\label{fig:white_pert_adv10}
\end{figure*}


\begin{table*}
	\centering
	\begin{tabular}{l|c|c|c|c|c|c|c}
		\hline
		\multirow{2}{*}{Methods} & \multirow{2}{*}{Clean} & \multicolumn{6}{c}{Attacked} \\
		\cline{3-8}
		&    & \multicolumn{2}{c|}{$\eta=0.01$} & \multicolumn{2}{c|}{$\eta=0.05$} & \multicolumn{2}{c}{$\eta=0.1$} \\
		\cline{3-8}
		& Absrel   & Absrel & Rel (\%)             & Absrel & Rel (\%)             & Absrel & Rel (\%) \\
		\hline
		SFM~\cite{zhou2017unsupervised}                      & 0.1755                 & 0.1955 & 12               & 0.4069 & 132               & 0.5913 & 237   \\
		DDVO~\cite{wang2018learning}                     & 0.1488                 & 0.1568 & 6                & 0.7291 & 390               & 1.2197 & 720    \\
		B2F~\cite{janai2018unsupervised}                      & 0.1358                 & 0.1448  & 7               & 0.3725 & 175               & 0.4574 & 237   \\
		SCSFM~\cite{bian2019unsupervised}                    & 0.1283                 & 0.133 & 4                 & 0.3389 & 165               & 0.4874 & 280   \\
		Mono1~\cite{godard2017unsupervised}                    & 0.1095                 & 0.1115 & 2                & 0.2678 & 145               & 0.4341 & 297   \\
		Mono2~\cite{godard2019digging}                    & 0.1123                 & 0.1192 & 7                & 0.4037 & 260               & 0.4503 & 301   \\
		
		\hline
	\end{tabular}
	\caption{Absolute Relative Error for white-box test on image specific perturbation attack.}
	\label{whitebox_pert_absrel}
\end{table*}

\begin{table*}
	\centering
	\begin{tabular}{l|c|c|c|c|c|c|c}
		\hline
		\multirow{2}{*}{Methods} & \multirow{2}{*}{Clean} & \multicolumn{6}{c}{Attacked} \\
		\cline{3-8}
		&    & \multicolumn{2}{c|}{$\eta=0.01$} & \multicolumn{2}{c|}{$\eta=0.05$} & \multicolumn{2}{c}{$\eta=0.1$} \\
		\cline{3-8}
		& RMSE   & RMSE & Rel (\%)             & RMSE & Rel (\%)             & RMSE & Rel (\%) \\
		\hline
		SFM~\cite{zhou2017unsupervised}                      & 6.1711                 & 6.6009 & 7               & 10.4535 & 70         & 12.3833 & 101           \\
		DDVO~\cite{wang2018learning}                     & 5.5072                 & 5.7871 & 6               & 13.4231 & 144        & 18.662 & 239           \\
		B2F~\cite{janai2018unsupervised}                      & 5.1615                 & 5.4437  & 6              & 10.2015 & 98         & 12.2744 & 138           \\
		SCSFM~\cite{bian2019unsupervised}                    & 5.2271                 & 5.3731 & 3               & 9.35 & 79            & 12.5616 & 141          \\
		Mono1~\cite{godard2017unsupervised}                    & 5.1973                 & 5.1973 & 0               & 8.8863 & 71          & 13.4931 & 160           \\
		Mono2~\cite{godard2019digging}                    & 4.9099                 & 5.0419 & 3               & 13.2947 & 171        & 14.7999 & 202           \\
		\hline
	\end{tabular}
	\caption{RMSE for white-box test on image specific perturbation attack.}
	\label{tab:whitebox_pert_rmse}
\end{table*}

\begin{table*}
	\centering
	\begin{tabular}{l|c|c|c|c|c|c|c}
		\hline
		\multirow{2}{*}{Methods} & \multirow{2}{*}{Clean} & \multicolumn{6}{c}{Attacked} \\
		\cline{3-8}
		&    & \multicolumn{2}{c|}{$\eta=0.01$} & \multicolumn{2}{c|}{$\eta=0.05$} & \multicolumn{2}{c}{$\eta=0.1$} \\
		\cline{3-8}
		& Absrel   & Absrel & Rel (\%)             & Absrel & Rel (\%)             & Absrel & Rel (\%) \\
		\hline
		SFM~\cite{zhou2017unsupervised}                      & 0.1755                 & 0.1783 & 2             & 0.2651 & 52               & 0.3975 & 127   \\
		DDVO~\cite{wang2018learning}                     & 0.1488                 & 0.1526 & 3             & 0.5703 & 284              & 1.1942 & 703    \\
		B2F~\cite{janai2018unsupervised}                      & 0.1358                 & 0.1411 & 4             & 0.2213 & 63               & 0.3783 & 179   \\
		SCSFM~\cite{bian2019unsupervised}                    & 0.1283                 & 0.13 & 2               & 0.1834 & 43               & 0.3272 & 156   \\
		Mono1~\cite{godard2017unsupervised}                    & 0.1095                 & 0.1102 & 1             & 0.1794 & 64               & 0.3659 & 235   \\
		Mono2~\cite{godard2019digging}                    & 0.1123                 & 0.1167 & 4             & 0.2826 & 152              & 0.3984 & 225   \\
		
		\hline
	\end{tabular}
	\caption{Absolute Relative Error for white-box test on global perturbation attack.}
	\label{whitebox_gpert_absrel}
\end{table*}

\begin{table*}
	\centering
	\begin{tabular}{l|c|c|c|c|c|c|c}
		\hline
		\multirow{2}{*}{Methods} & \multirow{2}{*}{Clean} & \multicolumn{6}{c}{Attacked} \\
		\cline{3-8}
		&    & \multicolumn{2}{c|}{$\eta=0.01$} & \multicolumn{2}{c|}{$\eta=0.05$} & \multicolumn{2}{c}{$\eta=0.1$} \\
		\cline{3-8}
		& Absrel   & Absrel & Rel (\%)             & Absrel & Rel (\%)             & Absrel & Rel (\%) \\
		\hline
		SFM~\cite{zhou2017unsupervised}                      & 0.1755                 & 0.1802 & 3               & 0.2023 & 16               & 0.2316 & 32   \\
		DDVO~\cite{wang2018learning}                     & 0.1488                 & 0.1558 & 5               & 0.1763 & 19               & 0.1947 & 31    \\
		B2F~\cite{janai2018unsupervised}                      & 0.1358                 & 0.1491  & 10             & 0.1905 & 41               & 0.2158 & 59   \\
		SCSFM~\cite{bian2019unsupervised}                    & 0.1283                 & 0.1436 & 12              & 0.1707 & 34               & 0.1846 & 44   \\
		Mono1~\cite{godard2017unsupervised}                    & 0.1095                 & 0.1175 & 8               & 0.1472 & 35               & 0.1711 & 57   \\
		Mono2~\cite{godard2019digging}                    & 0.1123                 & 0.133 & 19               & 0.1696 & 52               & 0.2052 & 83   \\
		
		\hline
	\end{tabular}
	\caption{Absolute relative error of white-box FGSM attack.}
	\label{whitebox_fgsm_absrel}
\end{table*}

\begin{table*}
	\centering
	\begin{tabular}{l|c|c|c|c|c|c|c}
		\hline
		\multirow{2}{*}{Methods} & \multirow{2}{*}{Clean} & \multicolumn{6}{c}{Attacked} \\
		\cline{3-8}
		&    & \multicolumn{2}{c|}{$\eta=0.01$} & \multicolumn{2}{c|}{$\eta=0.05$} & \multicolumn{2}{c}{$\eta=0.1$} \\
		\cline{3-8}
		& RMSE   & RMSE & Rel (\%)             & RMSE & Rel (\%)             & RMSE & Rel (\%) \\
		\hline
		SFM~\cite{zhou2017unsupervised}                      & 6.1711                 & 6.2857 & 2              & 6.8239 & 11        & 7.4746 & 22           \\
		DDVO~\cite{wang2018learning}                     & 5.5072                 & 5.7242 & 4              & 6.2509 & 14        & 6.6626 & 21           \\
		B2F~\cite{janai2018unsupervised}                      & 5.1615                 & 5.349  & 4              & 5.9803 & 16        & 6.4383 & 25           \\
		SCSFM~\cite{bian2019unsupervised}                    & 5.2271                 & 5.3107 & 2              & 5.5715 & 7         & 5.9258 & 14          \\
		Mono1~\cite{godard2017unsupervised}                    & 5.1973                 & 5.347 & 3               & 5.9713 & 15        & 6.5298 & 26           \\
		Mono2~\cite{godard2019digging}                    & 4.9099                 & 5.4985 & 12             & 6.1803 & 26        & 6.891 & 41           \\
		\hline
	\end{tabular}
	\caption{RMSE of white-box FGSM attack.}
	\label{whitebox_fgsm_rmse}
\end{table*}

\clearpage

\section{Black-box attack}

\begin{table*}[!ht]
	\begin{adjustbox}{max width=\textwidth}
	\begin{tabular}{l|c|c|c|c|c|c|c|c|c|c|c}
		\hline
		\multicolumn{1}{l}{} & \multicolumn{1}{|l}{} & \multicolumn{10}{|c}{Patch}  \\
		\cline{3-12}
		\multicolumn{1}{l|}{} & \multicolumn{1}{c|}{Clean} & \multicolumn{2}{c|}{SFM~\cite{zhou2017unsupervised}} & \multicolumn{2}{c|}{DDVO~\cite{wang2018learning}} & \multicolumn{2}{c|}{SCSFM~\cite{bian2019unsupervised}} & \multicolumn{2}{c|}{Mono1~\cite{godard2017unsupervised}} & \multicolumn{2}{c}{Mono2~\cite{godard2019digging}}\\
		\cline{2-12}
		Methods  & Absrel & Absrel   & Rel (\%)     & Absrel       & Rel (\%)     & Absrel   & Rel (\%)      & Absrel      & Rel (\%)      & Absrel  & Rel (\%)    \\
		\hline
		SFM~\cite{zhou2017unsupervised}      & 0.1755 & -            & -        & 0.1962        & 12       & 0.1858         & 6        & 0.191         & 9         & 0.1803         & 3        \\
		DDVO~\cite{wang2018learning}     & 0.1488 & 0.1488       & 0        & -             & -        & 0.1514         & 2        & 0.1496        & 1         & 0.1577         & 6        \\
		B2F~\cite{janai2018unsupervised}      & 0.1358 & 0.1422       & 5        & 0.1425        & 5        & 0.1432         & 6        & 0.1401        & 4         & 0.1444         & 7        \\
		SCSFM~\cite{bian2019unsupervised}    & 0.1283 & 0.1448       & 13       & 0.1477        & 16       & -              & -        & 0.1462        & 14        & 0.1343         & 5        \\
		Mono1~\cite{godard2017unsupervised}    & 0.1095 & 0.1265       & 16       & 0.12          & 10       & 0.119          & 9        & -             & -         & 0.115          & 6        \\
		Mono2~\cite{godard2019digging}    & 0.1123 & 0.1136       & 2        & 0.116         & 4        & 0.1189         & 6        & 0.1146        & 3         & -              & -        \\
		\hline
	\end{tabular}
	\end{adjustbox}
	\caption{Black-box patch attack with patch size $72 \times 72$.}
	\label{trans_patch_relabs}
\end{table*}

\begin{table*}[!ht]
	\begin{adjustbox}{max width=\textwidth}
	\begin{tabular}{l|c|c|c|c|c|c|c|c|c|c|c|c}
		\hline
		\multicolumn{1}{l}{} & \multicolumn{10}{|c}{Perturbation}  \\
		\cline{2-13}
		\multicolumn{1}{l|}{} & \multicolumn{2}{c|}{SFM~\cite{zhou2017unsupervised}} & \multicolumn{2}{c|}{DDVO~\cite{wang2018learning}} & \multicolumn{2}{c|}{B2F~\cite{janai2018unsupervised}} & \multicolumn{2}{c|}{SCSFM~\cite{bian2019unsupervised}} & \multicolumn{2}{c|}{Mono1~\cite{godard2017unsupervised}} & \multicolumn{2}{c}{Mono2~\cite{godard2019digging}}\\
		\cline{2-13}
		Methods    & Absrel   & Rel (\%)    & Absrel      & Rel (\%)  & Absrel       & Rel (\%)      & Absrel   & Rel (\%)  & Absrel     & Rel (\%)  & Absrel  & Rel (\%) \\
		\hline
		SFM~\cite{zhou2017unsupervised}      & -            & -       & 0.1842        & 5        & 0.181          & 4        & 0.1775       & 2         & 0.1775      & 2     & 0.1859      & 6   \\
		DDVO~\cite{wang2018learning}     & 0.1799       & 21      & -             & -        & 0.1675         & 13       & 0.1534       & 4         & 0.1598      & 8     & 0.1667      & 13  \\
		B2F~\cite{janai2018unsupervised}      & 0.1675       & 24      & 0.5703        & 284      & -              & -        & 0.1724       & 27        & 0.1553      & 15    & 0.1643      & 21   \\
		SCSFM~\cite{bian2019unsupervised}    & 0.1512       & 18      & 0.161         & 26       & 0.165          & 29       & -            & -         & 0.1413      & 11    & 0.1565      & 22    \\
		Mono1~\cite{godard2017unsupervised}    & 0.1147       & 5       & 0.124         & 14       & 0.1242         & 14       & 0.1255       & 15        & -           & -     & 0.1228      & 13   \\
		Mono2~\cite{godard2019digging}    & 0.1401       & 25      & 0.1452        & 30       & 0.1454         & 30       & 0.1519       & 36        & 0.129       & 15    & -           & -   \\
		\hline
	\end{tabular}
	\end{adjustbox}
	\caption{Black-box perturbation attack when $\eta=0.05$}
	\label{trans_pert_relabs}
\end{table*}

\section{Feature visualization}

\begin{figure*}[!htbp]
	\centering
	
	\begin{tabular}{ccccccccc}
		
		\multicolumn{9}{c}{Clean features} \\
		\hline
		
		conv1 & conv2 & conv3 & conv4 & conv5 & conv6 & conv7 & iconv1 & iconv2 \\
		
		\includegraphics[width=0.08\linewidth]{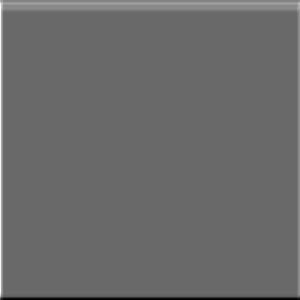} &
		\includegraphics[width=0.08\linewidth]{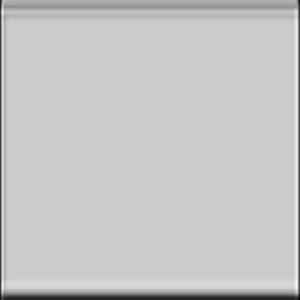} &
		\includegraphics[width=0.08\linewidth]{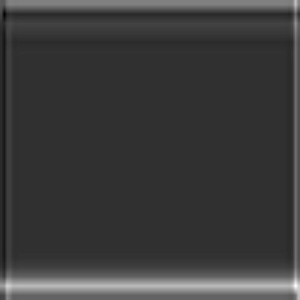} &
		\includegraphics[width=0.08\linewidth]{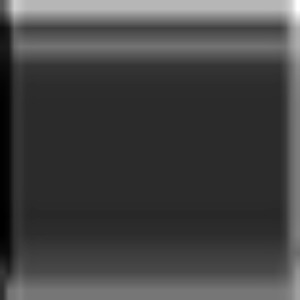} &
		\includegraphics[width=0.08\linewidth]{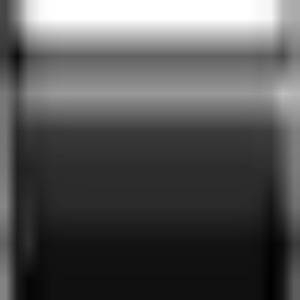} &
		\includegraphics[width=0.08\linewidth]{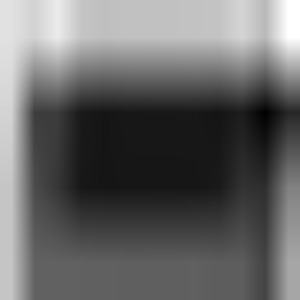} &
		\includegraphics[width=0.08\linewidth]{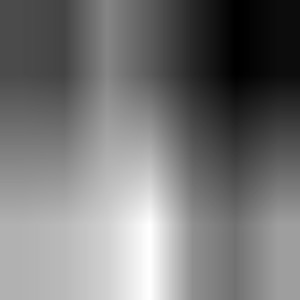} &
		\includegraphics[width=0.08\linewidth]{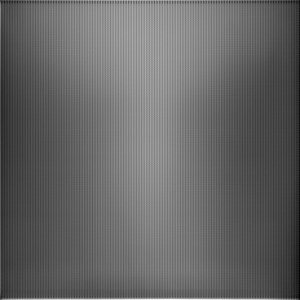} &
		\includegraphics[width=0.08\linewidth]{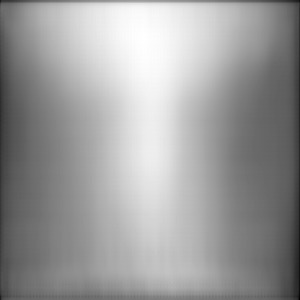} \\
		
		\includegraphics[width=0.08\linewidth]{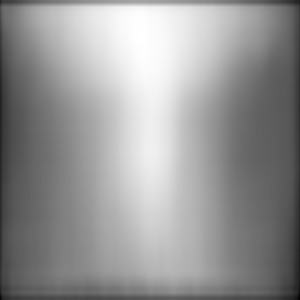} &
		\includegraphics[width=0.08\linewidth]{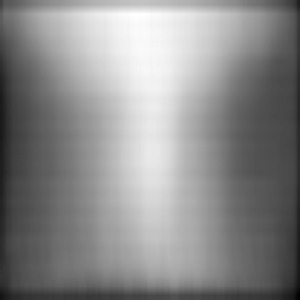} &
		\includegraphics[width=0.08\linewidth]{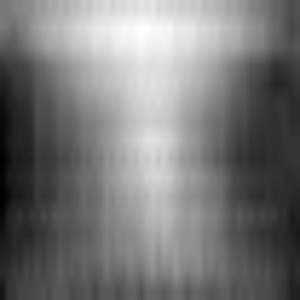} &
		\includegraphics[width=0.08\linewidth]{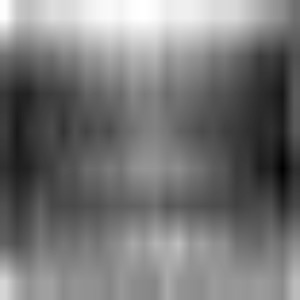} &
		\includegraphics[width=0.08\linewidth]{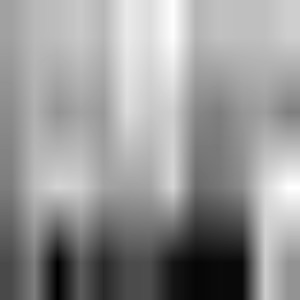} &
		\includegraphics[width=0.08\linewidth]{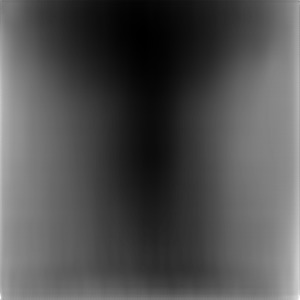} &
		\includegraphics[width=0.08\linewidth]{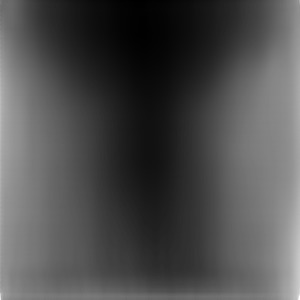} &
		\includegraphics[width=0.08\linewidth]{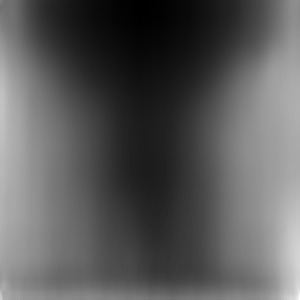} &
		\includegraphics[width=0.08\linewidth]{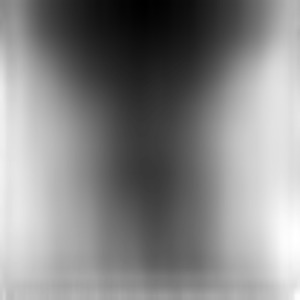} \\
		
		iconv3 & iconv4 & iconv5 & iconv6 & iconv7 & disp1 & disp2 & disp3 & disp4 \\
		
		\\
		
		\multicolumn{9}{c}{Attacked features} \\
		\hline
		
		conv1 & conv2 & conv3 & conv4 & conv5 & conv6 & conv7 & iconv1 & iconv2 \\
		
		\includegraphics[width=0.08\linewidth]{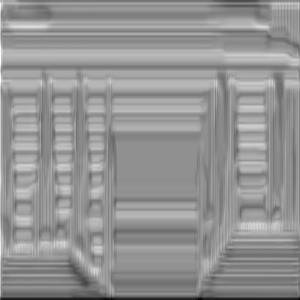} &
		\includegraphics[width=0.08\linewidth]{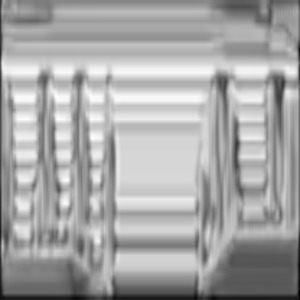} &
		\includegraphics[width=0.08\linewidth]{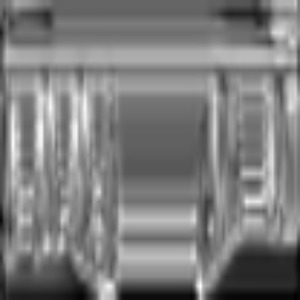} &
		\includegraphics[width=0.08\linewidth]{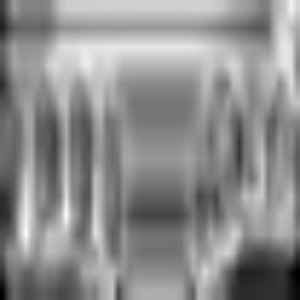} &
		\includegraphics[width=0.08\linewidth]{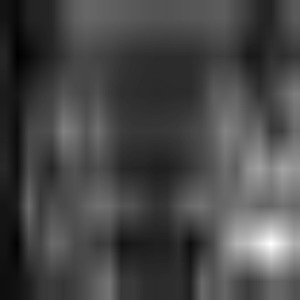} &
		\includegraphics[width=0.08\linewidth]{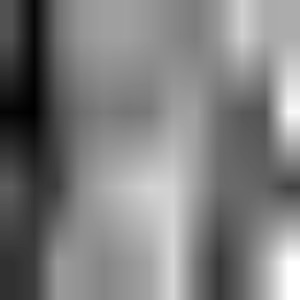} &
		\includegraphics[width=0.08\linewidth]{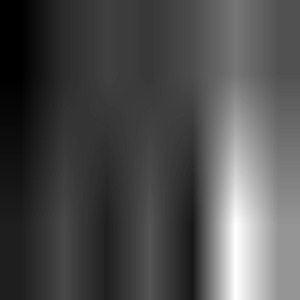} &
		\includegraphics[width=0.08\linewidth]{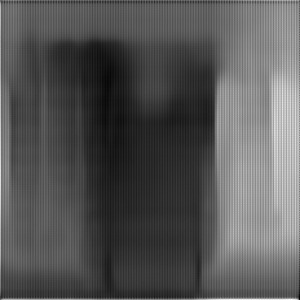} &
		\includegraphics[width=0.08\linewidth]{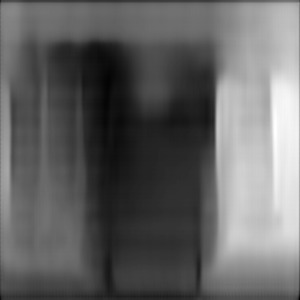} \\
		
		\includegraphics[width=0.08\linewidth]{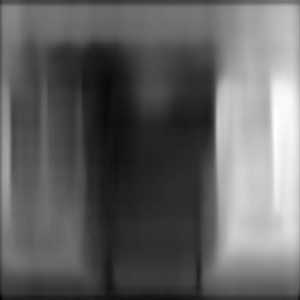} &
		\includegraphics[width=0.08\linewidth]{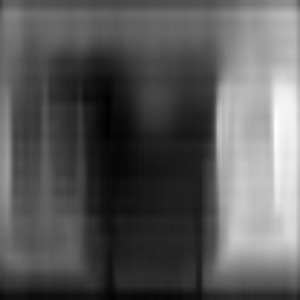} &
		\includegraphics[width=0.08\linewidth]{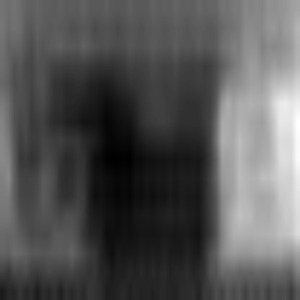} &
		\includegraphics[width=0.08\linewidth]{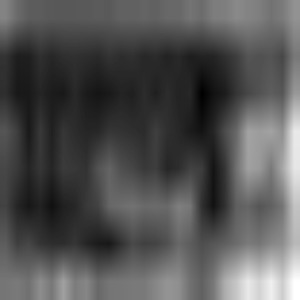} &
		\includegraphics[width=0.08\linewidth]{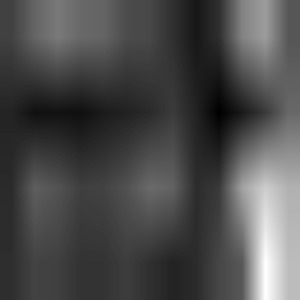} &
		\includegraphics[width=0.08\linewidth]{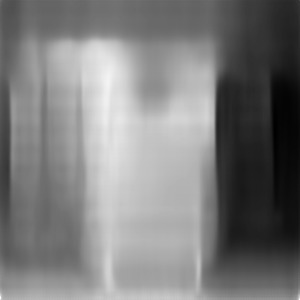} &
		\includegraphics[width=0.08\linewidth]{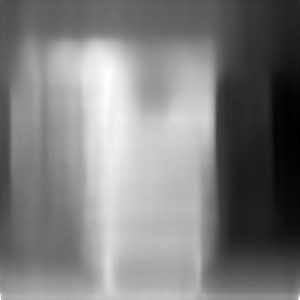} &
		\includegraphics[width=0.08\linewidth]{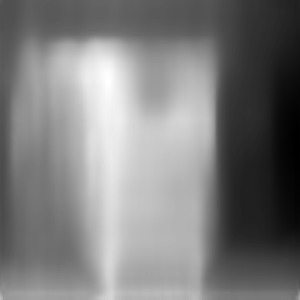} &
		\includegraphics[width=0.08\linewidth]{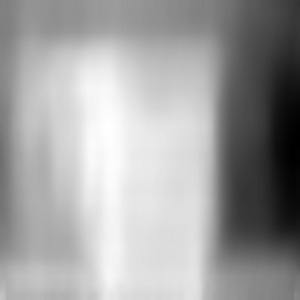} \\
		
		iconv3 & iconv4 & iconv5 & iconv6 & iconv7 & disp1 & disp2 & disp3 & disp4 \\
		
	\end{tabular}
	\caption{Layer-wise B2F~\cite{janai2018unsupervised} feature visualization. `convx' are encoder layers and `iconvx' are decoder layers.}
	\label{fig:feat_lvis_b2f}
\end{figure*}
	
\end{document}